\documentclass[10pt,twocolumn,letterpaper]{article}

\usepackage[pagenumbers]{cvpr}

\usepackage{graphicx}
\usepackage{amsmath}
\usepackage{amssymb}
\usepackage{booktabs}

\usepackage{comment}
\usepackage{array}

\usepackage{color}
\usepackage{enumitem}

\usepackage{pifont}

\usepackage[table,dvipsnames]{xcolor}

\usepackage{makecell}
\usepackage{multirow}
\usepackage{tabularx}

\newcommand{\cmark}{\ding{51}}%
\newcommand{\xmark}{\ding{55}}%
\newcommand{\mysize}{1.0}

\makeatletter

\definecolor{notetext}{rgb}{0.7,0,0}

\definecolor{ubpubColor}{rgb}{0.43, 0.5, 0.5}
\definecolor{backboneColor}{rgb}{0.423, 0.325, 0.365}
\definecolor{fpnColor}{rgb}{0.255, 0.498, 0.416}

\newcommand{\PAR}[1]{\vskip4pt \noindent {\bf #1~}}

\newcommand{\pointvos}{Point-VOS}
\newcommand{\dynamite}{DynaMITe}
\newcommand{\vidln}{VidLN}
\newcommand{\stcn}{STCN}
\newcommand{\pointstcn}{Point-STCN}

\newcommand{\pvoops}{PV-Oops}
\newcommand{\pvkinetics}{PV-Kinetics}
\newcommand{\pvdavis}{PV-DAVIS}
\newcommand{\pvytvos}{PV-YT}

\newcommand{\J}{\mathcal{J}}
\newcommand{\F}{\mathcal{F}}
\newcommand{\JnF}{\mathcal{J}\&\mathcal{F}}

\newcolumntype{Y}{>{\centering\arraybackslash}X}

\definecolor{cvprblue}{rgb}{0.21,0.49,0.74}
\usepackage[pagebackref,breaklinks,colorlinks,citecolor=cvprblue]{hyperref}
\newcommand\blfootnote[1]{
  \begingroup
  \renewcommand\thefootnote{}\footnote{#1}
  \addtocounter{footnote}{-1}
  \endgroup
}

\title{\pointvos{}: Pointing Up Video Object Segmentation}

\author{Idil Esen Zulfikar$^{1,}$\textcolor{red}{\footnotemark[1]}
\quad
Sabarinath Mahadevan$^{1,}$\textcolor{red}{\footnotemark[1]}
\quad
Paul Voigtlaender$^{2,}$\textcolor{red}{\footnotemark[1]}
\quad
Bastian Leibe$^{1}$\\
$^1${RWTH Aachen University, Germany}\quad 
$^2${Google Research} \\
\footnotesize{$^1${\tt\small \{zulfikar, mahadevan, leibe\}@vision.rwth-aachen.de}}\quad 
\footnotesize{$^2${\tt\small voigtlaender@google.com}} \\
\small{\texttt{\url{pointvos.github.io}}}
}

\begin{document}
\maketitle
\begin{abstract}
    Current state-of-the-art Video Object Segmentation (VOS) methods rely on dense per-object mask annotations both during training and testing. This requires time-consuming and costly video annotation mechanisms. We propose a novel \pointvos{} task with a spatio-temporally sparse point-wise annotation scheme that substantially reduces the annotation effort. We apply our annotation scheme to two large-scale video datasets with text descriptions and annotate over $19M$ points across $133K$ objects in $32K$ videos. 
    Based on our annotations, we propose a new \pointvos{} benchmark, and a corresponding point-based training mechanism, which we use to establish strong baseline results.
    We show that existing VOS methods can easily be adapted to leverage our point annotations during training, and can achieve results close to the fully-supervised performance when trained on pseudo-masks generated from these points. In addition, we show that our data can be used to improve models that connect vision and language, by evaluating it on the Video Narrative Grounding (VNG) task. We will make our code and annotations available at \texttt{\url{https://pointvos.github.io}}.
\end{abstract}
\blfootnote{\textcolor{red}{*} Equal contribution. The ordering of the authors was determined by a last-minute coin flip.}
\section{Introduction}
\label{sec:intro}
Video Object Segmentation (VOS) has grown into a very popular research field~\cite{Tsai2010MotionCT, Brox2010ObjectSB, Bro11c, Perazzi2016} that has shown considerable progress over the past few years~\cite{cheng2022xmem,cheng2021stcn, yang2021aot}, branching out into new downstream tasks with language referring expressions~\cite{khoreva2019video} or user interactions~\cite{Caelles_arXiv_2018}. New datasets have been instrumental in advancing progress in VOS \cite{xu2018youtube,Voigtlaender21WACV,athar2023burst,tokmakov2023breaking,ding2023mose,hong2022lvos}. However, the relatively costly annotation process necessary for creating VOS datasets has so far been a major limiting factor.
The traditional VOS task requires temporally dense object segmentation masks for the frames of each training video. As a result, existing video segmentation datasets~\cite{Perazzi2016, Pont-Tuset_arXiv_2017,  tokmakov2023breaking, hong2022lvos} are usually relatively small in scale, and past community efforts to scale them up to at least several thousand videos required substantial annotation effort~\cite{xu2018youtube,athar2023burst,ding2023mose,darkhalil2022visor}. Given the community's clear trend to connect vision to language~\cite{Voigtlaender23CVPR,radford21CLIP} and the consequent need for even larger datasets~\cite{schuhmann2022laion}, there is thus an urgent need to reduce the annotation cost for videos. 

\begin{figure}
    \centering
    \includegraphics[width=0.999\linewidth]{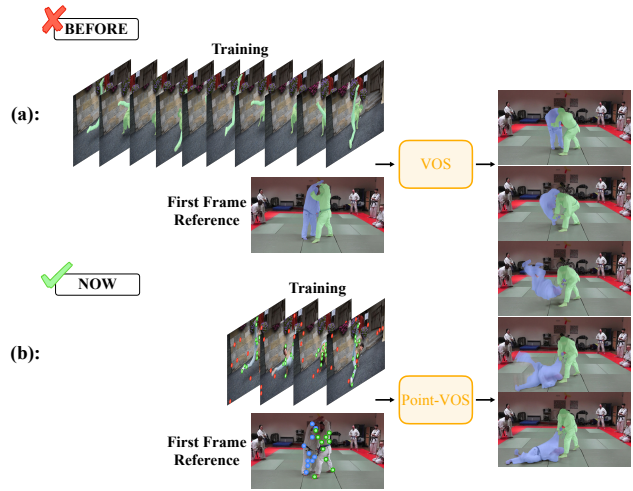}
    \vspace{-0.7cm}
    \caption{\textbf{Comparison of the 
    conventional VOS task with our new \pointvos{} task.} (a) The conventional VOS task utilizes dense segmentation mask for each frame during training and initializes the first-frame reference with dense masks. (b) We propose to change this paradigm and use only spatially sparse point annotations on a sparse subset of frames during training, and only a few points for the first-frame reference initialization. \textcolor{green}{Green} and \textcolor{blue}{blue} dots represent foreground points and \textcolor{red}{red} dots background points.}
    \label{fig:teaser}
\end{figure}

Some approaches try to mitigate this problem by reducing the reliance of vision models on annotated training data, \eg, by 
self-supervised learning~\cite{vondrick2018eccv,wang2019cvpr} or 
exploiting image-level mask annotations~\cite{athar2022hodor}. Another strategy has been to create semi-automatic annotation pipelines \cite{Voigtlaender21WACV} that generate pseudo ground-truth 
masks from more readily available data, such as existing bounding box annotations. Nevertheless, neither of those presents a general solution.

In this work, we address the annotation cost problem by proposing an entirely point-based VOS framework, \pointvos{}. Inspired by recent point-guided image segmentation methods~\cite{benenson2022colouring, cheng2022pointly}, \pointvos{} moves away from using full mask supervision and instead relies on spatio-temporally sparse point annotations as weak supervision signals for VOS (see \cref{fig:teaser}). Our point-based formulation enables us to design an efficient semi-automatic annotation pipeline (see \cref{fig:annotation_pipeline}) that requires substantially less annotation effort to create human-validated ground-truth video annotations.

We demonstrate the value of our proposed annotation pipeline by annotating two large-scale video datasets, \pointvos{} Oops~\cite{Epstein_2020_CVPR} (\pvoops) and \pointvos{} Kinetics~\cite{kay2017kinetics} (\pvkinetics), with altogether $19M$ points for $133K$ objects in $32K$ videos.
Our annotations cover almost an order of magnitude more videos and objects than major previous VOS datasets~\cite{Perazzi2016, Pont-Tuset_arXiv_2017, xu2018youtube, Voigtlaender21WACV, athar2023burst, ding2023mose, tokmakov2023breaking, hong2022lvos} (see \cref{tab:dataset-statistics}). In particular, we show how our annotation pipeline can make use of existing information from the Video Localized Narratives corpus (VidLN~\cite{Voigtlaender23CVPR}) in order to bootstrap the annotation process by automatically converting mouse traces from VidLN into \pointvos{} initializations.

We launch a new \pointvos{} benchmark based on these datasets, where VOS methods are expected to use only point annotations both as training supervision and as test-time initialization. We also develop two strong baselines by (i) adapting the state-of-the-art VOS method STCN~\cite{cheng2021stcn} to work directly with points instead of masks, and (ii) training STCN on pseudo-masks generated from point annotations.
Our experiments show that, despite the weaker level of supervision, this \pointvos{} STCN baseline already reaches more than 90\% of the performance of the original STCN when applied to the DAVIS benchmark \cite{Pont-Tuset_arXiv_2017} (see \cref{subsec:ablations}).

Finally, we also show a direct use case of our point annotations for language-guided VOS. 
As a consequence of our use of VidLN data to bootstrap the point annotation pipeline, the point annotations in \pvoops{} and \pvkinetics{} are connected to nouns in longer language captions (the Video Localized Narratives), describing the referred object's actions in the video. Thus, our annotations are multi-modal and bridge the gap between open-vocabulary language object descriptions and the corresponding video object segmentations. We showcase the usefulness of the multi-modal annotations by training a Video Narrative Grounding (VNG)~\cite{Voigtlaender23CVPR} model using our datasets, resulting in significant improvements on two VNG benchmarks.

In summary, we make the following contributions: 
\textbf{(1)} We propose the new \pointvos{} task for point-guided VOS, that includes weakly supervised training on spatio-temporally sparse point annotations.
\textbf{(2)} We propose a novel and efficient semi-automatic annotation pipeline for \pointvos{} that substantially reduces the annotation effort for creating human-validated ground-truth video annotations.
\textbf{(3)} We demonstrate the value of our proposed annotation pipeline by annotating and releasing two large video datasets, \pvoops{} and \pvkinetics{}. By design, those datasets feature multi-modal vision-language annotations that connect open-vocabulary language object descriptions to the corresponding video object annotations.
\textbf{(4)} We establish a new benchmark based on these datasets, where we train and test VOS methods either on point annotations or on pseudo masks, and present strong baselines.
\textbf{(5)} We realize the potential of multi-modal vision-language annotations in our proposed datasets and showcase their use for language-guided VOS.
\section{Related Work}
\begin{table}[t]
\renewcommand{\arraystretch}{1.25}
\center
\scriptsize{
\tabcolsep=0.10cm
\begin{tabular}{lcccccc}
\toprule 
Dataset & Videos & Objects & Annotations &\makecell[c]{Positive \\ Points} & \makecell[c]{Negative \\ Points}& \makecell[c]{Ambiguous \\ Points}\\
\midrule 
DAVIS’16~\cite{Perazzi2016} & 50 & 50 & 3.4K & - & - & - \\
DAVIS’17~\cite{Pont-Tuset_arXiv_2017} & 90  & 205 & 13.5K & - & - & - \\
YT-VOS~\cite{xu2018youtube}  & 4.4K & 7.7K & 197K & - & - & - \\
BURST~\cite{athar2023burst} & 2.9K & 16K & 600K & - & - & - \\
VISOR~\cite{darkhalil2022visor} & 7.8K & \dag & 271K & - & - & - \\
VOST~\cite{tokmakov2023breaking} & 713 & \dag &  175K & - & - & - \\
MOSE~\cite{ding2023mose} & 2.1K & 5.2K & 431K & - & - & - \\
\midrule
PV-Oops & 8.4K & 13.1K & 93K & 548K & 1.2M & 18K \\
PV-Kinetics & 23.9K & 120K & 965K & 5.2M & 12.6M & 253K \\
\bottomrule
\end{tabular}
\vspace{-0.3cm}
\caption{\textbf{Comparison of VOS datasets with ours.} Our \pointvos{} data is much larger compared to existing VOS datasets.
``Annotations'' counts for each object in how many frames it is annotated.
\dag: The number of objects is not reported.}
\label{tab:dataset-statistics}
}
\end{table}

\PAR{Video Object Segmentation Datasets.} 
DAVIS~\cite{Perazzi2016, Pont-Tuset_arXiv_2017} is one of the first densely annotated VOS datasets, with 90 videos. Later, the YouTube-VOS (YT-VOS) dataset~\cite{xu2018youtube} with $4.4K$ videos further advanced the state-of-the-art. Later datasets~\cite{qi2022occluded,athar2023burst,darkhalil2022visor,tokmakov2023breaking,ding2023mose} focused on specific VOS sub-challenges; among them, VISOR~\cite{darkhalil2022visor} is the largest in terms of the number of videos with $7.8K$ kitchen videos. Despite the growing interest in the VOS task, VOS datasets are still small in scale mainly due to their expensive annotation process. In contrast, we introduce a much more efficient point-wise annotation scheme that enables us to annotate about $32K$ videos, 4 times more than VISOR.

\PAR{Fully-Supervised VOS Methods.} 
Early VOS methods~\cite{Cae+17,voigtlaender17DAVIS,voigtlaender2017online, Luiten2018PReMVOSPR,Sun_2020} use online learning at test time which makes them very slow.
The following methods~\cite{Yang2019vis,oh2018fast,voigtlaender2019feelvos} alleviate this by propagating predictions frame-by-frame, using limited context and accumulating errors on the way.
Recent methods~\cite{oh2019video,cheng2021stcn,cheng2022xmem} address this by incorporating a larger temporal context using an external memory (\eg, STM~\cite{oh2019video} and its extensions \cite{seong2020kernelized,cheng2021modular,cheng2021stcn,cheng2022xmem}).
More recent works~\cite{yang2021aot,athar2023tarvis,athar2022hodor} use Transformers with spatio-temporal attention.
All of these methods rely on dense segmentation masks. 
In contrast, we use much cheaper point annotations in our work.

\definecolor{tablegreen}{RGB}{67, 255, 100}
\definecolor{tablered}{RGB}{255, 0, 0}
\definecolor{tableyellow}{RGB}{255, 251, 115}
\begin{table}[t]
\renewcommand{\arraystretch}{1.5}
\center
\footnotesize{
\tabcolsep=0.10cm
 \begin{tabular}{l  c  c  c c} 
 \toprule
 {} & VOS & PET & TAP-Vid & \pointvos{} \\
 \midrule
 Point annotations for training & \cellcolor{tablered!50}\xmark & \cellcolor{tablered!50}\xmark & \cellcolor{tablegreen!50}\cmark &  \cellcolor{tablegreen!50}\cmark\\ 
 Point annotations for test-time &\cellcolor{tablered!50}\xmark & \cellcolor{tablegreen!50}\cmark & \cellcolor{tablegreen!50}\cmark & \cellcolor{tablegreen!50}\cmark \\
 Arbitrary point location on object & \cellcolor{tablered!50}\xmark & \cellcolor{tablegreen!50}\cmark & \cellcolor{tablered!50}\xmark & \cellcolor{tablegreen!50}\cmark\\ 
 Temporally sparse annotations & \cellcolor{tablered!50}\xmark & \cellcolor{tablered!50}\xmark & \cellcolor{tablered!50}\xmark & \cellcolor{tablegreen!50}\cmark \\ 
 Simple, fast \& efficient annotations & \cellcolor{tablered!50}\xmark & \cellcolor{tablered!50}\xmark & \cellcolor{tablered!50}\xmark & \cellcolor{tablegreen!50}\cmark \\
 Multi-modal annotations & 
 \cellcolor{tableyellow!50}(\cmark)& \cellcolor{tablered!50}\xmark & \cellcolor{tablered!50}\xmark & \cellcolor{tablegreen!50}\cmark \\
 \bottomrule
 \end{tabular}
\vspace{-0.3cm}
\caption{\textbf{Comparison of the design decisions of our new spatially-temporally sparse point annotation scheme in \pointvos{} with the annotation schemes in other tasks}: VOS~\cite{Perazzi2016}, Point Exemplar-guided Tracking (PET)~\cite{athar2023burst} and  Tracking Any Point in a
Video (TAP-Vid)~\cite{doersch2022tap}. (\cmark): Multi-modal annotations are only available for some extensions of VOS datasets~\cite{khoreva2019video,seo2020urvos,Ding_2023} that initialize by a referring expression.} 
\label{tab:annotation-pipeline-goals}
}
\end{table}

\PAR{Weakly-Supervised VOS Methods.} 
Weakly supervised VOS approaches can be mainly classified into two types: (i) The first type aims at reducing the first-frame supervision at test time by either using points~\cite{athar2023burst} or bounding boxes~\cite{zhao2021generating,voigtlaender2020siam,wang2019fast,sun2020fast, Lin_Xie_Li_Zhang_2021} instead of dense masks while still relying on dense annotations for training. (ii) The second type~\cite{athar2022hodor,khoreva2017cvmprw,Voigtlaender21WACV} reduces the training supervision by exclusively using image-level datasets or by using bounding boxes instead of masks, while still relying on dense masks for the reference at test time. Unlike these methods, we use weak point supervision both at training and test times.
\PAR{Point Supervision for Images.}
Point supervision during training has already been explored for various image tasks such as instance segmentation~\cite{benenson2022colouring, cheng2022pointly, mahadevan18bmvc,SofiiukArxiv21,Sofiiukf20CVPR,Chen22CVPR,kirillov2023sam,RanaMahadevan23Arxiv}, action recognition~\cite{mettes2018ijcv,mettes2016eccv} and object counting~\cite{laradji2018eccv}. 
~\cite{cheng2022pointly} annotate points on objects and show that instance segmentation methods can be effectively trained on them. ~\cite{benenson2022colouring} use an efficient mobile friendly point annotation scheme to collect a new image dataset. However, unlike our method, these methods work only with images and sometimes require object-level bounding boxes~\cite{cheng2022pointly} for annotating points. 
\PAR{Point Supervision for Videos.}
Existing video-level point based annotations~\cite{doersch2022tap,zheng2023odyssey} are mainly used for point tracking, which requires point correspondences that are again expensive to obtain and not relevant for tasks such as VOS. In contrast, we focus on annotating \textit{random} points, \ie, we want points that are on a certain object, but we do not require points in different frames to correspond to the exact same part of the object. To the best of our knowledge,~\cite{athar2023burst} is the only previous attempt that utilizes point annotations for VOS-related tasks. However, different from our work,~\cite{athar2023burst} makes use of point annotations only to initialize the first frame at test time while keeping the training supervision unchanged.

\PAR{Video Annotations.}
We also compare the design decisions of the point annotation scheme in \pointvos{} with existing video annotation schemes in \cref{tab:annotation-pipeline-goals}.
Conventional VOS uses dense masks that are expensive to annotate.
PET~\cite{athar2023burst} uses point initialization at test-time but still dense masks for training.
TAP-Vid~\cite{doersch2022tap} uses exact point correspondences for training that are extremely costly to annotate.
In contrast, \pointvos{} expects that only random points on objects are annotated and the annotation effort is even further reduced by making use of temporal sparsity.

\section{Point Annotations for VOS}
\begin{figure}
    \centering
    \includegraphics[width=0.9\linewidth]{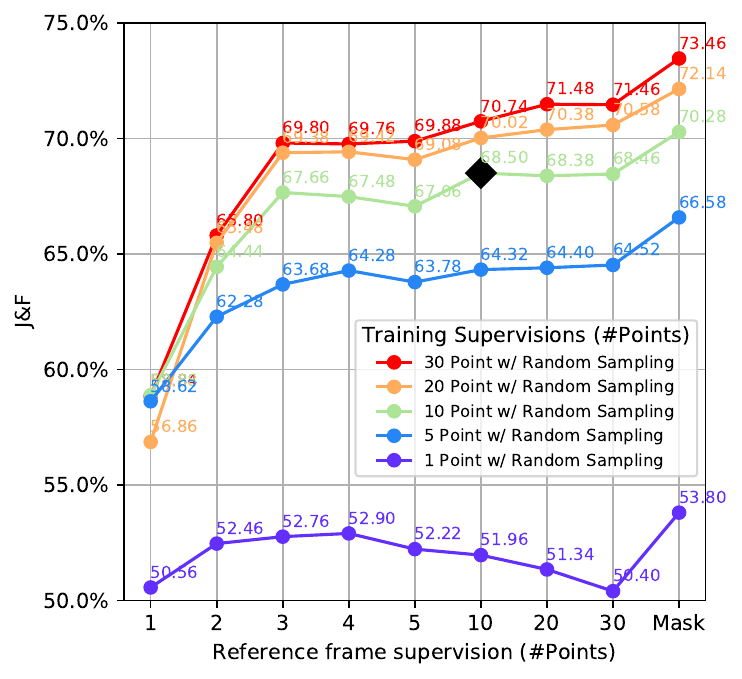}
    \vspace{-0.5cm}
    \caption{\textbf{Training \vs test-time point supervision results using simulated points on the DAVIS validation set.} \textcolor{black}{$\blacklozenge$} represents our chosen setting, i.e. 10 points for training supervision and 10 points for test-time supervision. We run each experiment 5 times and report the mean score.}
    \label{fig:point-sampling}
\end{figure}

In the conventional VOS task, the training set contains $N$ videos $V\scriptstyle=\left\{v_1, v_2,\dots, v_N\right\}$, where each video $v\in V$ with $T_v$ frames and $O_v$ objects consist of a set of images $I_v\scriptstyle=\left\{I_1,I_2,\dots,I_{T_v}\right\}$ and a set of dense segmentation masks $M_v\scriptstyle=\left\{m_{t,o}|t \in \{1,\dots,T_v\}, o \in \{1,\dots,O_v\}\right\}$.
At test-time, the input is a video $v \notin V$ and the corresponding reference segmentation masks $M^R\scriptstyle=\{m^R_1,m^R_2,\dots,m^R_{O^R}\}$ for $O^R$ objects in a single frame (usually the first).
The conventional VOS method then has to generate temporally consistent segmentation masks $\hat{M}_v\scriptstyle=\left\{\hat{m}_{t,o}|t \in \{1,\dots,T_v\}, o \in \{1,\dots,O^R\}\right\}$ for the $O^R$ foreground objects for each frame of the video.

In our proposed \pointvos{} task, we update this task by replacing the training masks $M$ and reference masks $M^R$ with point annotations, and by working with a sparse set of frames. Hence, in a \pointvos{} training set, each video $v\in V$ has a set of images $I_v\scriptstyle=\left\{I_1,I_2,\dots,I_{T_v}\right\}$ and a set of point annotations $P_v\scriptstyle=\left\{P_{t,o}| t \in T^{sparse}_v, o \in \{1,\dots,O_v\}\right\}$ with $T^{sparse}_v\scriptstyle\subset\left\{1,\dots,T_v\right\}$, $|T^{sparse}_v|  \ll T_v$,
where each $P_{t,o}$ is a set of points for object $o$ in frame $t$. 
At test-time, a~\pointvos{} method is initialized with reference points $P^R\scriptstyle=\left\{P^R_1, P^R_2,\dots,P^R_{O^R}\right\}$ on $O^R$ objects in the reference frame. The expected output $\hat{M}_v$ is the same as for the original VOS task, \ie, a predicted segmentation mask for each frame of each object.

\subsection{Simulating Point Annotations}
\label{subsec:point_simulation}
To study the effect of training and initializing with points rather than masks, we perform a series of experiments where we train jointly on the DAVIS and YT-VOS training sets and evaluate on the DAVIS validation set.

First, we analyze the number of points required for training supervision and for test-time initialization. For this, we sample points randomly from each of the available ground-truth masks in every annotated video frame such that these points are at least $d=20$ pixels apart from each other. When the required number of points under the distance constraint cannot be sampled, \eg, when the ground-truth masks are very small, we retain the maximum possible number of points under the constraint. 

\definecolor{figgreen}{RGB}{173, 228, 152}  
\definecolor{figpink}{RGB}{231, 97, 97}

We then use an STCN~\cite{cheng2021stcn} version that is adapted to work with points (see~\cref{sec:benchmark}), and train multiple models with varying numbers of points used for training supervision. We evaluate each model on the DAVIS validation set with different numbers of points for initialization on the reference frame. \cref{fig:point-sampling} shows 
that training with 10 points (\textcolor{figgreen}{\textbf{---}}) achieves good results and adding more points during training only gives minor gains. When we have less than 10 points during training, the performance strongly degrades.
For inference, we find that more points on the reference frame lead to better results. 
However, this increases the test-time annotation burden and is a trade-off which is highly application dependent. As a result, we propose to evaluate different numbers of points (up to 10) on the reference frame (\textcolor{black}{$\blacklozenge$}). We also tried other sampling strategies such as farthest-point sampling, but random sampling gave better results (see supplement for details).

\begin{figure}
    \centering
    \includegraphics[width=0.92\linewidth]{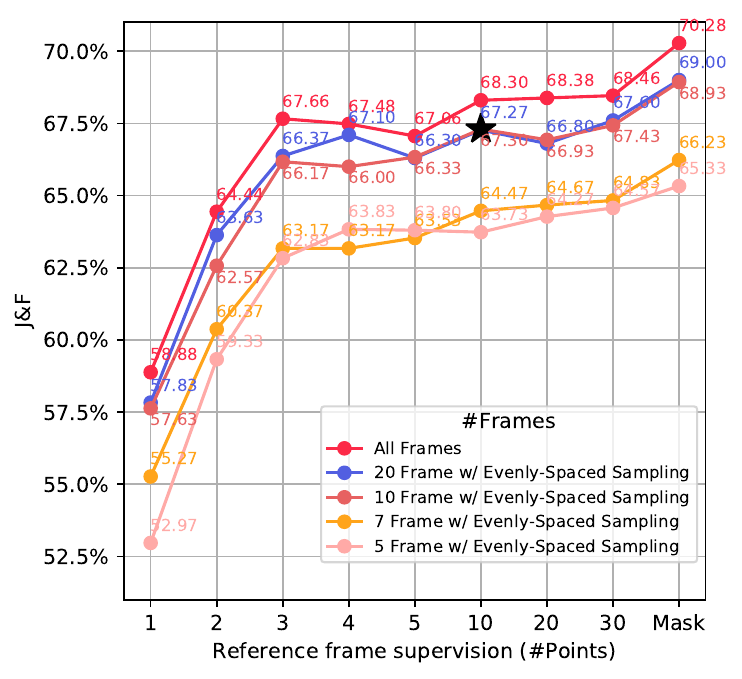}
    \vspace{-0.5cm}
    \caption{\textbf{STCN results on DAVIS validation set for varying temporal sparsity, when trained on 10 randomly sampled points per frame per object.} \textcolor{black}{$\bigstar$} represents our chosen setting, \ie 10 points for training supervision and 10 points for test-time supervision, on 10 frames. We run each experiment 3 times and report the mean score.}
    \label{fig:temp_evenly_spaced_10_point}
\end{figure}

Next, we analyze the number of frames required for training on randomly sampled points 10 per frame per object. Starting from all frames, we sub-sample (evenly-spaced) up to 20 frames for each video.
\cref{fig:temp_evenly_spaced_10_point} shows the results of STCN trained on such a temporally sparse training set.
We find that the performance of STCN saturates at 10 frames (\textcolor{figpink}{\textbf{---}}), where increasing it further does not yield any noticeable performance improvements, and having less than 10 frames deteriorates the performance. Here again, we try different frame sampling techniques, such as random sampling, but do not observe any performance difference (details in the supplement). In summary, we find 10 points per object on 10 frames to be a good setting (\textcolor{black}{$\bigstar$}).

\subsection{Semi-Automatic Annotation Scheme}
\label{subsec:annotation-pipeline}

\begin{figure*}
    \centering
    \includegraphics[width=0.99\linewidth]{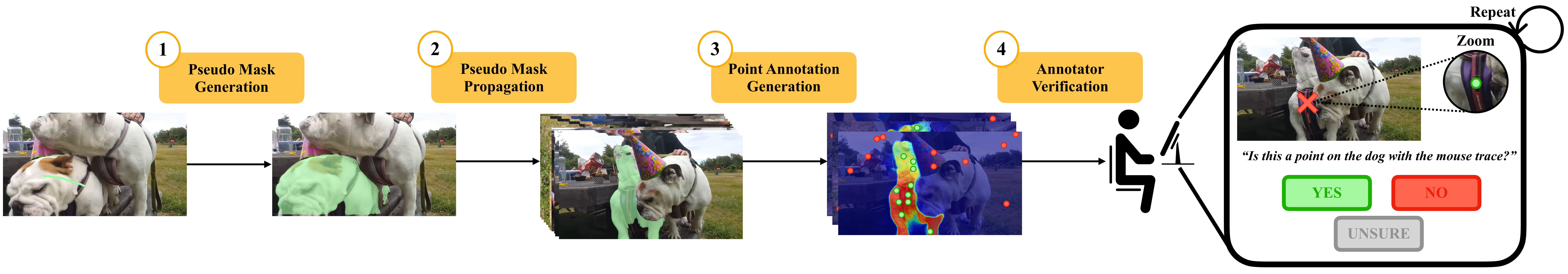}
    \vspace{-0.3cm}
    \caption{\textbf{Semi-automatic annotation pipeline used to annotate~\vidln{} data.} We first extract a mouse trace segment for each noun in~\vidln{} captions, and convert it into a pseudo mask using~\dynamite{}. We then use STCN to propagate the pseudo-mask across the video. We then use the STCN output probability maps to sample sparse point annotations and let annotators verify them. \textcolor{green}{Green} circles represent foreground points and \textcolor{red}{red} circles background points.}
    \label{fig:annotation_pipeline}
\end{figure*}

We design a very efficient semi-automatic point annotation pipeline to annotate videos with points (\cref{fig:annotation_pipeline}). Instead of annotating points in multiple frames manually from scratch, we aim to generate point candidates automatically that then only need to be quickly verified by human annotators. 

\PAR{Pseudo-mask Generation.}
To annotate an object, as a starting point we require only a rough localization of it (\eg, by a few points) in a single frame of the video.
We then convert this rough localization into a pseudo-mask using the interactive segmentation method~\dynamite{}~\cite{RanaMahadevan23Arxiv}.  called~\dynamite{}~\cite{RanaMahadevan23Arxiv}. 

\PAR{Pseudo-mask Propagation.}
We feed the pseudo-mask from the previous step into \stcn{}~\cite{cheng2021stcn} to propagate it both forward and backward in time to all other frames of the video and obtain pixel-wise binary probability maps $\mathcal{P}\scriptstyle=\left\{p_1,p_2,...,p_T\right\}$ for all frames $1,\dots,T$ of the video.

\PAR{Point Sampling.}
Because both~\dynamite{} and~\stcn{} can introduce errors in the previous process, we do not use the resulting pseudo-masks directly, but instead, we use the~\stcn{} output probability maps to sample points and let human annotators verify them.
We sub-sample the probability maps temporally equally-spaced to 10 frames. Then, for each object and each retained frame, we threshold the probability map into (i) a high probability region that likely represents the foreground object, (ii) a low probability region that likely represents the background, and (iii) an uncertain region.
Points from the uncertain region are hard for STCN and hence might provide a valuable learning signal after being manually annotated.
For each of the 10 frames, we then randomly sample $10$ potential background points from the low-probability region, $7$ potential foreground points from the high-probability region, and $3$ ambiguous points from the uncertain region.
Here, we again ensure that each of these points are at least $d$ distance apart from each other, and we obtain in total up to 200 points per object. 

\PAR{Annotator Verification.} 
We show the annotators the rough localization information used to generate the points, overlaid on the image, so that they understand which object should be considered.
Next, we show them the foreground point candidates one by one overlaid on the image (see \cref{fig:annotation_pipeline}, right). They use a hotkey to either accept or reject this foreground point candidate or to indicate it is ambiguous.
We repeat the same procedure with the background point candidates, and finally with the points with high uncertainty.
By batching points of the same type (\eg, foreground candidates) together, the annotators can very quickly verify them.

\subsection{\pointvos{} Datasets}

For our annotations, we choose two large datasets from  Video Localized Narratives (\vidln{}~\cite{Voigtlaender23CVPR}). \vidln{} provides annotations, where annotators speak to provide a caption for the video, and while speaking, they move their mouse pointer over the object they refer to in multiple key-frames to provide a rough localization. Leveraging \vidln{} annotations has primarily two advantages for us: (i) the mouse traces can be used to automatically select foreground objects in a video, and correspondingly give us a free rough localization as starting point for our annotation scheme; and (ii) the associated text description can be used to develop multi-modal VOS algorithms.
We build on the ``location-output question'' annotations from Oops~\cite{Epstein_2020_CVPR} because they provide a set of mouse traces for nouns that are already verified to have good quality. Additionally, we choose Kinetics~\cite{kay2017kinetics}, because it is by far the largest VidLN dataset.

To convert the continuous mouse traces into segments, we first use the NLP toolkit spaCy to find nouns in the VidLN captions, and then for each noun retrieve a rough localization by mouse trace segments $\mathcal{T}\scriptstyle=\left\{t_1,t_2,...,t_n\right\}$ on key-frames $F\scriptstyle=\left\{f_1,f_2,...,f_n\right\}$ provided by \vidln{}~\cite{Voigtlaender23CVPR}. We extract the mouse trace segment $t_k$ on the key frame $f_k$ on which it has the largest area.
Each noun is thus localized with a corresponding mouse trace segment on a single key-frame. 

\PAR{Instance Segmentation from Mouse Traces.} 
We adapt \dynamite{} to work with mouse traces instead of user clicks and hence our version of \dynamite{} takes a mouse trace segment $t_k$ and the corresponding frame $f_k$ as input, and generates a binary segmentation mask $m_k$ as output. More details can be found in the supplement.

\PAR{Point Verification.} On average, we get 147 points per object to be verified and, following this procedure, an annotator spends on average 140 seconds per object, \ie, 0.95 seconds per point. 
In contrast, annotating a single dense mask can take $\sim\!\!80s$~\cite{lin2014eccv}. If we consider annotating an object with a mask in each frame of a single video in the DAVIS training set with an average of 70 frames, it takes about 5600s, which is 40 times slower than our annotation scheme. This demonstrates that our annotation scheme achieves an extreme speedup and lets us annotate much larger datasets than existing VOS benchmarks. Moreover, \cite{doersch2022tap} report that annotating point correspondences over 250 frames for 10 objects with 3 points per object takes 3.3 hours, \ie, $1,\!188s$ per object. 
This is 8 times slower than our annotation scheme, which shows that our point annotations are also much faster than annotating point correspondences.

The statistics of our point annotations in \cref{tab:dataset-statistics} show that in total we annotated more than $19M$ points for $133K$ objects in $32K$ videos. Thus, we annotated significantly more videos and objects than the largest existing VOS datasets VISOR~\cite{darkhalil2022visor} and  BURST~\cite{athar2023burst}. Our annotations cover 4 times more videos than VISOR, consisting of $7.8K$ videos, and 8 times more objects than BURST with $16K$ objects.

\PAR{\pvoops{}.} 
Oops~\cite{Epstein_2020_CVPR} consists of fail videos of unintentional action, often filmed by amateurs in diverse environments. They contain a lot of camera jitter and motion blur, making tracking and segmentation challenging.
We create the~\pointvos{} Oops dataset (\pvoops{}), and annotate more than $13K$ objects in $8.4K$ videos that are split into a training and a validation set.
We also create a ~\pvoops{} benchmark to measure \pointvos{} performance on this challenging domain. For $991$ objects in the validation set, we annotate points in the first frame for initialization and dense pixel masks for up to 3 frames per object for evaluation. By conducting simulation experiments on DAVIS, we found that the results evaluated against either masks in only 3 frames, or in all frames, correlate extremely well, meaning that 3 frames are sufficient for evaluation purposes (see supplement for details).

\begin{figure*}[t]
    \centering
    \setlength{\fboxsep}{0.24pt}
    \begin{subfigure}[b]{0.24\linewidth}
    \includegraphics[width=\mysize\linewidth]{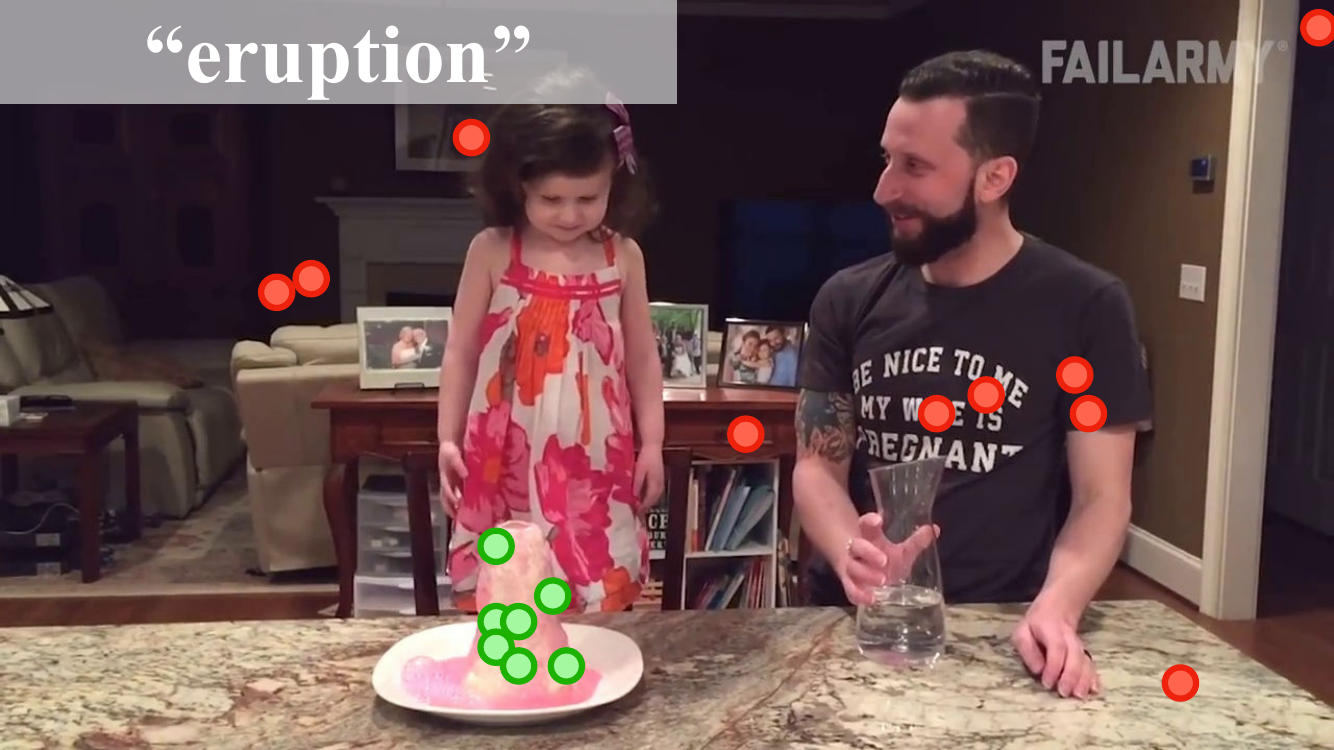}
    \end{subfigure}
    \begin{subfigure}[b]{0.24\linewidth}
        \includegraphics[width=\mysize\linewidth]{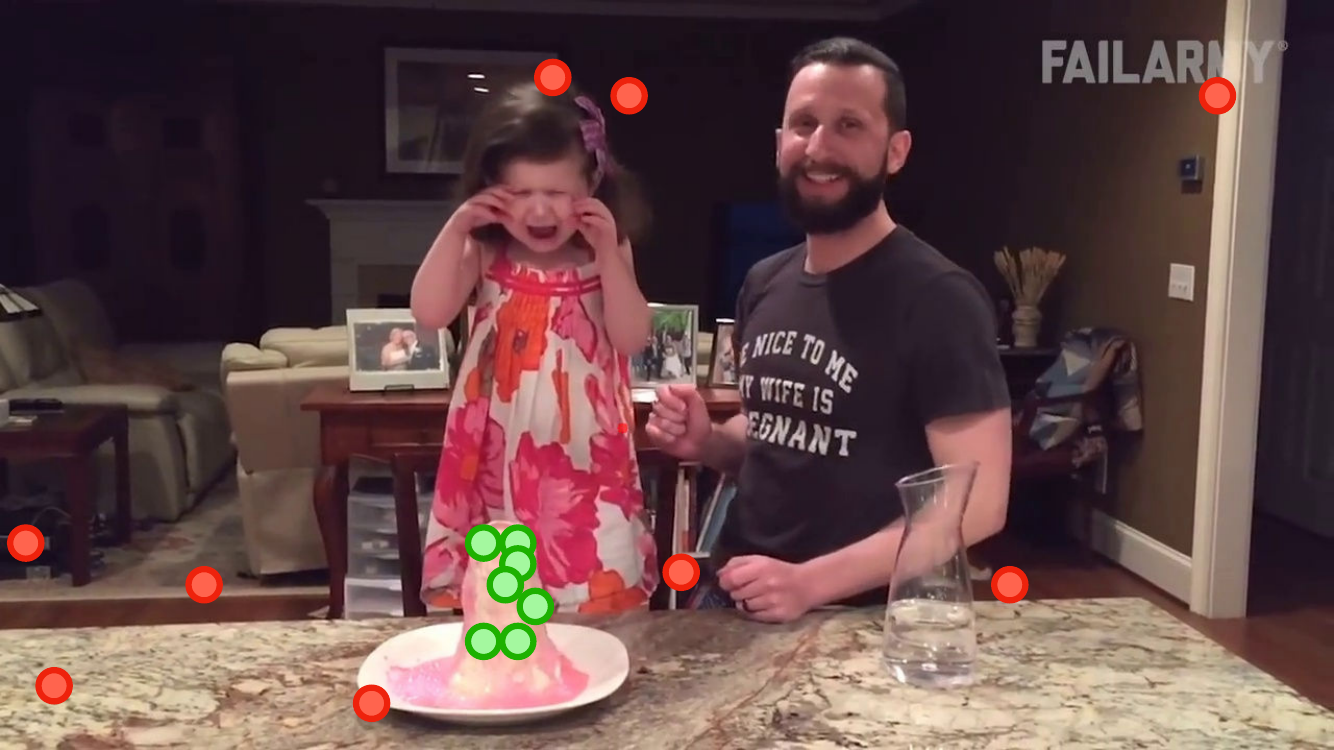}
    \end{subfigure}
    \begin{subfigure}[b]{0.24\linewidth}
        \includegraphics[width=\mysize\linewidth]{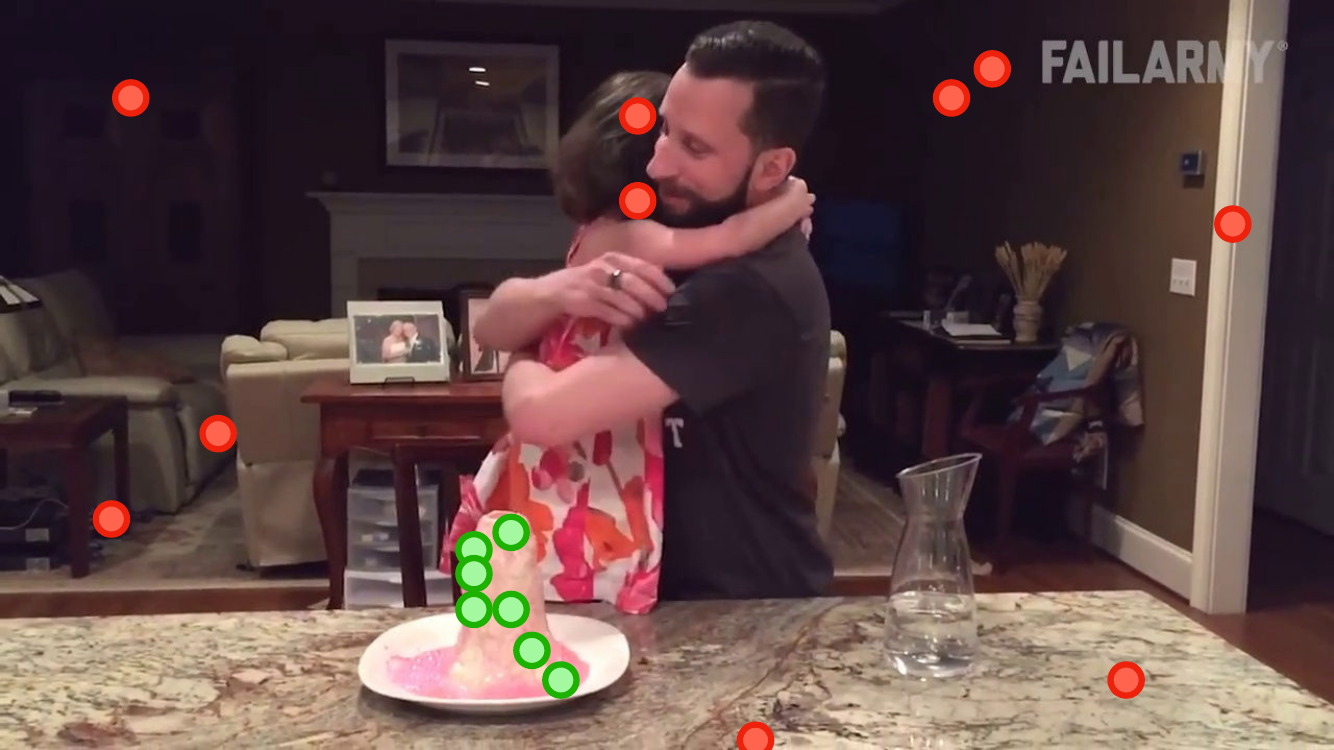}
    \end{subfigure} 
    \begin{subfigure}[b]{0.24\linewidth}
        \includegraphics[width=\mysize\linewidth]{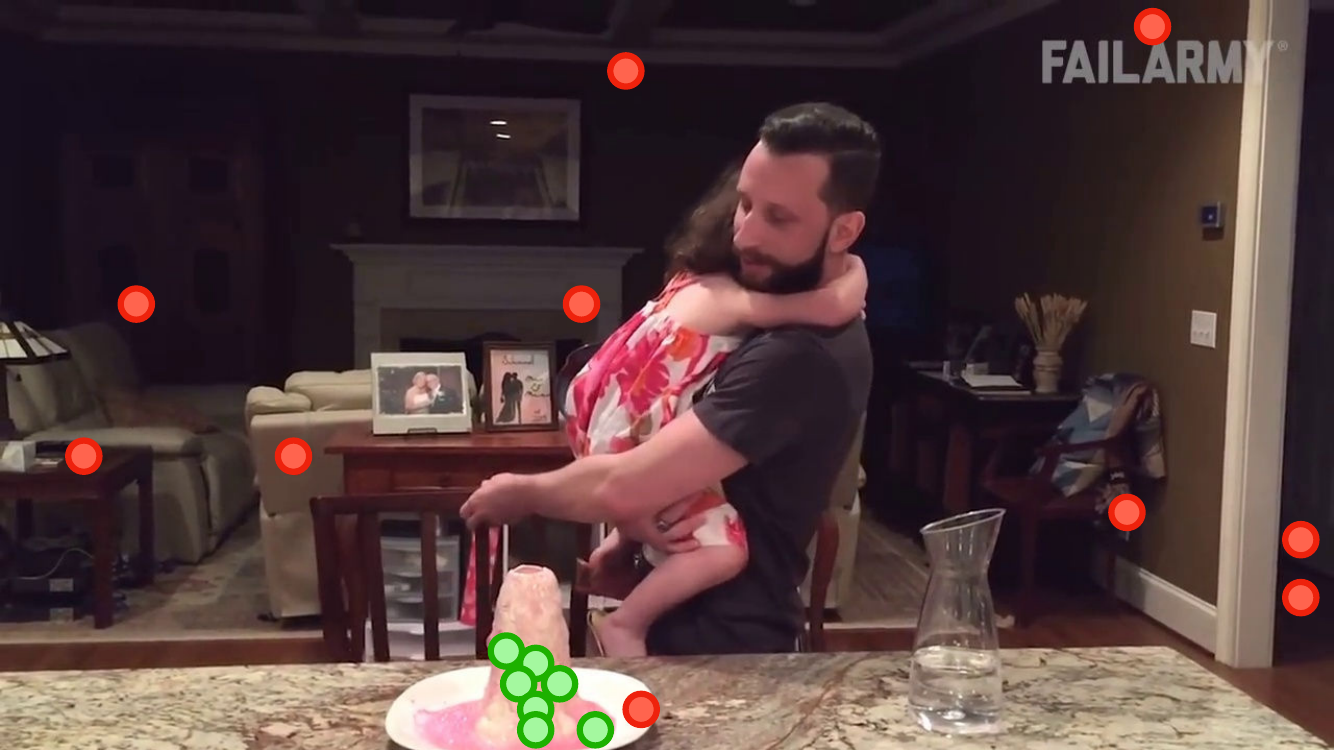}
    \end{subfigure} 

    \begin{subfigure}[b]{0.24\linewidth}
        \includegraphics[width=\mysize\linewidth]{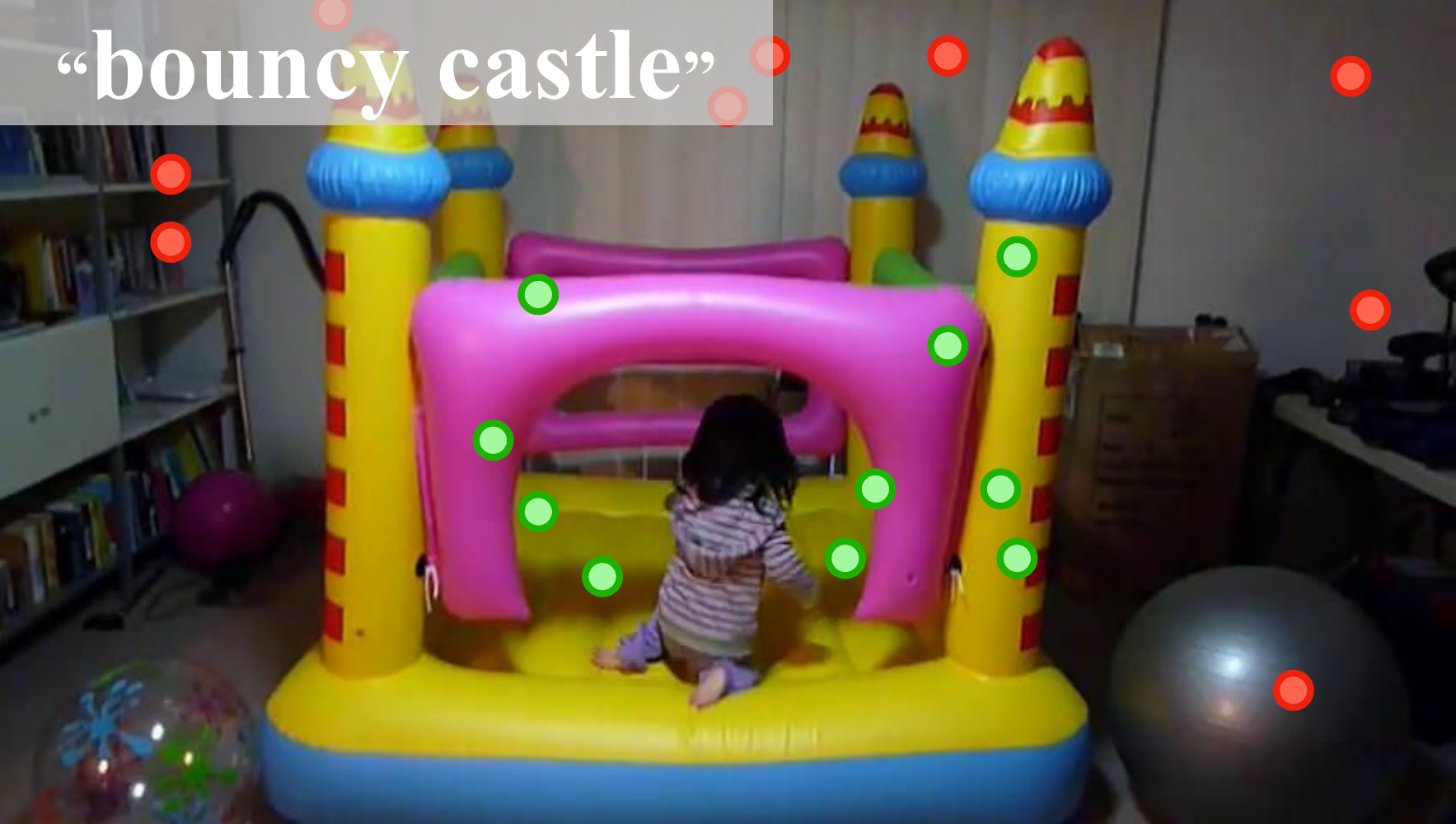}
    \end{subfigure}
    \begin{subfigure}[b]{0.24\linewidth}
        \includegraphics[width=\mysize\linewidth]{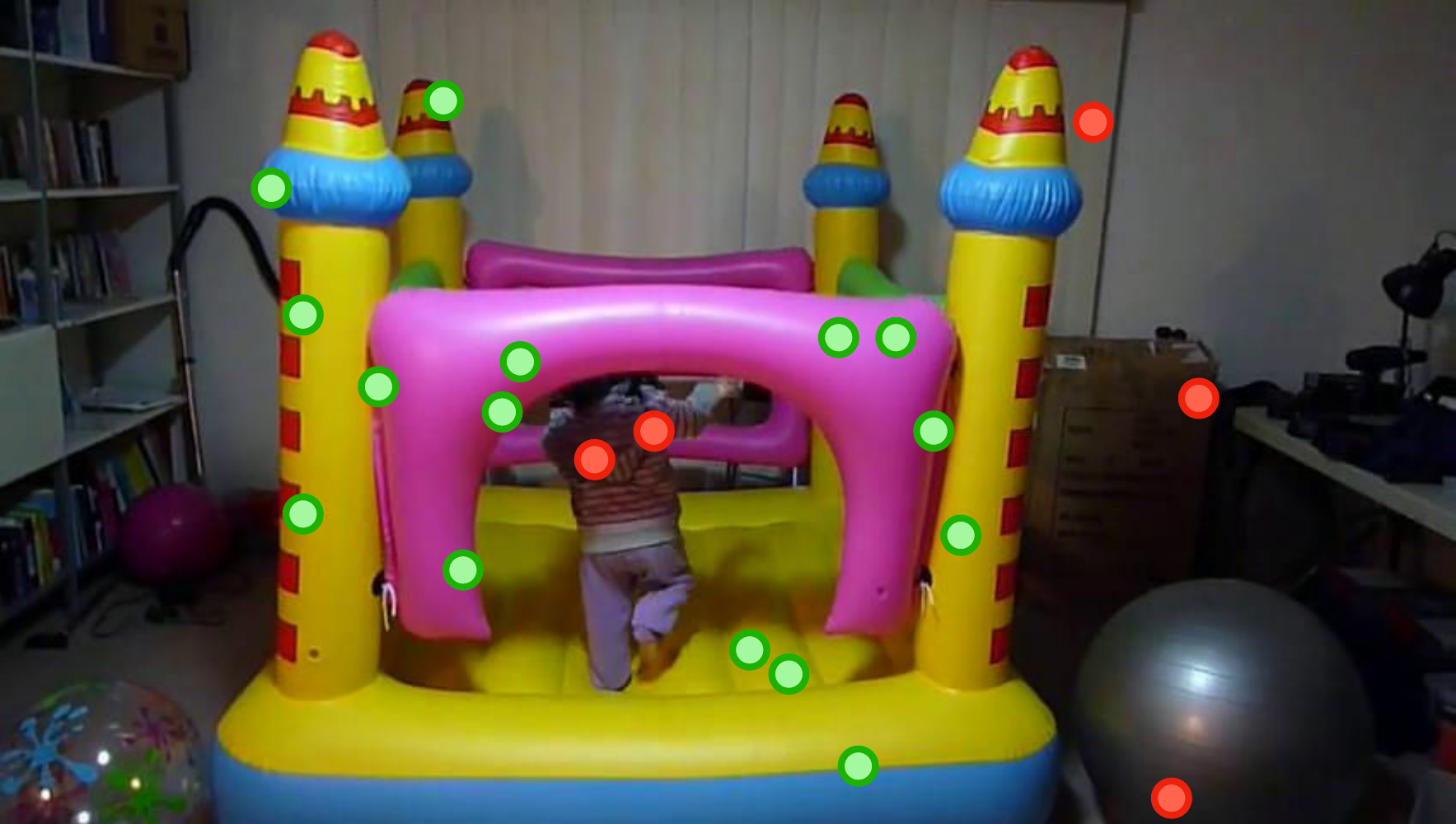}
    \end{subfigure}
    \begin{subfigure}[b]{0.24\linewidth}
     \includegraphics[width=\mysize\linewidth]{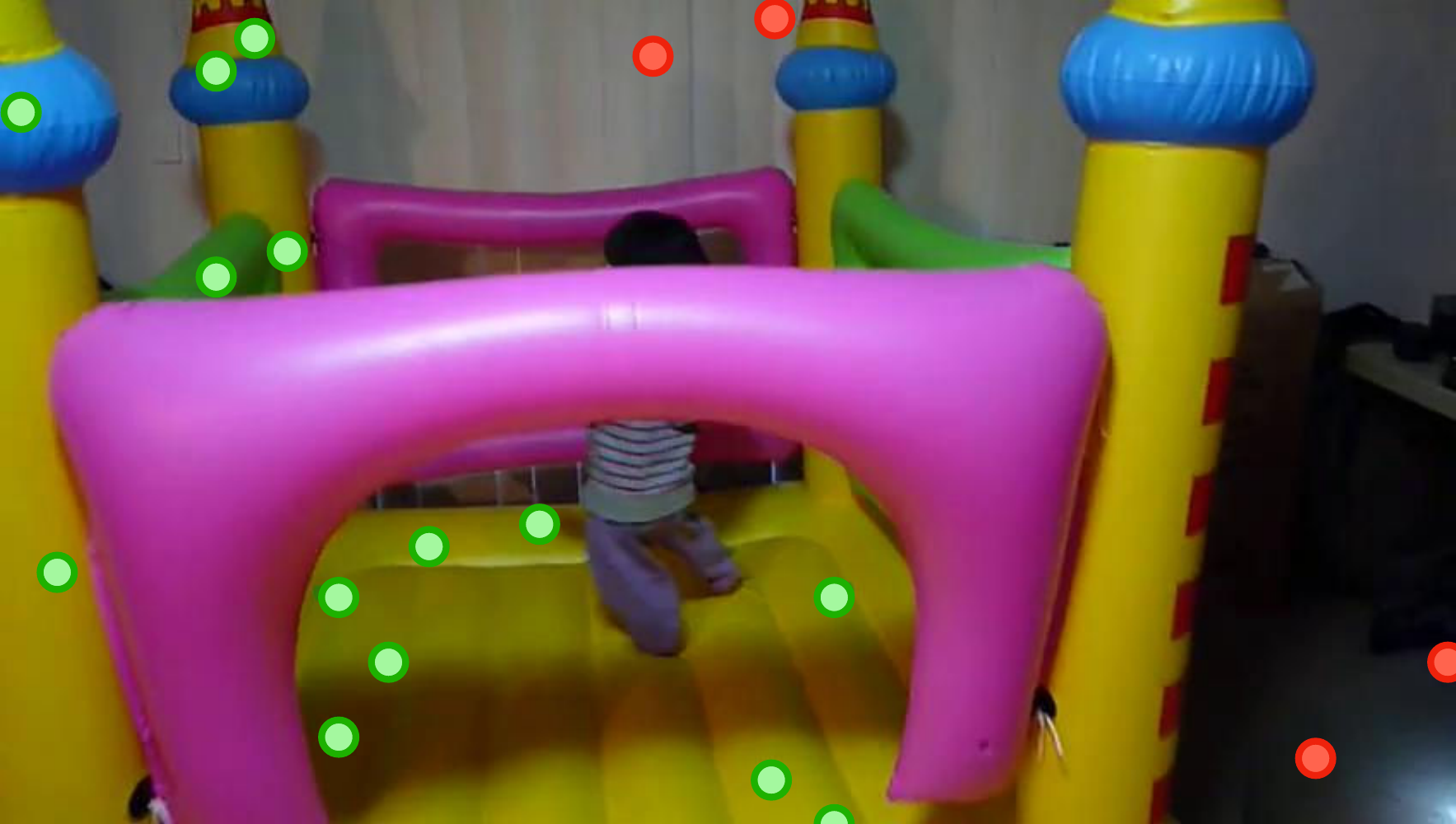}
    \end{subfigure} 
    \begin{subfigure}[b]{0.24\linewidth}
     \includegraphics[width=\mysize\linewidth]{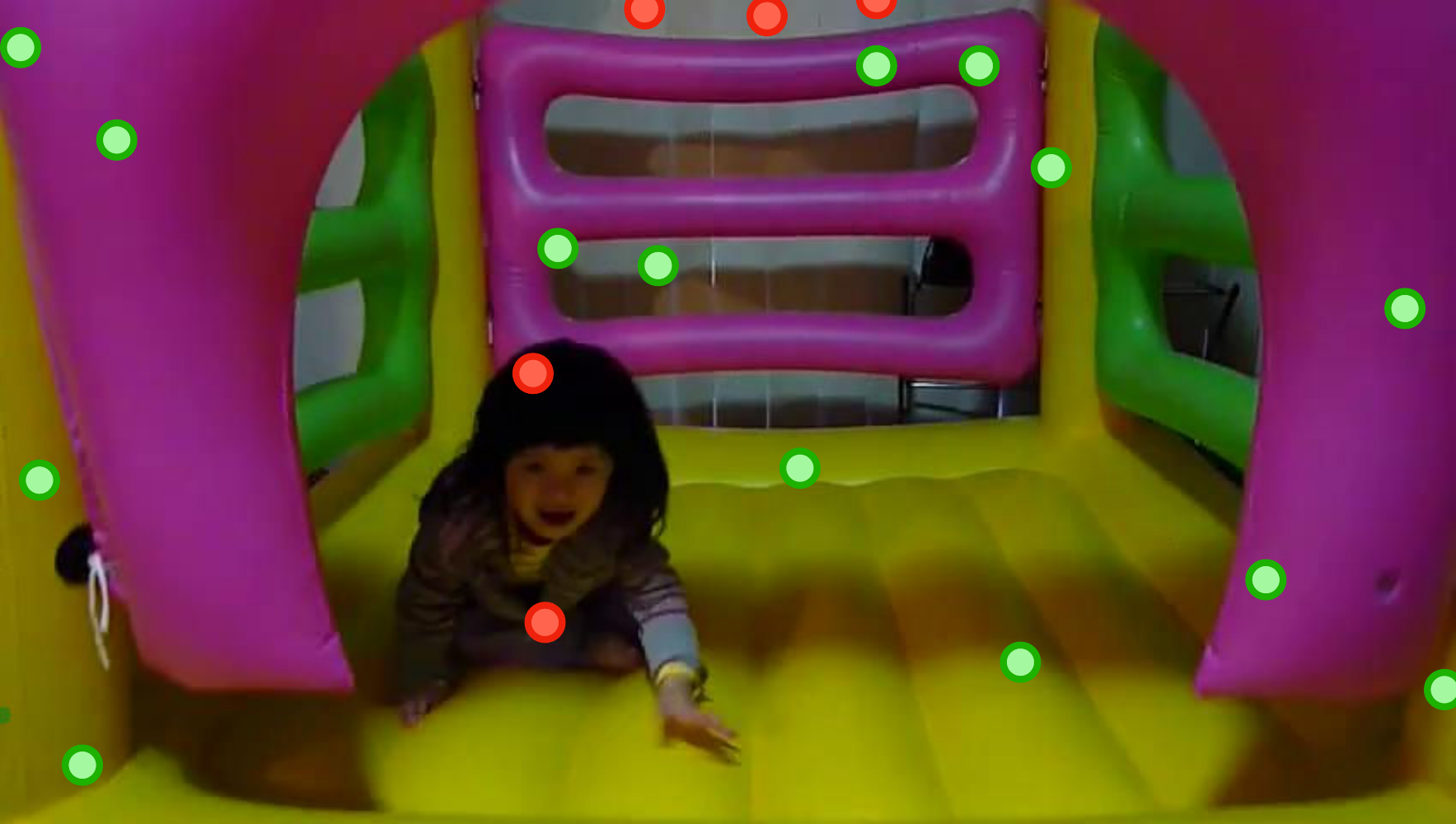}
    \end{subfigure} 
    \caption{\textbf{Example point annotations for \pvoops{} (top) and \pvkinetics{} (bottom)}. The objects are connected to nouns from a large vocabulary. \textcolor{green}{Green} dots represent foreground points and \textcolor{red}{red} dots background points.}
    \label{fig:example_point_annotations}
\end{figure*}

\PAR{\pvkinetics{}.} Kinetics~\cite{kay2017kinetics} is a very large-scale action recognition dataset with $650K$ videos that cover 700 action classes. The Kinetics videos are approximately 10s long and are annotated with action labels. Similar to Oops, for the~\pointvos{} Kinetics (\pvkinetics{}) dataset, we use the subset of
videos with \vidln{} annotations and annotate $120K$ objects with points across $23.9K$ videos. 

With our new annotations, we obtain the largest VOS-related dataset in terms of the number of videos that cover a wide range of human actions. 
\cref{fig:example_point_annotations} shows some example point annotations from \pvoops{} and \pvkinetics{}. More detailed statistics and more annotation visualizations are available in the supplement.

\section{Experiments}
\label{sec:benchmark}
\subsection{\pointvos{} Benchmark}

We propose a new benchmark for the \pointvos{} task in order to evaluate what a method can achieve by using point annotations for training and testing. At test time, for each foreground object, we provide multiple sets of point initializations with varying degrees of sparsity (1, 2, 5, or 10 points) on the corresponding reference frame, and also report their mean scores. This reflects different trade-offs at test-time between annotation effort and result quality.

For training, we use point annotations from our annotated \pvoops{} and \pvkinetics{} datasets, in addition to \pointvos{} versions \pvdavis{} and \pvytvos{} of the popular VOS training sets, that we create by sub-sampling the object masks both spatially and temporally, as explained in~\cref{subsec:point_simulation}. The methods are then evaluated on the validation sets of \pvdavis{}, \pvytvos{}, and \pvoops{}. For~\pvdavis{} and~\pvoops{}, we use the popular $\JnF$ metric, and report the mean score for the different point initilizations. On the~\pvytvos{} validation set, consistent with the original task, we report the $\J$ and $\F$ scores for both seen and unseen classes, along with their overall mean $\mathcal{G}$.
Owing to the limited evaluations permitted by the YT-VOS evaluation server, we only consider initialization with 10 points. We compute all scores on dense ground truth masks.

\PAR{\pointstcn{} Baseline.}
As a first baseline, we adapt STCN to work with points both for training supervision and test-time initialization, and we call this adaptation~\pointstcn{}. \pointstcn{} makes minimal changes to the original STCN model, showing that existing VOS methods can be easily adapted to work with our~\pointvos{} datasets. 

\begin{table}[ht!]
\renewcommand{\arraystretch}{1.25}
\footnotesize{
\centering
\tabcolsep=0.02cm
\begin{tabularx}{\linewidth}{l c | Y | Y Y Y Y} 
\toprule
\makecell[l]{\textbf{Pre-training}} & {FT} &  
\makecell[c]{Mean} & 
\makecell[c]{1-point}& 
\makecell[c]{2-point} & 
\makecell[c]{5-point} & 
\makecell[c]{10-point} \\
\midrule
{\pvoops}  & \xmark & \textbf{59.8} & \textbf{48.6} & \textbf{57.8} & \textbf{65.5} & {67.7} \\
{\pvkinetics} & \xmark &  {50.4} & {29.5} & {41.5} & {60.7} & {69.7}\\
{\pvoops\ + \pvkinetics} & \xmark & {54.2} & {35.2} & {45.9} & {63.6} & \textbf{71.9}  \\
\midrule
{---} & \cmark & {61.3} & {49.4} & {60.8} & {67.2} & {67.7} \\
{\pvoops}  & \cmark & {62.8} & \textbf{50.6} & \textbf{62.4} & {67.7} & {70.4} \\
{\pvkinetics} & \cmark &  {62.8}  & {48.0} & {61.7} & \textbf{69.6} & {72.0}\\
{\pvoops\ + \pvkinetics} & \cmark & \textbf{63.1} & {48.4} & {61.4} & {69.5} & \textbf{72.9} \\
\bottomrule
\end{tabularx}
\caption{\textbf{\pvdavis\ benchmark results ($\JnF$) of~\pointstcn{}.} \texttt{FT}: fine-tuning on \pvdavis{} and \pvytvos{}.}
\label{tab:pointvos_davis}
}
\end{table}
\begin{table}[ht!]
\renewcommand{\arraystretch}{1.25}
\footnotesize{
\centering
\tabcolsep=0.02cm
\begin{tabularx}{\linewidth}{l c | Y | Y Y Y Y} 
\toprule
\makecell[l]{\textbf{Pre-training}} & {FT} & 
\makecell[c]{$\mathcal{G}$} & 
\makecell[c]{$\mathcal{J_S}$} & 
\makecell[c]{$\mathcal{F_S}$} & 
\makecell[c]{$\mathcal{J_U}$} & 
\makecell[c]{$\mathcal{F_U}$} \\
\midrule
{\pvoops}  & \xmark & {51.6} & \textbf{60.8} & \textbf{62.0} & {40.1} & {43.5} \\
{\pvkinetics} & \xmark & {49.6} & {48.2} & {50.6} & {46.3} & {53.4}\\
{\pvoops\ + \pvkinetics} & \xmark & \textbf{52.2} & {52.4} & {54.2} & \textbf{47.7} & \textbf{54.5} \\
\midrule
{---} & \cmark & {51.9} & {59.2} & {60.5} & {41.7} & {46.5} \\
{\pvoops}  & \cmark & {54.4} & {61.1} & {62.6} & {44.6} & {49.5} \\
{\pvkinetics} & \cmark & {56.6} & {61.5} & {64.4} & {46.9} & {53.6} \\
{\pvoops + \pvkinetics} & \cmark & \textbf{57.2} & \textbf{62.5} & \textbf{64.7} & \textbf{47.7} & \textbf{53.7} \\
\bottomrule
\end{tabularx}
\caption{\textbf{\pvytvos\ benchmark results of~\pointstcn{}} when initialized with 10-points. \texttt{FT}: fine-tuning on \pvdavis{} and \pvytvos{}.}
\label{tab:pointvos_ytvos}
}
\end{table}
\begin{table}[t]
\renewcommand{\arraystretch}{1.25}
\center
\footnotesize{
\tabcolsep=0.02cm
\begin{tabularx}{\linewidth}{ l | Y | Y Y Y Y} 
\toprule 
\makecell[l]{\textbf{Training}} & \makecell[c]{Mean} & \makecell[c]{1-point} & \makecell[c]{2-point} & \makecell[c]{5-point} & \makecell[c]{10-point}  \\
\midrule
{\pvdavis\ + \pvytvos{}} & {48.6} & {40.5} & {47.4} & {52.7} & {53.8} \\
{\pvkinetics} & {42.5} & {27.3} & {35.7} & {50.3} & {56.7} \\
{\pvoops} & \textbf{61.2} & \textbf{54.6} & \textbf{60.2} & \textbf{64.4} & \textbf{65.5}\\
\bottomrule
\end{tabularx}
\caption{\textbf{\pvoops\ benchmark results ($\JnF$) of~\pointstcn{}.}
}
\label{tab:point-training-oops-results}
}
\end{table}

The original STCN method is first pre-trained on synthetic video sequences created by augmenting static images, and then fine-tuned on the densely labelled DAVIS and YT-VOS video datasets. Additionally, STCN also uses a synthetic dataset called BL30K~\cite{cheng2021modular}, which we do not use in our work.
For \pointstcn{}, we also start from static image pre-training, and then directly train on our spatially-temporally sparse~\pointvos{} data. Here, we start by training on \pvdavis{} and \pvytvos{}, and then further explore the benefits of adding our new \pvoops{} and \pvkinetics{} data as an additional pre-training step. More details about the implementation of \pointstcn{} are in the supplement.

\cref{tab:pointvos_davis,tab:pointvos_ytvos} demonstrate that the newly annotated~\pvoops{} and~\pvkinetics{} data bring large improvements as compared to starting from static images, especially for the settings where we fine-tune these models on \pvytvos{} and \pvdavis{}. \Eg, the mean $\JnF$ on \pvdavis{} improves from 61.3\% to 63.1\%, and $\mathcal{G}$ on \pvytvos{} improves from 51.9\% to 57.2\% when pre-training with both \pvoops{} and \pvkinetics{}. This demonstrates that our new \pvoops{} and \pvkinetics{} annotations serve as good initilizations for adapting models to other target domains. For results without fine-tuning, we observe that training on~\pvkinetics{} improves the scores only when sufficiently many points ($>\!5$) are available as test-time initialization. This could be attributed to the domain mismatch between Kinetics and YT-VOS/DAVIS, hence requiring more test-time information.  

\cref{tab:point-training-oops-results} shows the results on \pvoops{}. Using our~\pvoops{} annotations improve the performance from 48.6\% to 61.2\% $\JnF$ as compared to~\pointstcn{} trained on just~\pvdavis{} and~\pvytvos{}. This shows that using points, VOS models can be adapted to target domains with a strongly reduced annotation effort.

\PAR{Pseudo-mask Baseline.}
As an alternative to training on points directly, we consider using the points to first generate pseudo-masks and then train STCN on those pseudo-masks. For this, we use the same training procedure that was used to train original STCN, but replace the bootstrapped cross-entropy loss with a more robust \textit{Huberised cross-entropy loss}~\cite{Menon2020iclr,Voigtlaender21WACV} to reduce the influence of errors in the pseudo-masks.  At test-time, we first generate pseudo-masks from the different point initializations and then use these masks as reference. The pseudo-masks provide much more useful information than points, but require an additional model at test-time and increase the run-time.

\definecolor{Gray}{gray}{0.9}
\definecolor{LightCyan}{rgb}{0.88,1,1}
\begin{table}[ht!]
\renewcommand{\arraystretch}{1.25}
\footnotesize{
\centering
\tabcolsep=0.03cm
\begin{tabularx}{\linewidth}{l c | Y | Y Y c Y} 
\toprule
\makecell[l]{\textbf{Pre-training}} & {FT} &
\makecell[c]{{Mean}} & 
\makecell[c]{{1-point}} & 
\makecell[c]{{2-point}} & 
\makecell[c]{{5-point}} &
\makecell[c]{{10-point}} \\
\midrule
{---} & \xmark & {65.6} & {55.2} & {67.4} & {69.5} & {70.4} \\
{\pvoops}  & \xmark & {67.2} & {59.0} & {69.8} & {69.1} & {70.9}  \\
{\pvkinetics} & \xmark & {68.9} & {59.9} & {71.1} & {71.6} & {73.0} \\
{\pvoops\ + \pvkinetics} & \xmark & \textbf{70.4} & \textbf{61.1} & \textbf{72.5} & \textbf{73.2} & \textbf{74.8} \\
\midrule
{---} & \cmark & {70.3} & {61.6} & {72.0} & {72.7} & {75.0} \\
{\pvoops}  & \cmark & {70.6} & {61.0} & {72.4} & {73.1} & {75.8} \\
{\pvkinetics} & \cmark & {71.0} & {62.4} & {72.9} & {73.6} & {75.0} \\
{\pvoops\ + \pvkinetics} & \cmark & \textbf{71.6} & \textbf{63.0} & \textbf{73.4} & 74.4 & {75.8}  \\
\midrule
{\pvoops\ + \pvkinetics{} $^*$} & \cmark & {67.4} & {44.8} & {69.1} & \textbf{77.0} & \textbf{78.8} \\
\bottomrule
\end{tabularx}
\caption{\textbf{\pvdavis\ benchmark results ($\JnF$) of STCN~\cite{cheng2021stcn} trained with pseudo-masks}. \texttt{FT}: fine-tuning on~\pvdavis\ and~\pvytvos{}, *: using SAM pseudo-masks.}
\label{tab:pseudomasks_davis}
}
\end{table}
\begin{table}[ht!]
\renewcommand{\arraystretch}{1.25}
\footnotesize{
\centering
\tabcolsep=0.05cm
\begin{tabularx}{\linewidth}{l c | Y | Y Y Y Y} 
\toprule
\makecell[l]{\textbf{Pre-training}} & {FT} & 
\makecell[c]{$\mathcal{G}$} & 
\makecell[c]{$\mathcal{J_S}$} & 
\makecell[c]{$\mathcal{F_S}$} & 
\makecell[c]{$\mathcal{J_U}$} & 
\makecell[c]{$\mathcal{F_U}$} \\
\midrule
{---} & \xmark & {63.0} & {63.8} & {66.0} & {58.2} & {63.7} \\
{\pvoops}  & \xmark & {63.6} & {67.9} & {69.9} & {55.5} & {61.0}\\
{\pvkinetics} & \xmark & {67.3} & {67.8} & {70.5} & {62.0} & \textbf{69.0}\\
{\pvoops\ + \pvkinetics} & \xmark & \textbf{68.3} & \textbf{70.0} & \textbf{72.7} & \textbf{62.1} & {68.5}  \\
\midrule
{---} & \cmark & {67.7} & {69.4} & {72.7} & {60.9} & {67.8}\\
{\pvoops}  & \cmark & {67.7} & {69.5} & {73.0} & {60.3} & {68.0}\\
{\pvkinetics} & \cmark & {68.1} & {70.6} & {73.7} & {60.4} & {67.6} \\
{\pvoops\ + \pvkinetics} & \cmark & {68.7} & {70.5} & {73.7} & {61.8} & {68.8} \\
\midrule
{\pvoops\ + \pvkinetics\ $^*$} & \cmark & \textbf{73.7} & \textbf{75.5} & \textbf{77.6} & \textbf{68.1} & \textbf{73.9} \\
\bottomrule
\end{tabularx}
\caption{\textbf{\pvytvos\ benchmark results of STCN~\cite{cheng2021stcn} trained with pseudo-masks, and evaluated on 10-points.} \texttt{FT}: fine-tuning on~\pvdavis\ and~\pvytvos{}, *: using SAM pseudo-masks.}
\label{tab:pseudomasks_ytvos}
}
\end{table}

\begin{table}
\renewcommand{\arraystretch}{1.25}
\center
\footnotesize{
\tabcolsep=0.05cm
\begin{tabularx}{\linewidth}{ p{2.84cm} | Y | Y Y Y Y} 
\toprule
\makecell[l]{\textbf{Training}} & \makecell[c]{Mean} & \makecell[c]{1-point} & \makecell[c]{2-point} & \makecell[c]{5-point} & \makecell[c]{10-point}  \\
\midrule
{---} & {57.1} & {50.7} & {56.7} & {60.4} & {60.8} \\{\pvdavis +\pvytvos} & {61.3} & {55.9} & {61.2} & {63.9} & {64.3} \\
{\pvkinetics} & {61.3} & {55.4} & {60.9} & {64.2} & {64.7} \\
{\pvoops} & \textbf{64.9} & \textbf{59.7} & \textbf{64.9} &\textbf{67.4} & \textbf{67.7} \\
\midrule
{PV-Oops$^*$} & {57.7} & {40.5} & {57.6} & {65.9} & {66.7} \\
\bottomrule
\end{tabularx}
\caption{\textbf{\pvoops\ benchmark results ($\JnF$) of STCN~\cite{cheng2021stcn} trained with pseudo-masks, starting from static-image pre-training.} *: using SAM pseudo-masks.}
\label{tab:pseudo_mask_oops}
}
\end{table}
\label{tab:pseudo-mask-training-oops-results}

We generally use \dynamite{}~\cite{RanaMahadevan23Arxiv} to generate pseudo-masks from the point annotations. For the training set, we feed both positive and negative points for each object in every annotated frame to \dynamite{}, while for the validation set, we only use the available foreground point initialization.
Recently, the very strong SAM~\cite{kirillov2023sam} segmentation model became available, so for some additional setups we create masks with SAM with ViT-H backbone instead of~\dynamite{} and also report those results.

Similar to the previous baseline setup, we first train STCN on pseudo-masks generated from~\pvdavis{} and~\pvytvos{}, and then later include the additional data from~\pvoops{} and~\pvkinetics{} as an additional pre-training step.
\cref{tab:pseudomasks_davis,tab:pseudomasks_ytvos} show that the performance of STCN consistently improves with additional~\pointvos{} data on both~\pvdavis\ and~\pvytvos. Without fine-tuning, the additional~\pointvos{} data improves the mean $\JnF$ for~\pvdavis{} from 65.6\% to 70.4\% as compared to just using the static image pre-training, showing that the additional video annotations are very helpful. Likewise, the results on~\pvytvos{} improve the $\mathcal{G}$ from 63.0\% to 68.3\%. Also after fine-tuning, consistent improvements can be seen on both datasets. On DAVIS, using SAM pseudo-masks instead of \dynamite{} is beneficial when more points are available at test-time, but leads to significantly worse results for initialization with only 1 or 2 points. This is likely because SAM tends to generate part-based segmentations for a low number of points.

\cref{tab:pseudo_mask_oops} again shows that fine-tuning STCN on the \pvoops{} training data significantly improves the results on the~\pvoops{} benchmark from 57.1\% mean $\JnF$ to 64.9\%, as compared to just using static images, further boosting the baseline performance. 

\subsection{Ablation Study}
\label{subsec:ablations}

\newcommand{\numberone}{\normalsize\ding{192}}%
\newcommand{\numbertwo}{\normalsize\ding{193}}%
\newcommand{\numberthree}{\normalsize\ding{194}}%
\newcommand{\numberfour}{\normalsize\ding{195}}%
\newcommand{\numberfive}{\normalsize\ding{196}}%

\begin{table}
\footnotesize
\tabcolsep=0.05cm
\begin{centering}
\begin{tabularx}{\linewidth}{l | lp{0.2cm} lp{0.2cm} lp{0.4cm} | Y}
\toprule 
 & \textbf{Task} && Training && Testing && $\mathcal{J}\&\mathcal{F}$\tabularnewline
\midrule
\numberone & VOS && Masks && Masks && 85.3\tabularnewline
\midrule
\numbertwo & Point-VOS && Points$^\dag$ && Points && 67.7\tabularnewline
\numberthree & Point-VOS* && Pseudo-Masks$^\dag$ && Pseudo-Masks && 76.9\tabularnewline
\midrule
\numberfour & Hybrid && Masks && Points && 78.3\tabularnewline
\numberfive & Hybrid* && Masks && Pseudo-Masks && 80.4 \tabularnewline
\bottomrule
\end{tabularx}
\par\end{centering}
\begin{centering}
\caption{\label{tab:baselines}\textbf{Ablations on DAVIS using different training and test-time supervisions.} *: SAM pseudo-masks, \dag: temporally sparse. Hybrid: masks during training, points at test-time.}
\par\end{centering}
\end{table}

In \cref{tab:baselines}, we compare the conventional VOS task with~\pointvos{} on the DAVIS validation set. The training and testing columns denote the annotations available at training and test-time, respectively for each task setup. 

In the original VOS setup (\numberone), \stcn{} achieves 85.3\% $\JnF$. The \pointvos{} setup (\numbertwo{}) just on points yields 67.7\% $\JnF$, which is around 80\% of the original VOS quality (\numberone). However, when we use pseudo-masks (\numberthree), the gap closes more and we achieve 76.9\% $\JnF$, which is more than 90\% of the original VOS quality (\numberone). This result is remarkable, considering that \numberthree{} has a three-fold disadvantage compared to \numberone: 1) weak point supervision instead of masks during training, 2) temporal sparsity during training, 3) sparse point initialization instead of masks at test-time.

We also consider a \textit{Hybrid} task setting, which uses dense mask annotations for training but point-based initialization for testing (see supplement for implementation details). Results on the Hybrid setup show that switching from VOS to point-based training has a larger negative effect on quality than switching to point-based initialization (\numberone{} $\rightarrow$ \numberfour{} \vs \numberfour{} $\rightarrow$ \numbertwo{}). Again, the use of pseudo-masks improves results (\numberfive{}) and shrinks the gap towards the original VOS setup.

\begin{table}
\renewcommand{\arraystretch}{1.25}
\center
\footnotesize{
\tabcolsep=0.08cm
\begin{tabularx}{\linewidth}{l c| p{0.1cm}Y Yp{0.01cm} Y} 
\toprule
{\textbf{Pre-Training}} & {UVO-FT} && \multicolumn{2}{c}{OVIS-VNG} && {UVO-VNG} \\
\cmidrule{4-5}
\makecell[l]{} & \makecell[c]{} && {No-FT} & {FT} && {} \\
\midrule
{Static} & \xmark && {28.5} & {32.4} && {39.6} \\
{Static + \pvoops} & \xmark && {24.5} & {31.4} && {32.9} \\
{Static + \pvkinetics} & \xmark && \textbf{33.9} & \textbf{35.1} && \textbf{51.8} \\
\midrule
{Static} & \cmark && {32.0} & {32.7} && {46.4} \\
{Static + \pvoops} & \cmark && {31.8} & {32.6} && {48.0} \\
{Static + \pvkinetics} & \cmark && {32.0} & \textbf{35.0} && \textbf{52.8} \\
\midrule
{Static + \pvkinetics\ $^*$} & \cmark && \textbf{32.9} & {34.4} && {52.5} \\
\bottomrule
\end{tabularx}
\caption{\textbf{OVIS-VNG and UVO-VNG results ($\JnF$) of RF-VNG}. All models start with COCO-PNG pre-training. \texttt{UVO-FT}: fine-tuning on UVO-VNG data, \texttt{FT}: fine-tuning on OVIS-VNG data, \texttt{No-FT}: no fine-tuning. *: using SAM pseudo-masks.}
\label{tab:referformer-vng-experiments}
}
\end{table}
\subsection{Language-Guided Tasks}
\label{subsec:language-guided}
As described in \cref{subsec:annotation-pipeline}, each object that we annotated is linked to a noun in a sentence.
Hence, those multi-modal annotations can be used to improve models connecting vision and language, \eg, for the Video Narrative Grounding (VNG)~\cite{Voigtlaender23CVPR} task. In VNG, the input to a method is a text description (\eg, ``A green parrot with a red-black neckline is playing with the other parrot''~\cite{Voigtlaender23CVPR}) in which the position of certain nouns (\eg, both instances of ``parrot'') is marked. For each marked noun, the VNG method has to segment the corresponding noun in each frame of the video.

Our multi-modal point annotations link each point to a noun in a sentence, which matches the setup of VNG. We combine the pseudo-masks generated by \dynamite{} based on the annotated points with the language annotations to create our new Oops-VNG and Kinetics-VNG training sets that cover 133K objects in 32K videos. This is more than 3 times larger than the existing VNG datasets OVIS-VNG~\cite{qi2022occluded, Voigtlaender23CVPR} and UVO-VNG~\cite{wang2021unidentified, Voigtlaender23CVPR} that span 45K objects in 8K videos.

We conduct experiments using the state-of-the-art VNG method RF-VNG~\cite{Voigtlaender23CVPR}. The original RF-VNG is trained in 3 steps: 1) pre-training on static images of the COCO-PNG~\cite{gonzalez2021cocopng} training set, 2) optional training on UVO-VNG, 3) optional fine-tuning on the OVIS-VNG training set for evaluation on the OVIS-VNG validation set. We use the same training recipe, but insert another pre-training step between steps 1) and 2), where we train RF-VNG~\cite{Voigtlaender23CVPR} on our new Oops-VNG or our new Kinetics-VNG training sets. 

\cref{tab:referformer-vng-experiments} demonstrates that adding our new data significantly improves the baseline results for the VNG task.
\Eg, the best result on OVIS-VNG improves from 32.7\% $\JnF$ to 35.0\%, and for UVO-VNG, from 46.4\% to 52.8\%, \ie, we achieve an improvement of 6.4 percentage points.

\section{Conclusion}
\label{sec:conclusion}

In this work, we have proposed a point-based VOS task, \pointvos{}, and a point-wise annotation scheme, which is much more efficient than the existing dense-mask annotation scheme. We use this to annotate two large-scale multi-modal VOS datasets that are much larger than the largest available densely annotated VOS datasets. In addition, we also introduce a point-based training strategy for the VOS methods and correspondingly show that existing VOS methods can be easily adapted to leverage our point annotations. Finally, our experiments show the benefits of our newly annotated point data by advancing the state-of-the-art performance for various uni-modal and multi-modal (vision+language) benchmarks.

\footnotesize{\PAR{Acknowledgments.} We would like to thank Rodrigo Benenson, Amit Kumar Rana, and Vittorio Ferrari for their helpful feedback and discussions. We also would like to thank all our annotators. This project was partially funded by BMBF project NeuroSys-D (03ZU1106DA) and EU Network of Excellence TAILOR (H2020-ICT-2019-952215). The compute resources for several experiments were granted by the Gauss Centre for Supercomputing
e.V. through the John von Neumann Institute for Computing
on the GCS Supercomputer JUWELS at Julich Supercomputing Centre.}

{
    \small
    \bibliographystyle{ieeenat_fullname}
    \bibliography{main}
}

\clearpage

\appendix
\twocolumn[
\begin{center}
  {\Large \bf \Large{\pointvos{}: Supplementary Material } \par}
  \vspace*{12pt}
\end{center}
]

\begin{abstract}
In this supplementary, we provide the experimental results of the additional simulations in ~\cref{sec:additional_simulations}, the details for annotating the \pointvos{} Oops validation set in~\cref{sec:annotate_pointvos_oops_val_set}, the statistics for \pointvos{} datasets in ~\cref{sec:statistics_datasets}, the implementation details in ~\cref{sec:implementation_details} and the additional qualitative results in~\cref{sec:additional_qualitative}.

\end{abstract}

\section{Additional Simulations}
\label{sec:additional_simulations}

\PAR{Farthest Point Sampling Strategy.} In Sec. 3.1 of the main paper, we ran a number of point simulation experiments on DAVIS~\cite{Pont-Tuset_arXiv_2017} and YT-VOS~\cite{xu2018youtube} to analyse the effect of using point annotations both during training and testing. For these experiments, the simulated points are sampled randomly from the available ground truth segmentation masks for each frame. 

In addition to sampling the points randomly, we also consider using the farthest-point sampling (FPS) technique. The FPS algorithm starts from some random initial point in the given input point set, and then iteratively selects a single point that has the largest distance from all the previously sampled ones.
For our point simulations, instead of starting from a random point, we always start from a point that best represents the center of the input mask. 
To sample this center, we first generate Euclidean distance transforms from the ground-truth segmentation masks for each foreground object and the common background. We then sample the point that has the largest distance from each of these distance transforms and further use these as the starting points for the FPS algorithm. The FPS algorithm is then separately applied on the points that represent each of the foreground objects and the background starting from the corresponding center point.

\begin{figure}
    \centering
    \includegraphics[width=0.92\linewidth]{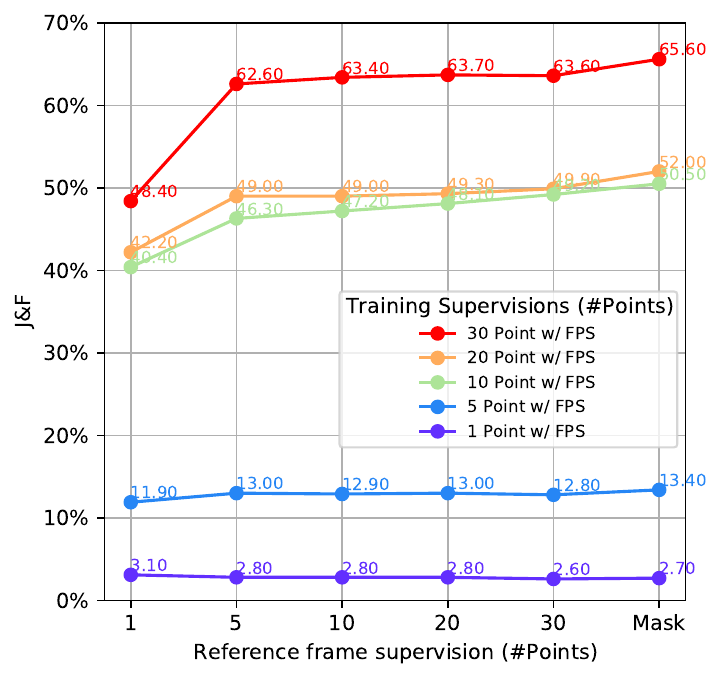}
    \vspace{-0.4cm}
    \caption{\textbf{FPS point sampling results on the DAVIS validation set.} We vary the number of sampled points per object for training supervision and the number of sampled points on the reference frame.}
    \label{fig:point_sampling_fps}
\end{figure}

Similar to the point simulation experiments presented in Sec. 3.1, we again use ~\pointstcn{} to train multiple models on different number of simulated points. Here again, we do not apply any temporal sparsity. Also, note that in both random and FPS point sampling strategies, we run each experiment 5 times and report the mean score. In~\cref{fig:point_sampling_fps}, we show the results for the FPS point sampling strategy on the DAVIS validation set.
It can be seen that the performance of~\pointstcn{} is much worse when we use FPS instead of the random sampling strategy (see Fig. 2 in the main paper),~\eg, for FPS we achieve the best result by training with 30 points, which is almost on par with using 10 points as training supervision in the random point sampling strategy. Thus, we decided to use the random point sampling strategy for our point annotations.

\begin{figure}
    \centering
    \includegraphics[width=0.92\linewidth]{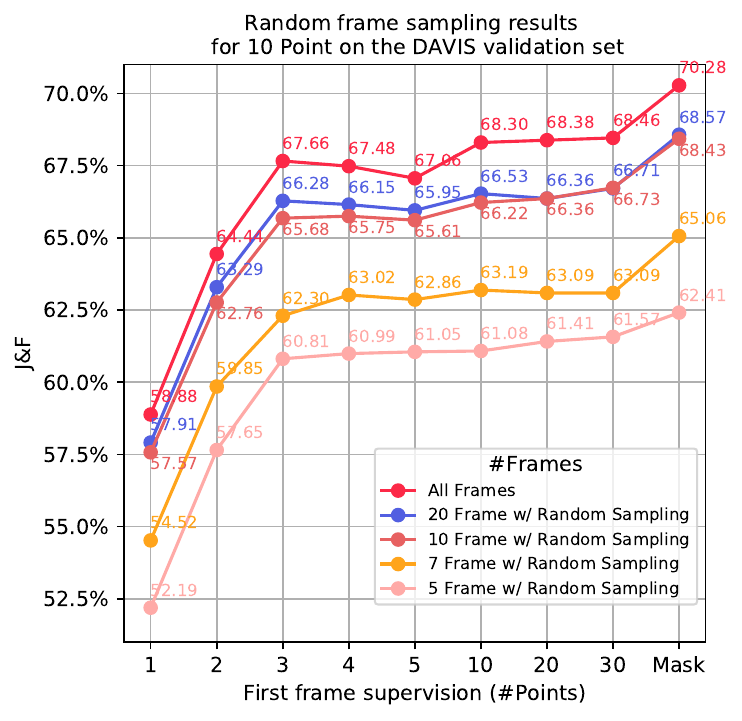}
    \vspace{-0.4cm}
    \caption{\textbf{Results on the DAVIS validation set for randomly sub-sampling frames.} We vary the number of randomly sampled frames.}
    \label{fig:random_frame_subsampling}
\end{figure}

\begin{figure*}[t]
    \centering
      \begin{subfigure}[h]{0.49\textwidth}
        \centering
        \includegraphics[width=\mysize\linewidth]{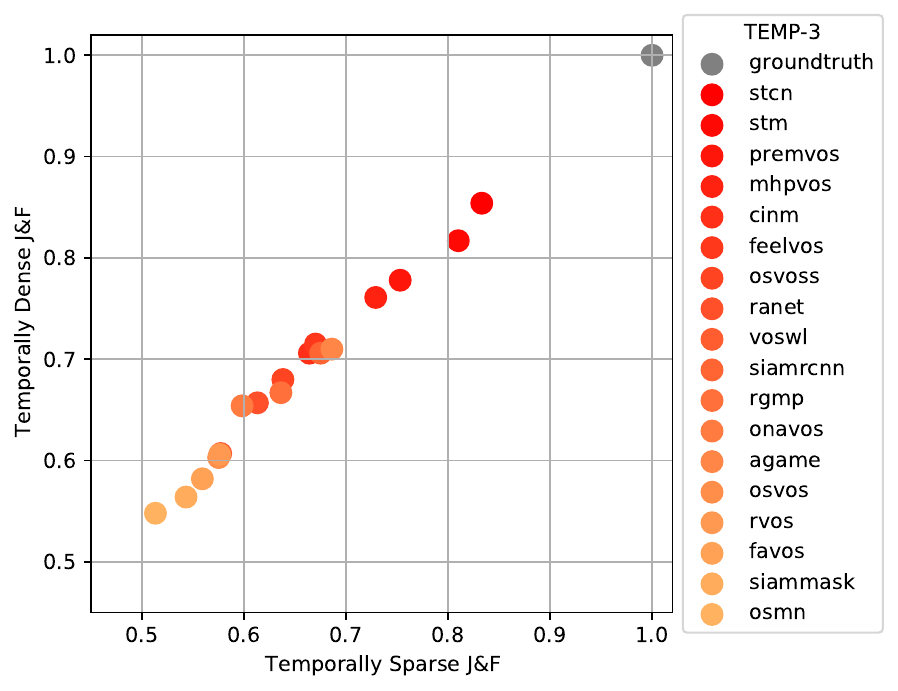}
       \end{subfigure}
       \begin{subfigure}[h]{0.49\textwidth}
        \centering
        \includegraphics[width=\mysize\linewidth]{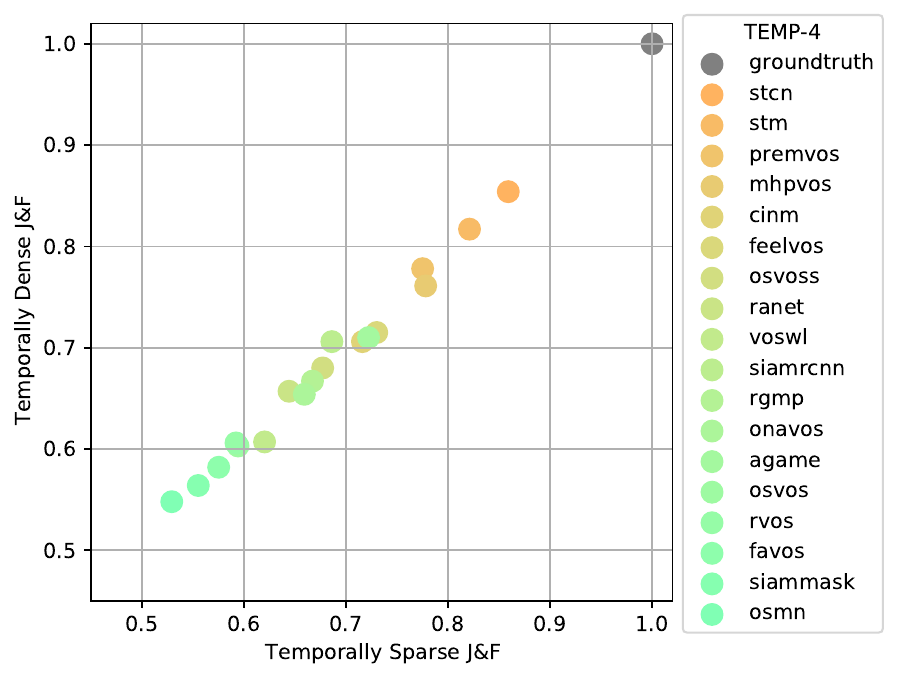}
       \end{subfigure}
       \begin{subfigure}[h]{0.49\textwidth}
        \centering
       \includegraphics[width=\mysize\linewidth]{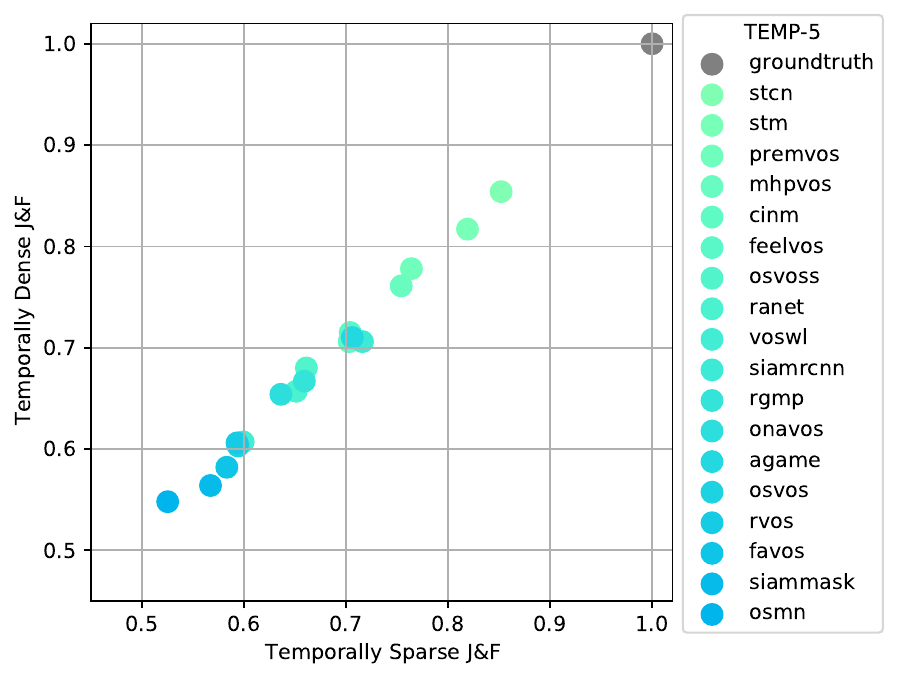}
       \end{subfigure}
        \begin{subfigure}[h]{0.49\textwidth}
        \centering
       \includegraphics[width=\mysize\linewidth]{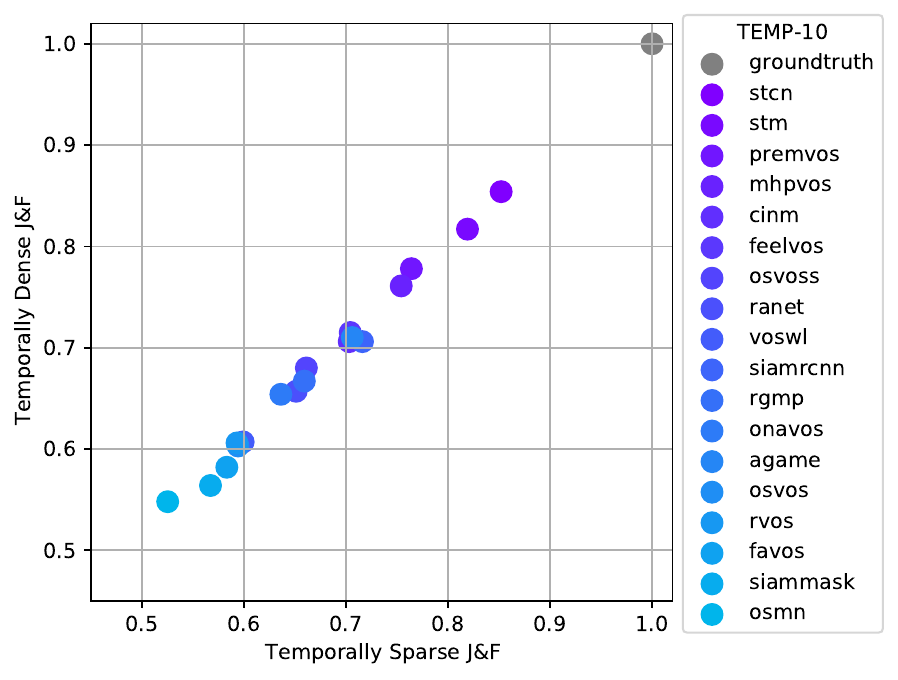}
       \end{subfigure}
       \vspace{-8pt}
       \caption{\textbf{Temporally dense $\JnF$ \vs temporally sparse $\JnF$ results.} We get 18 methods (colored dots) from the leaderboard of the DAVIS benchmark and evaluate them on the 4 different temporally sparse DAVIS validation sets. TEMP-3 shows the results evaluated on 3 sub-sampled frames, TEMP-4 on 4 sub-sampled frames, TEMP-5 on 5 sub-sampled frames, and TEMP-10 on 10 sub-sampled frames.}
       \label{fig:temporally-sparse-eval}
\end{figure*}

\PAR{Randomly Sub-sampling Frames.} Along with the evenly-spaced sub-sampling strategy explained in Sec.~3.1 of the main paper, we also try to sub-sample frames randomly. Similarly to evenly-spaced sub-sampling, starting from all frames, we randomly sub-sample up to 20 frames for each video. For training supervision, we keep again the setup of 10 randomly sampled points per frame per object. Also, we run each experiment 3 times for both evenly-spaced and random sub-sampling strategies and report the mean score. 

We demonstrate the results for the random sub-sampling strategy on the DAVIS validation set in~\cref{fig:random_frame_subsampling}. We obtain very similar results for both strategies. We cannot observe a notable difference compared to Fig. 3 in the main paper, so we decided to make use of the evenly-spaced sub-sampling strategy. 

\PAR{Evaluating on temporally sparse videos.} To annotate the ground-truth segmentation masks for the evaluation of the \pointvos{} Oops (\pvoops{}) benchmark, we also make the following key design decision. As the consecutive frames are extremely correlated and redundant, we question whether evaluating the results on a sparse subset of frames is sufficient. In that way, we would diminish the annotation effort while annotating the validation set as well, increasing cost and time-efficiency.

To this end, we run analysis experiments on DAVIS benchmark results. First, we generate temporally sparse validation sets from the DAVIS validation set by sub-sampling 3, 4, 5, and 10 frames evenly spaced,~\ie, we obtain 4 temporally sparse validation sets consisting of sub-sampled 3, 4, 5, or 10 frames. Then,we get the methods~\cite{cheng2021stcn,oh2019video,Luiten2018PReMVOSPR,xu2019mhp,bao2018cnn,voigtlaender2019feelvos,Man+18b,Ziqin2019RANet,khoreva2019video,voigtlaender2020siam,oh2018fast,voigtlaender17DAVIS,johnander2019generative,Cae+17,ventura2019rvos,Cheng_favos_2018,wang2019fast,Yang2018EfficientVO} from the DAVIS benchmark leaderboard, evaluate them on a sparse set of frames. Finally, we compare these results with the results on the temporally dense validation set,~\ie, the original DAVIS validation set with all frames. As seen in~\cref{fig:temporally-sparse-eval}, the results are extremely correlated for all temporally sparse validation sets. In other words, even with only 3 ground-truth frames per object for evaluation, the ranking between methods does not change in almost all cases (except when their performance is extremely close to each other).

\section{Annotating \pointvos{} Oops Validation Set}
\label{sec:annotate_pointvos_oops_val_set}
We start annotating the \pointvos{} Oops (\pvoops{}) validation set by first annotating the reference frame with points. We generate point-wise annotations on the sub-sampled (evenly-spaced) 10 frames from each video and ask human annotators to verify them in the same way as for the training point annotations. Then, we check each video to decide the reference frame. In each video, we assign the first frame that contains at least 7 foreground points as the reference frame and remove all frames before the reference frame.
In case, we cannot find a frame in the video with at least 7 foreground points, we eliminate the video. We also check whether we have enough frames after the reference frame. If there is no frame after the reference frame with at least 3 foreground points and 1 background point, we also drop the video. 

Afterwards, we annotate the ground-truth segmentation masks for the evaluation of the \pvoops{} benchmark. Informed by the simulation experiment for evaluating on a sparse subset of frames (see~\cref{sec:additional_simulations}), we decided to annotate temporally sparse segmentation masks for the evaluation of the \pvoops{} benchmark with 
3 ground-truth frames.

While annotating 3 ground-truth frames, we start by first annotating the frame with the mouse trace segment for each video. Note that the mouse trace comes from the location-output questions of VidLN~\cite{Voigtlaender23CVPR} for the \pvoops{} dataset. In the original VidLN location-output task (which we do not consider in our work), a mask in the frame with the mouse trace is approximately evaluated by comparing it to the mouse trace.
By annotating a segmentation mask for this frame, we make sure that our annotations can be used to replace the original VidLN evaluation, that compares the predicted mask with the mouse trace, with a more precise evaluation, that compares the predicted mask with the annotated mask.

After annotating the frame with mouse trace, we check each video and eliminate the videos, if the frame with the mouse trace is temporally before the reference frame, or exactly on the reference frame. From the remaining videos, we sub-sample (evenly-spaced) 3 frames from the frames coming after the reference frame with point annotations, and we check whether the frame with the mouse trace is in the 3 sub-sampled frames. If the frame with the mouse trace is in the 3 sub-sampled frames, we keep the other 2 sub-sampled frames and annotate them with ground-truth masks. If the frame with the mouse trace is not in the 3 sub-sampled frames, we drop the frame that is temporally closest to the mouse trace frame and send the other 2 frames to annotation. 

\section{\pointvos{} Datasets Statistics}
\label{sec:statistics_datasets}

\PAR{Overview.}
In~\cref{tab:annotation_stats}, we present the detailed statistics for the training and validation splits of the \pointvos{} datasets.

\pointvos{} Oops (\pvoops{}) and \pointvos{} Kinetics (\pvkinetics{}) are the datasets that we annotated with new points. In total, we collected $19.7M$ points where $5.8M$ points are annotated as positive points and $13.9M$ points as negative points. Also, $271K$ points are annotated as ambiguous points. We do not use any ambiguous annotations in our experiments.

In \pvoops{}, there are $541K$ positive points and $1.2M$ negative points in the training split, and also $7.3K$ positive points and $9.9K$ negative points in the validation split. In \pvkinetics{}, there are $5.2M$ positive points and $12.6M$ negative points.
 
In addition to the \pvoops{} and \pvkinetics{} datasets, we also generated the \pointvos{} versions of the DAVIS and YouTube-VOS (YT-VOS) datasets. For \pointvos{} DAVIS (\pvdavis{}) and \pointvos{} YouTube (\pvytvos{}), we sample the spatially temporally sparse points from the ground truth masks. Since the original DAVIS and YT-VOS datasets are massively smaller than \pvoops{} and \pvkinetics{}, the total positive and negative points are also very much less in \pvdavis{} and \pvytvos{}. There are $9.7K$ positive points and $6K$ negative points in the \pvdavis{} training split, $558$ positive and $300$ negative points in the \pvdavis{} validation split. \pvytvos{} contains $472K$ positive and $346K$ negative points in the training split, and $9.8K$ positive and $6K$ negative points in the validation split. Note that there are fewer annotations in both \pvdavis{} and \pvytvos{} compared to the original DAVIS and YT-VOS datasets as we sub-sample 10 frames.

\begin{table}[t]
\renewcommand{\arraystretch}{1.5}
\center
\scriptsize{
\tabcolsep=0.03cm
\begin{tabular}{llcccccc}
\toprule 
& \textbf{Dataset} & Videos & Annotations & Objects &\makecell[c]{Positive \\ Points} & \makecell[c]{Negative \\ Points}& \makecell[c]{Ambiguous \\ Points}\\
\midrule 
 \parbox[t]{3mm}{\multirow{4}{*}{\rotatebox[origin=c]{90}{\scriptsize{\textbf{train}}}}} 
 & \pointvos{} Oops & 7.4K & 93K & 12K & 541K & 1.2M & 18K \\
 & \pointvos{} Kinetics & 23.9K & 965K & 120K & 5.2M & 12.6M &  253K \\
 \cline{2-8}
 &\pointvos{} DAVIS  & 60  &  600 & 145  & 9.7K & 6K & --- \\
 &\pointvos{} YouTube  &  3471 & 34.6K  & 6.4K & 472K & 346K & --- \\
\midrule
 \parbox[t]{3mm}{\multirow{3}{*}{\rotatebox[origin=c]{90}{\scriptsize{\textbf{val}}}}} 
 & \pointvos{} Oops & 991 & 3.5K & 991 &  7.3K & 9.9K & 91 \\
 \cline{2-8}
 &\pointvos{} DAVIS  & 30  &  1.9K & 61  & 558 & 300 & --- \\
 &\pointvos{} YouTube  &  507 &  614 & 1K & 9.8K & 6K & --- \\
\bottomrule
\vspace{-0.7cm}
\end{tabular}\caption{\textbf{Statistics for the \pointvos{} datasets.} Annotations here means summing up frames containing at least one annotated object. Note that for \pointvos{} DAVIS and \pointvos{} YouTube, we sampled the points from the ground truth masks, while for all other datasets, we annotated new points.}
\label{tab:annotation_stats}
}
\end{table}

\PAR{Frame Distribution.} In addition to the detailed statistics, we also analyze the distribution of frames in the training splits of \pvoops{} and \pvkinetics{}. During the annotation process, we provided 10 frames to the human annotators for annotations. Here, the distribution of frames means, we check each video after the annotation process and sum up the frames in each video, which have at least one positive point annotation.

\definecolor{piecolor9}{RGB}{0, 177, 29}  
\definecolor{piecolor8}{RGB}{236, 200, 111}
\definecolor{piecolor7}{RGB}{164, 248, 159}
\definecolor{piecolor6}{RGB}{90, 248, 200}

\begin{figure}[ht!]
    \centering
      \begin{subfigure}[h]{0.23\textwidth}
        \centering
        \includegraphics[trim={0 0.9cm 0 0.9cm},clip,width=\mysize\linewidth]{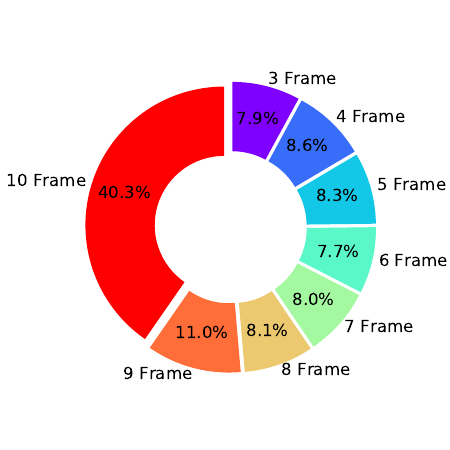}
         \caption{\pvoops{}}
        \label{fig:oops_video_dist_frame}
       \end{subfigure}
       \begin{subfigure}[h]{0.23\textwidth}
        \centering
       \includegraphics[trim={0 0.9cm 0 0.9cm},clip,width=\mysize\linewidth]{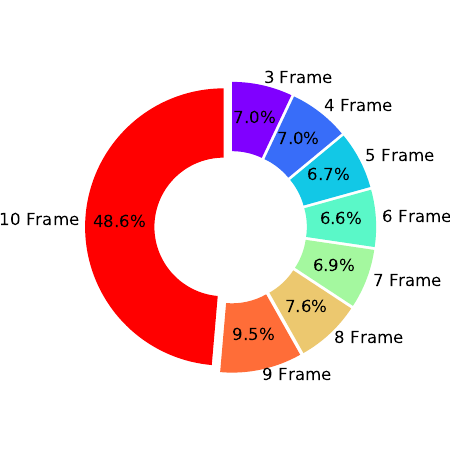}
       \caption{\pvkinetics{}}
       \label{fig:kinetics_video_dist_frame}
       \end{subfigure}
       \vspace{-0.cm}
       \caption{\textbf{The distribution of frames for \pvoops{} and \pvkinetics{}.} The distribution of frames means summing up frames in each video, which contains at least one positive point annotation.}
       \label{fig:frame_dist}
\end{figure}

~\cref{fig:frame_dist} shows the frame distribution for \pvoops{} (see ~\cref{fig:oops_video_dist_frame}) and \pvkinetics{} (see~\cref{fig:kinetics_video_dist_frame}). As seen, more than 40\% of the videos in both \pvoops{} and \pvkinetics{} have all frames with positive point annotations (see \textcolor{red}{red} slice). Also, more than 30\% of the videos in both \pvoops{} and \pvkinetics{} contain more than 5 frames with positive point annotations (see \textcolor{piecolor6}{chameleon}, \textcolor{piecolor7}{green}, \textcolor{piecolor8}{caramel macchiato} and \textcolor{orange}{orange} slices).

\PAR{Point Distribution.} Finally, we analyze the distribution of the positive and negative points in the training splits of \pvoops{} and \pvkinetics{}. Here, the distribution of points means reporting the total number of videos in the different ranges of the number of point annotations.

\begin{figure*}[ht!]
    \centering
      \begin{subfigure}[h]{0.99\textwidth}
        \centering
        \includegraphics[width=\mysize\linewidth]{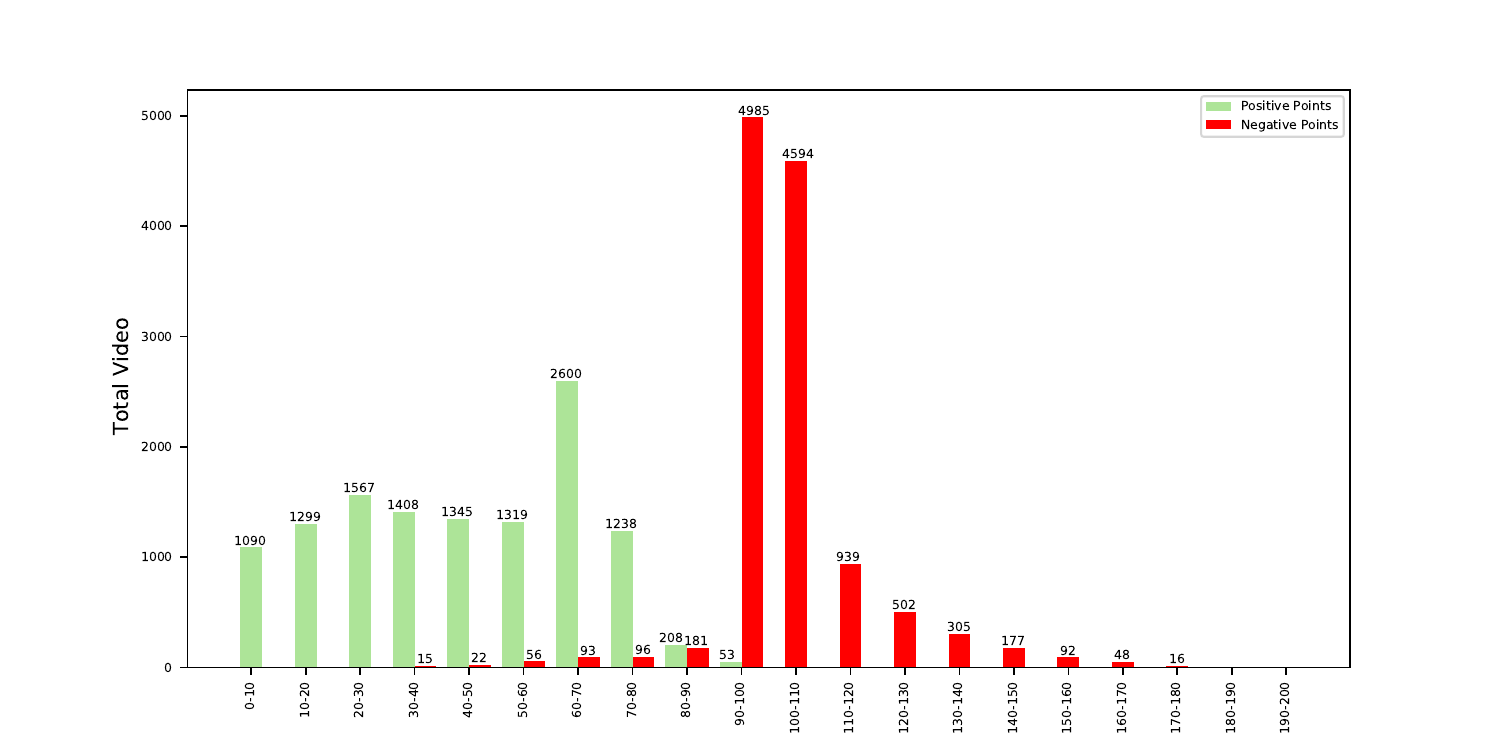}
         \caption{\pvoops{}}
        \label{fig:oops_point_distribution}
       \end{subfigure}
       \begin{subfigure}[h]{0.99\textwidth}
        \centering
       \includegraphics[width=\mysize\linewidth]{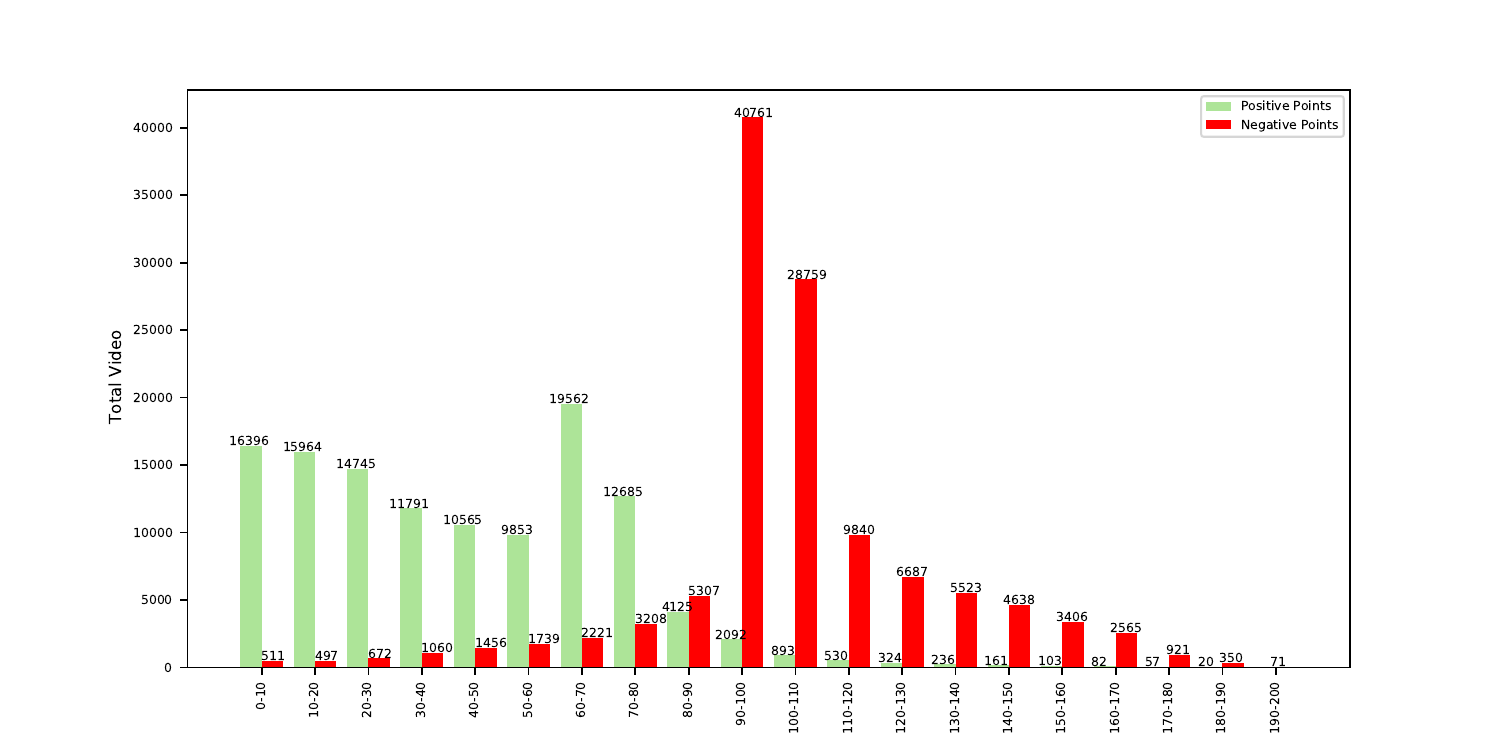}
       \caption{\pvkinetics{}}
     \label{fig:kinetics_point_distribution}
       \end{subfigure}
       \vspace{-0.3cm}
       \caption{\textbf{The distribution of the positive and negative points in \pvoops{} and \pvkinetics{}.} The x-axis represents the different ranges for the number of points, and the y-axis represents the total number of videos. Also, we show the precise numbers for the total videos at the top of the bars.}
       \label{fig:point_distribution}
\end{figure*}

We show the distribution of points in~\cref{fig:point_distribution} for \pvoops{} (see~\cref{fig:oops_point_distribution}) and \pvkinetics{} (see~\cref{fig:kinetics_point_distribution}). As seen, we observe similar point distributions in both \pvoops{} and \pvkinetics{}. As the size of the objects varies, the distribution of the positive points has more probability mass on the left than the distribution of the negative points in both \pvoops{} and \pvkinetics{}. Since we fixed the number of background points to 10 points for annotating, the distribution of the negative points has probability mass at the center for both \pvoops{} and \pvkinetics{}.

\section{Implementation Details}
\label{sec:implementation_details}

\PAR{\pointstcn{}.} 
A major advantage of using point annotations is that it can be used to train existing VOS models without making drastic changes to either the inherent model or the training strategy. We show this by easily adapting STCN to work with our point annotations while keeping most of the network structure intact. Specifically, we make the following modifications to STCN: (i) The value encoder of STCN now takes a set of sparse points (that we represent as a sparse segmentation mask) for each of the reference foreground objects in the first frame mask instead of the dense pixel-level masks. To leverage these point annotations, similar to the original STCN pre-processing pipeline, we apply augmentations like affine transformations and convert the points into a mask that has only non-zero elements on the locations of the points. We concatenate the point masks with the input image which is then processed by the value encoder. (ii) Instead of using a bootstrapped cross-entropy loss on the predicted dense posterior probabilities, we use a point-wise cross entropy loss where the loss is applied to only the output vectors at sparse point locations that are annotated in the ground-truth. We use bilinear interpolation on the output probability map to approximate the predictions on the precise point locations. During training, we use both the positive and the negative points for the loss computation. For each training sample, we sample 3 frames from a video. One of those frames is considered the reference frame which we used for initialization. The two other frames are considered the target frames, on which we calculate the loss. Only the positive (foreground) points are used as initialization in the reference (first) frame during both training and testing, while both positive and negative points are used to calculate the loss on the target frames.

\PAR{\dynamite{} Adaptation.}
\dynamite{}~\cite{RanaMahadevan23Arxiv} was originally designed to process user interactions in the form of user clicks. Since, for our point annotation scheme, only a mouse trace is available on the reference frames for each foreground object, we adapt~\dynamite{} to work with a trace as input for generating a reference mask. Those reference masks are later fed to~\stcn{} for propagation (see Sec. 3.2 of the main paper). 
To adapt \dynamite{}, we first sample the image features that correspond to each of the pixel-locations covered by the input mouse trace, and perform a \textit{global average pooling} operation to generate a single feature vector. This feature vector is then projected linearly to generate a query that corresponds with the trace, similar to the click features in~\dynamite{}. This query is then used by the \textit{Interactive Transformer} module in~\dynamite{} to generate the output mask for the object of interest.

\PAR{Training Details.}
We train \pointstcn{} with points and \stcn{} with pseudo-masks on \pointvos{} DAVIS (\pvdavis{}) and \pointvos{} YouTube-VOS (\pvytvos{}) jointly for a total of 38K iterations. The learning rate is reduced after 30K steps. On \pointvos{} Oops (\pvoops{}), \pointstcn{} and \stcn{} are trained in total 60K iterations, and the learning rate is reduced after 50K steps. On \pointvos{} Kinetics (\pvkinetics{}), and also joint training on \pvoops{} and \pvkinetics{}, we train \pointstcn{} and \stcn{} in total 190K iterations and reduce the learning rate after 150K steps.

Following the original \stcn{} setup, when training jointly on \pvdavis{} and \pvytvos{}, we build a combined dataset by repeating the \pvdavis{} dataset 5 times and \pvytvos{} 1 time, to compensate for the smaller size of \pvdavis{}. Similarly, when training jointly on \pvoops{} and \pvkinetics{}, we build a combined dataset by repeating \pvoops{} 5 times and \pvkinetics{} 1 time in order to compensate for the smaller size of \pvoops{}.

Moreover, for each training of \pointstcn{} and \stcn{}, we use Adam~\cite{kingma2014adam} and start with a learning rate of $10^{-5}$ and reduce it to $10^{-6}$ after a certain number of training steps as indicated above. We set the weight decay to $10^{-7}$ and the batch size to $4$. We conduct all \stcn{} and \pointstcn{} trainings with 8 V100 GPUs, and all inference experiments on a single 3090 GPU.

For training ReferFormer, we closely follow the setup used by VidLN~\cite{Voigtlaender23CVPR}.

\PAR{Hybrid Task.} 
In Sec.~4.1 of the main paper, we introduced the \textit{Hybrid} task (a task in between VOS and Point-VOS). In the VOS task, dense segmentation masks are used both during training and for test-time initialization, while, in the Point-VOS task, spatially temporally sparse point annotations are used in both cases.
For the Hybrid task, spatially and temporally dense masks are used during training, while only points are used on the reference frame at test-time. This means that the Hybrid task follows the setup from VOS at training time, while it follows the setup from Point-VOS at test-time.

In the Hybrid setup, we make use of dense masks to train Hybrid-STCN while we initialize the reference frame with sparse points. Recall that STCN uses 3 frames during training, from which one is the reference frame and two are the target frames. In the Hybrid setup, we initialize STCN with points in the reference frame and apply a full mask loss in the target frames. At test-time, Hybrid-STCN can then be initialized with points and achieves better results than Point-STCN, as we use more supervision during training.

\section{Additional Qualitative Results}
\label{sec:additional_qualitative}

\begin{figure*}[t]
    \centering
    \setlength{\fboxsep}{0.32pt}
    \begin{subfigure}[b]{0.32\linewidth}
     \includegraphics[width=\mysize\linewidth]{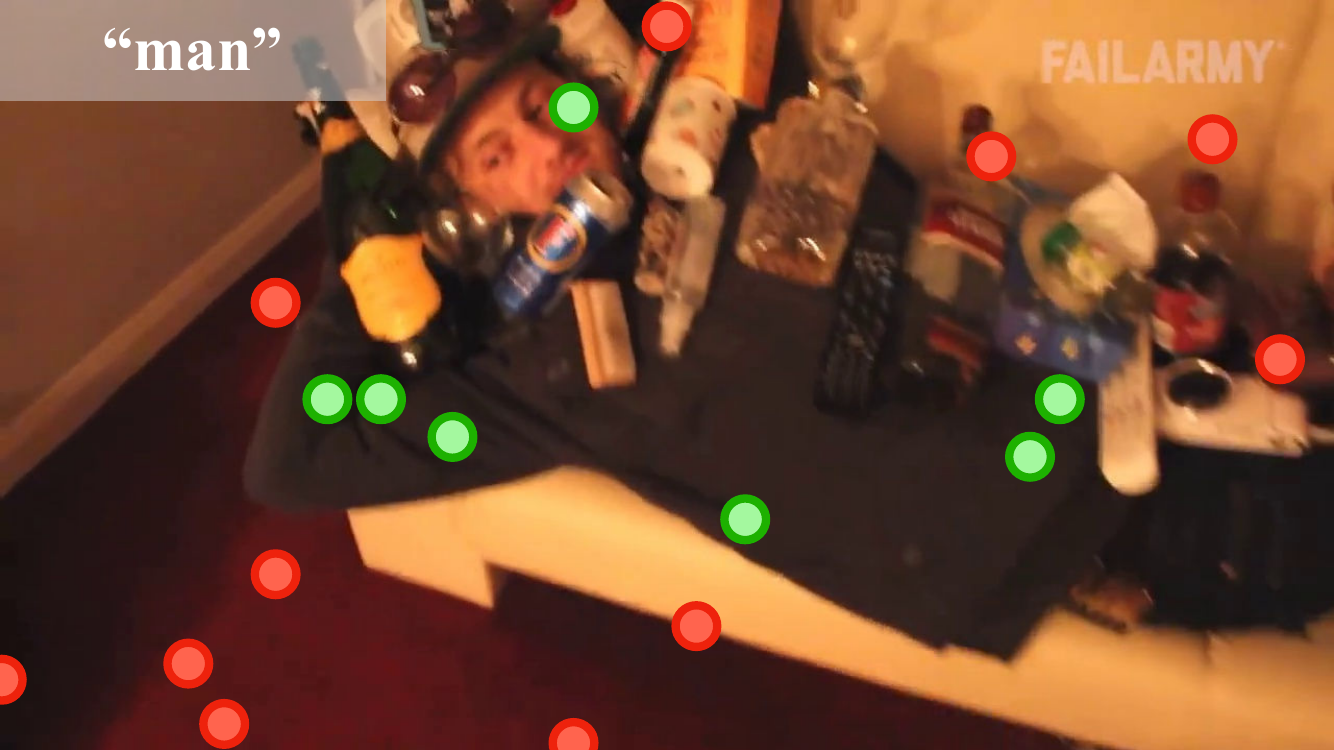}
        \vspace{-16pt}
    \end{subfigure}
    \begin{subfigure}[b]{0.32\linewidth}
        \includegraphics[width=\mysize\linewidth]{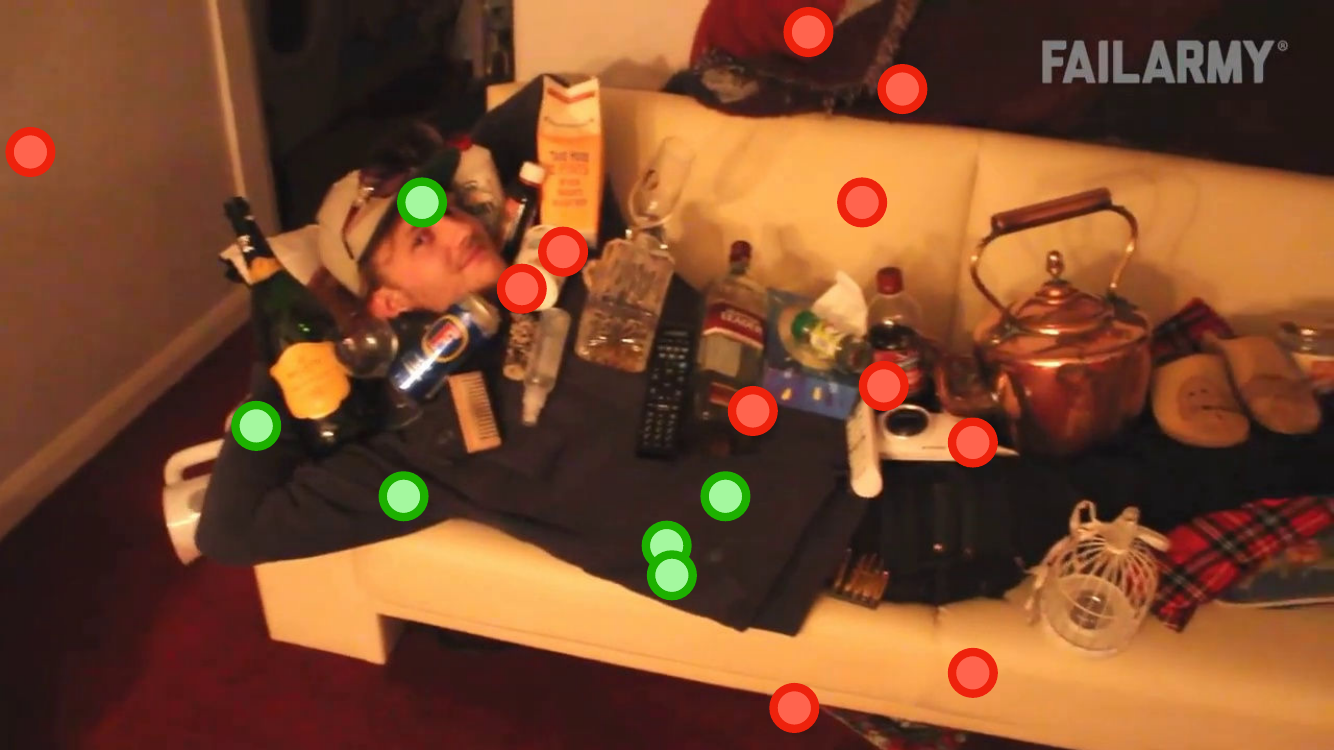}
        \vspace{-16pt}
    \end{subfigure}
    \begin{subfigure}[b]{0.32\linewidth}
     \includegraphics[width=\mysize\linewidth]{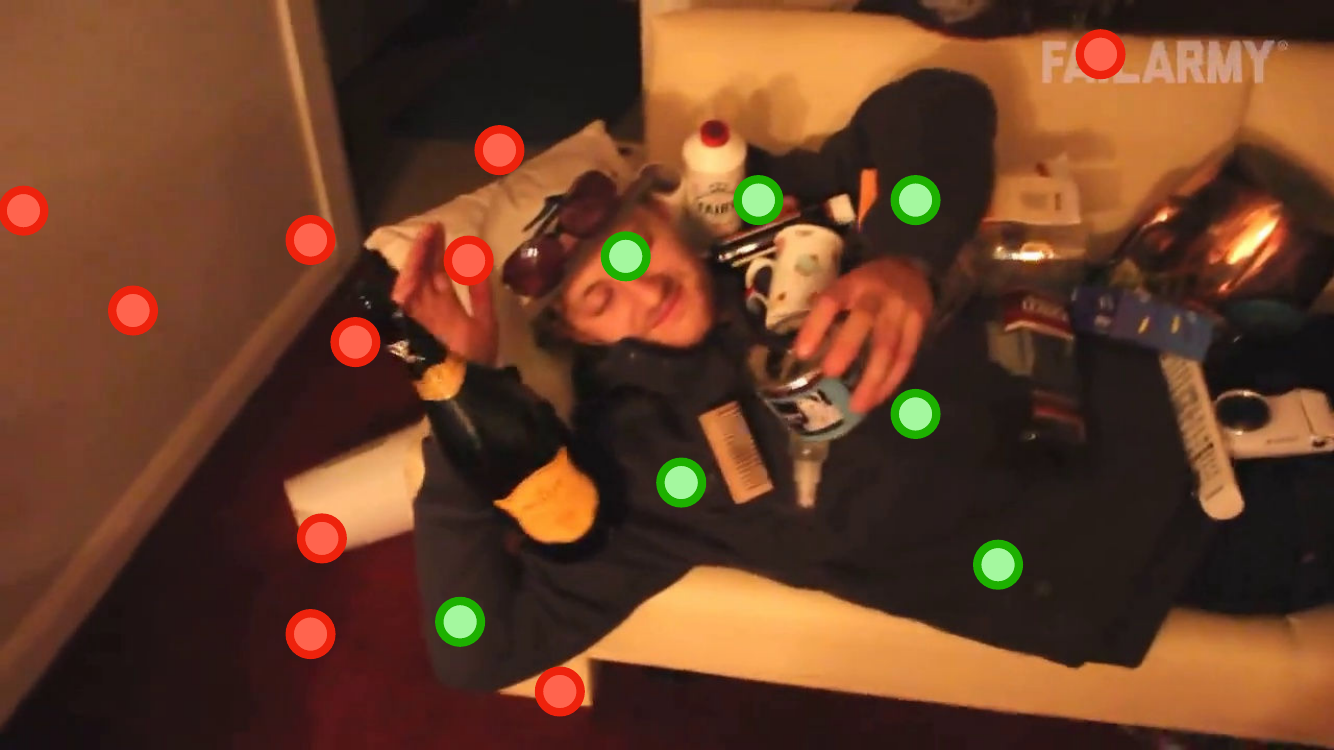}
        \vspace{-16pt}
    \end{subfigure}
    \vspace{5pt}
    
    \begin{subfigure}[b]{0.32\linewidth}
        \includegraphics[width=\mysize\linewidth]{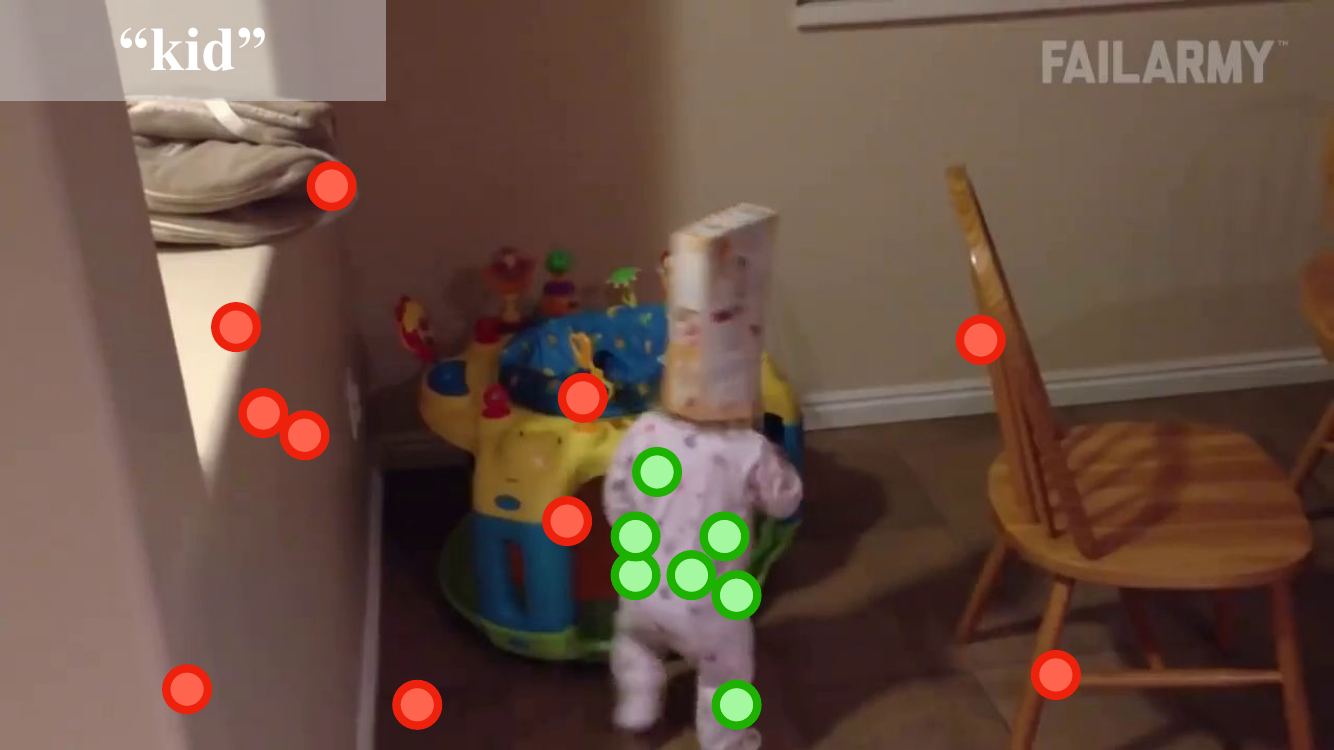}
        \vspace{-16pt}
    \end{subfigure}
    \begin{subfigure}[b]{0.32\linewidth}
        \includegraphics[width=\mysize\linewidth]{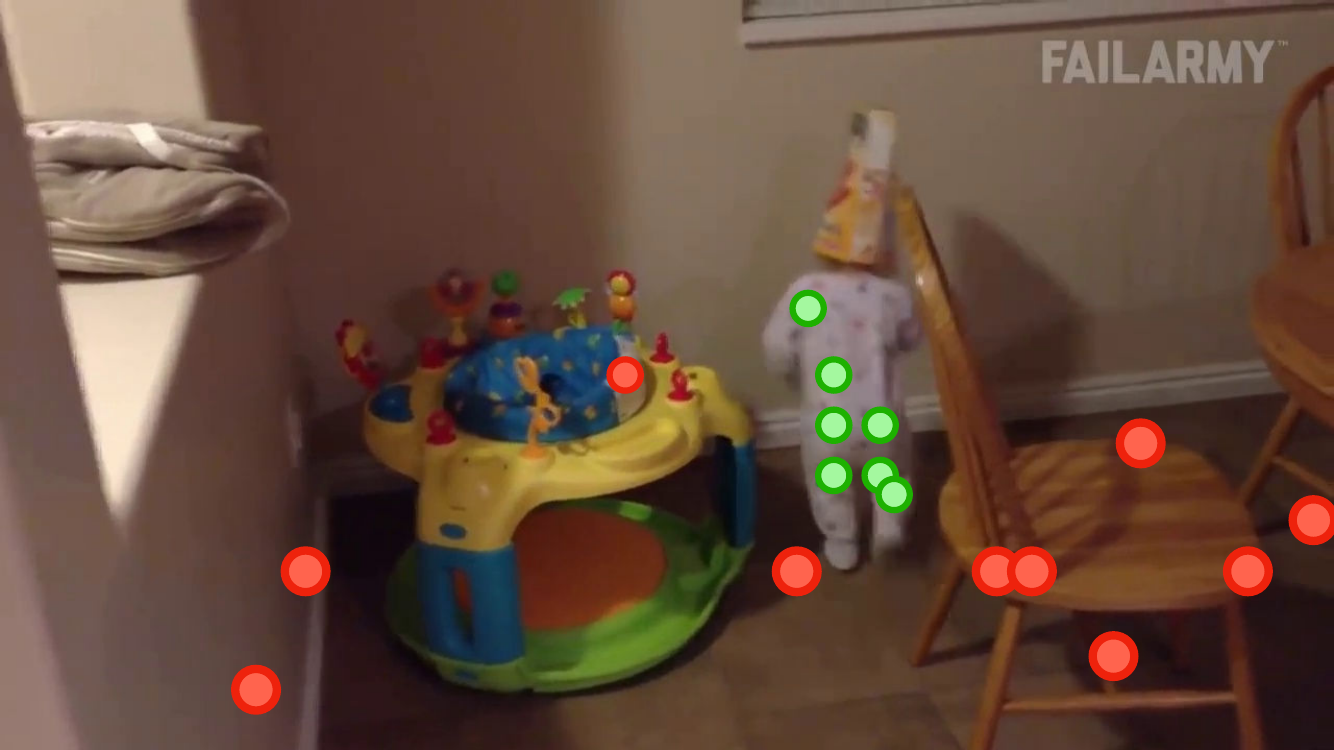}
        \vspace{-16pt}
    \end{subfigure}
    \begin{subfigure}[b]{0.32\linewidth}
     \includegraphics[width=\mysize\linewidth]{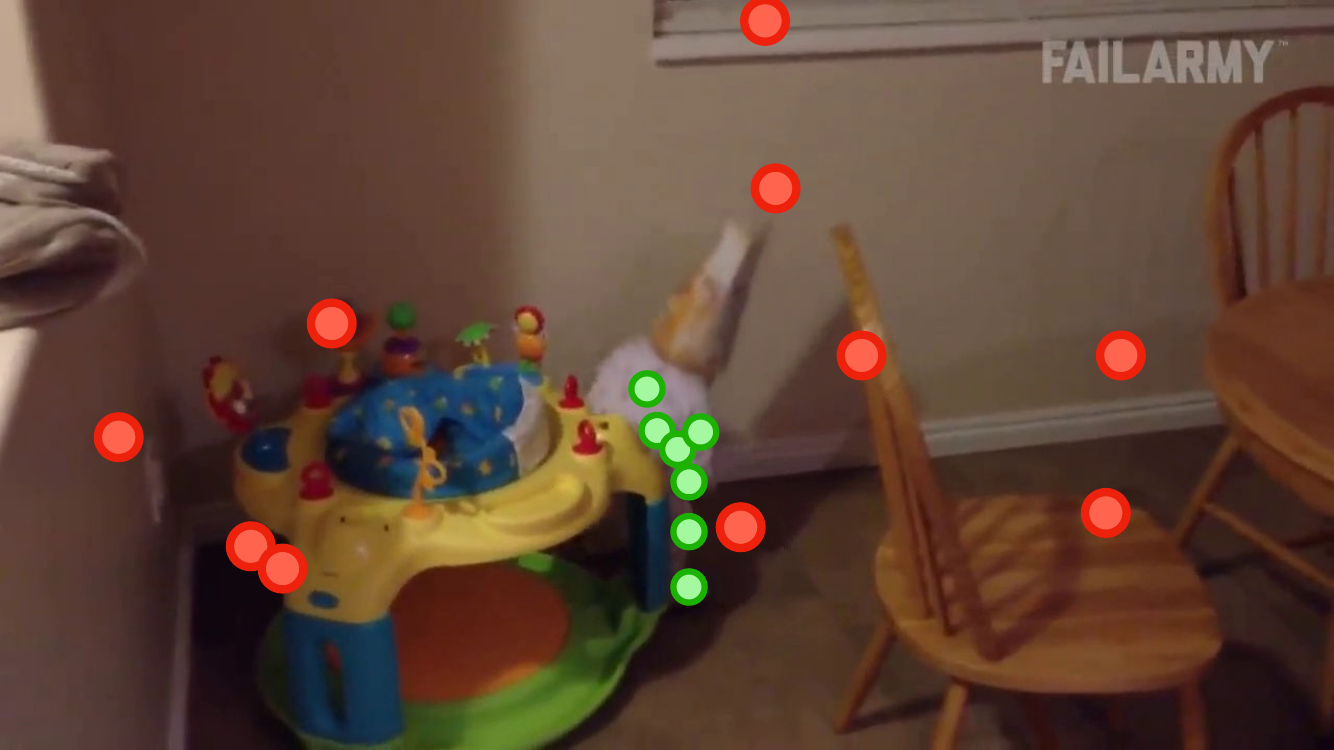}
        \vspace{-16pt}
    \end{subfigure}
    \vspace{5pt}

    \begin{subfigure}[b]{0.32\linewidth}
        \includegraphics[width=\mysize\linewidth]{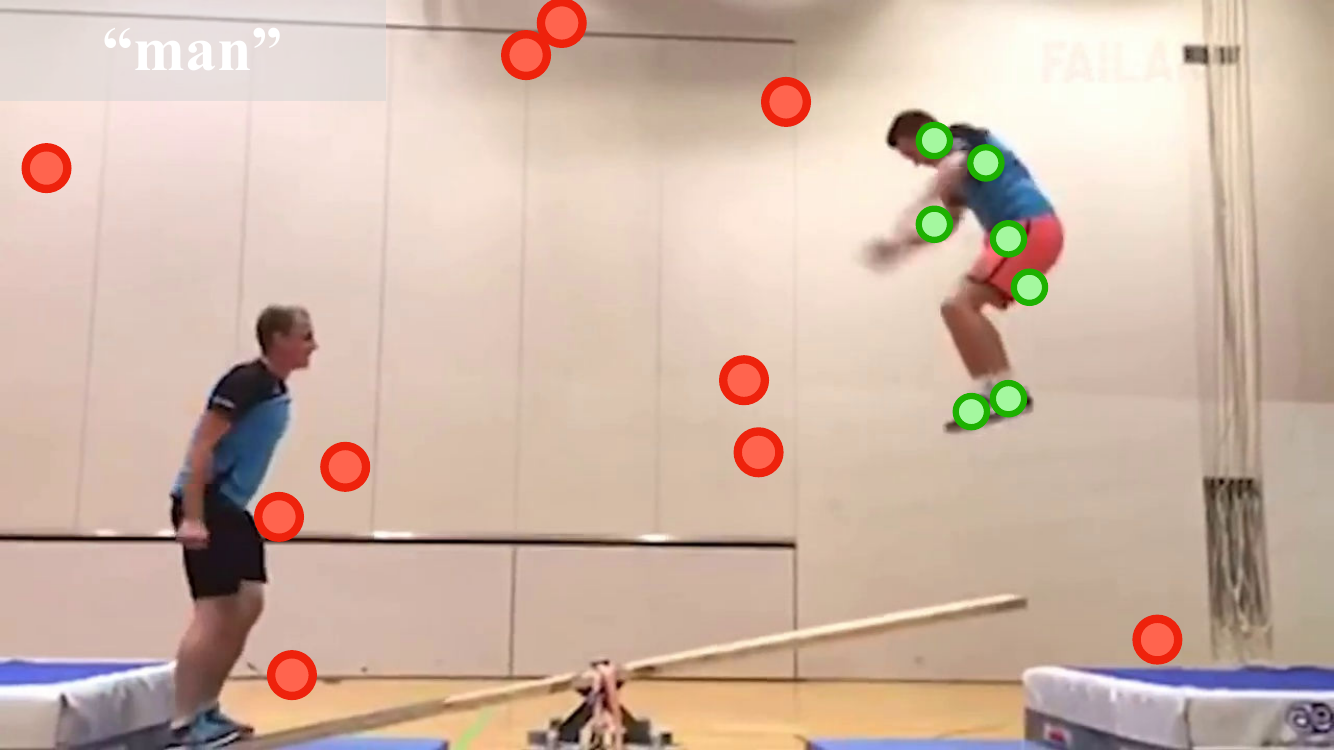}
        \vspace{-16pt}
    \end{subfigure}
    \begin{subfigure}[b]{0.32\linewidth}
     \includegraphics[width=\mysize\linewidth]{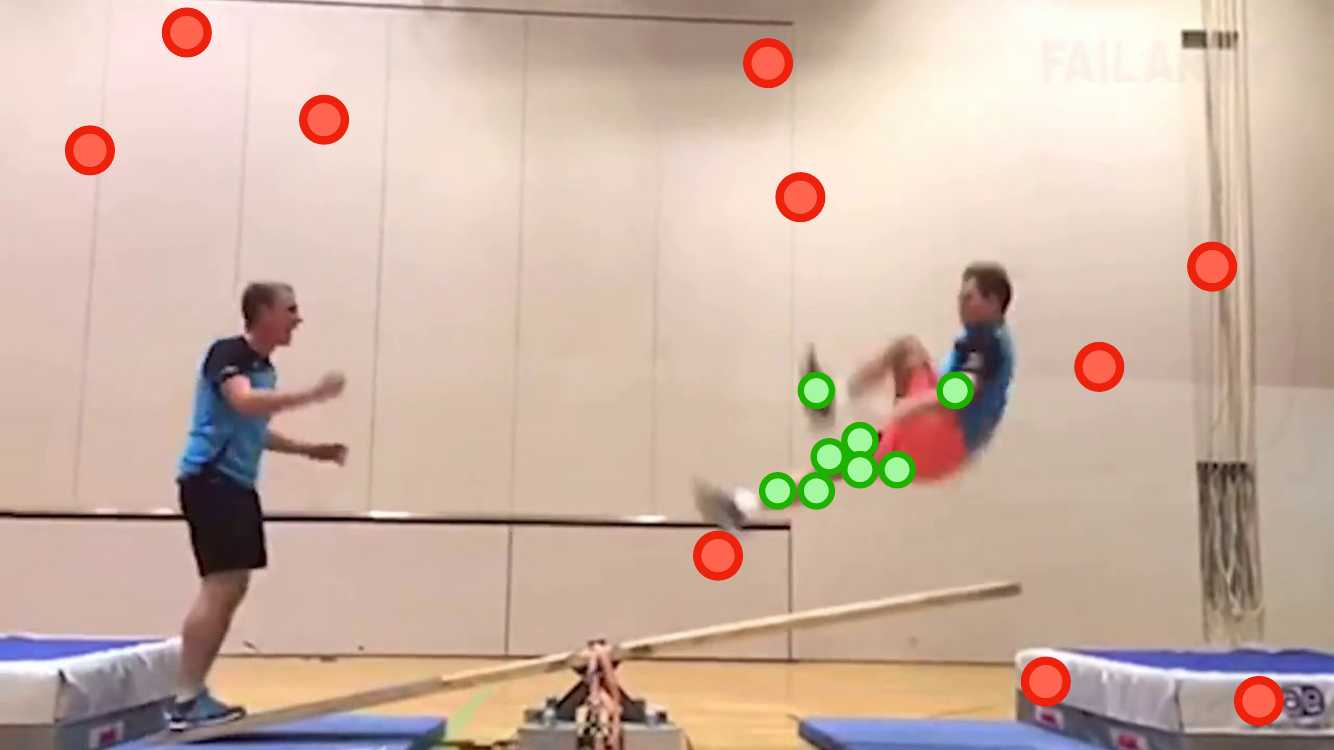}
        \vspace{-16pt}
    \end{subfigure}
    \begin{subfigure}[b]{0.32\linewidth}
     \includegraphics[width=\mysize\linewidth]{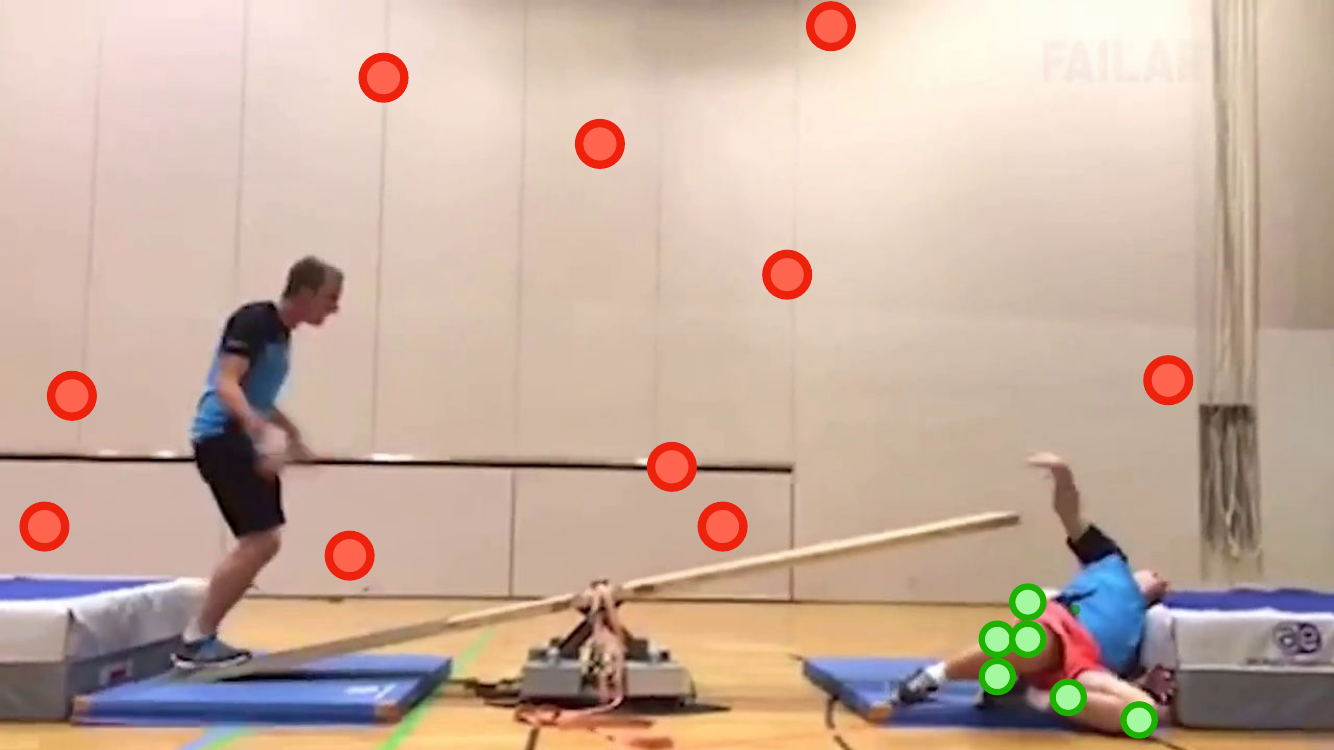}
        \vspace{-16pt}
    \end{subfigure}
    \vspace{5pt}

    \begin{subfigure}[b]{0.32\linewidth}
        \includegraphics[width=\mysize\linewidth]{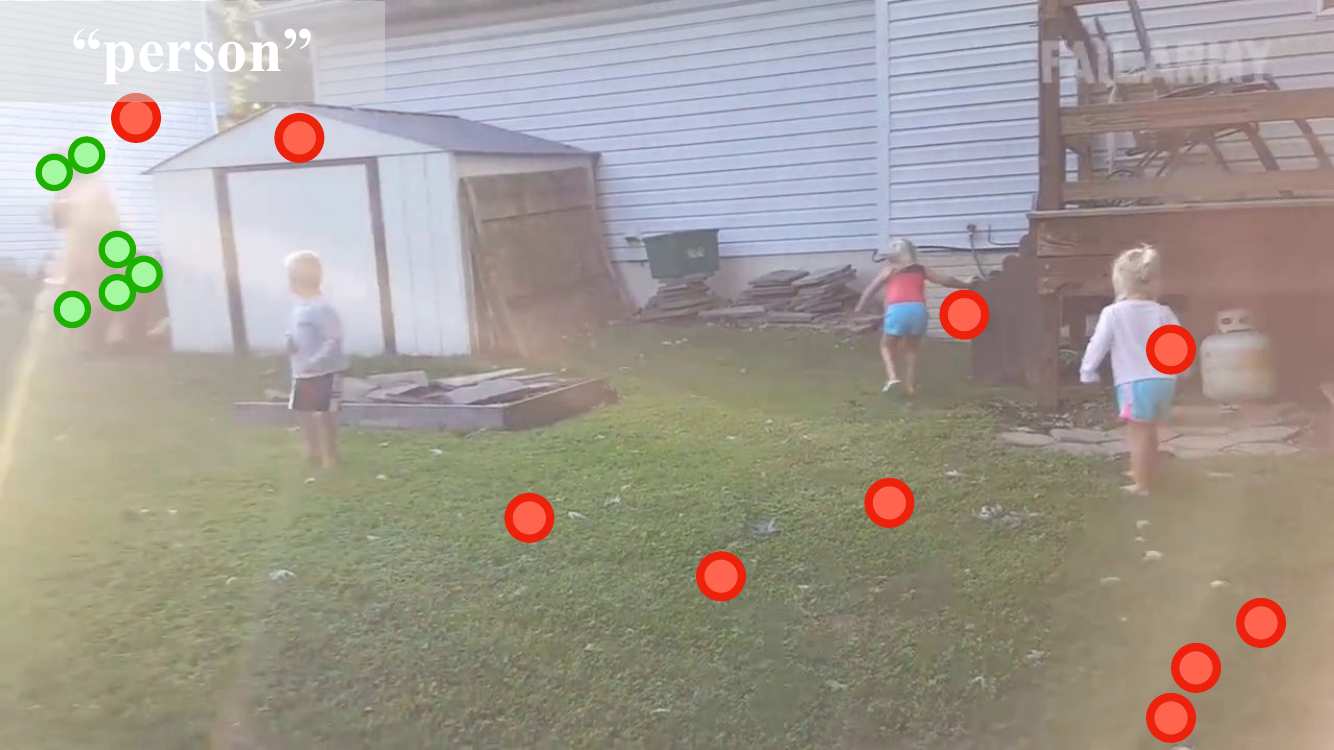}
        \vspace{-16pt}
    \end{subfigure}
    \begin{subfigure}[b]{0.32\linewidth}
     \includegraphics[width=\mysize\linewidth]{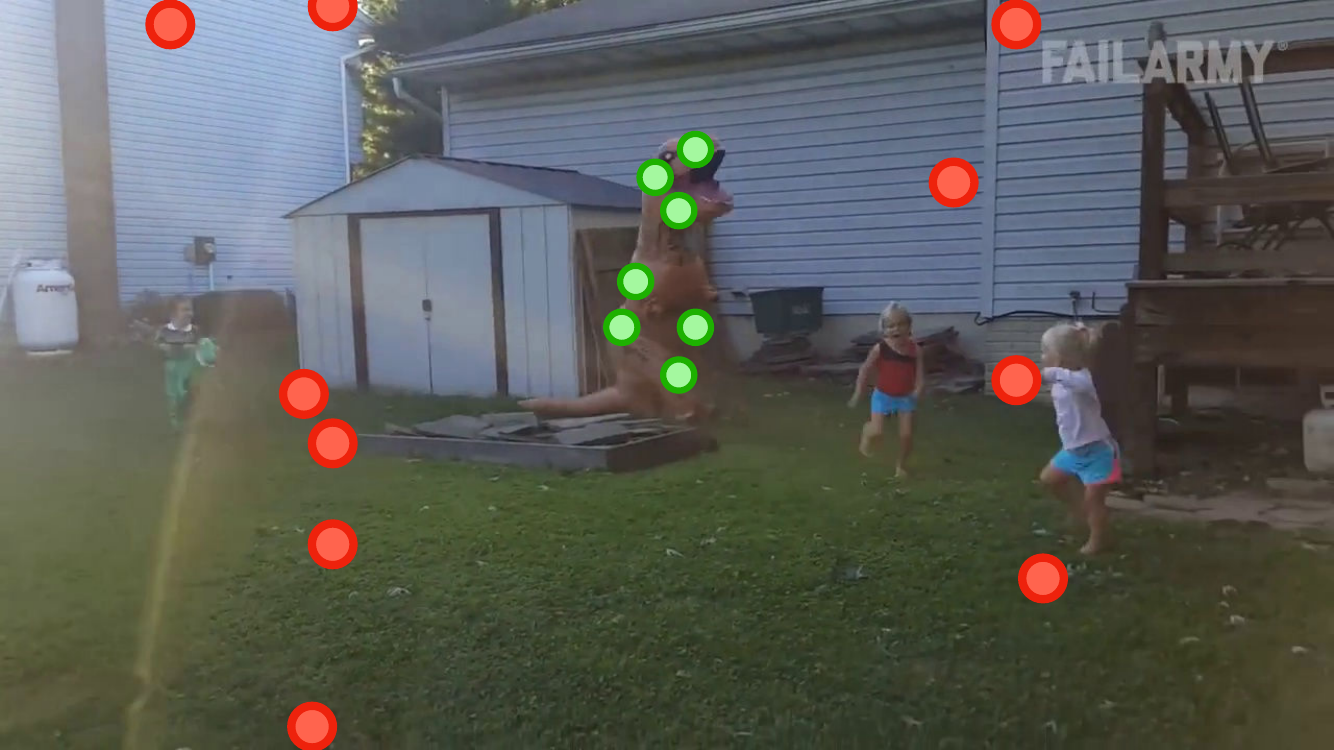}
        \vspace{-16pt}
    \end{subfigure}
    \begin{subfigure}[b]{0.32\linewidth}
     \includegraphics[width=\mysize\linewidth]{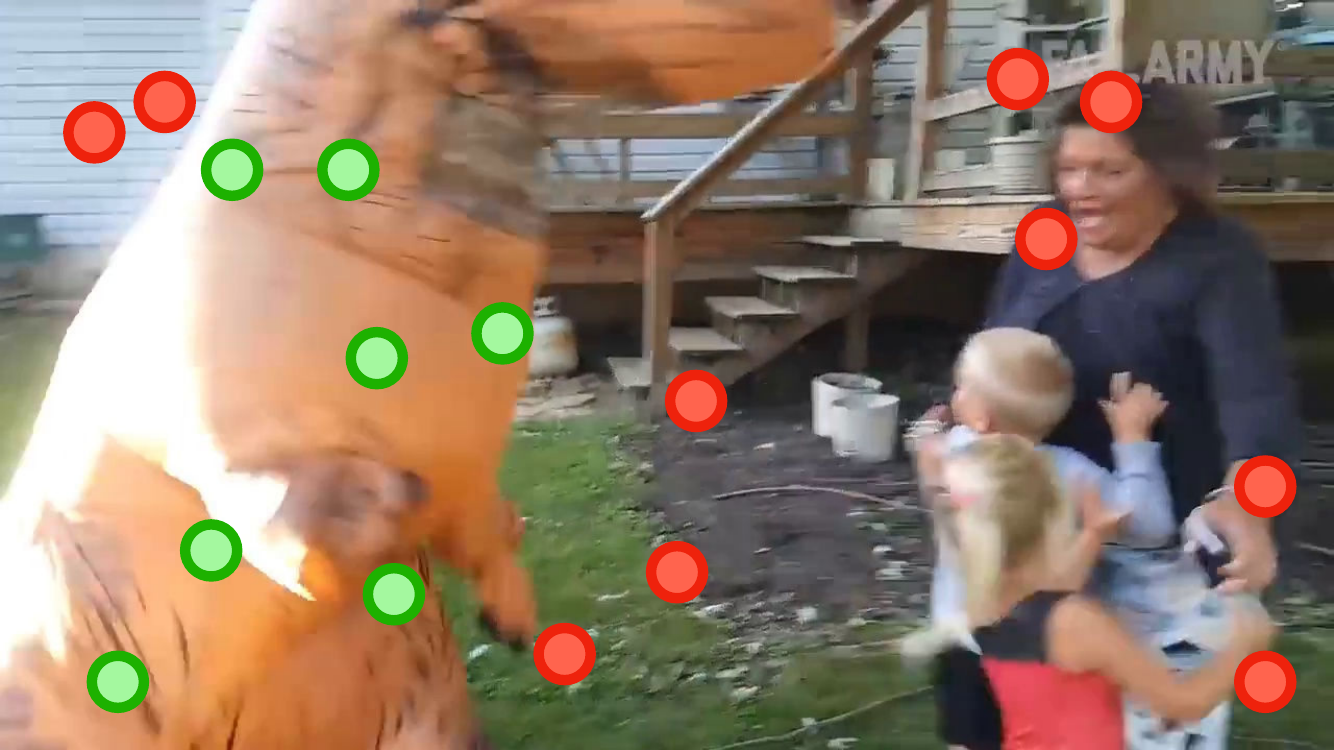}
        \vspace{-16pt}
    \end{subfigure}
    \vspace{5pt}

    \begin{subfigure}[b]{0.32\linewidth}
        \includegraphics[width=\mysize\linewidth]{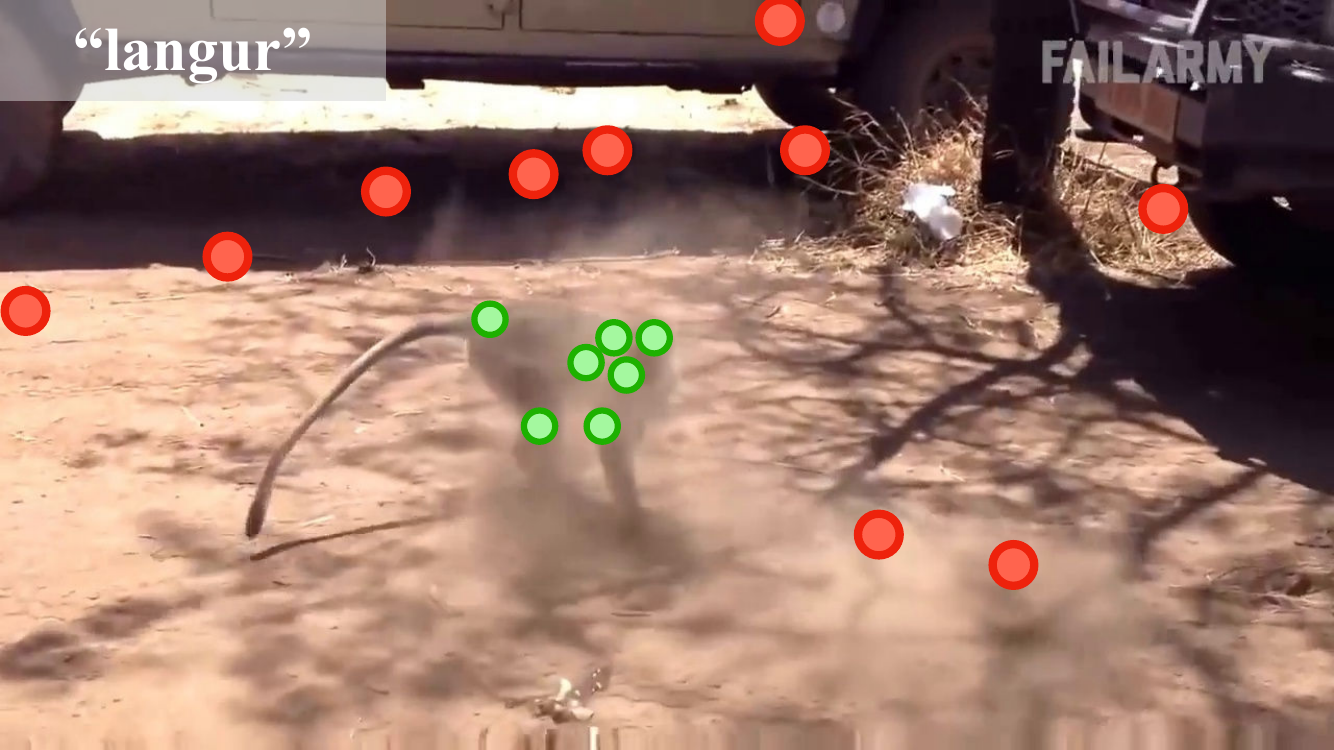}
        \vspace{-16pt}
    \end{subfigure}
    \begin{subfigure}[b]{0.32\linewidth}
     \includegraphics[width=\mysize\linewidth]{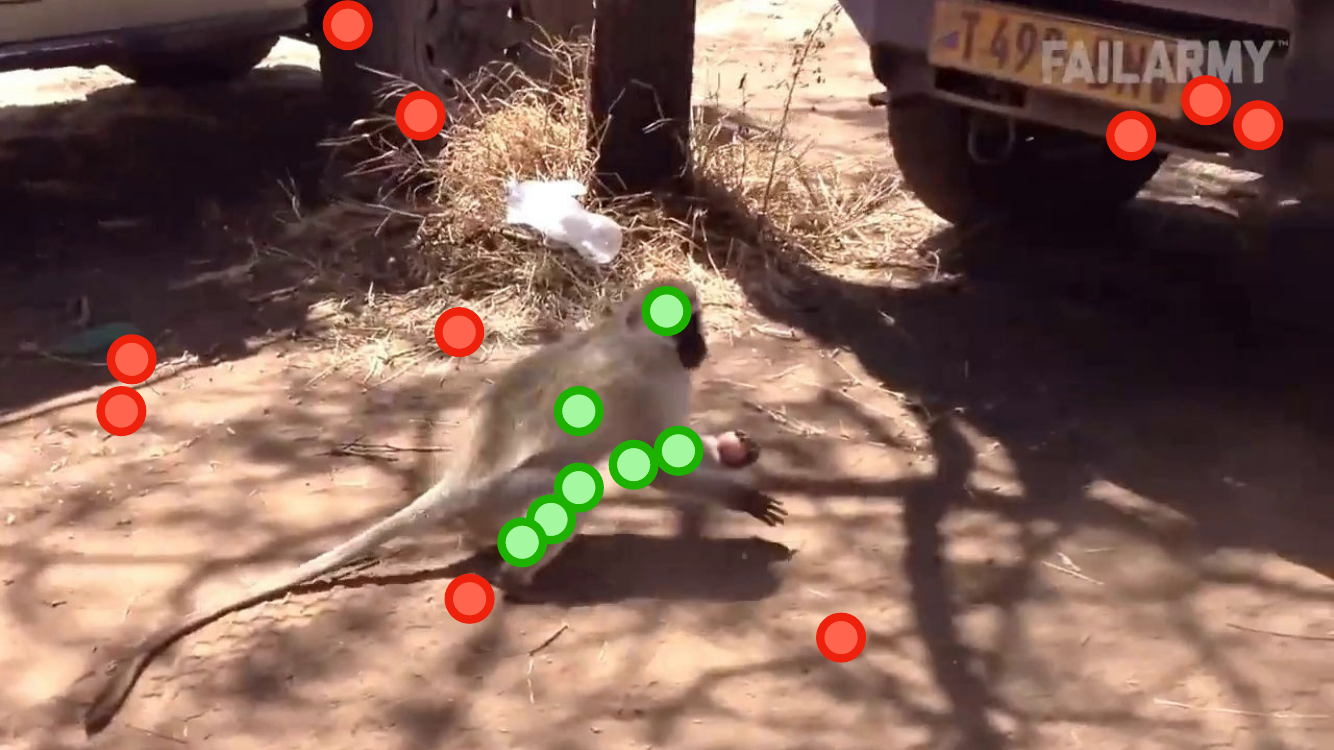}
        \vspace{-16pt}
    \end{subfigure}
    \begin{subfigure}[b]{0.32\linewidth}
     \includegraphics[width=\mysize\linewidth]{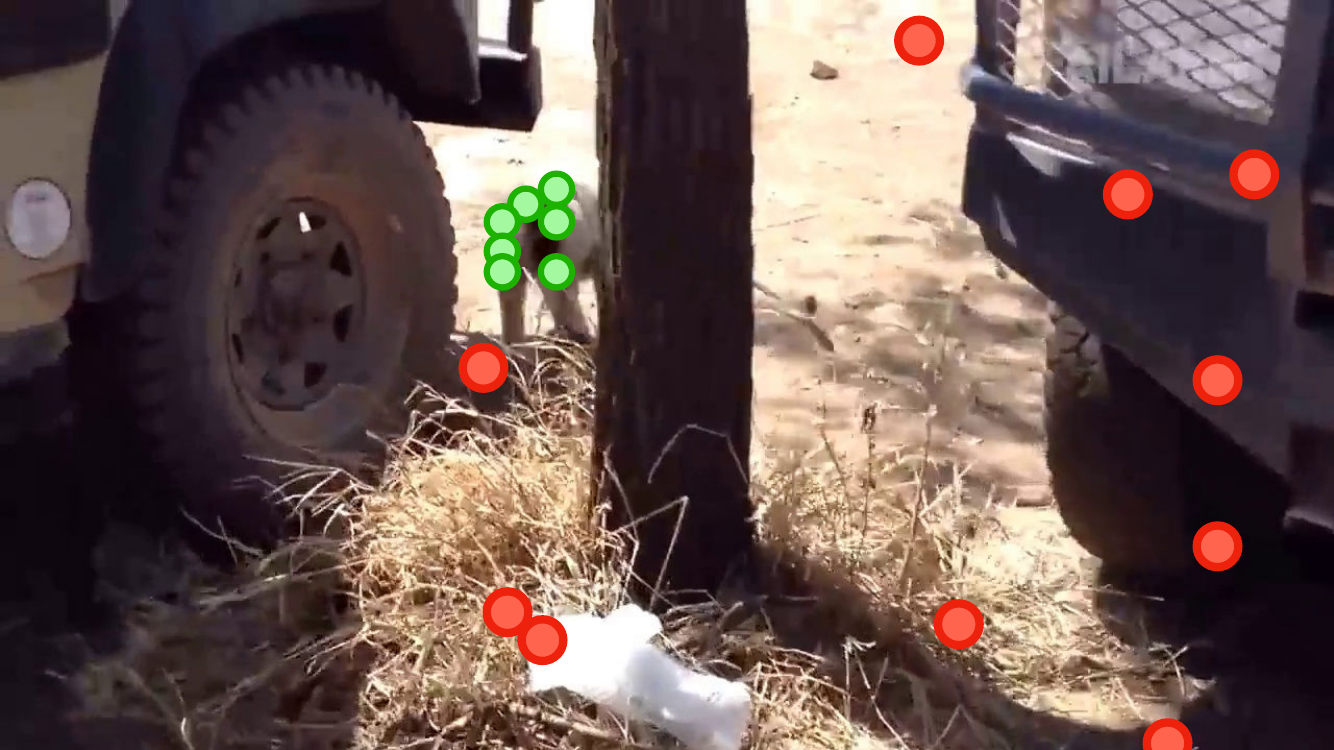}
        \vspace{-16pt}
    \end{subfigure}
    \vspace{5pt}
    
    \begin{subfigure}[b]{0.32\linewidth}
        \includegraphics[width=\mysize\linewidth]{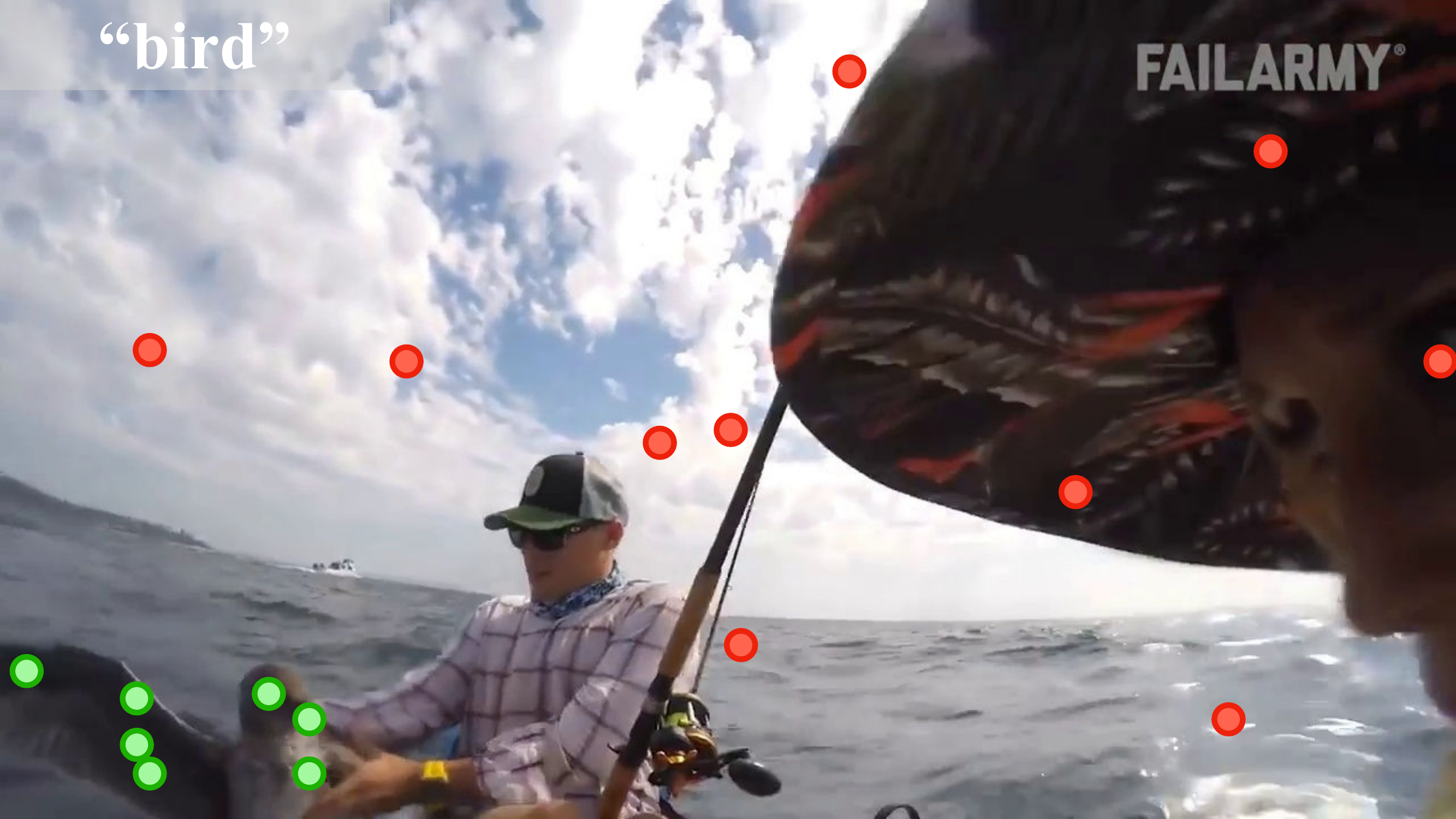}
        \vspace{-16pt}
    \end{subfigure}
    \begin{subfigure}[b]{0.32\linewidth}
     \includegraphics[width=\mysize\linewidth]{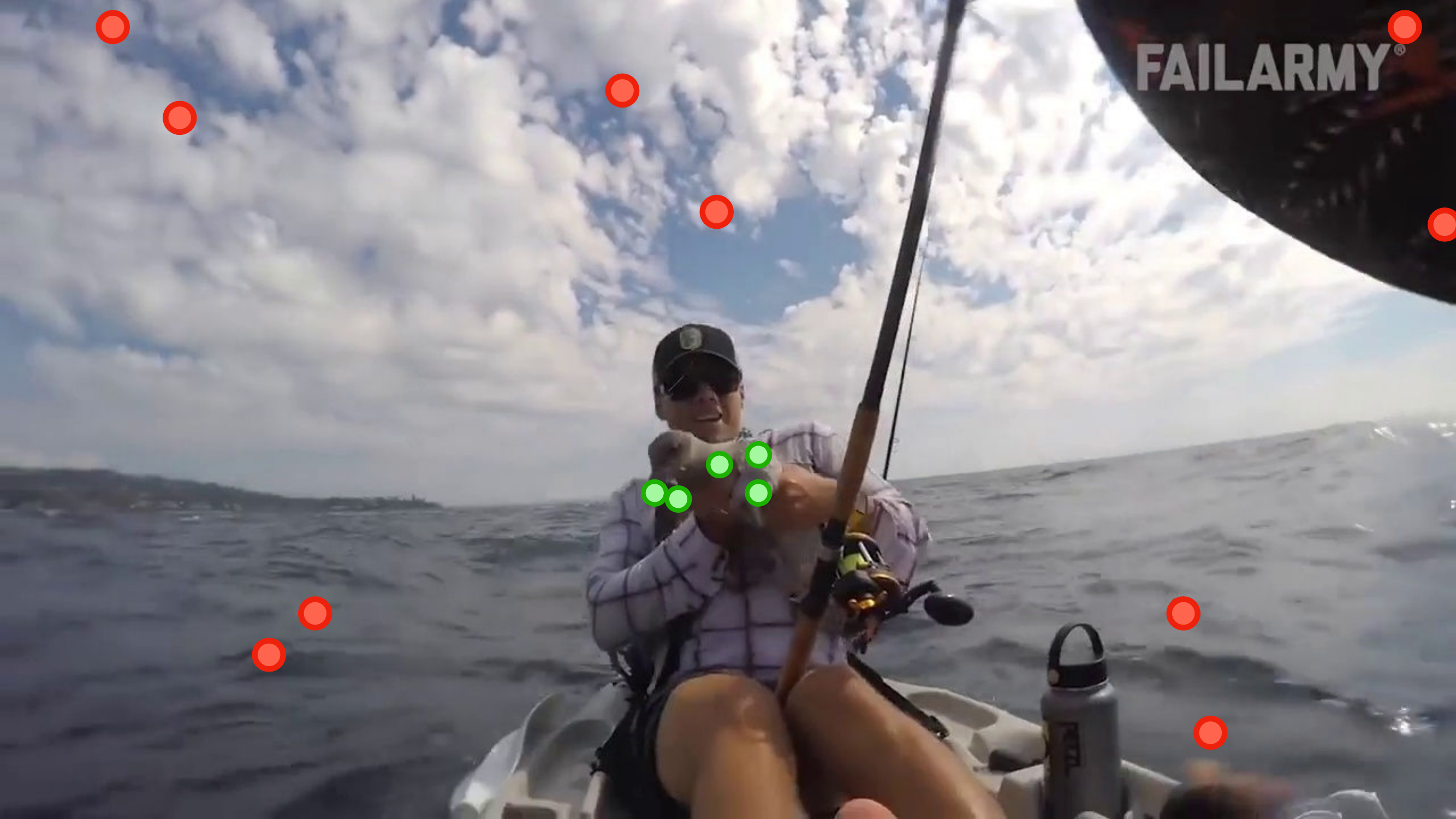}
        \vspace{-16pt}
    \end{subfigure}
    \begin{subfigure}[b]{0.32\linewidth}
     \includegraphics[width=\mysize\linewidth]{figures/supplementary/additional_example_point_annotations/oops/417c525c41f9ac2e35c91b389de6ea97/000064.pdf}
        \vspace{-16pt}
    \end{subfigure}
    \vspace{5pt}

    \caption{\textbf{Additional example point annotations for \pointvos{} Oops.} We are able to have multi-modal point annotations in cluttered scenes (\textit{first row}), fast motion (\textit{third row}), challenging lighting conditions (\textit{fourth row}), and motion blur (\textit{fifth row}). \textcolor{green}{Green} dots represent positive points and \textcolor{red}{red} dots negative points.}
    \label{fig:add_point_anns_oops}
\end{figure*}

\begin{figure*}[t]
    \centering
    \setlength{\fboxsep}{0.32pt}
    \begin{subfigure}[b]{0.32\linewidth}
        \includegraphics[width=\mysize\linewidth]{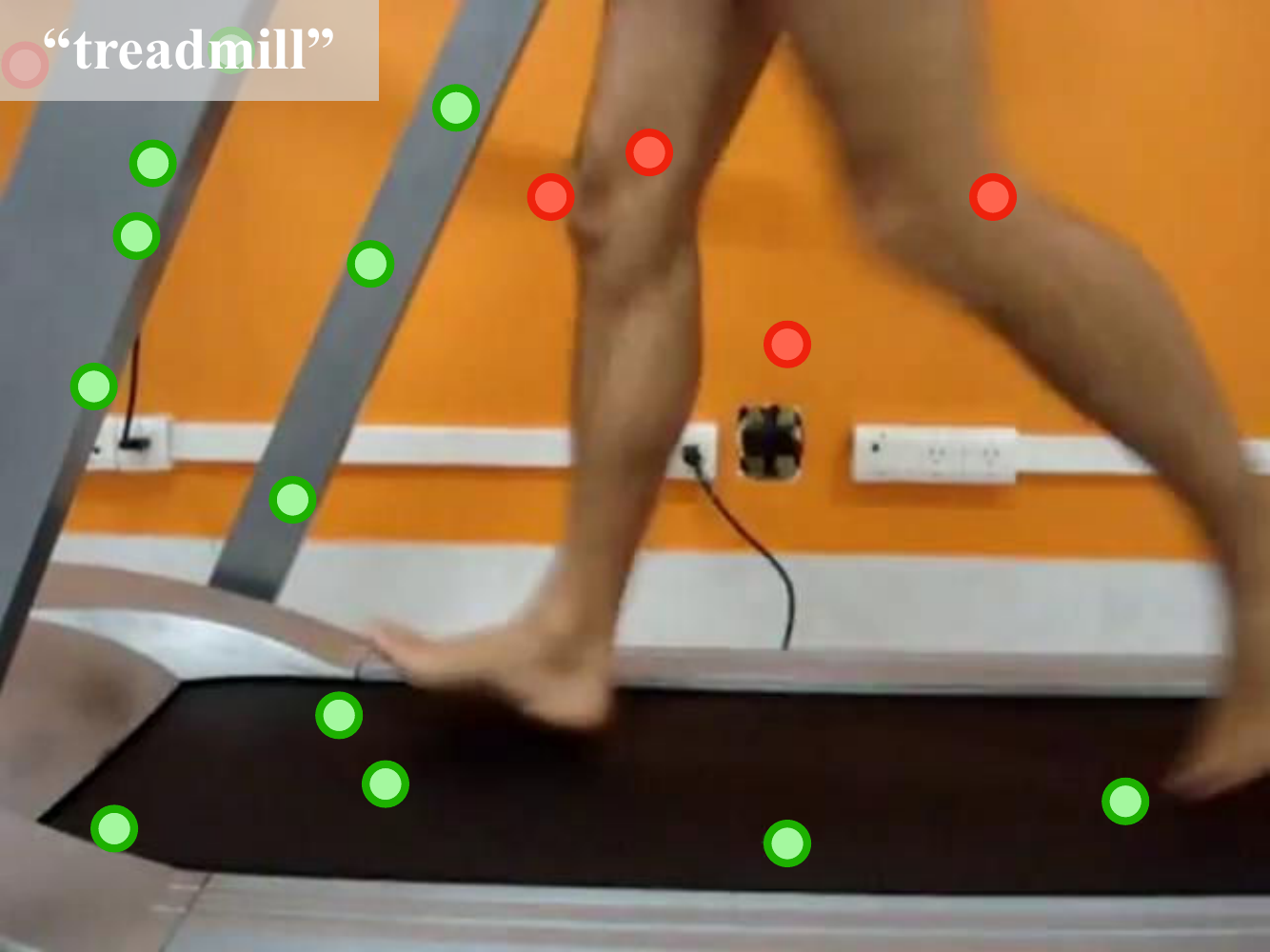}
        \vspace{-16pt}
    \end{subfigure}
    \begin{subfigure}[b]{0.32\linewidth}
     \includegraphics[width=\mysize\linewidth]{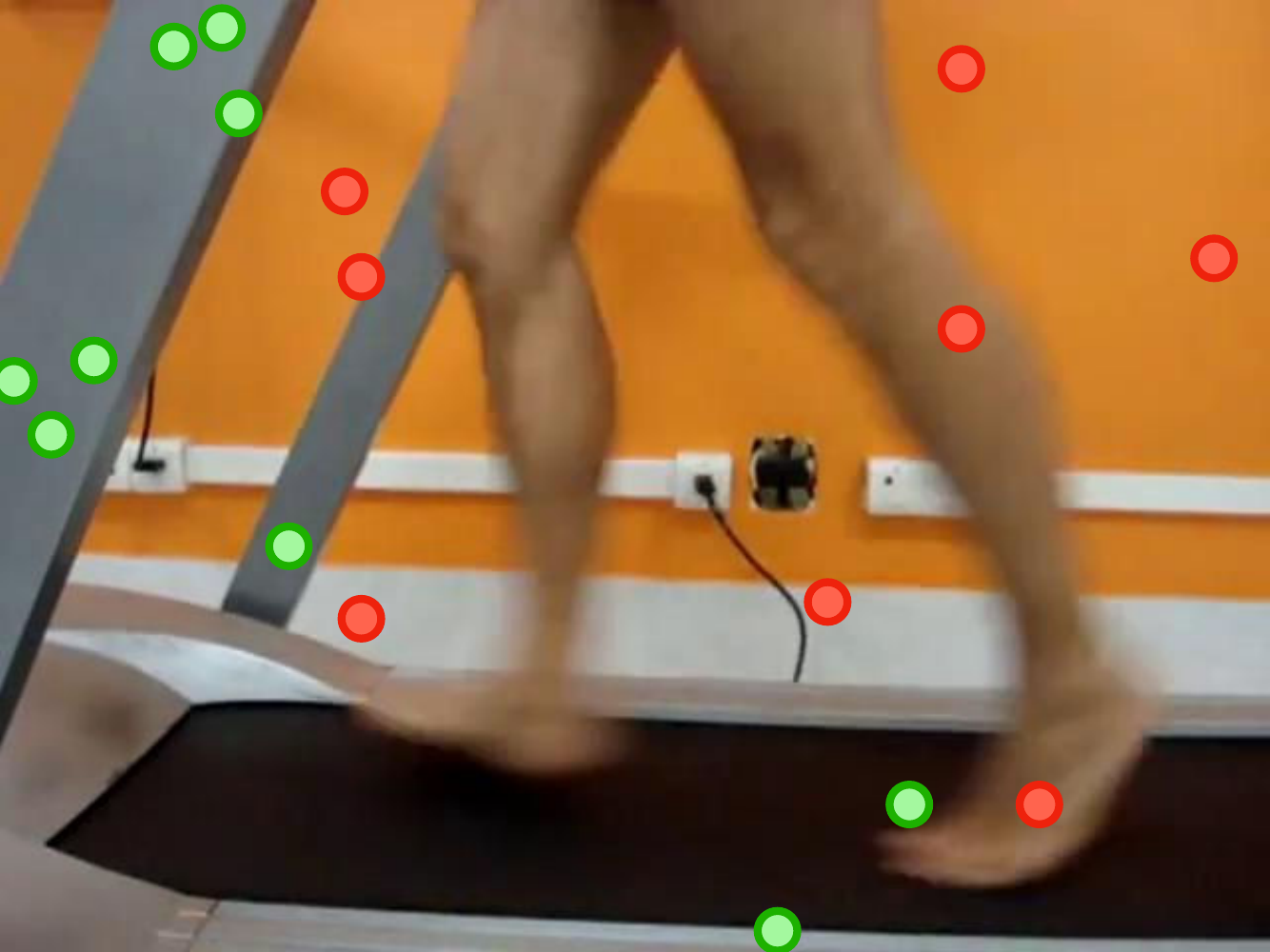}
        \vspace{-16pt}
    \end{subfigure}
    \begin{subfigure}[b]{0.32\linewidth}
     \includegraphics[width=\mysize\linewidth]{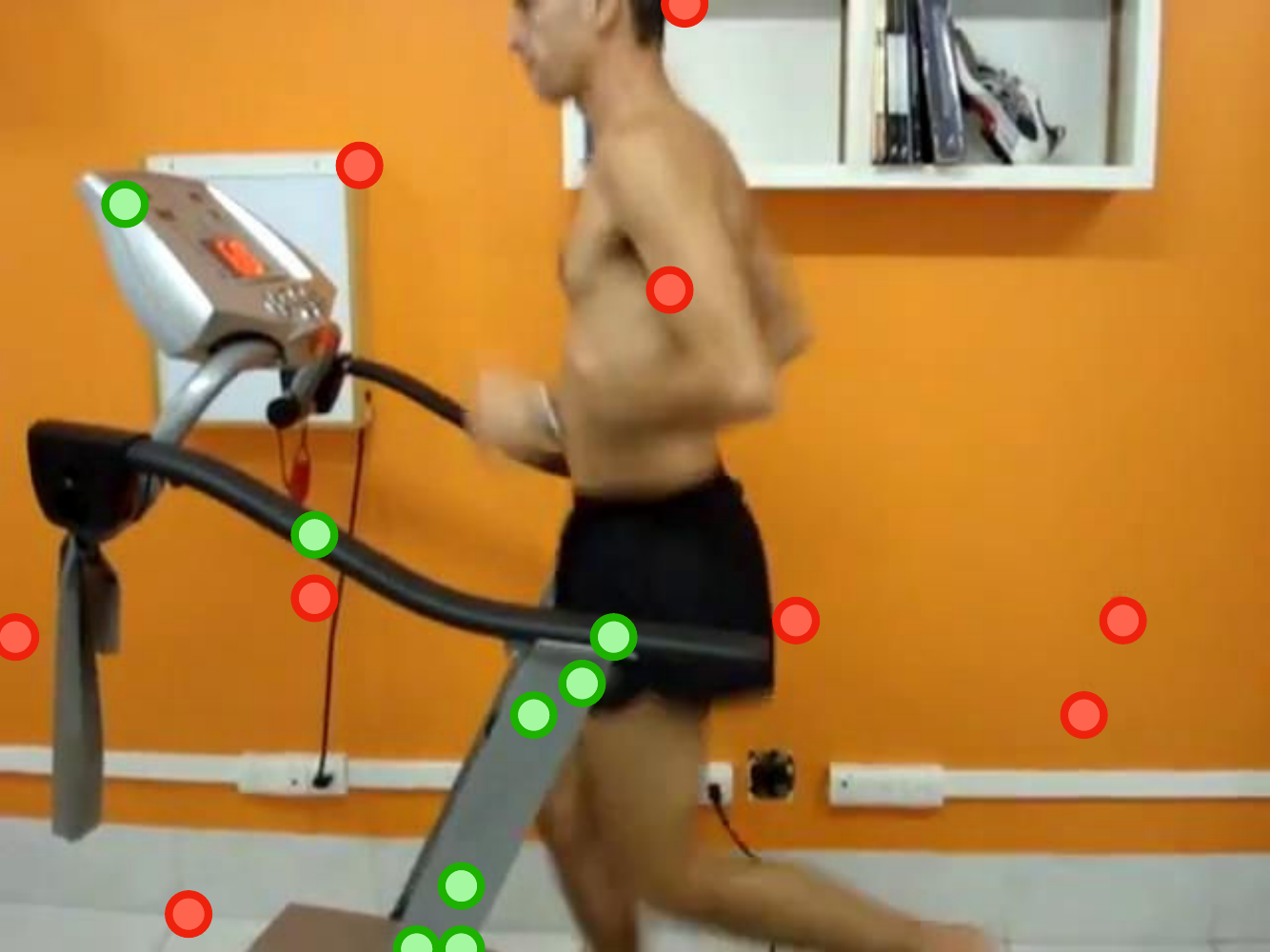}
        \vspace{-16pt}
    \end{subfigure}
    \vspace{5pt}

    \begin{subfigure}[b]{0.32\linewidth}
     \includegraphics[width=\mysize\linewidth]{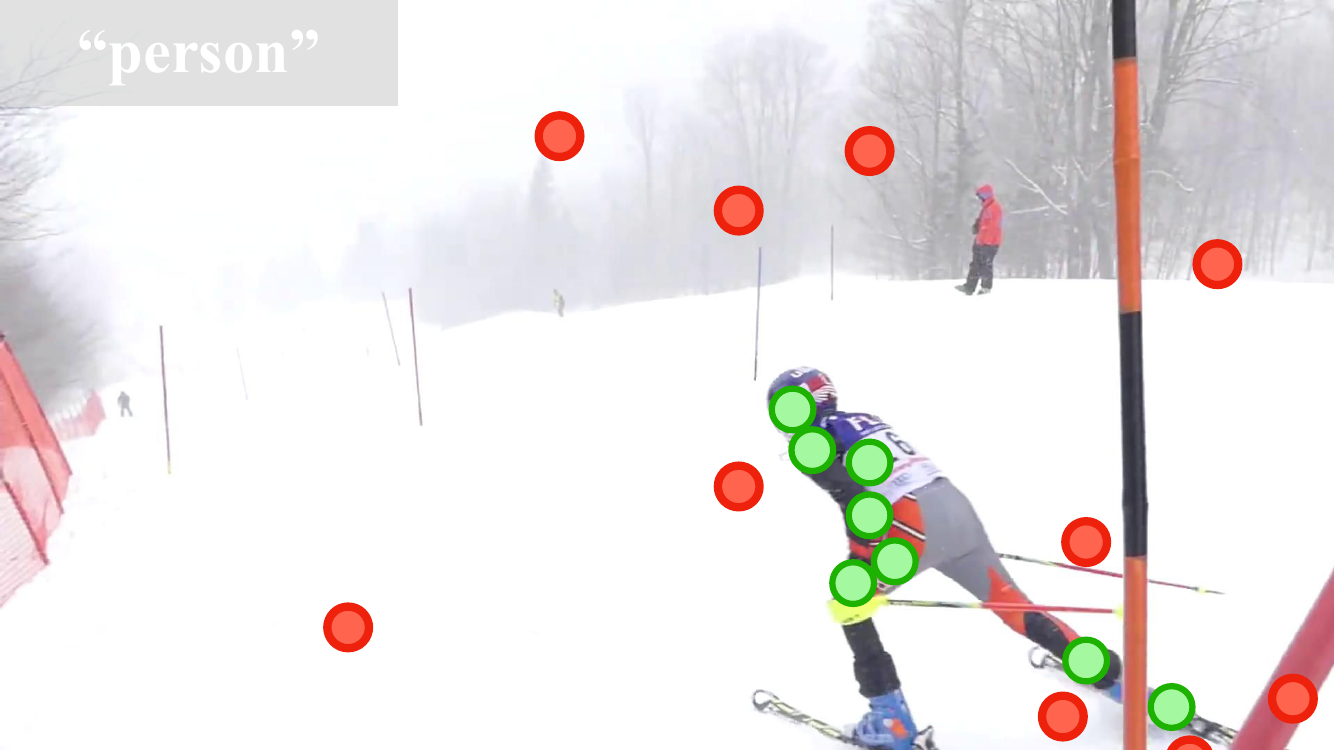}
        \vspace{-16pt}
    \end{subfigure}
    \begin{subfigure}[b]{0.32\linewidth}
        \includegraphics[width=\mysize\linewidth]{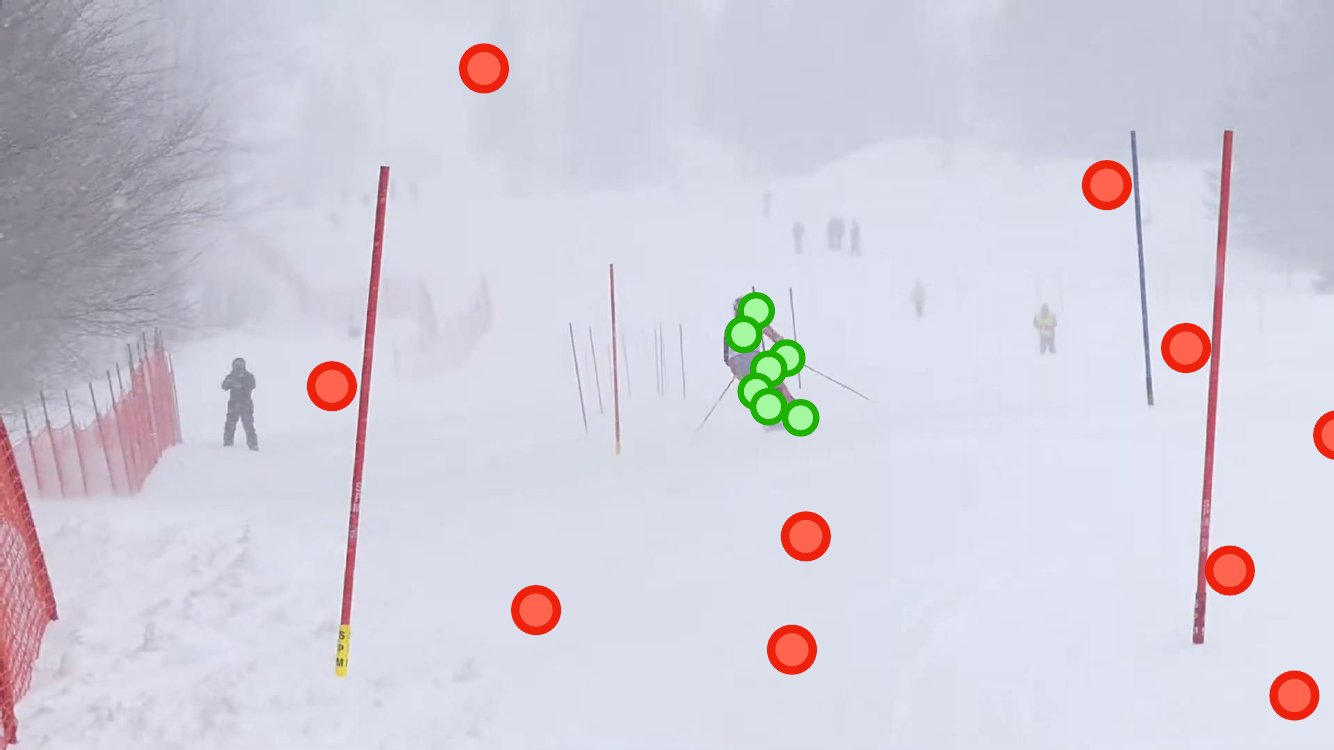}
        \vspace{-16pt}
    \end{subfigure}
    \begin{subfigure}[b]{0.32\linewidth}
     \includegraphics[width=\mysize\linewidth]{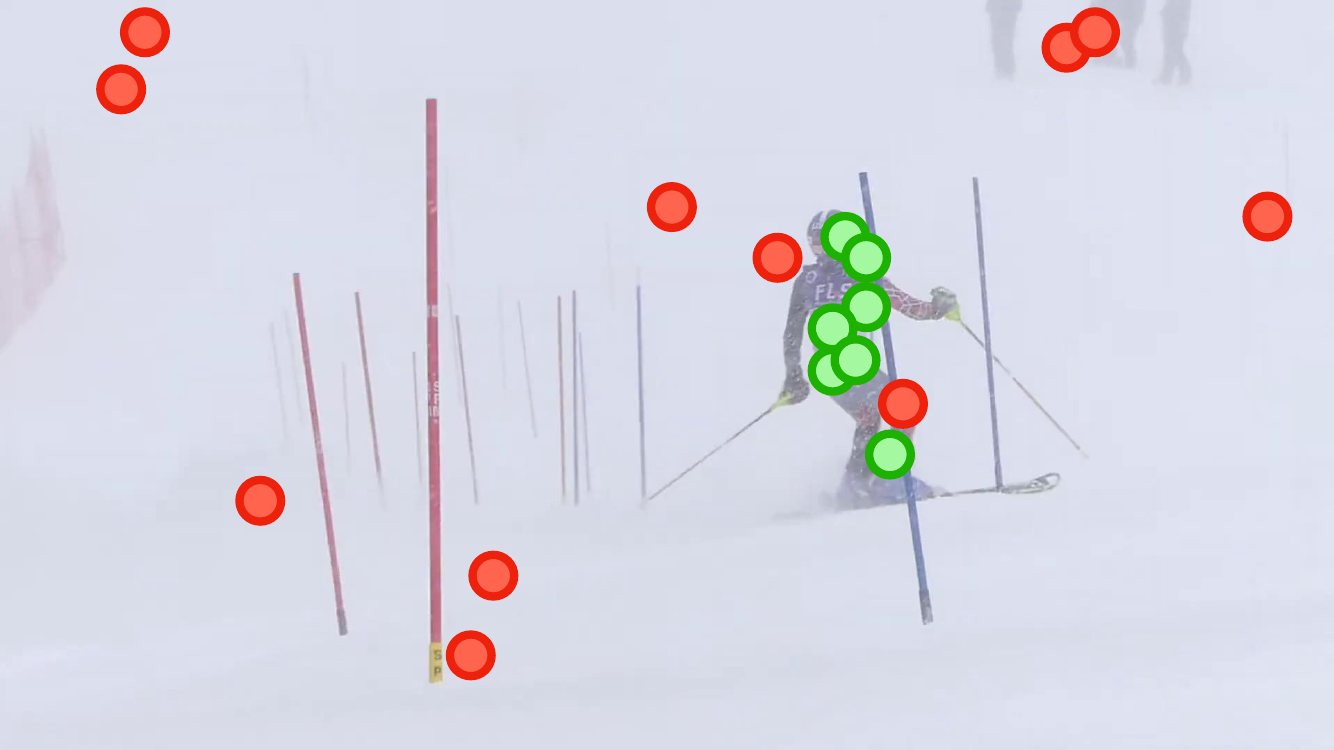}
        \vspace{-16pt}
    \end{subfigure}
    \vspace{5pt}
    
    \begin{subfigure}[b]{0.32\linewidth}
        \includegraphics[width=\mysize\linewidth]{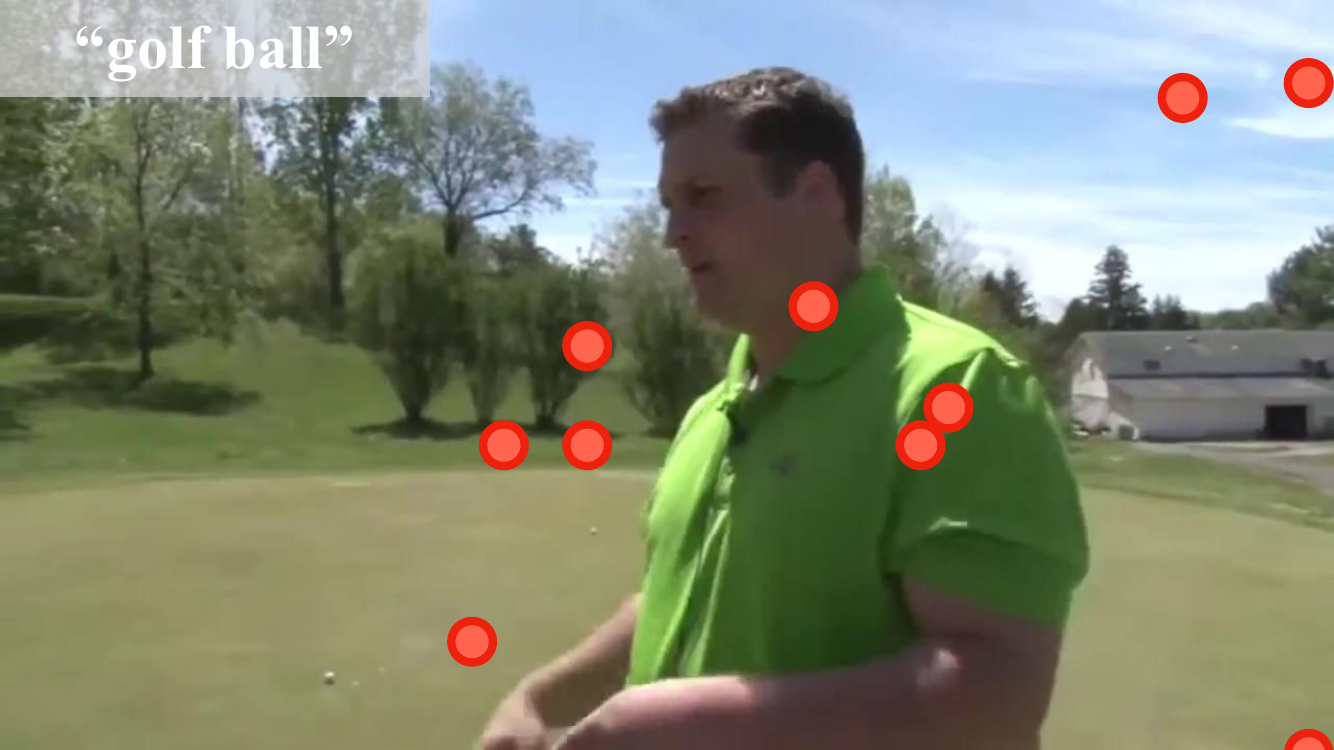}
        \vspace{-16pt}
    \end{subfigure}
    \begin{subfigure}[b]{0.32\linewidth}
        \includegraphics[width=\mysize\linewidth]{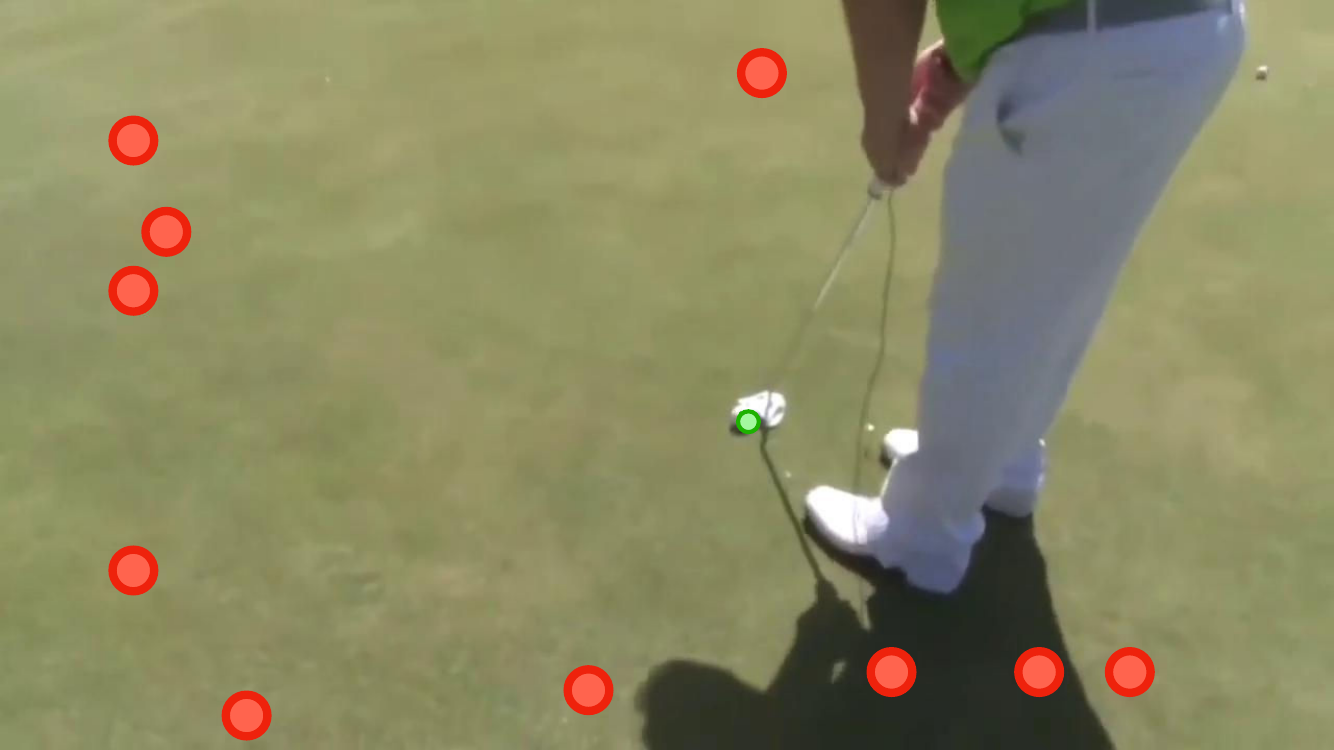}
        \vspace{-16pt}
    \end{subfigure}
    \begin{subfigure}[b]{0.32\linewidth}
     \includegraphics[width=\mysize\linewidth]{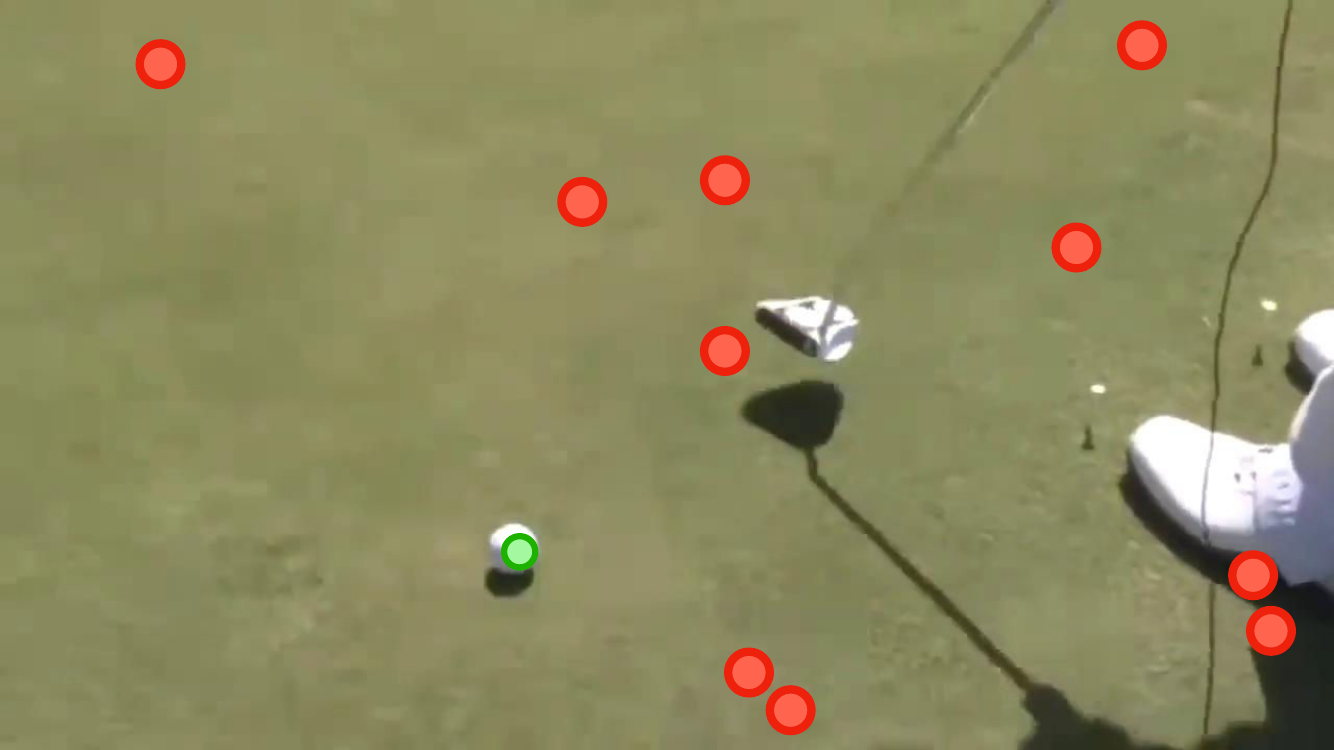}
        \vspace{-16pt}
    \end{subfigure}
    \vspace{5pt}

    \begin{subfigure}[b]{0.32\linewidth}
        \includegraphics[width=\mysize\linewidth]{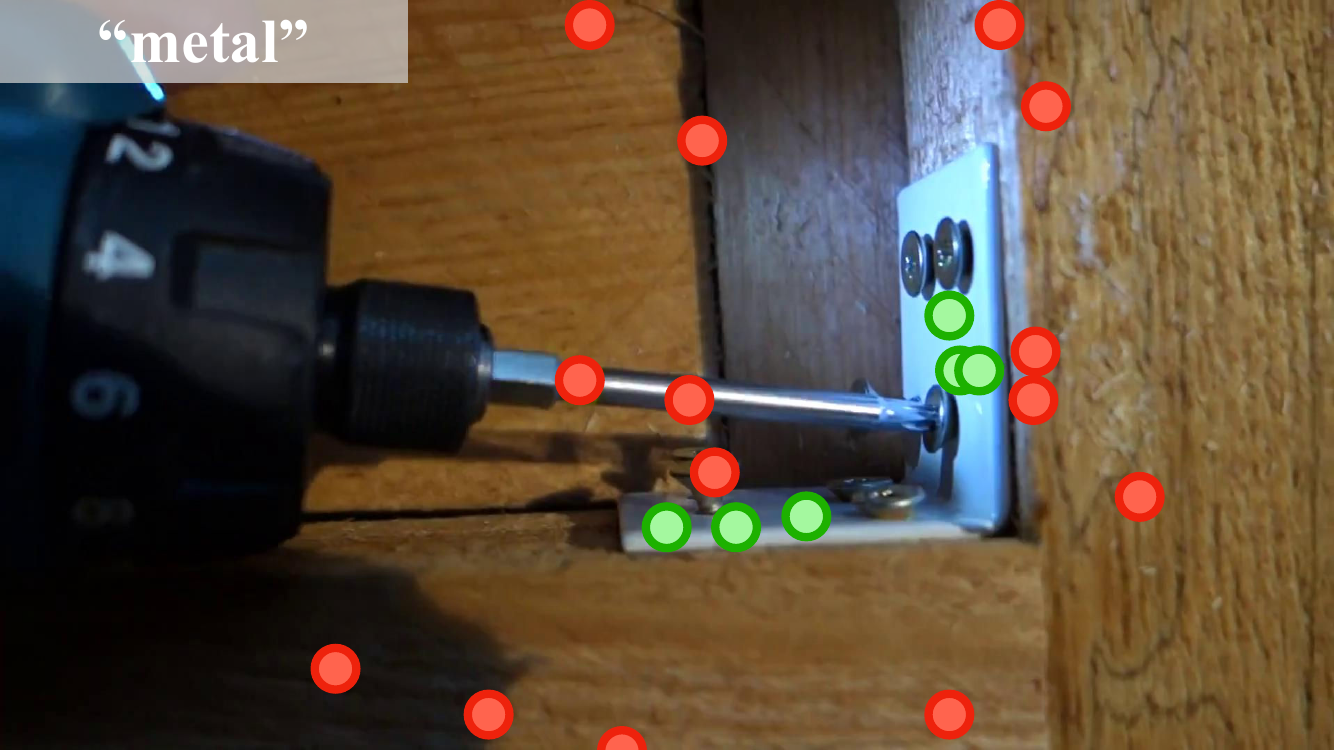}
        \vspace{-16pt}
    \end{subfigure}
    \begin{subfigure}[b]{0.32\linewidth}
     \includegraphics[width=\mysize\linewidth]{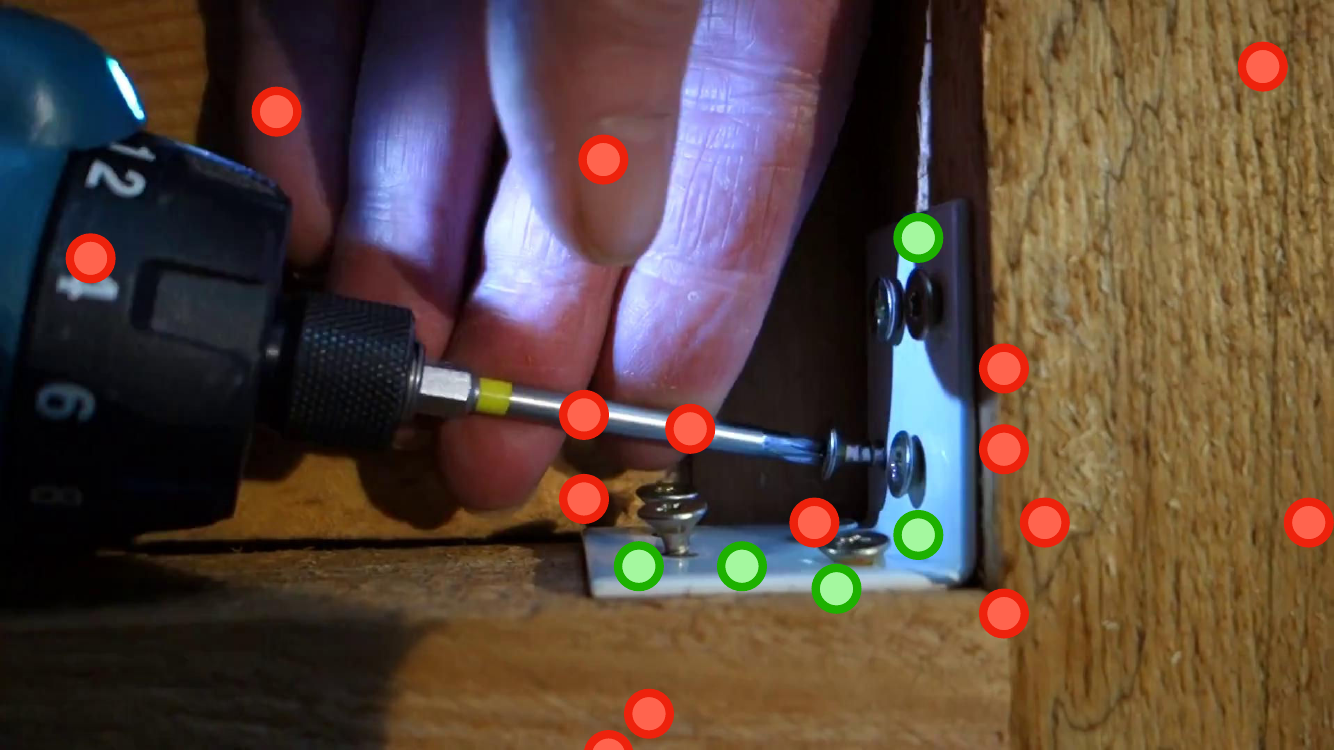}
        \vspace{-16pt}
    \end{subfigure}
    \begin{subfigure}[b]{0.32\linewidth}
     \includegraphics[width=\mysize\linewidth]{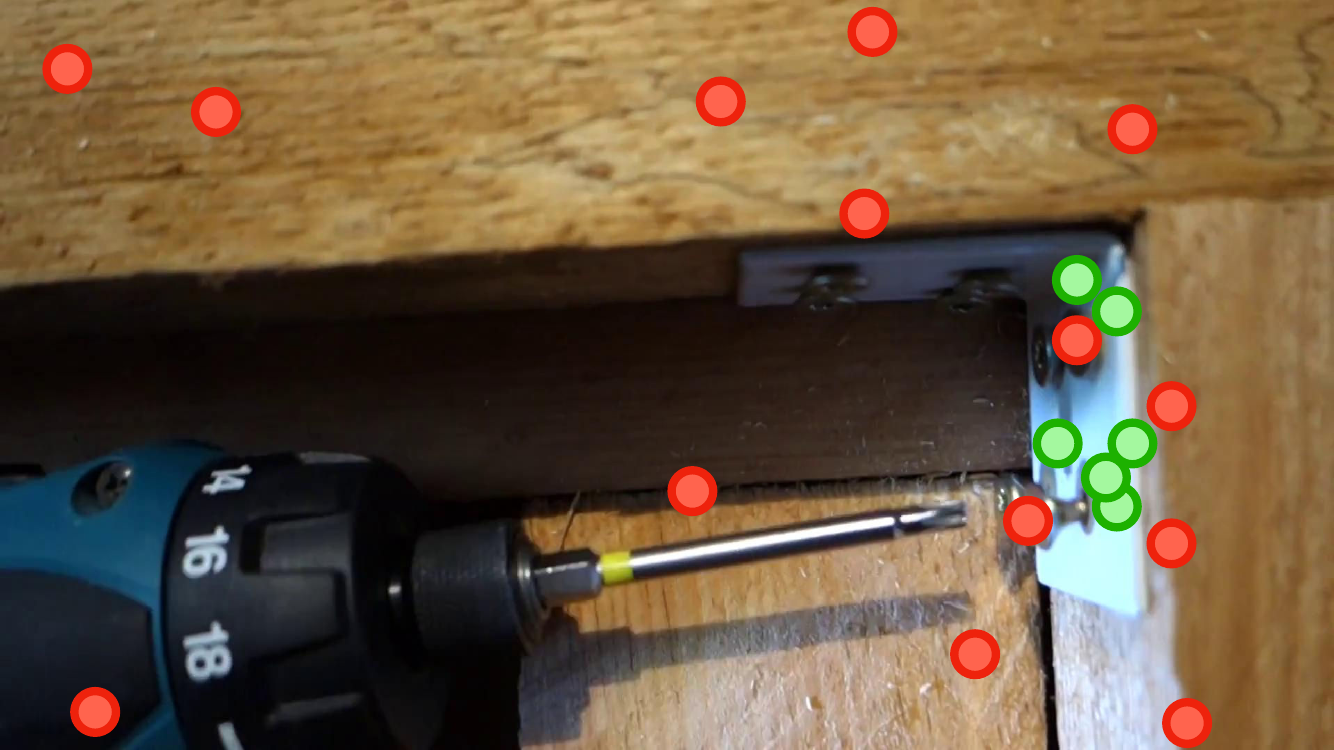}
        \vspace{-16pt}
    \end{subfigure}
    \vspace{5pt}

    \begin{subfigure}[b]{0.32\linewidth}
        \includegraphics[width=\mysize\linewidth]{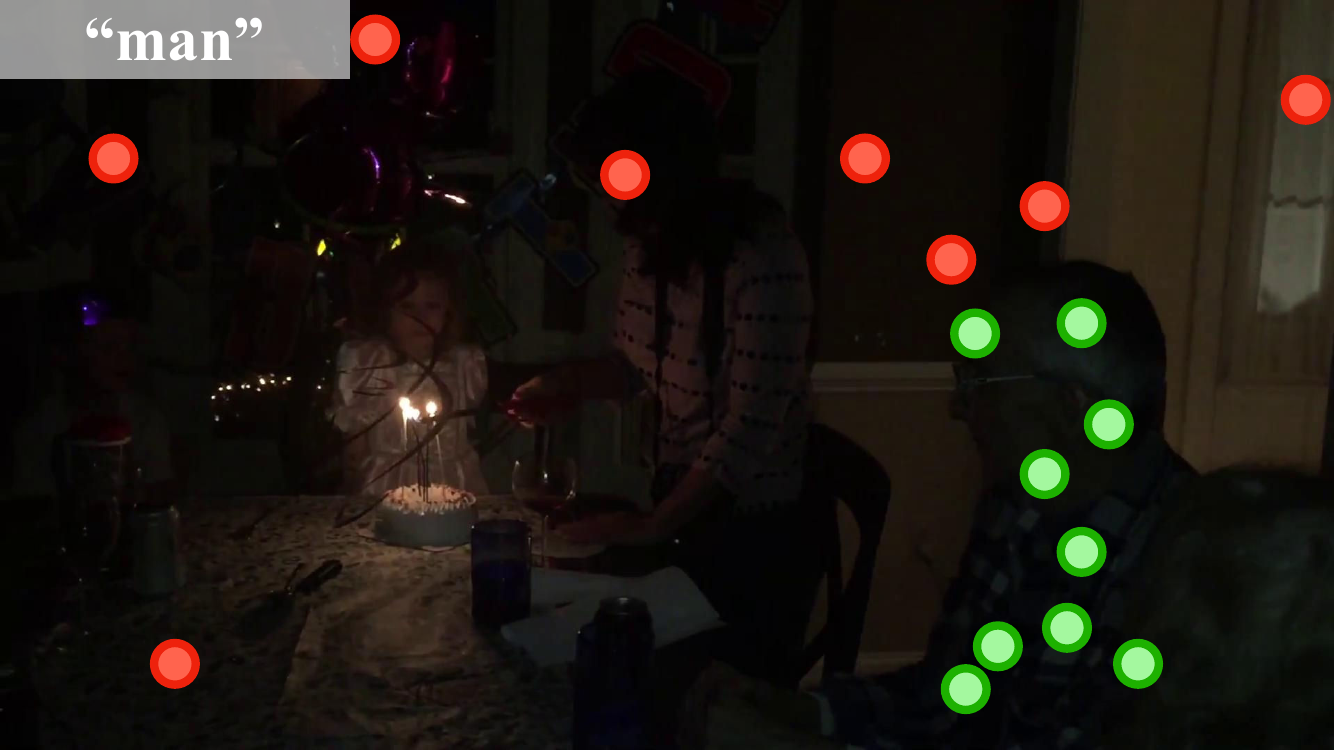}
        \vspace{-16pt}
    \end{subfigure}
    \begin{subfigure}[b]{0.32\linewidth}
     \includegraphics[width=\mysize\linewidth]{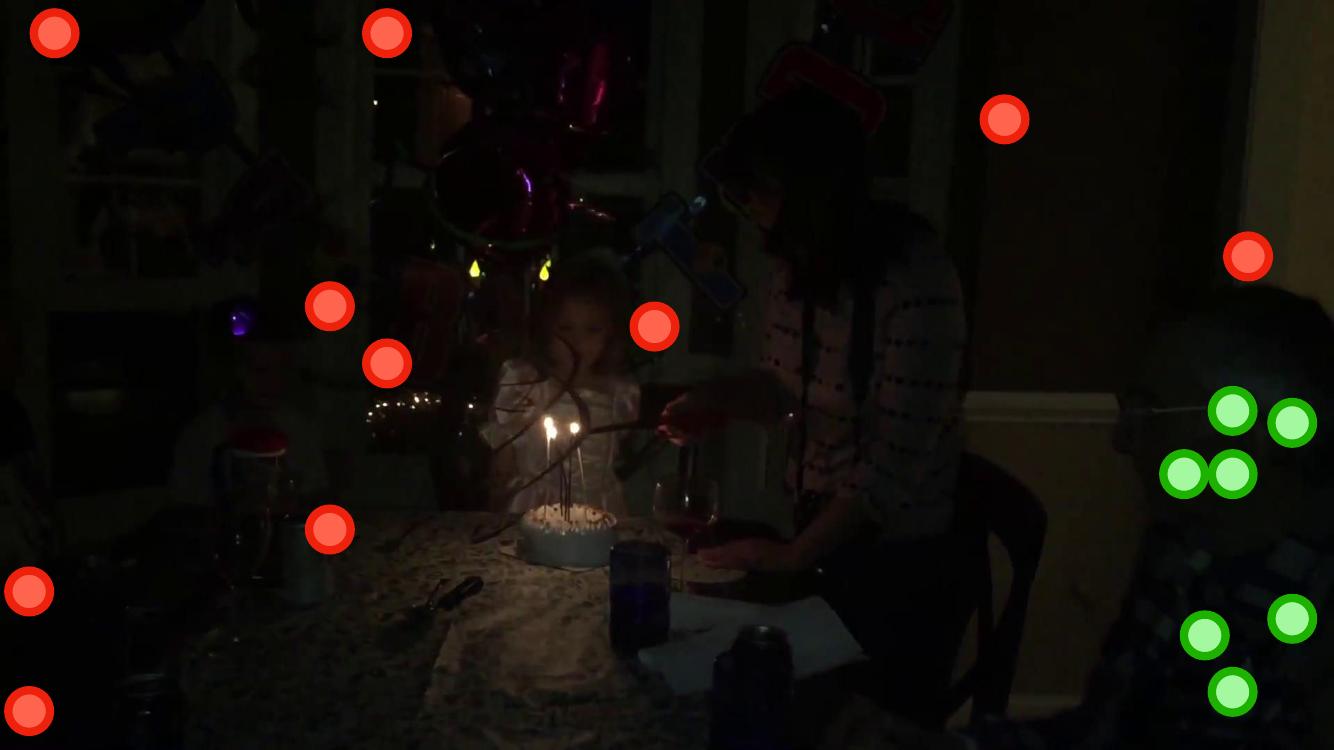}
        \vspace{-16pt}
    \end{subfigure}
    \begin{subfigure}[b]{0.32\linewidth}
     \includegraphics[width=\mysize\linewidth]{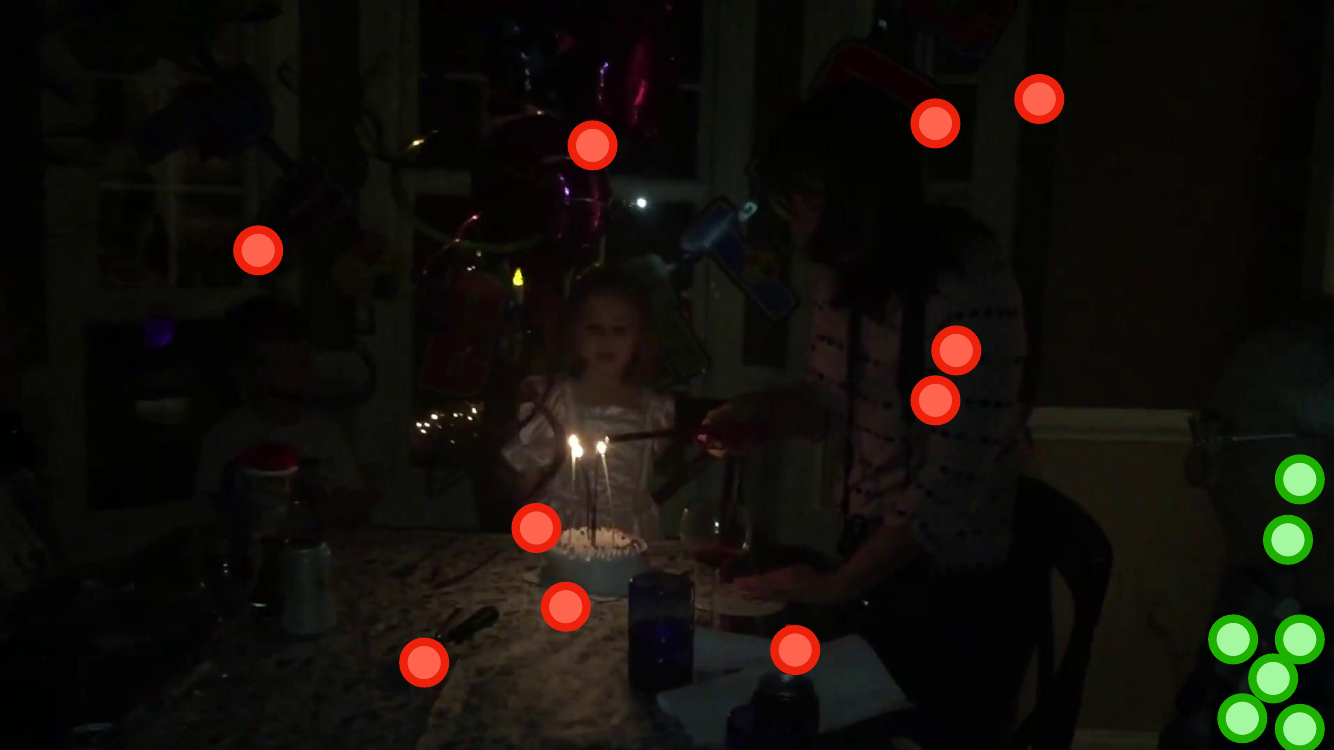}
        \vspace{-16pt}
    \end{subfigure}
    \vspace{5pt}

    \begin{subfigure}[b]{0.32\linewidth}
     \includegraphics[width=\mysize\linewidth]{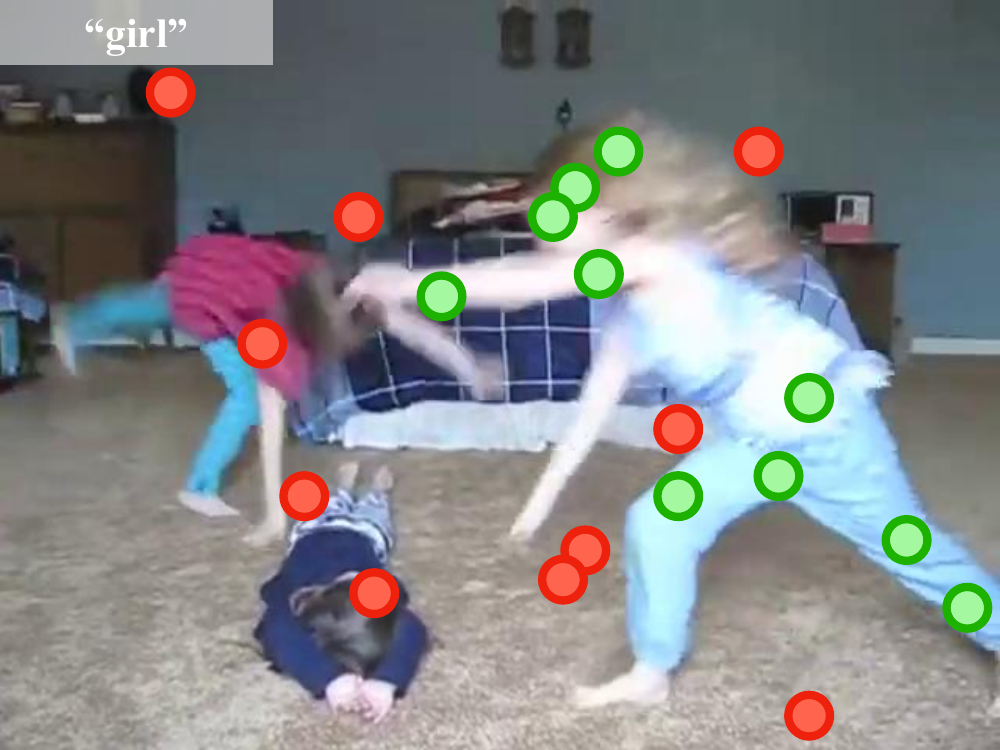}
        \vspace{-16pt}
    \end{subfigure}
    \begin{subfigure}[b]{0.32\linewidth}
     \includegraphics[width=\mysize\linewidth]{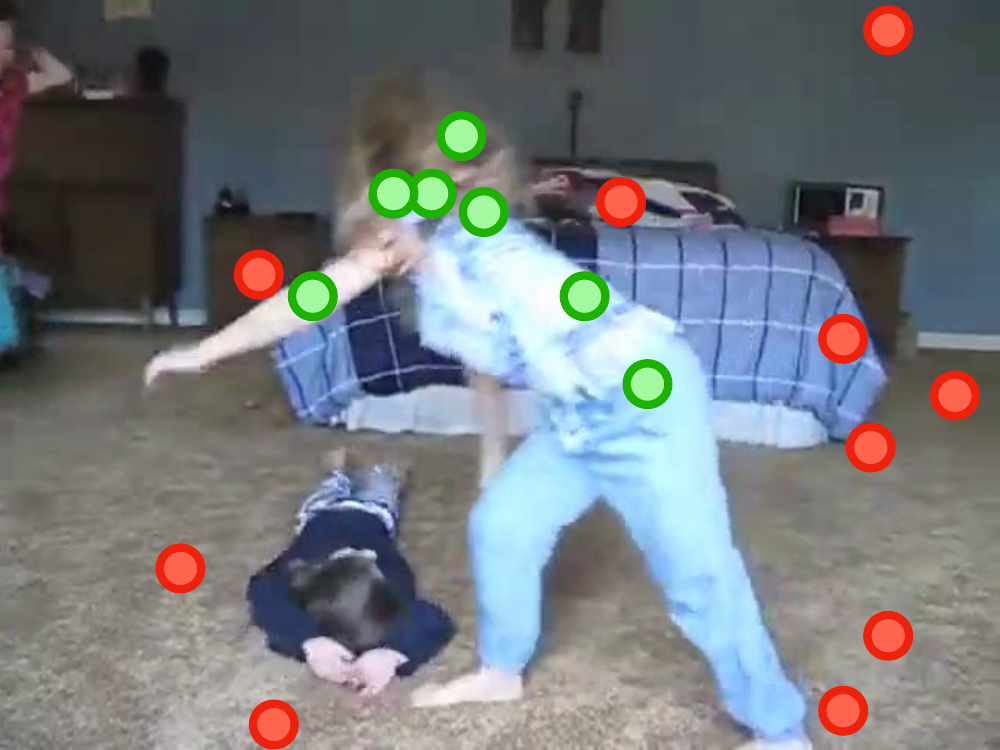}
        \vspace{-16pt}
    \end{subfigure}
    \begin{subfigure}[b]{0.32\linewidth}
     \includegraphics[width=\mysize\linewidth]{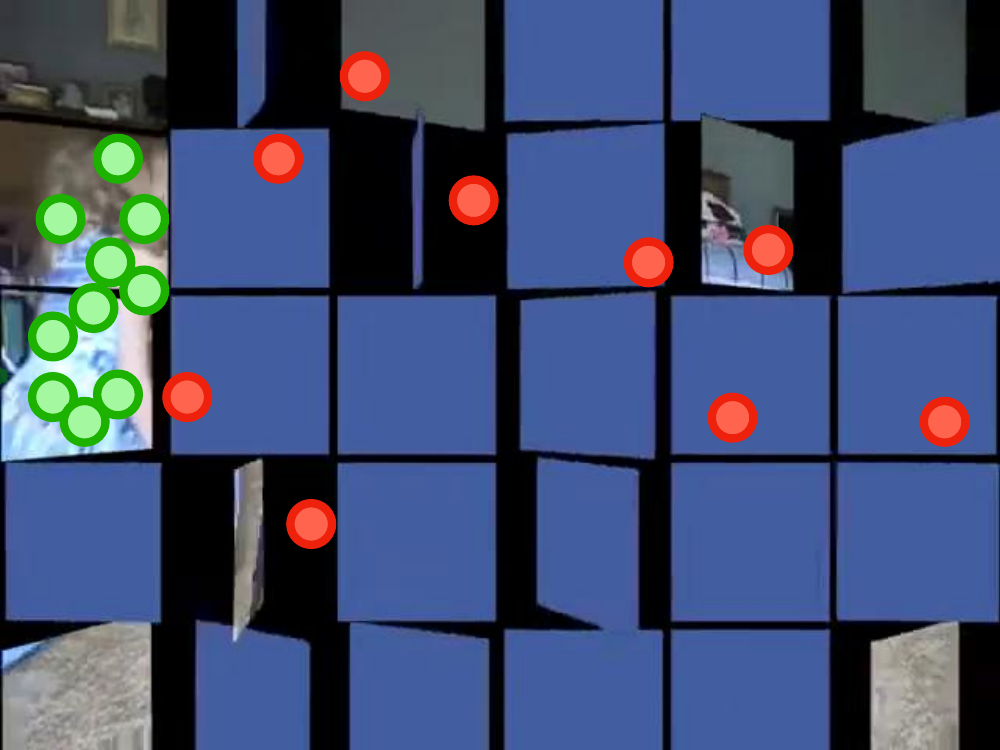}
        \vspace{-16pt}
    \end{subfigure}
    \vspace{-4pt}
    \caption{\textbf{Additional example point annotations for \pointvos{} Kinetics.} We can provide multi-modal point-wise annotations for objects from a large vocabulary (\textit{first and fourth rows}), scenes with fast motion (\textit{second row}), small objects (\textit{third row}), and also scenes with difficult lighting conditions (\textit{fourth row}) and motion blur (\textit{fifth row}). \textcolor{green}{Green} dots represent positive points and \textcolor{red}{red} dots negative points.}
    \label{fig:add_point_anns_kinetics}
\end{figure*}

In~\cref{fig:add_point_anns_oops} and~\cref{fig:add_point_anns_kinetics}, we provide the additional example point annotations for \pointvos{} Oops (\pvoops{}) and \pointvos{} Kinetics (\pvkinetics{}). We successfully annotated multi-modal points for different and challenging scenes, and also the objects from a large vocabulary.

\begin{figure}[t]
    \centering
    \setlength{\fboxsep}{0.49pt}
    \begin{subfigure}[b]{0.49\linewidth}
        \includegraphics[width=\mysize\linewidth]{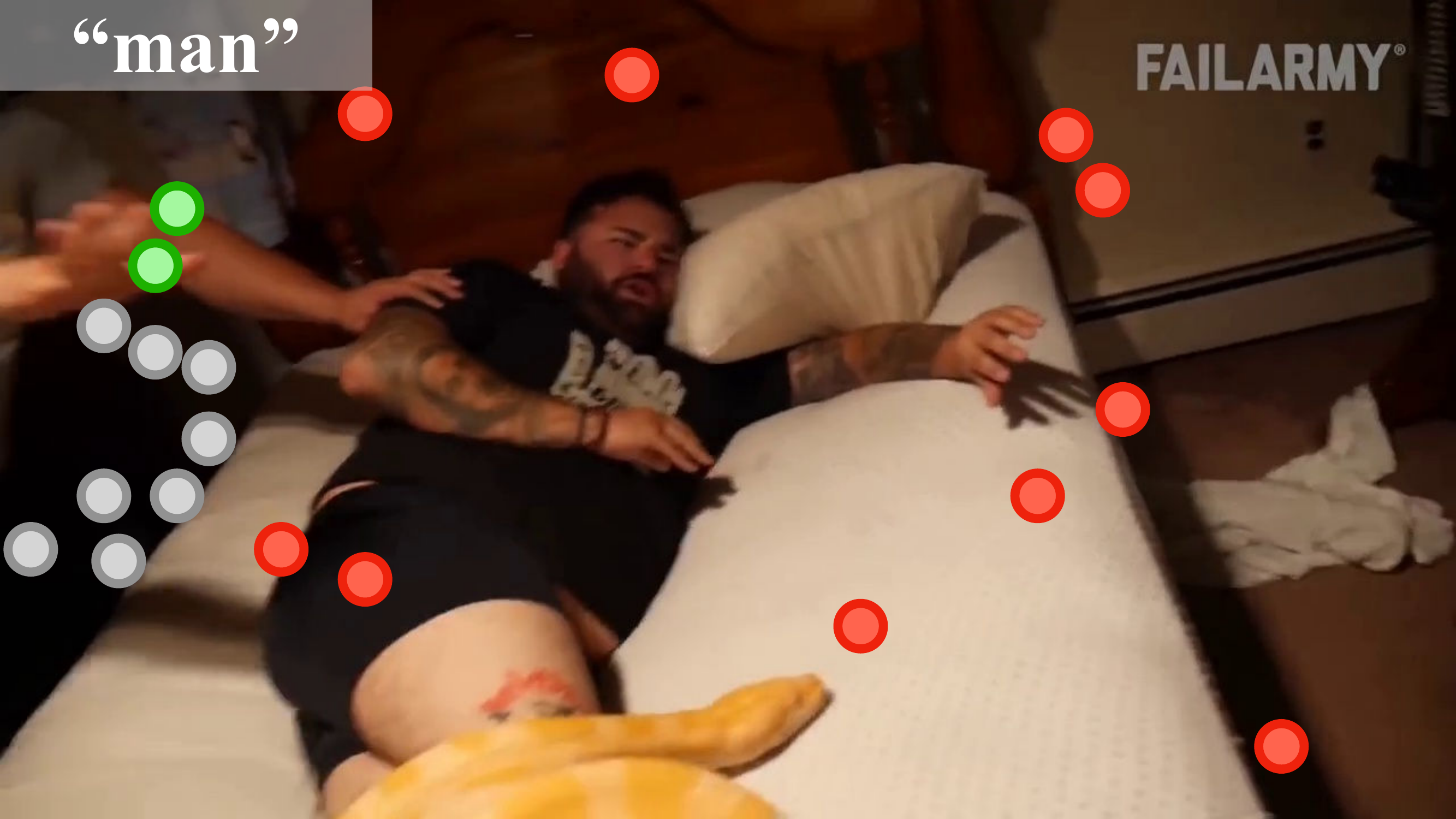}
        \vspace{-10pt}
    \end{subfigure}
    \begin{subfigure}[b]{0.49\linewidth}
    \includegraphics[width=\mysize\linewidth]{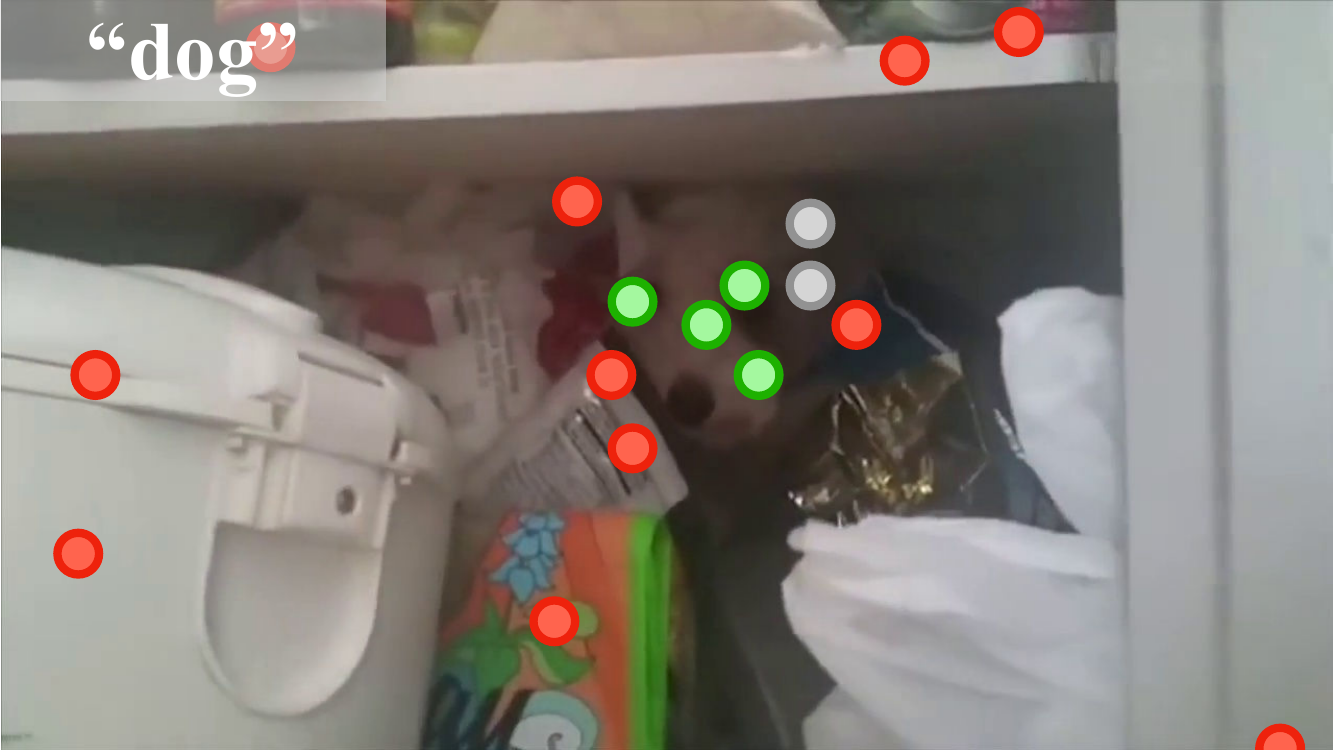}
        \vspace{-10pt}
    \end{subfigure}
    \begin{subfigure}[b]{0.49\linewidth}
     \includegraphics[width=\mysize\linewidth]{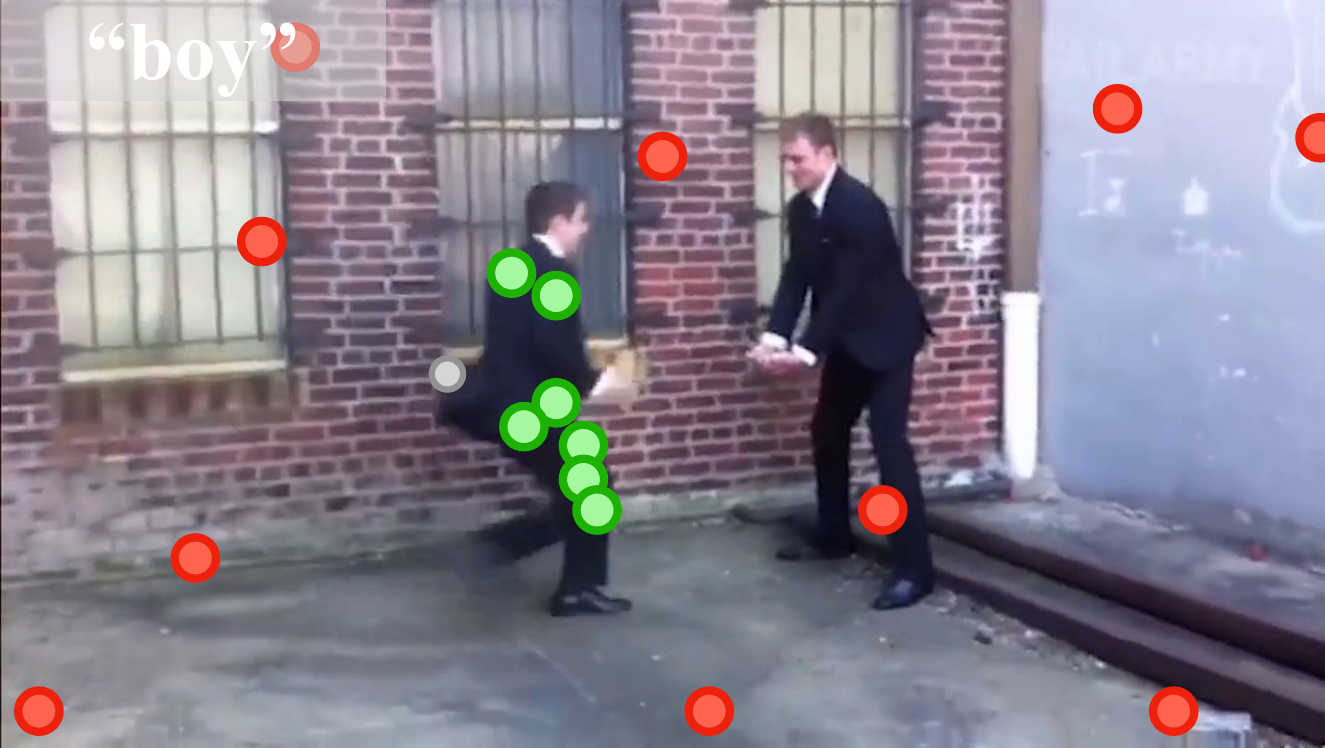}
        \vspace{-10pt}
    \end{subfigure}
    \begin{subfigure}[b]{0.49\linewidth}
        \includegraphics[width=\mysize\linewidth]{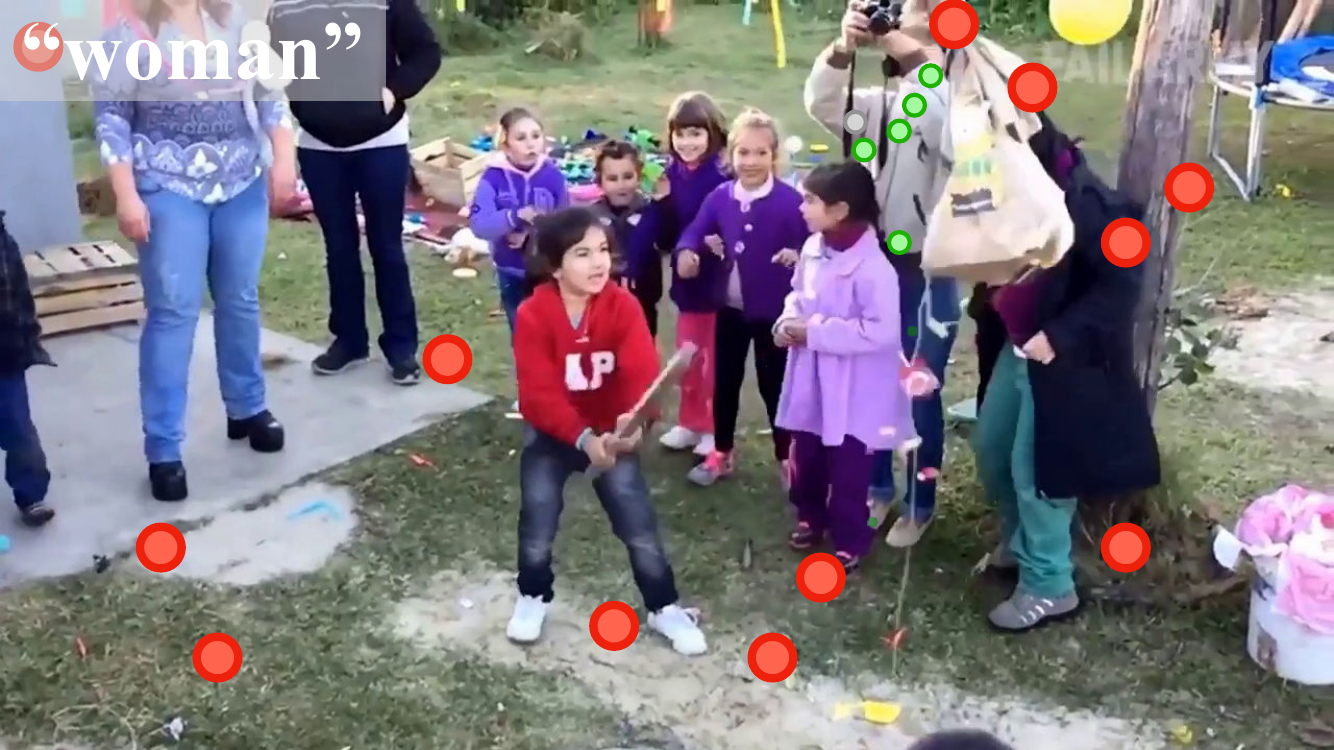}
        \vspace{-10pt}
    \end{subfigure}   
    \begin{subfigure}[b]{0.49\linewidth}
     \includegraphics[width=\mysize\linewidth]{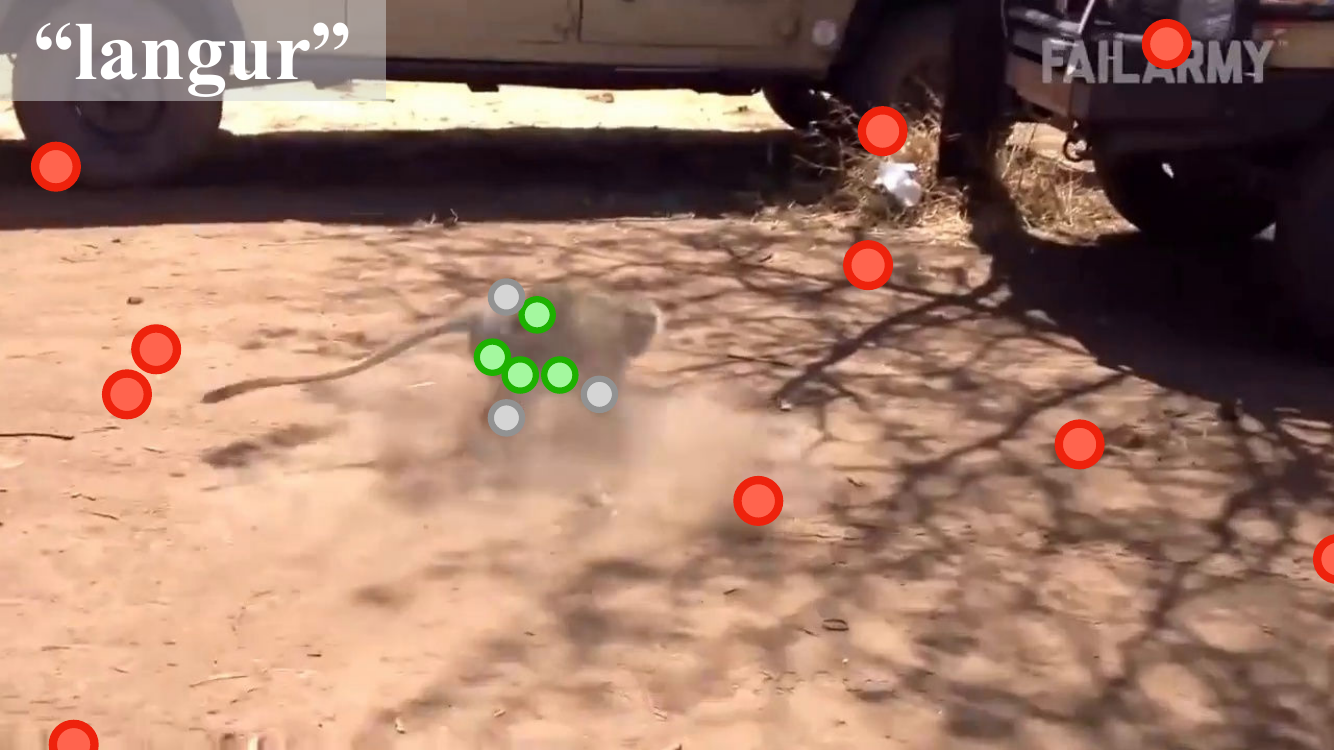}
        \vspace{-10pt}
    \end{subfigure} 
    \begin{subfigure}[b]{0.49\linewidth}
        \includegraphics[width=\mysize\linewidth]{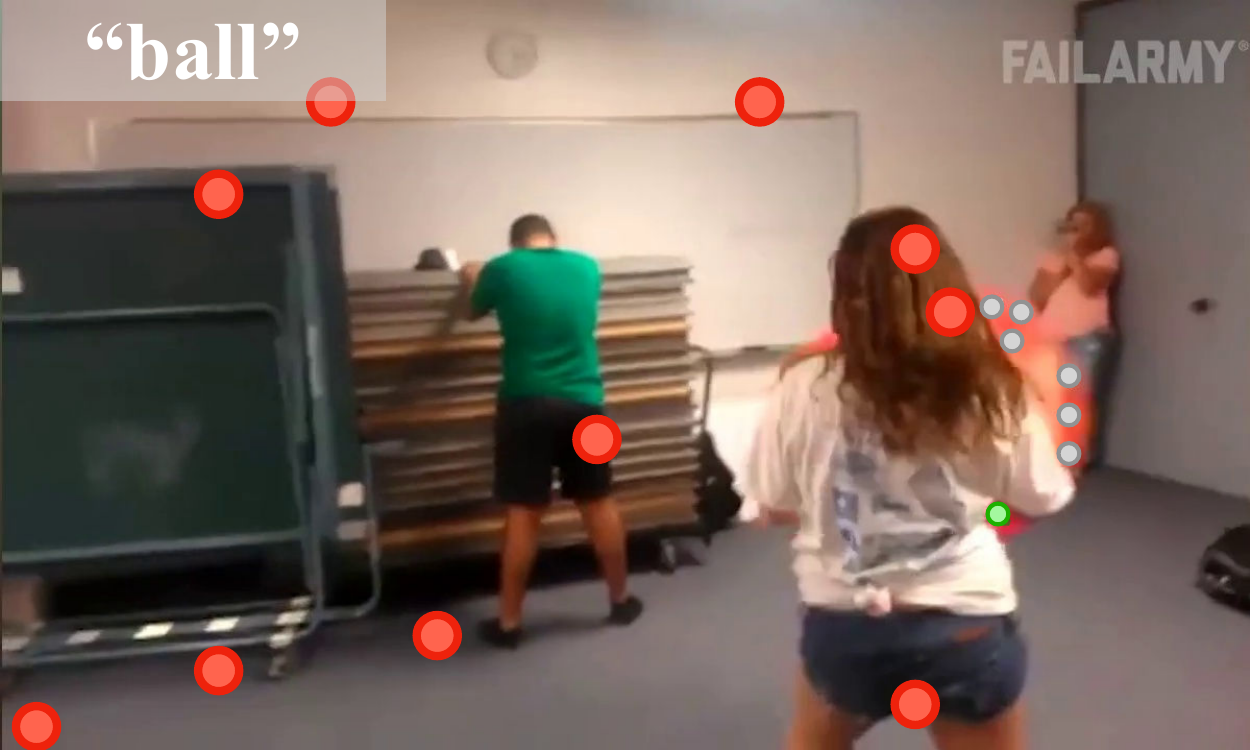}
        \vspace{-10pt}
    \end{subfigure}
    \begin{subfigure}[b]{0.49\linewidth}
     \includegraphics[width=\mysize\linewidth]{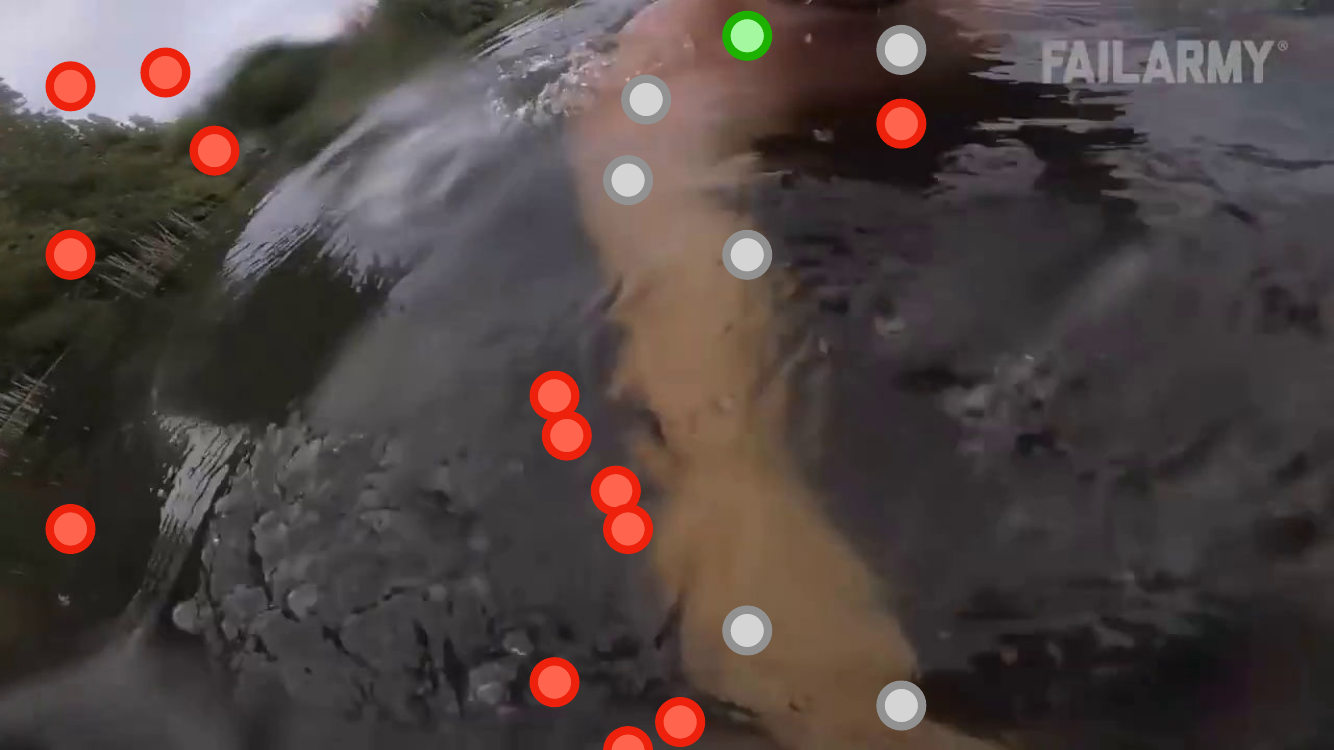}
        \vspace{-10pt}
   \end{subfigure}
    \begin{subfigure}[b]{0.49\linewidth}
        \includegraphics[width=\mysize\linewidth]{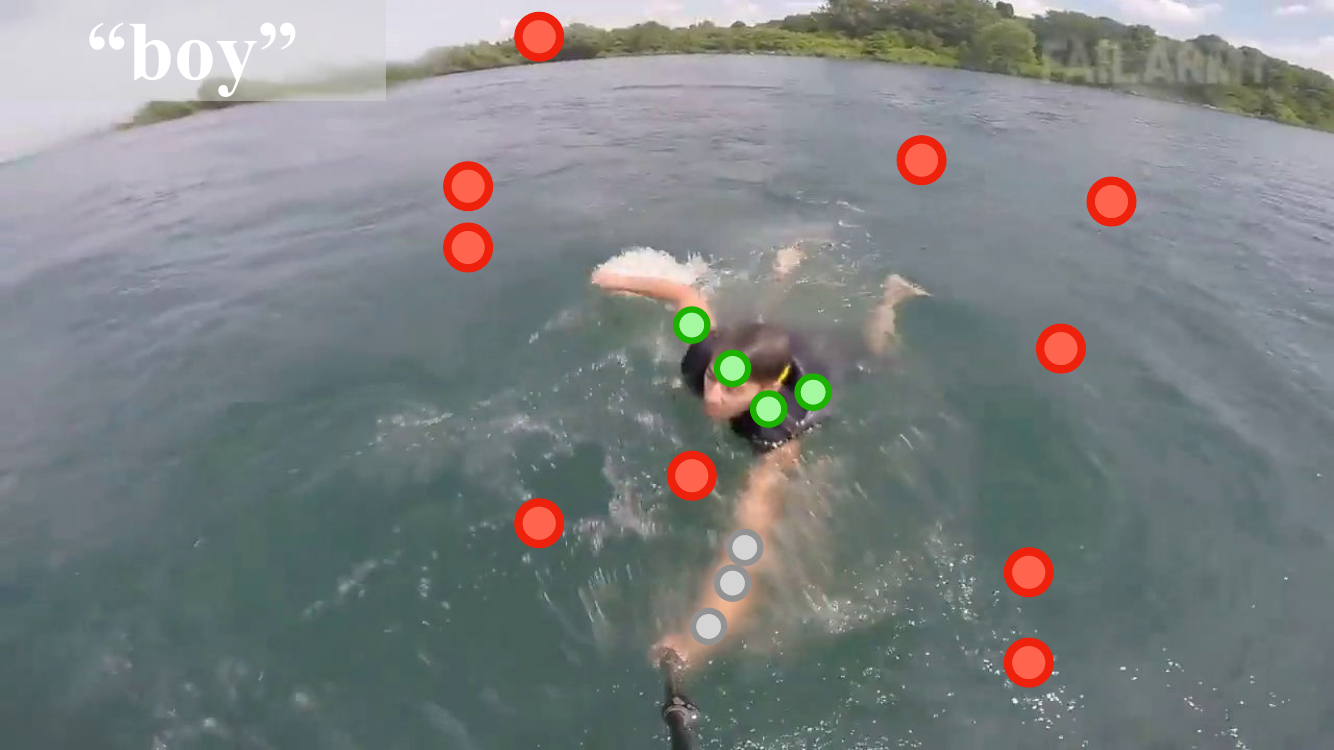}
        \vspace{-10pt}
    \end{subfigure}
    \vspace{5pt}

    \caption{\textbf{Example ambiguous point annotations from \pointvos{} Oops.} We observe that the human annotators indicate unsure if the given point is in challenging lighting condition (\textit{first row)}) or at border (\textit{second row}), or at motion blur (\textit{third row}), or if the object is ambiguous (\textit{fourth row}). \textcolor{green}{Green} dots represent positive annotations, \textcolor{red}{red} dots negative annotations, and \textcolor{gray}{gray} dots ambiguous annotations.}
    \label{fig:ambiguous_points_oops}
\end{figure}

\begin{figure}[t]
    \centering
    \setlength{\fboxsep}{0.49pt}
    \begin{subfigure}[b]{0.49\linewidth}
        \includegraphics[width=\mysize\linewidth]{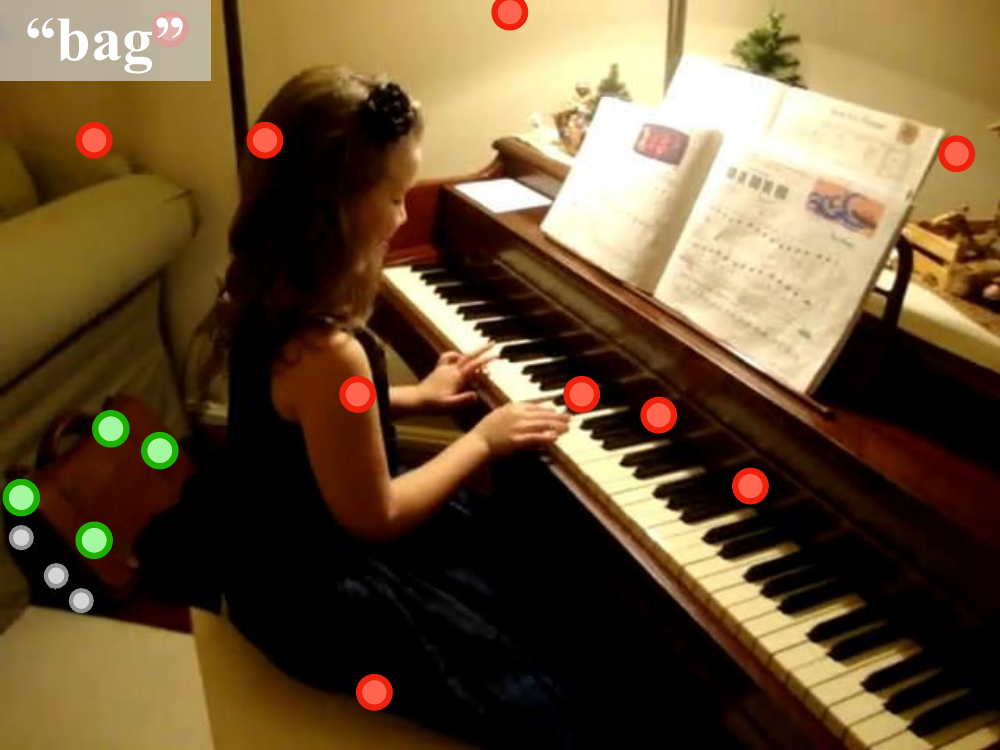}
        \vspace{-10pt}
    \end{subfigure}
    \begin{subfigure}[b]{0.49\linewidth}
    \includegraphics[width=\mysize\linewidth]{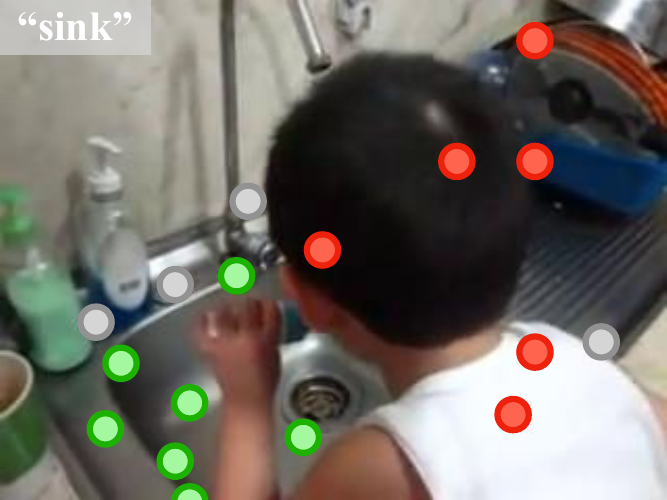}
        \vspace{-10pt}
    \end{subfigure}
    \begin{subfigure}[b]{0.49\linewidth}
     \includegraphics[width=\mysize\linewidth]{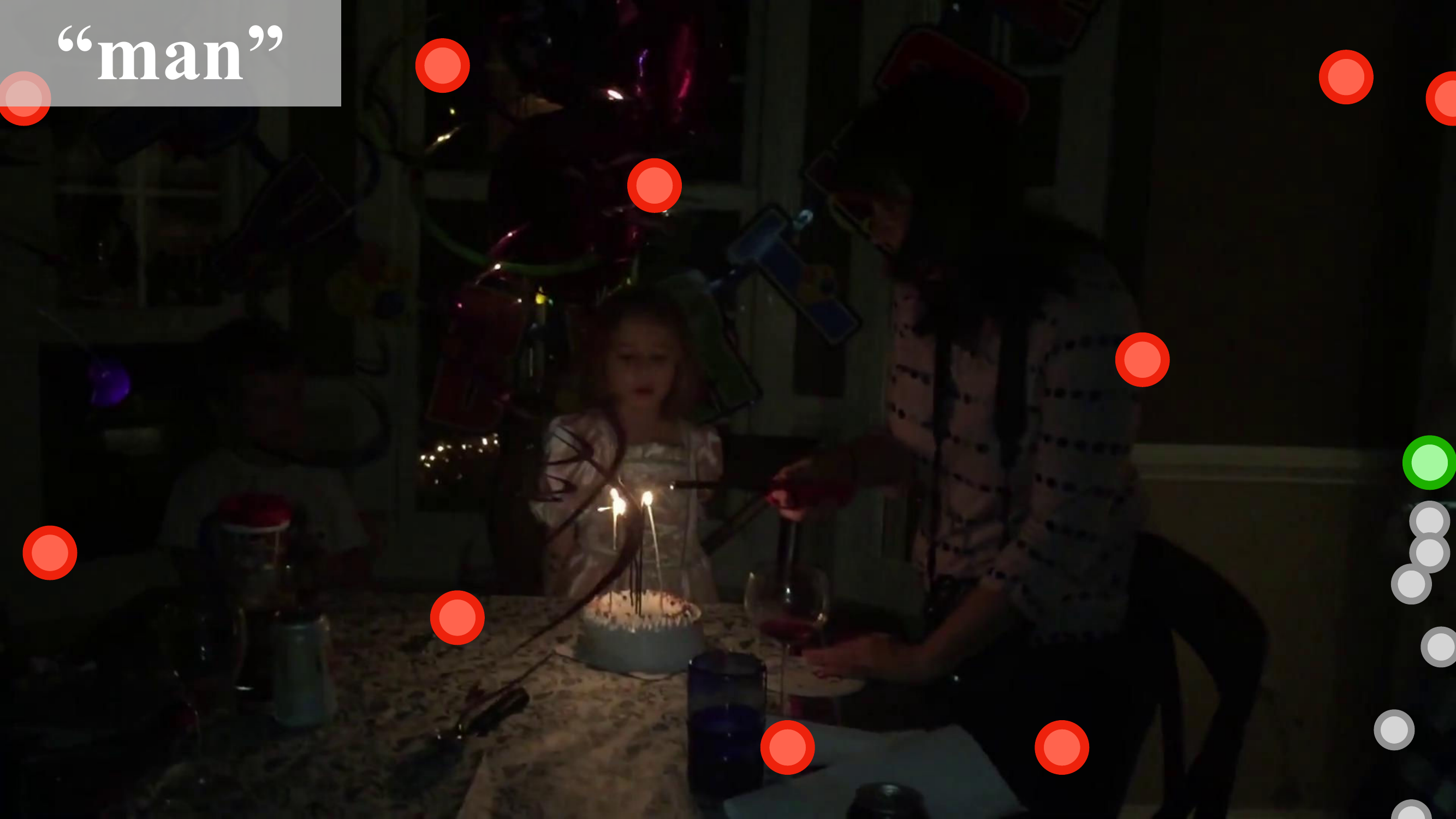}
        \vspace{-10pt}
    \end{subfigure}
    \begin{subfigure}[b]{0.49\linewidth}
        \includegraphics[width=\mysize\linewidth]{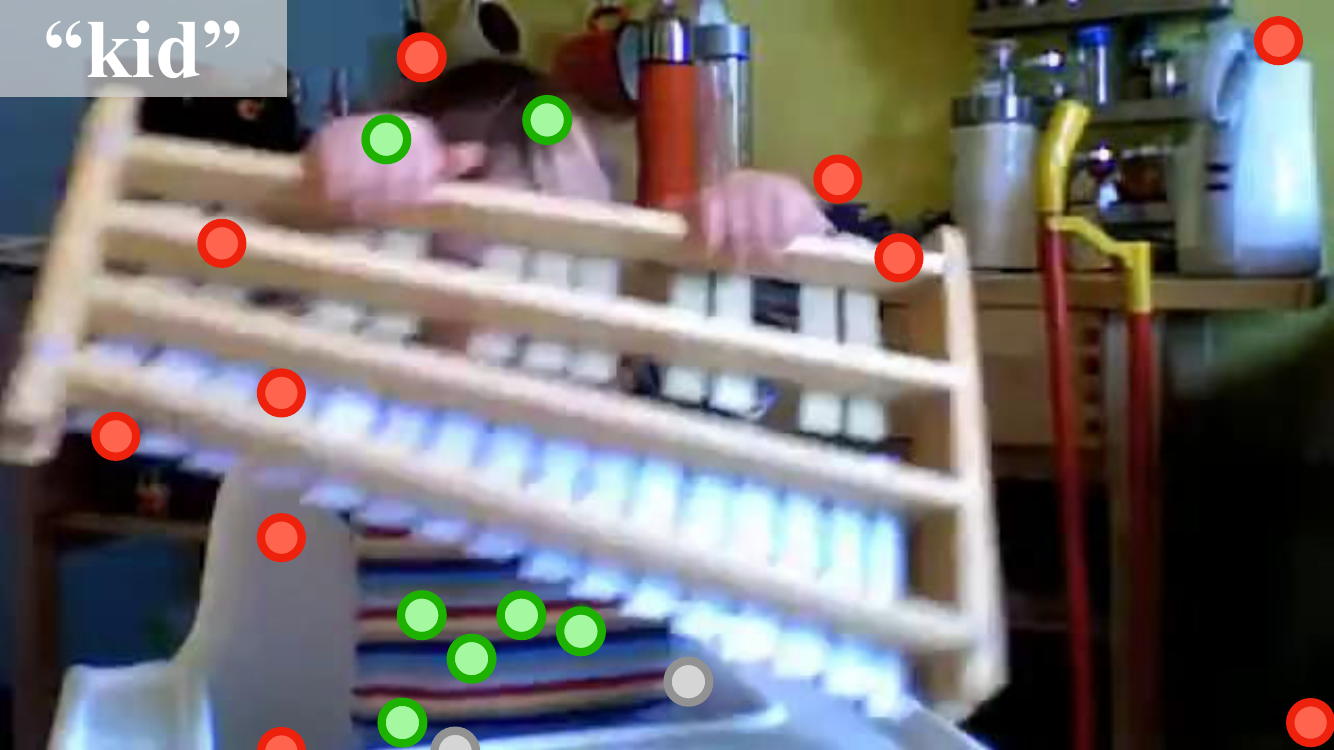}
        \vspace{-10pt}
    \end{subfigure}
    \begin{subfigure}[b]{0.49\linewidth}
        \includegraphics[width=\mysize\linewidth]{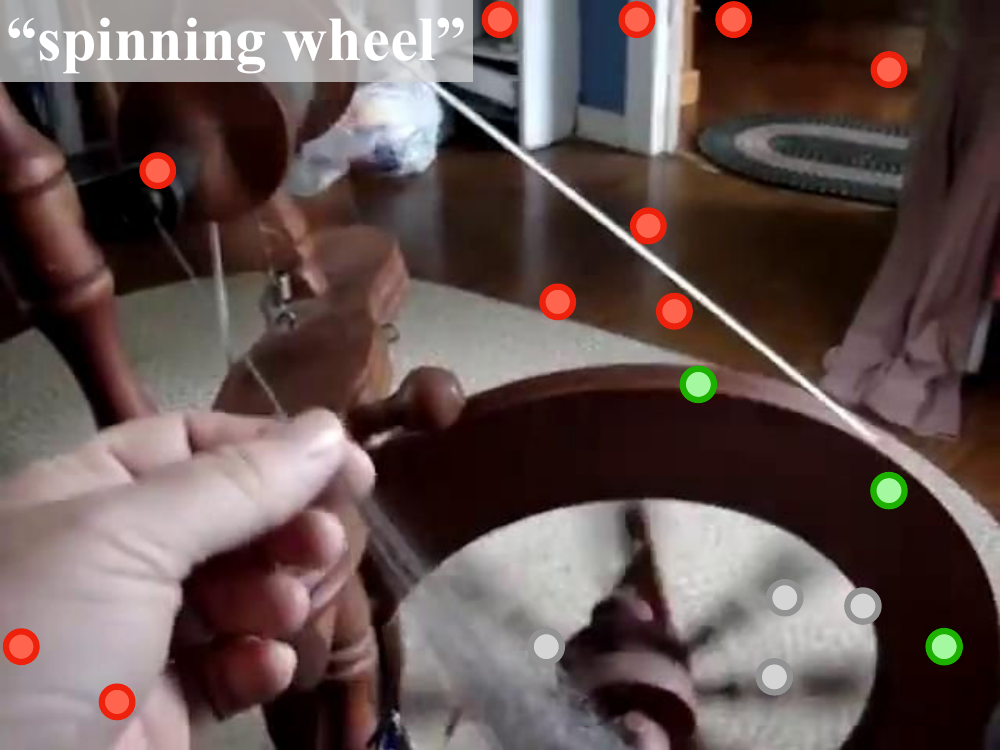}
        \vspace{-10pt}
    \end{subfigure}
    \begin{subfigure}[b]{0.49\linewidth}
     \includegraphics[width=\mysize\linewidth]{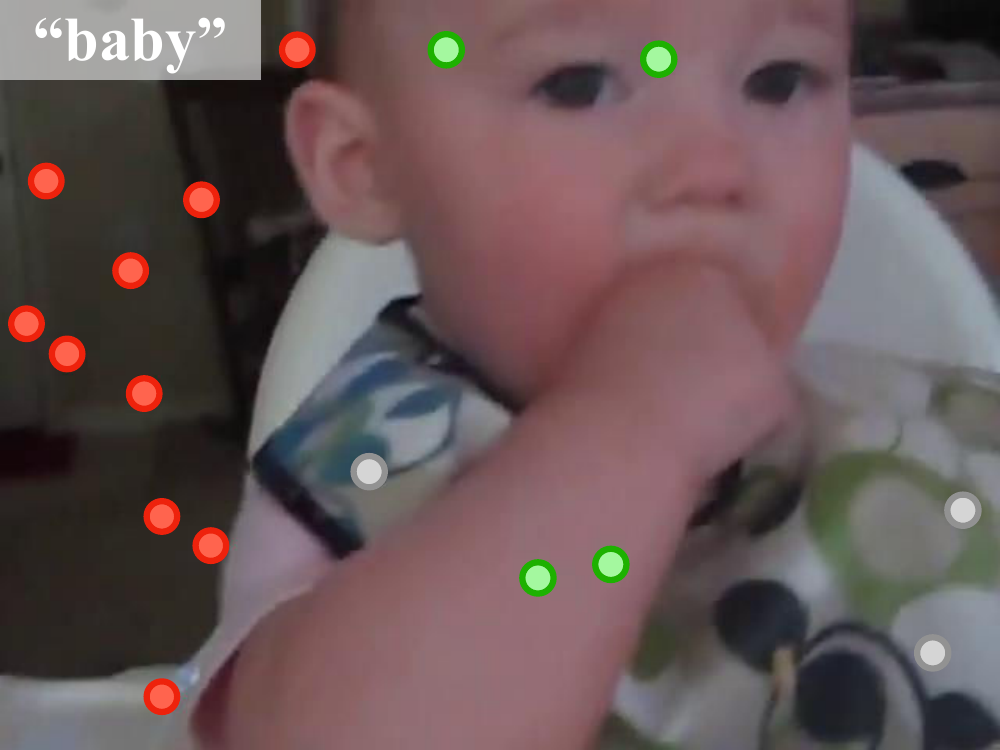}
        \vspace{-10pt}
    \end{subfigure} 
    \begin{subfigure}[b]{0.49\linewidth}
        \includegraphics[width=\mysize\linewidth]{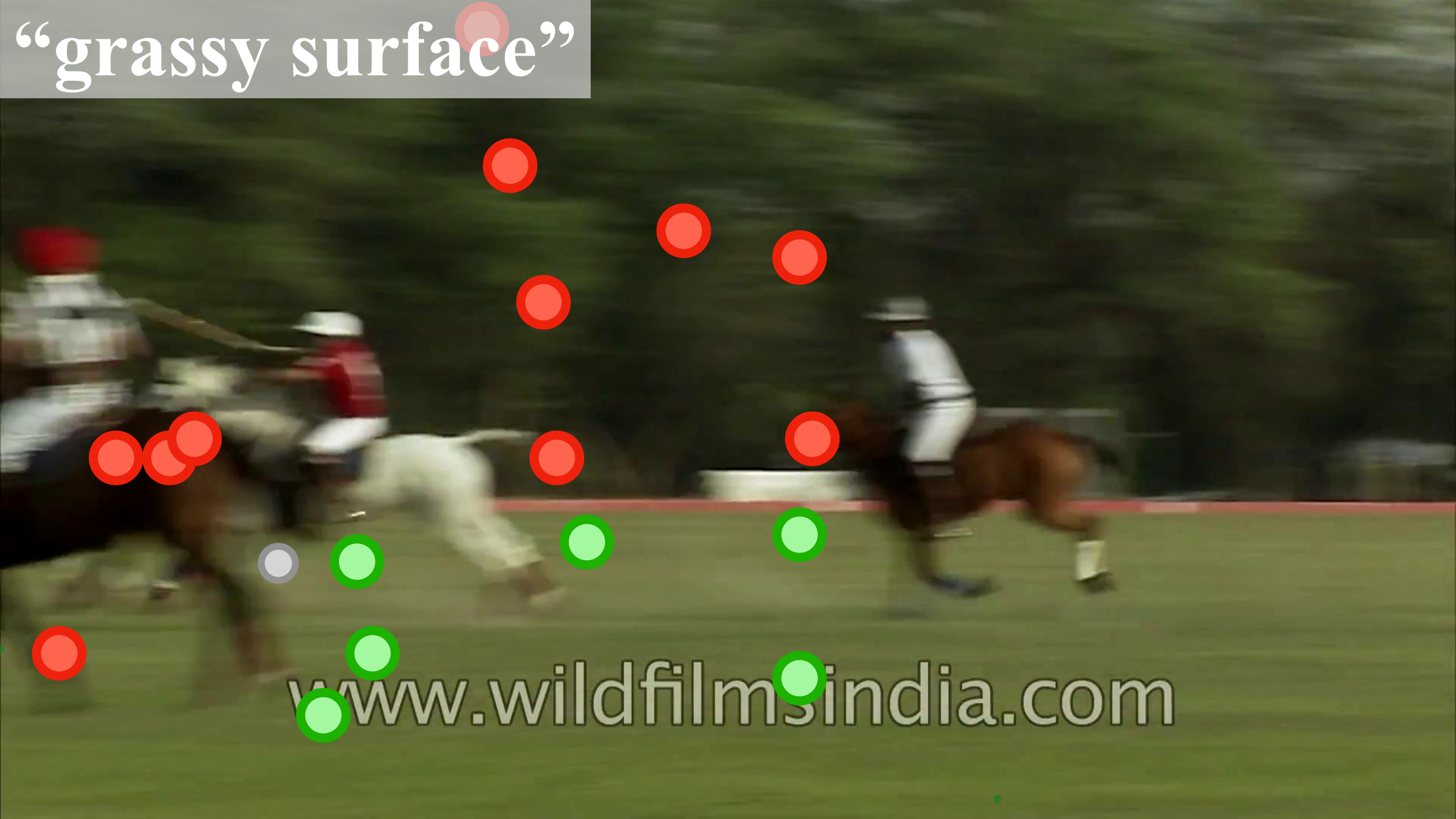}
        \vspace{-10pt}
    \end{subfigure}
    \begin{subfigure}[b]{0.49\linewidth}
     \includegraphics[width=\mysize\linewidth]{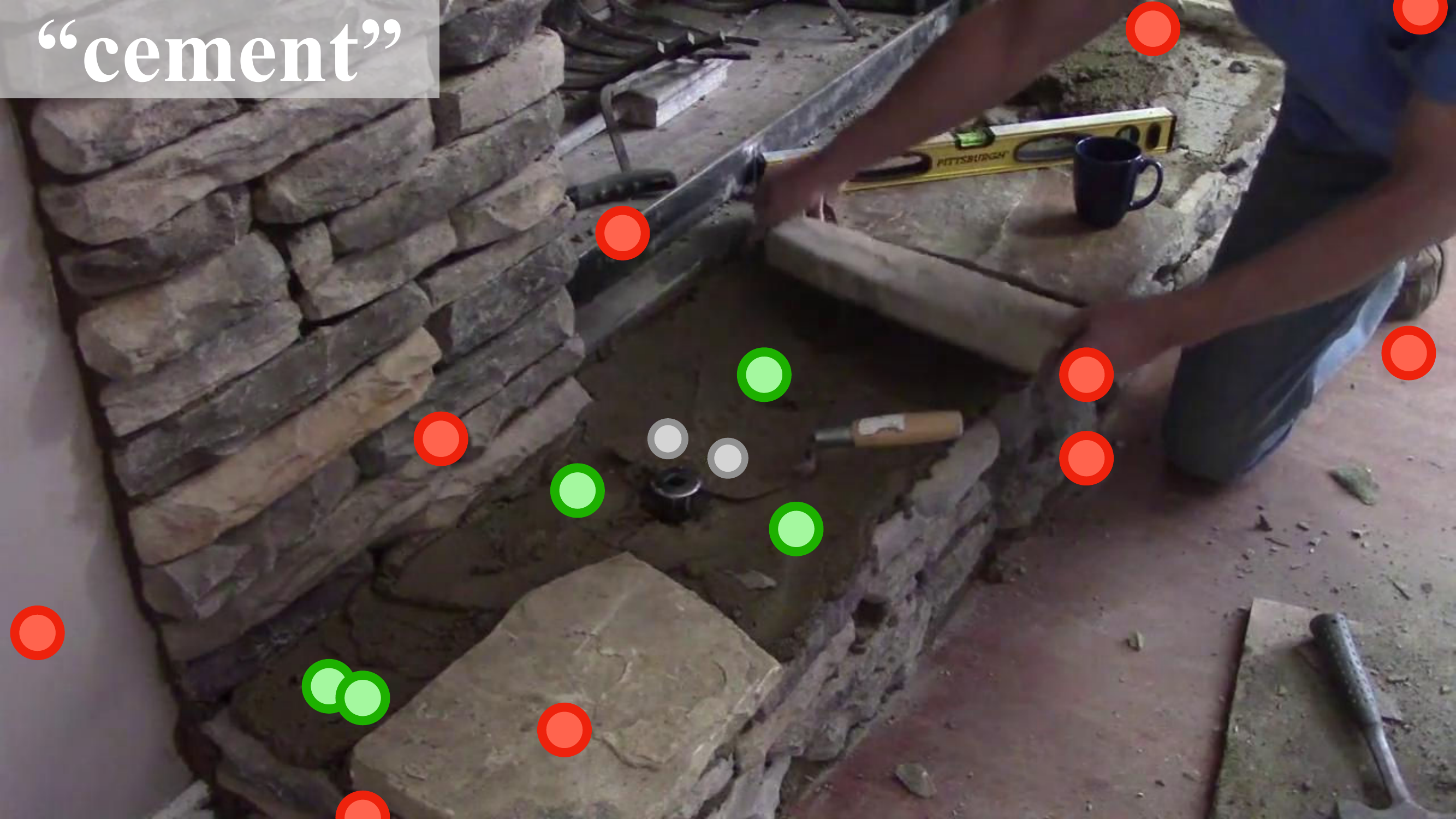}
        \vspace{-10pt}
   \end{subfigure}
    \vspace{5pt}
    \caption{\textbf{Example ambiguous point annotations from \pointvos{} Kinetics.} Similarly, the human annotators indicate unsure if the given point is in challenging lighting condition (\textit{first column, first two rows}) or at border (\textit{second column, first two rows}), or at motion blur (\textit{first column, last two rows}), or if the object is ambiguous (\textit{second column, last two rows}). \textcolor{green}{Green} dots represent positive annotations, \textcolor{red}{red} dots negative annotations, and \textcolor{gray}{gray} dots ambiguous annotations.}
    \label{fig:ambiguous_points_kinetics}
\end{figure}

In~\cref{fig:ambiguous_points_oops} and~\cref{fig:ambiguous_points_kinetics}, we also show the examples of ambiguous point annotations from \pvoops{} and \pvkinetics{},~\ie the point annotations where the human annotators indicated that they were unsure. We observe that we have ambiguous point annotations in particular cases for both \pvoops{} and \pvkinetics{},~\eg, if the given point is in a challenging lighting condition or at the border.

\begin{figure*}[ht!]
    \centering
    \setlength{\fboxsep}{0.32pt}
    \begin{subfigure}[b]{0.32\linewidth}
        \includegraphics[width=\mysize\linewidth]{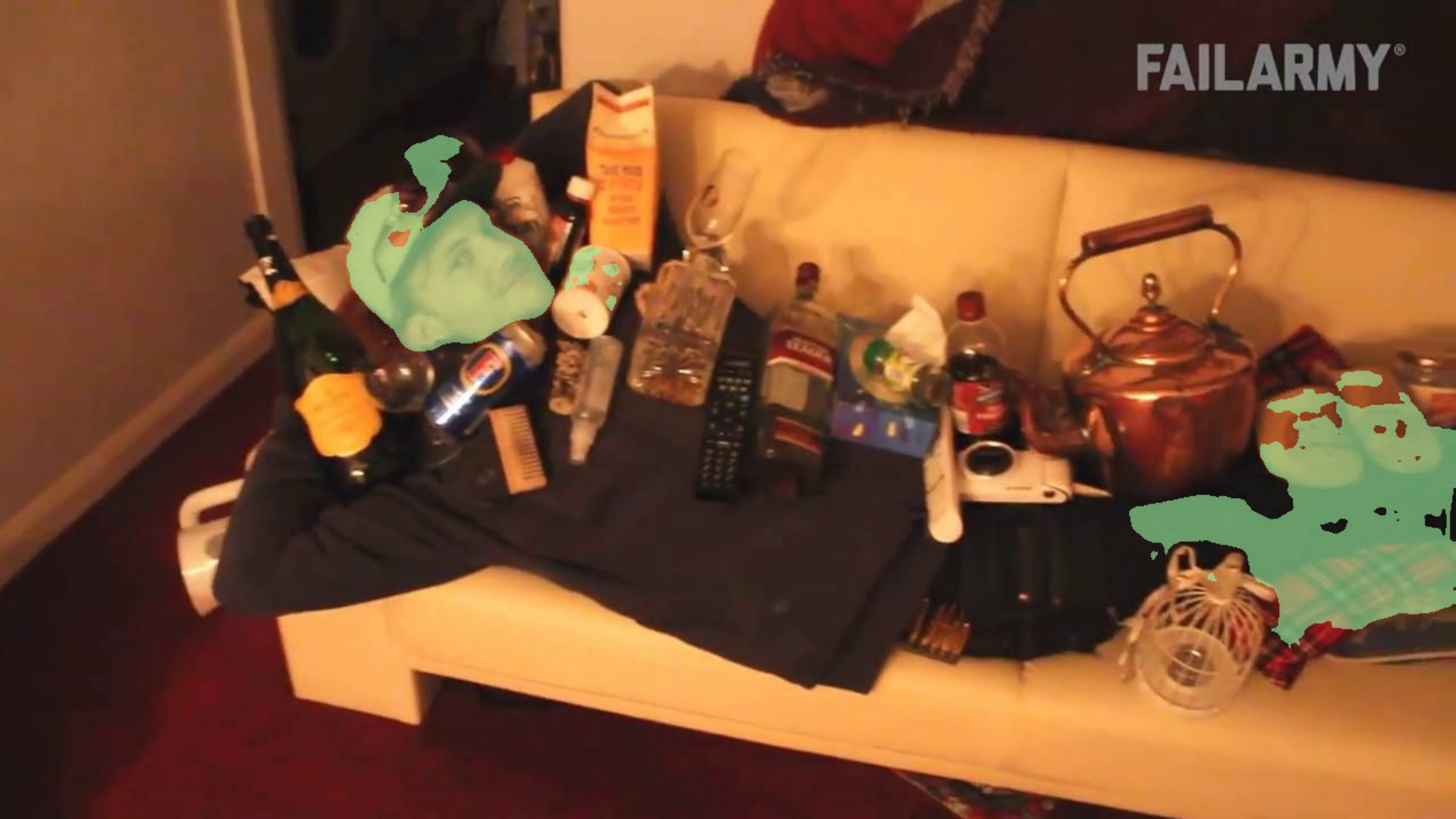}
        \vspace{-16pt}
    \end{subfigure}
    \begin{subfigure}[b]{0.32\linewidth}

     \includegraphics[width=\mysize\linewidth]{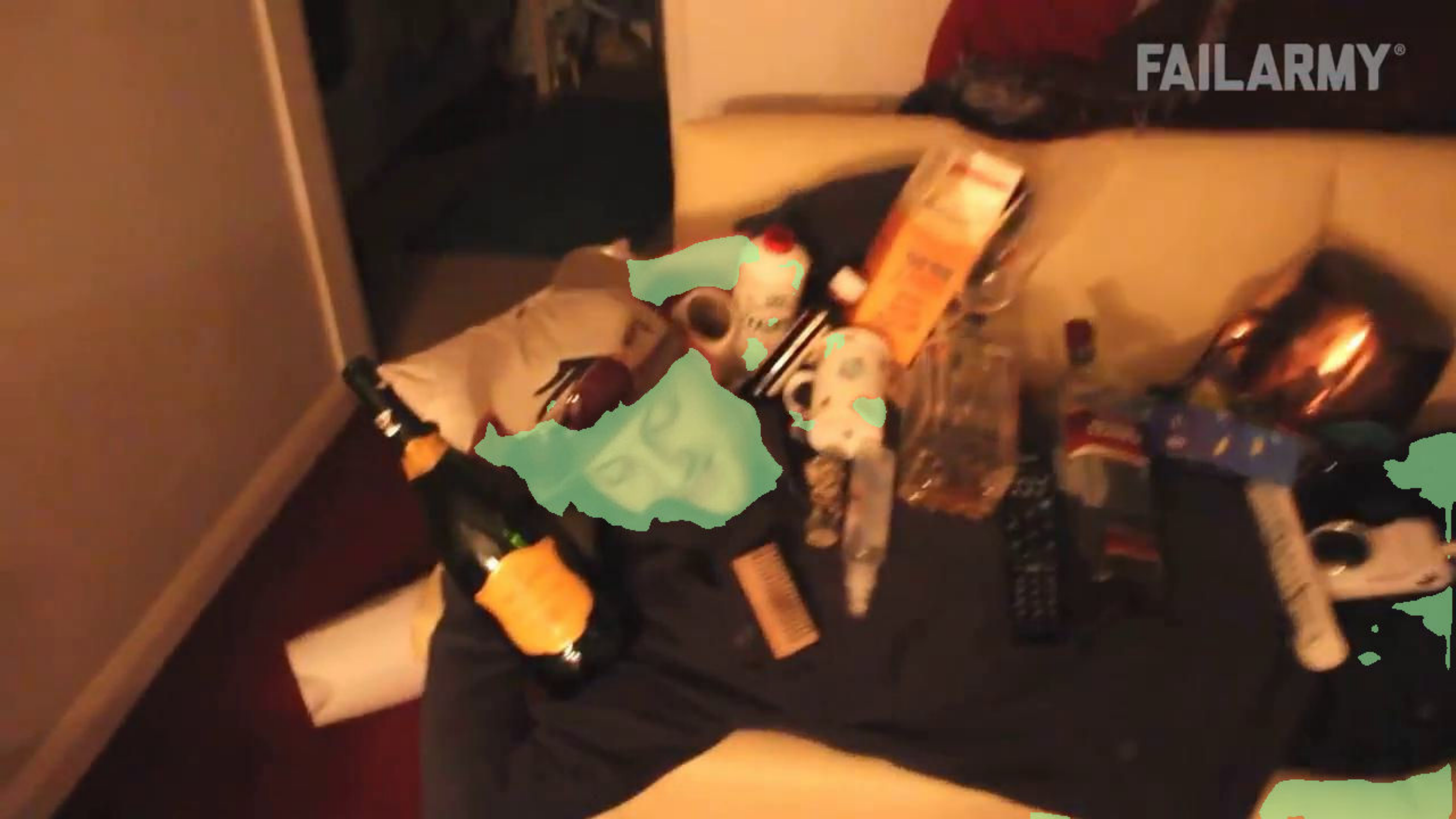}
        \vspace{-16pt}
    \end{subfigure}
    \begin{subfigure}[b]{0.32\linewidth}
     \includegraphics[width=\mysize\linewidth]{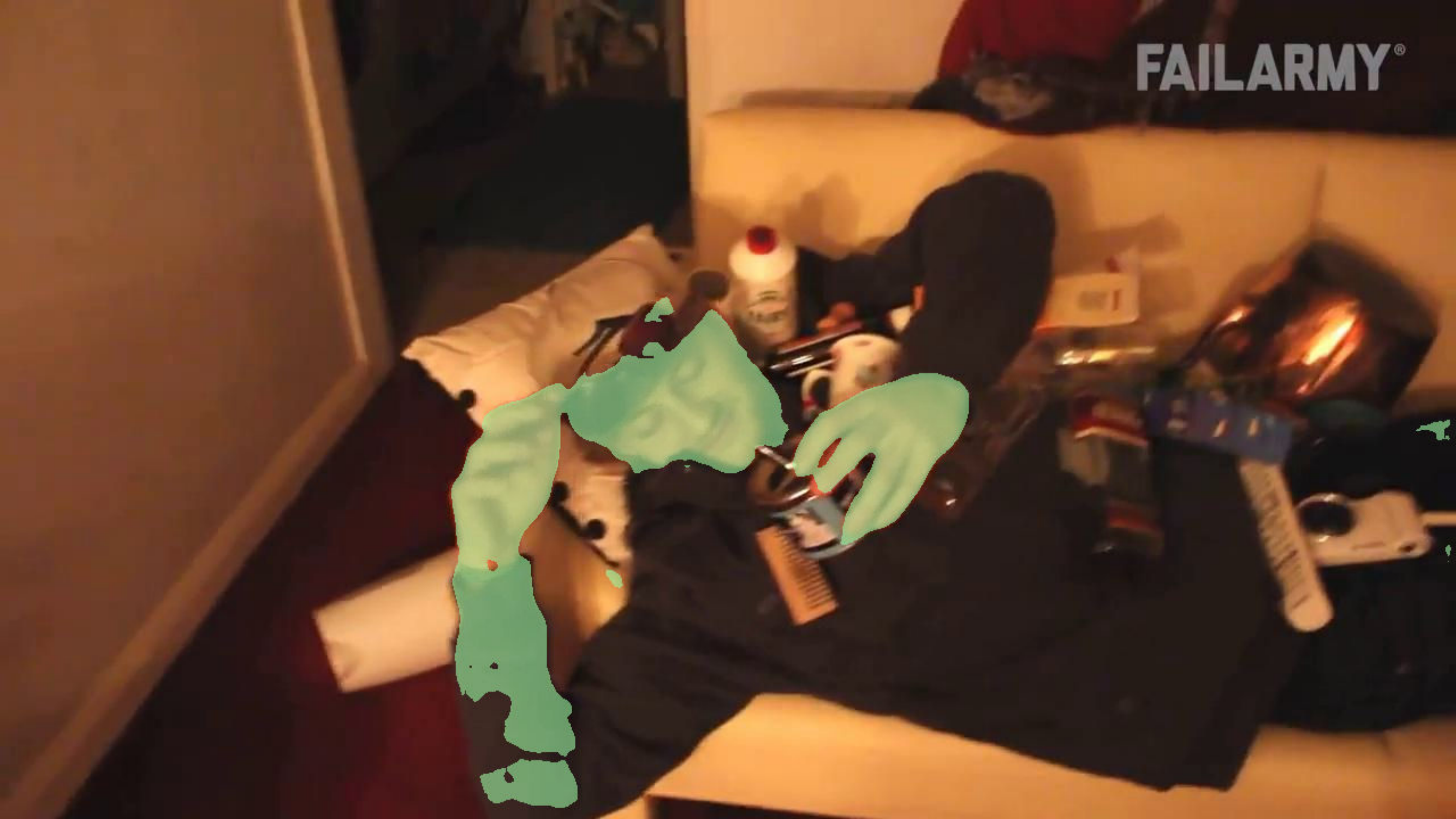}
        \vspace{-16pt}
    \end{subfigure}
    \vspace{5pt}

    \begin{subfigure}[b]{0.32\linewidth}
        \includegraphics[width=\mysize\linewidth]{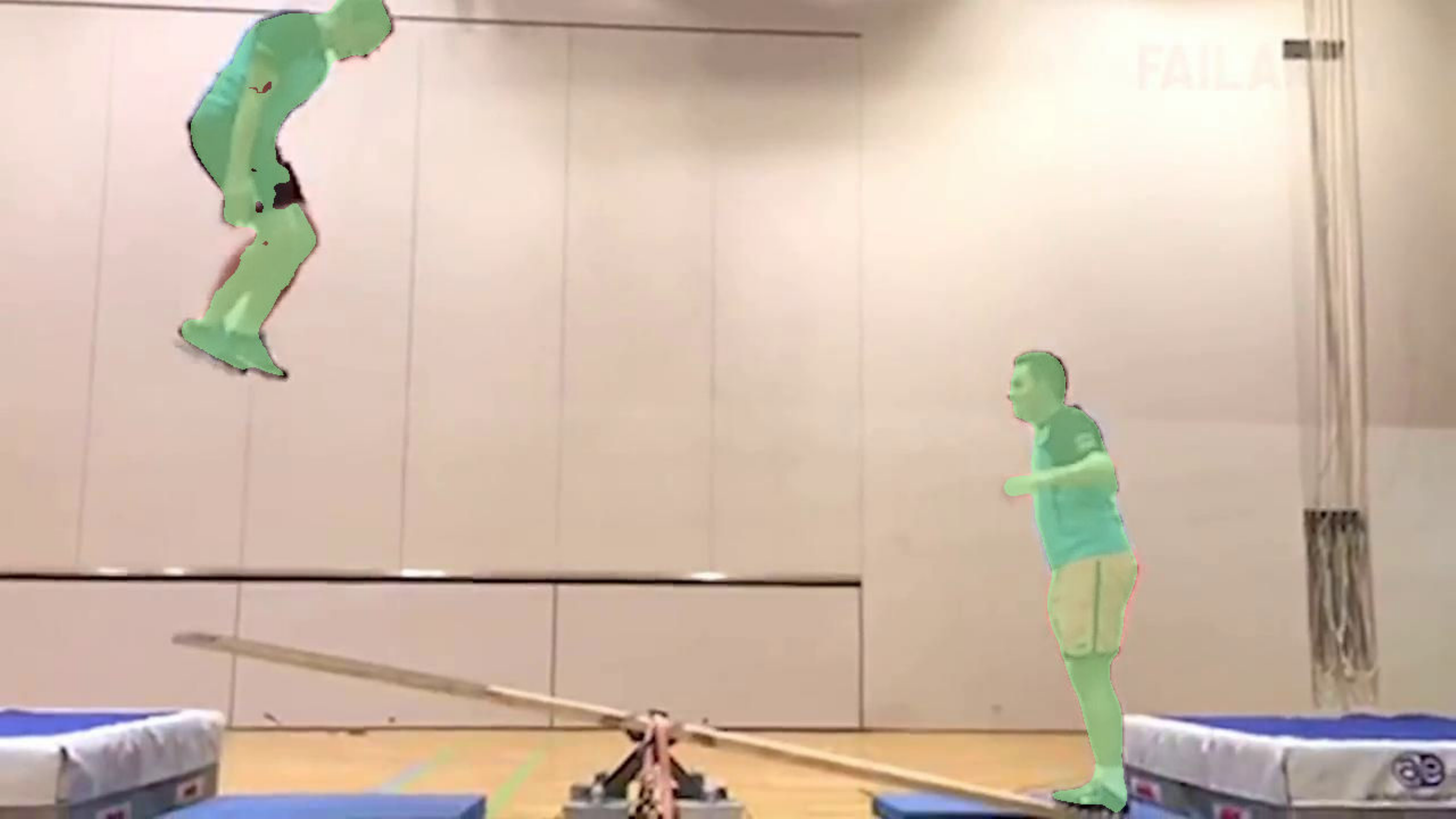}
        \vspace{-16pt}
    \end{subfigure}
    \begin{subfigure}[b]{0.32\linewidth}
     \includegraphics[width=\mysize\linewidth]{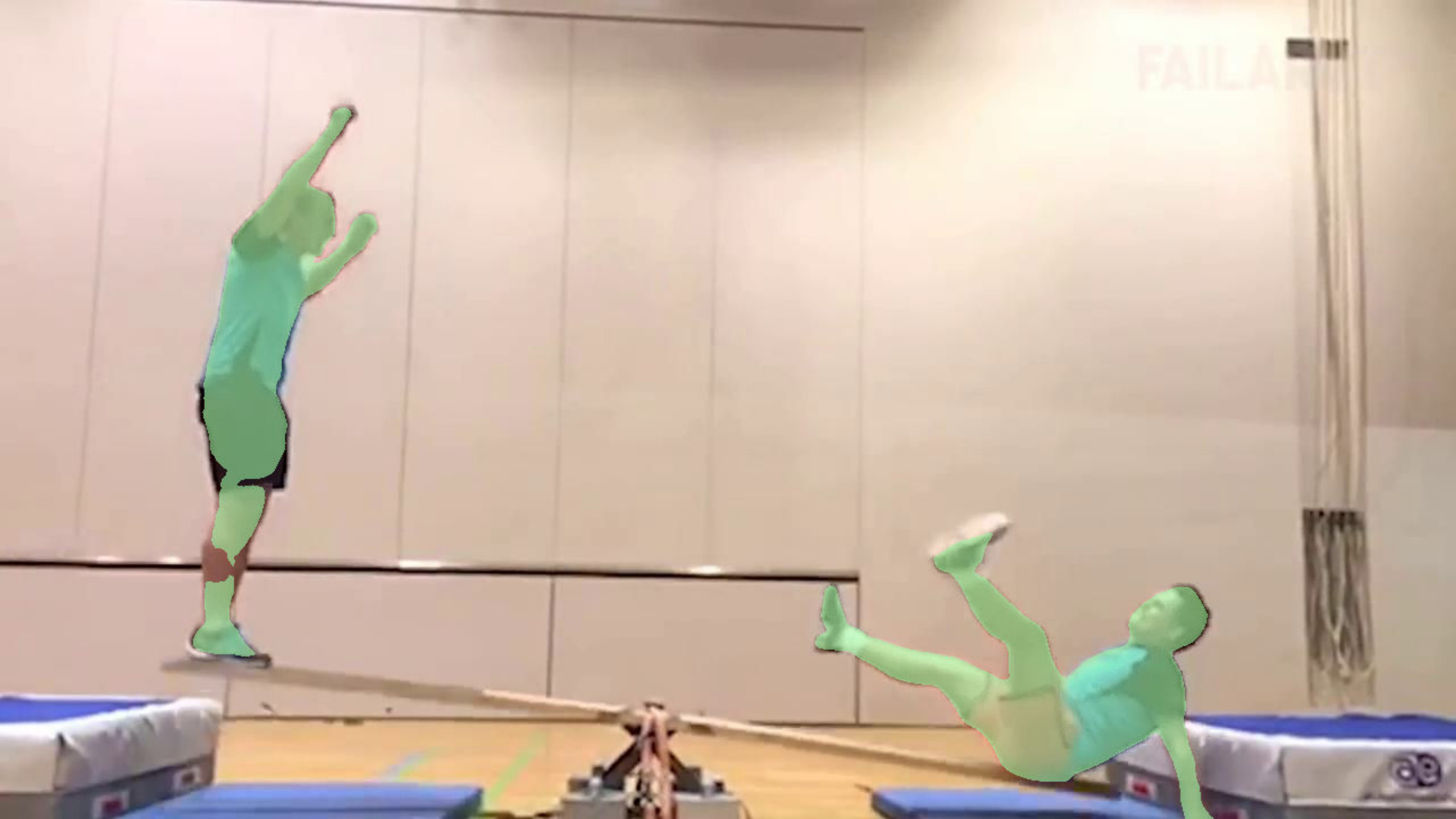}
        \vspace{-16pt}
    \end{subfigure}
    \begin{subfigure}[b]{0.32\linewidth}
     \includegraphics[width=\mysize\linewidth]{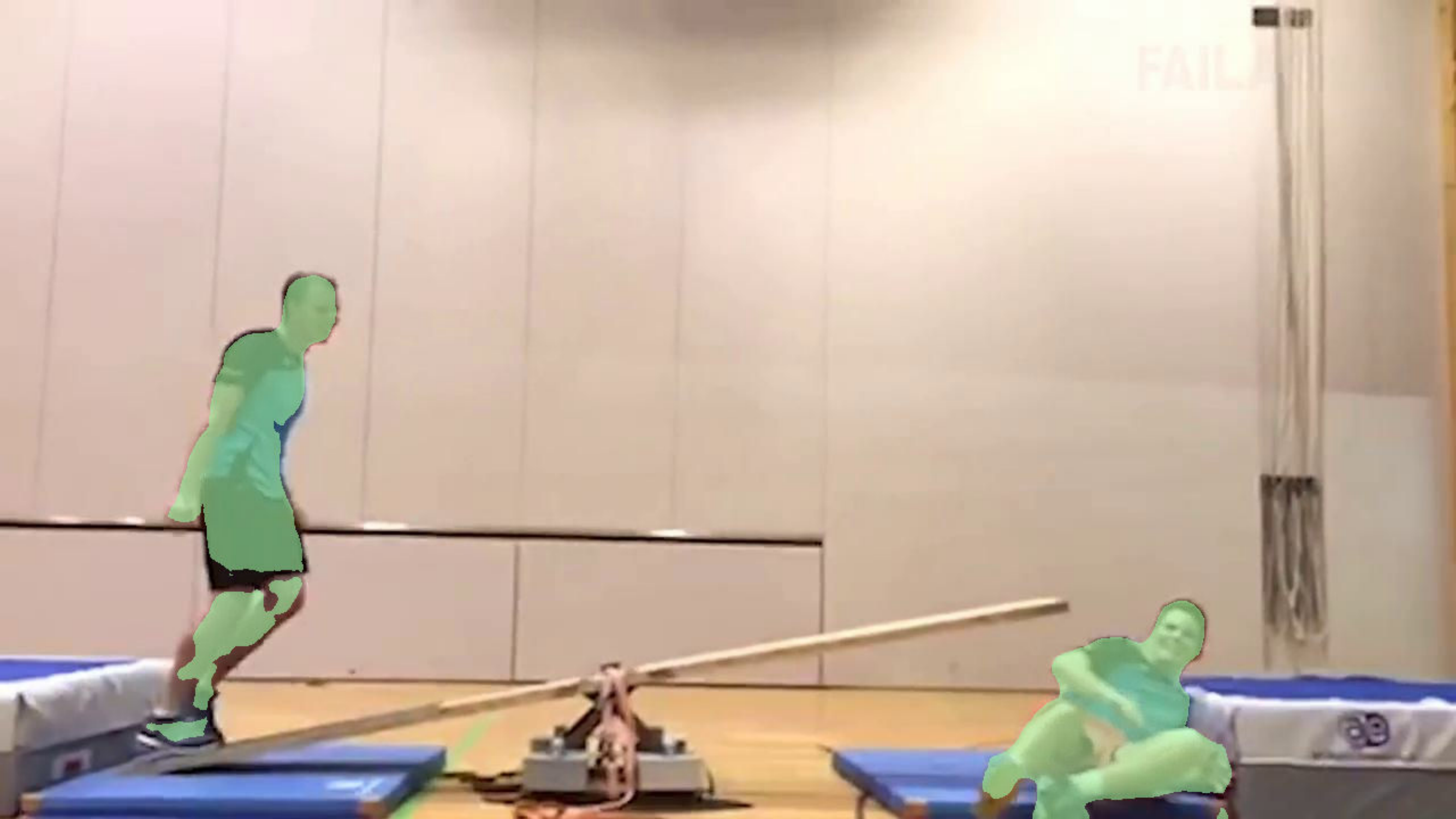}
        \vspace{-16pt}
    \end{subfigure}
    \vspace{5pt}
    
    \begin{subfigure}[b]{0.32\linewidth}
        \includegraphics[width=\mysize\linewidth]{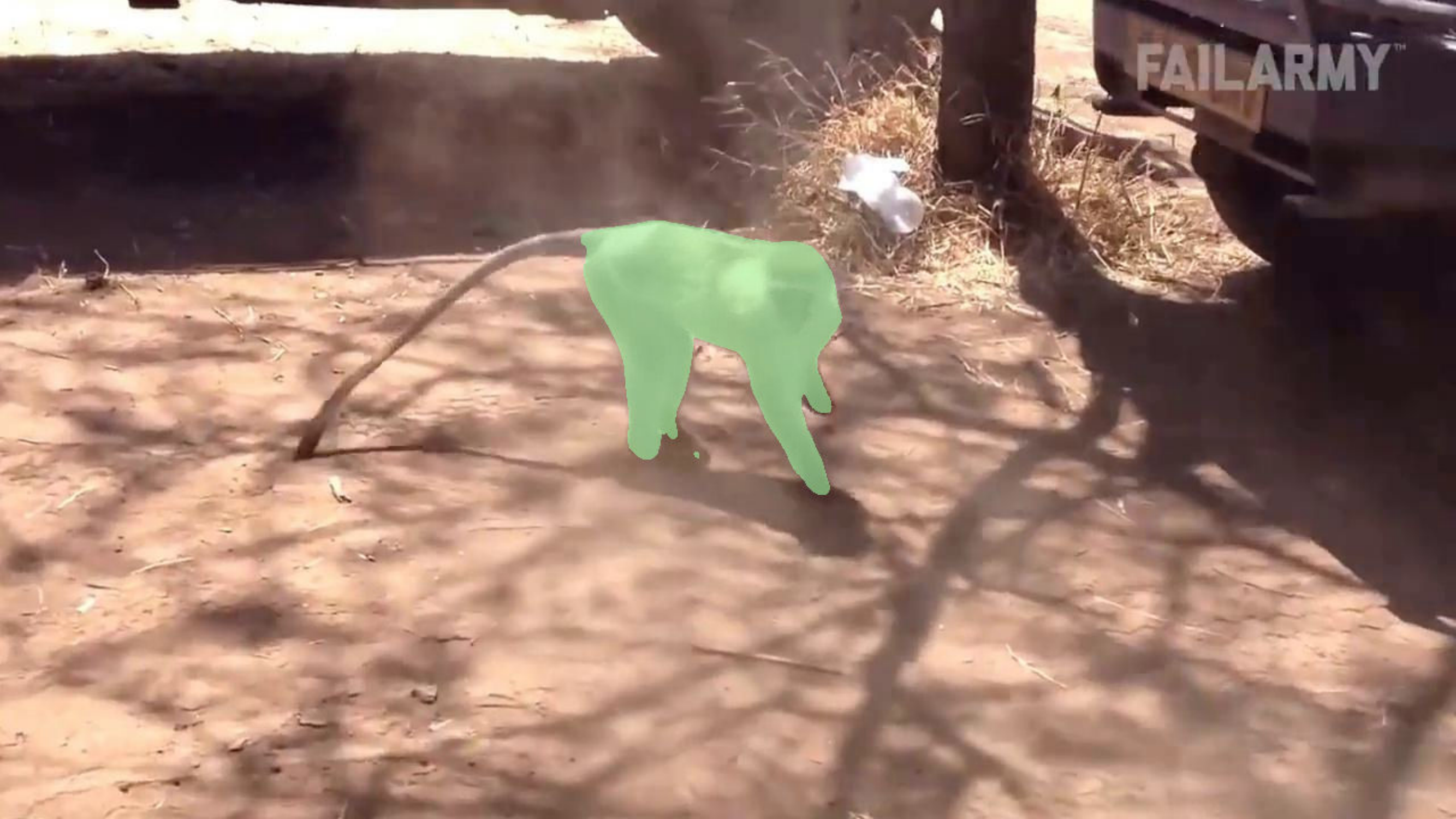}
        \vspace{-16pt}
    \end{subfigure}
    \begin{subfigure}[b]{0.32\linewidth}
        \includegraphics[width=\mysize\linewidth]{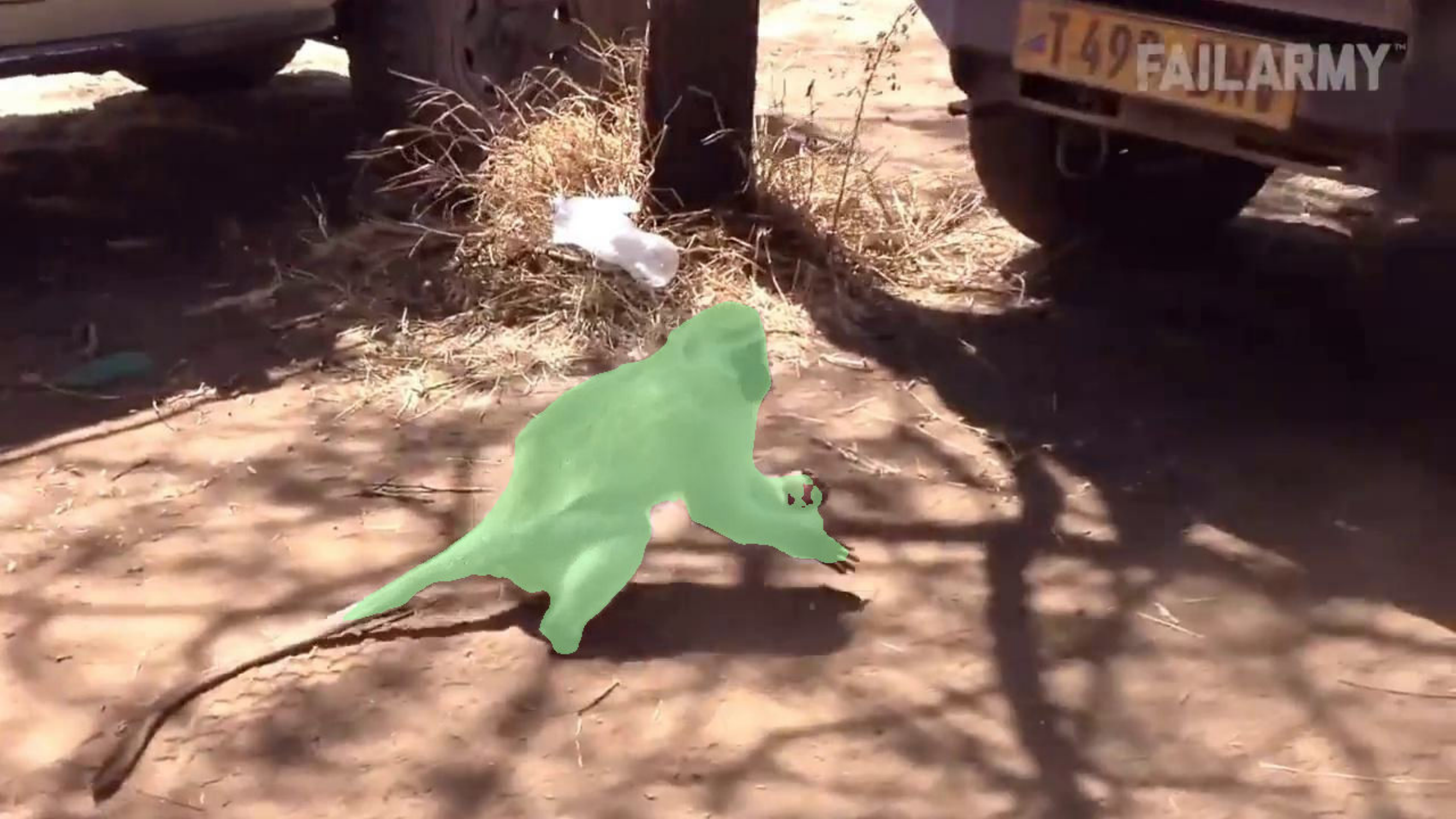}
        \vspace{-16pt}
    \end{subfigure}
    \begin{subfigure}[b]{0.32\linewidth}
     \includegraphics[width=\mysize\linewidth]{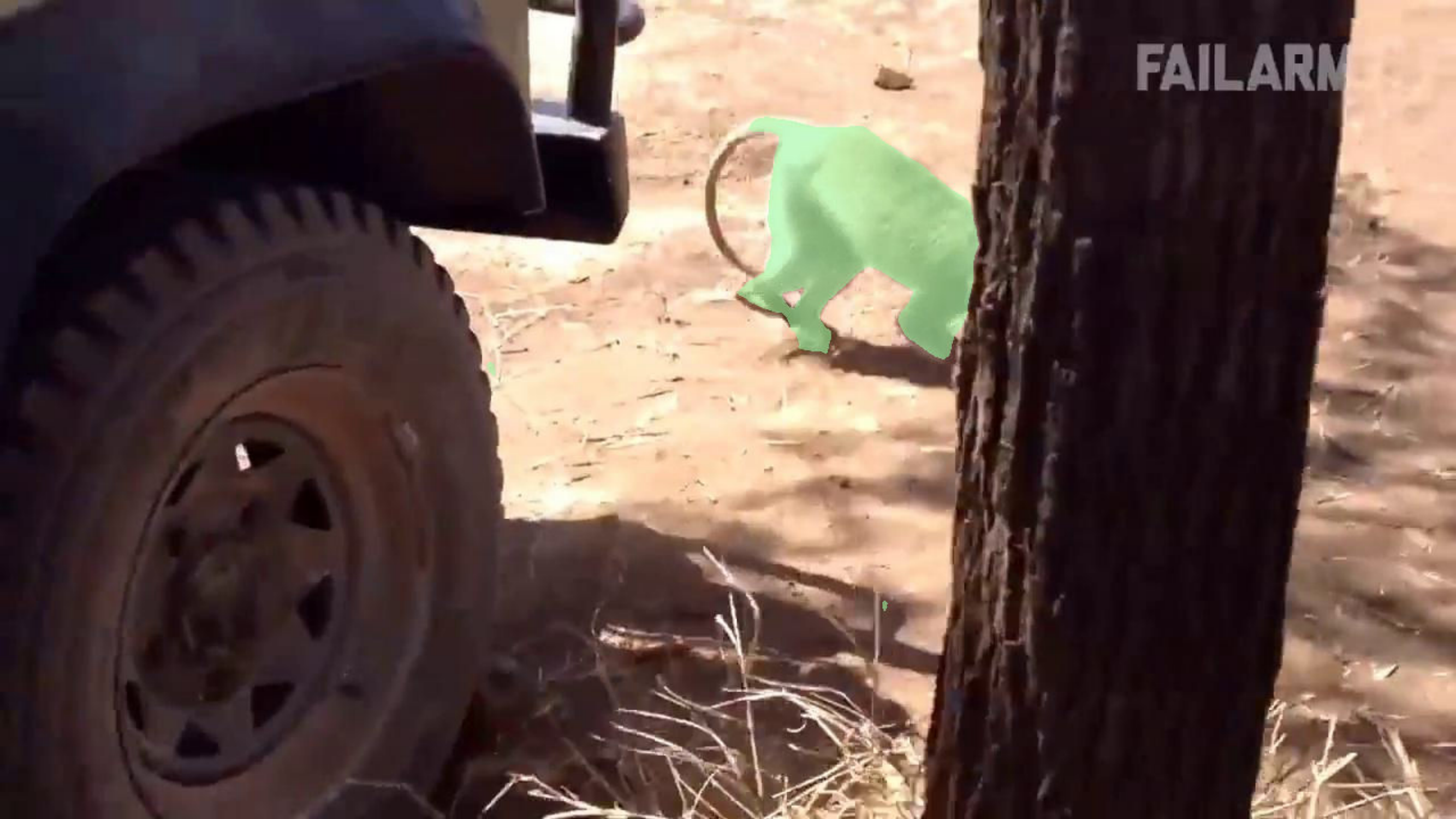}
        \vspace{-16pt}
    \end{subfigure}
    \vspace{5pt}
    \caption{\textbf{Tracking results of \pointstcn{} on \pvoops{}.} The model is trained on \pvoops{} with points, then evaluated on the 10-point setup.}
    \label{fig:point_results_oops}
\end{figure*}

\begin{figure*}[ht!]
    \centering
    \setlength{\fboxsep}{0.32pt}
    \begin{subfigure}[b]{0.32\linewidth}
        \includegraphics[width=\mysize\linewidth]{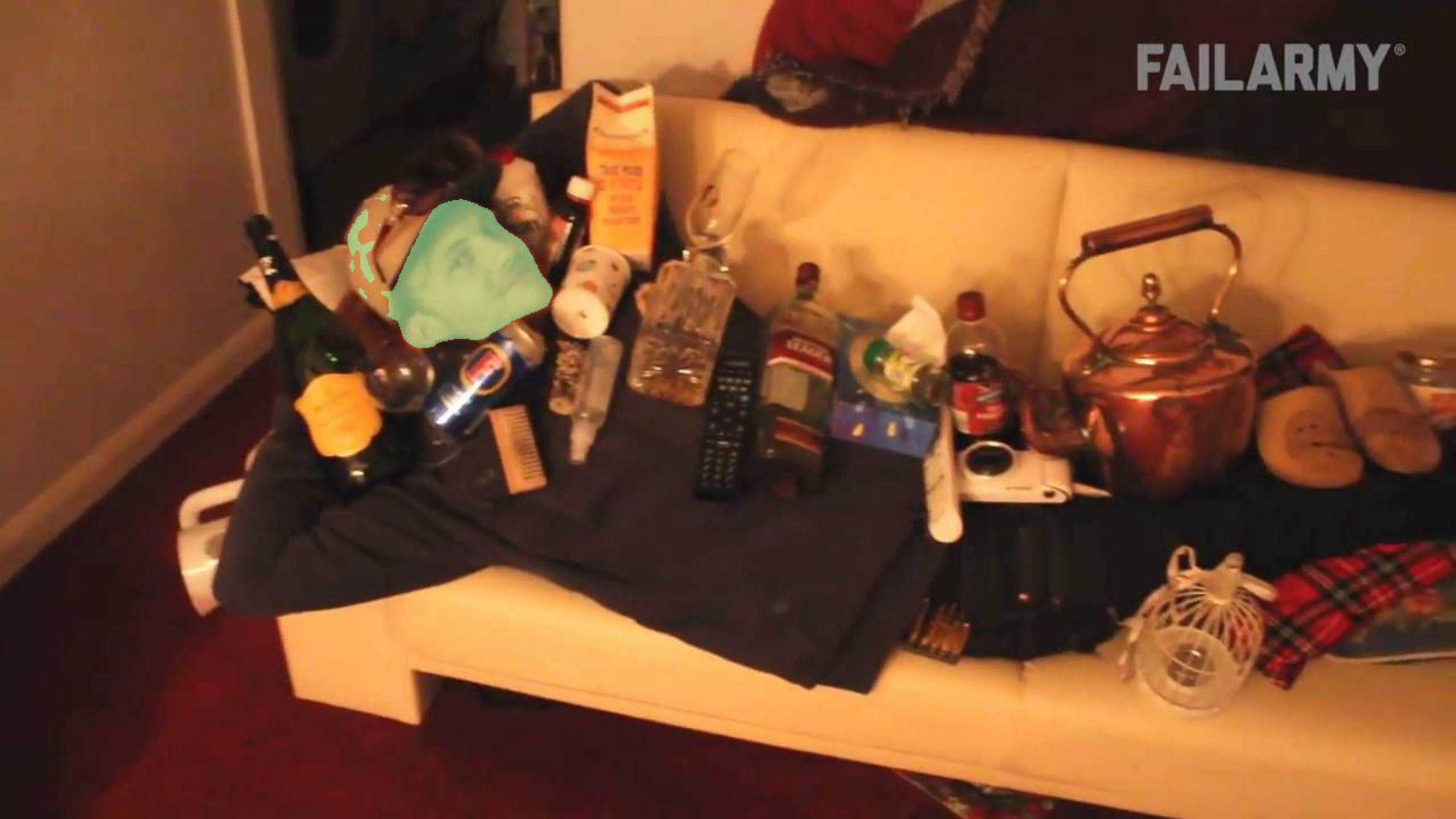}
        \vspace{-16pt}
    \end{subfigure}
    \begin{subfigure}[b]{0.32\linewidth}
     \includegraphics[width=\mysize\linewidth]{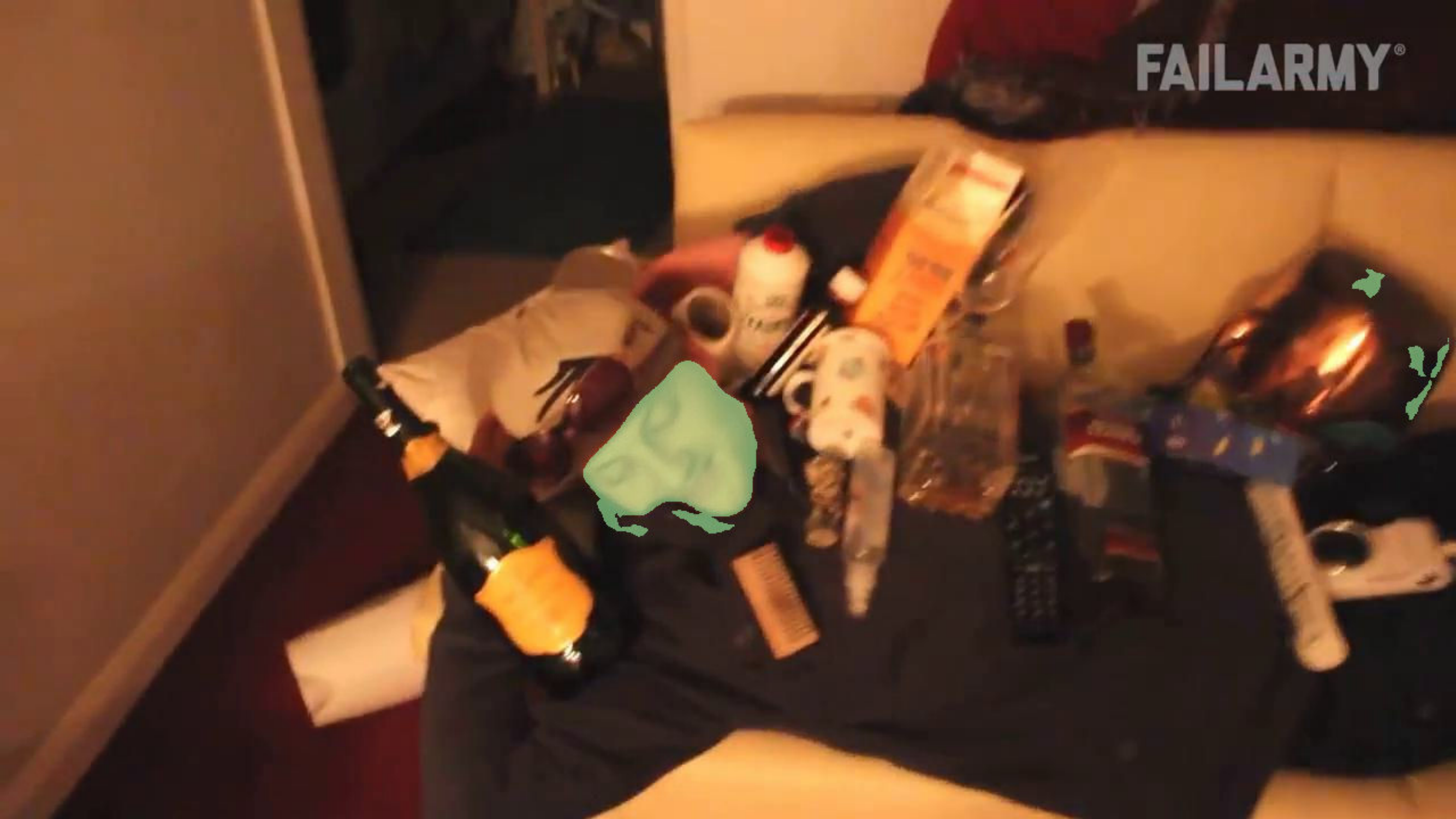}
        \vspace{-16pt}
    \end{subfigure}
    \begin{subfigure}[b]{0.32\linewidth}
     \includegraphics[width=\mysize\linewidth]{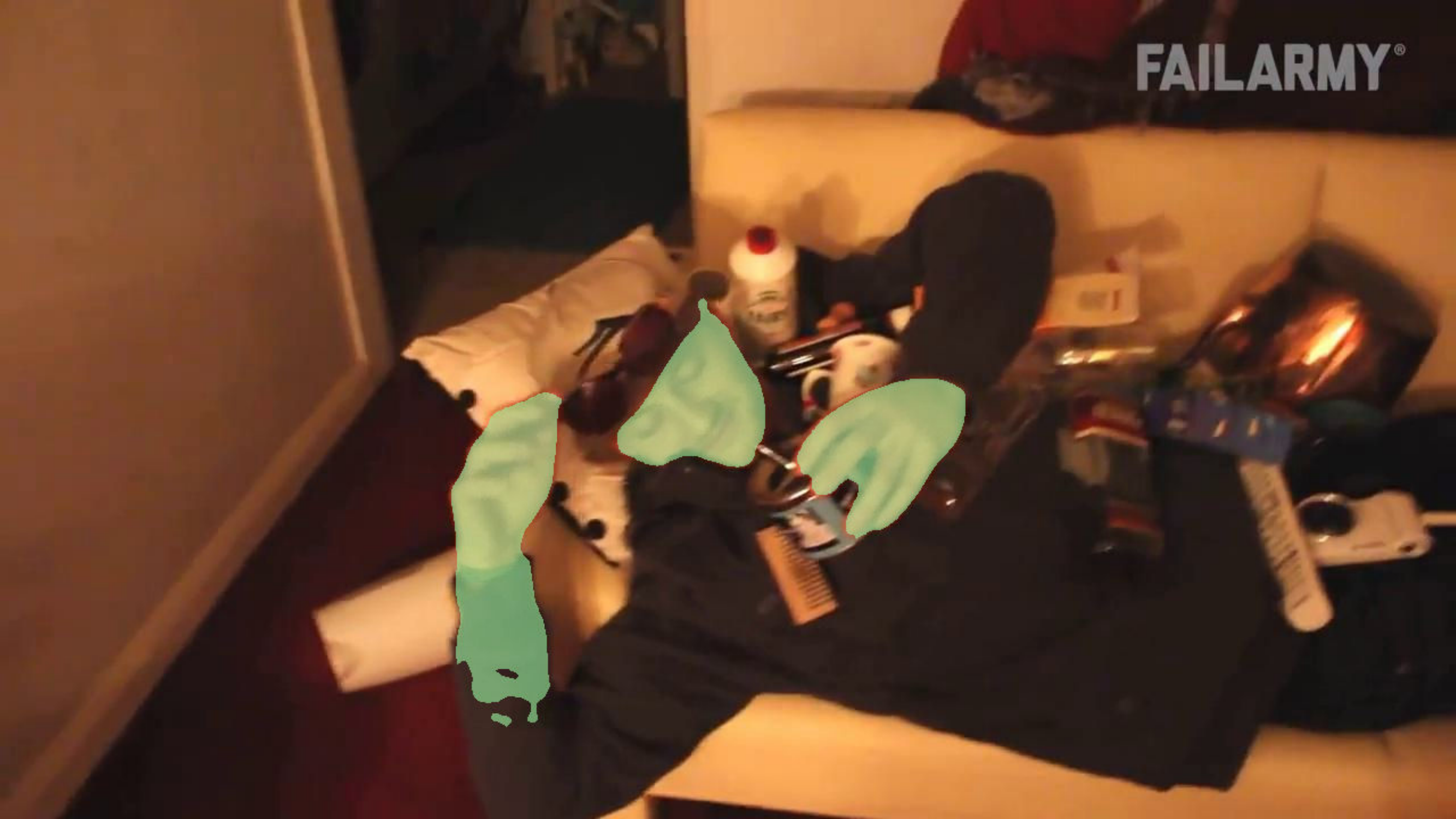}
        \vspace{-16pt}
    \end{subfigure}
    \vspace{5pt}

    \begin{subfigure}[b]{0.32\linewidth}
        \includegraphics[width=\mysize\linewidth]{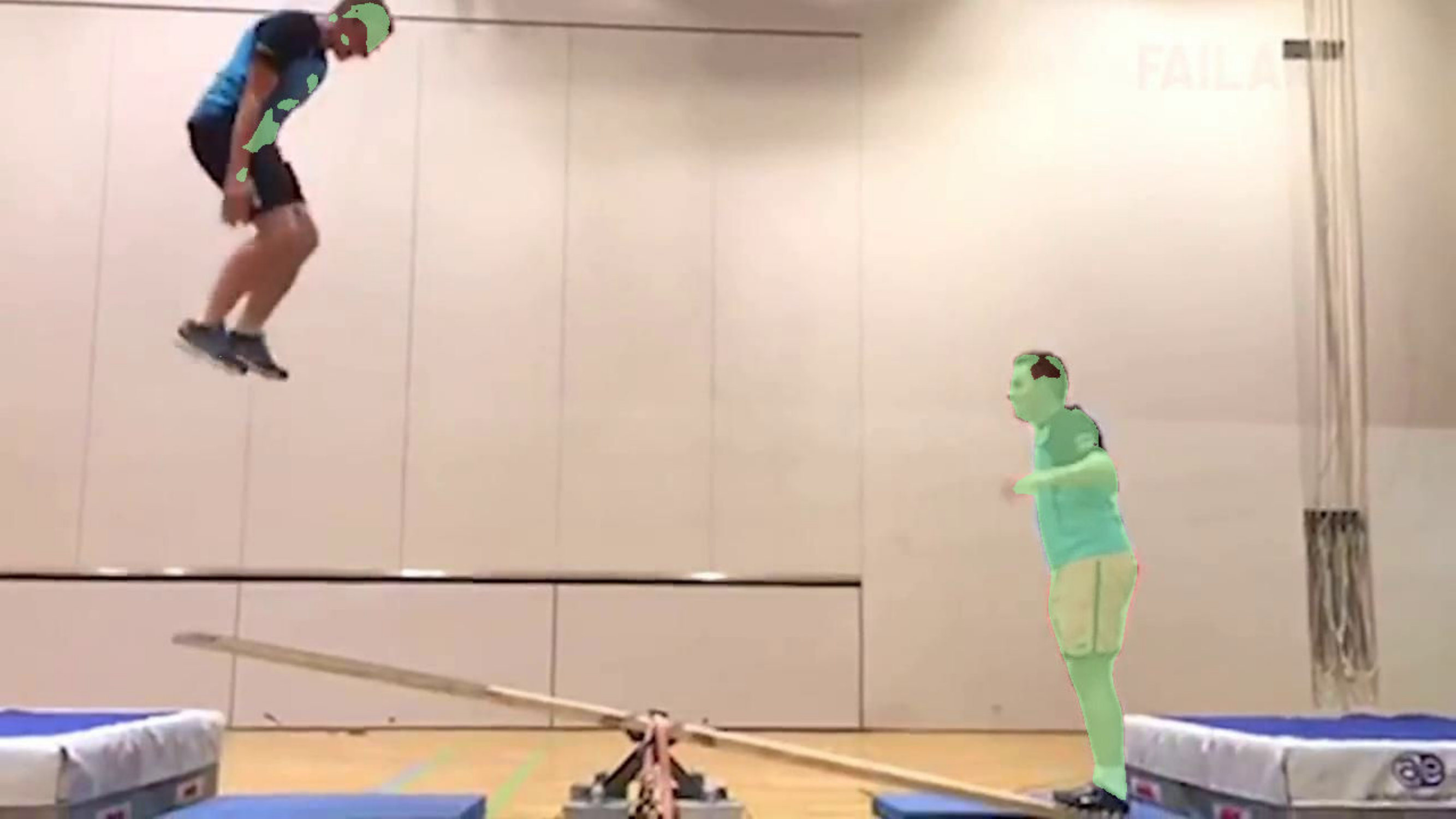}
        \vspace{-16pt}
    \end{subfigure}
    \begin{subfigure}[b]{0.32\linewidth}
     \includegraphics[width=\mysize\linewidth]{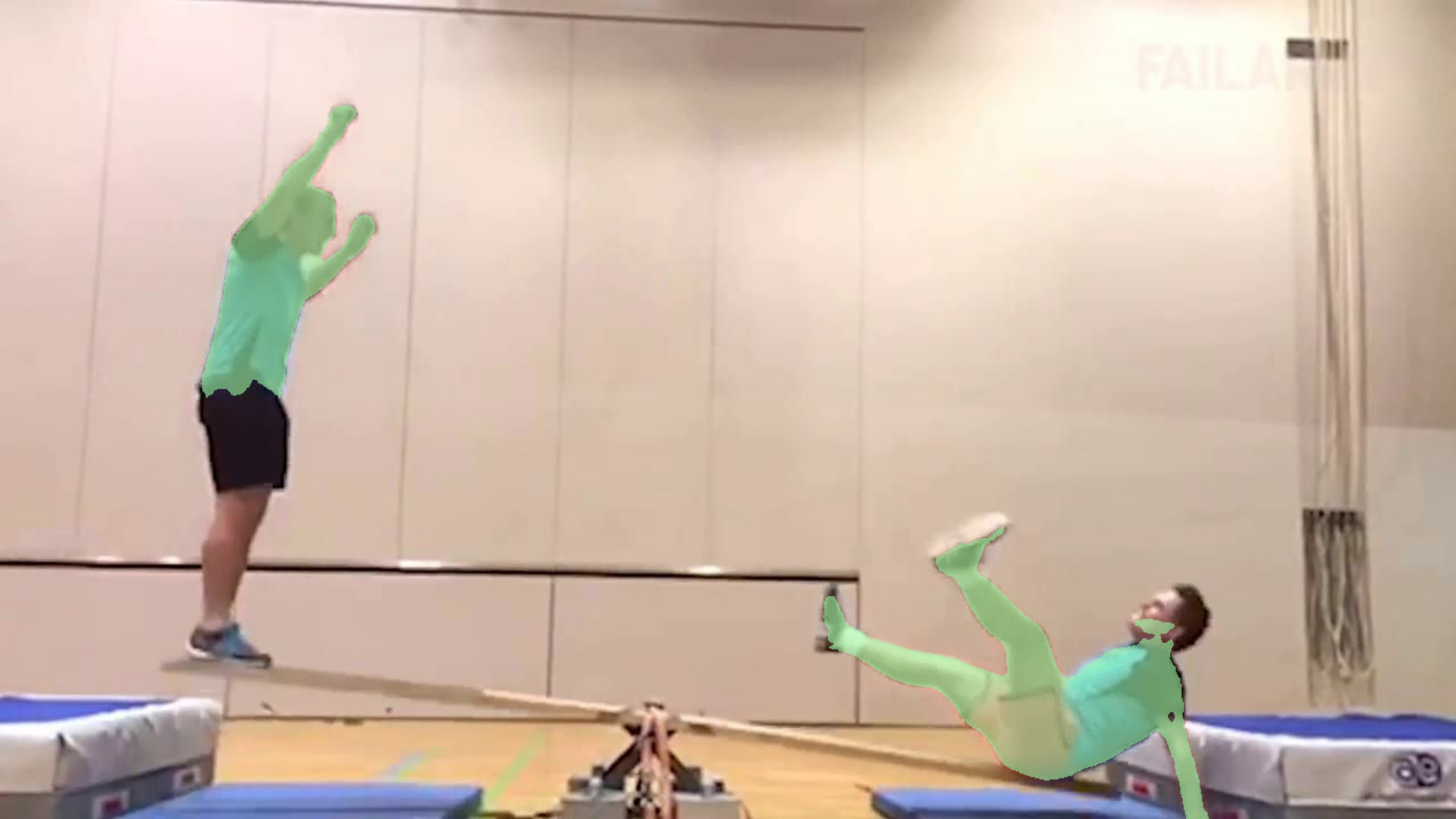}
        \vspace{-16pt}
    \end{subfigure}
    \begin{subfigure}[b]{0.32\linewidth}
     \includegraphics[width=\mysize\linewidth]{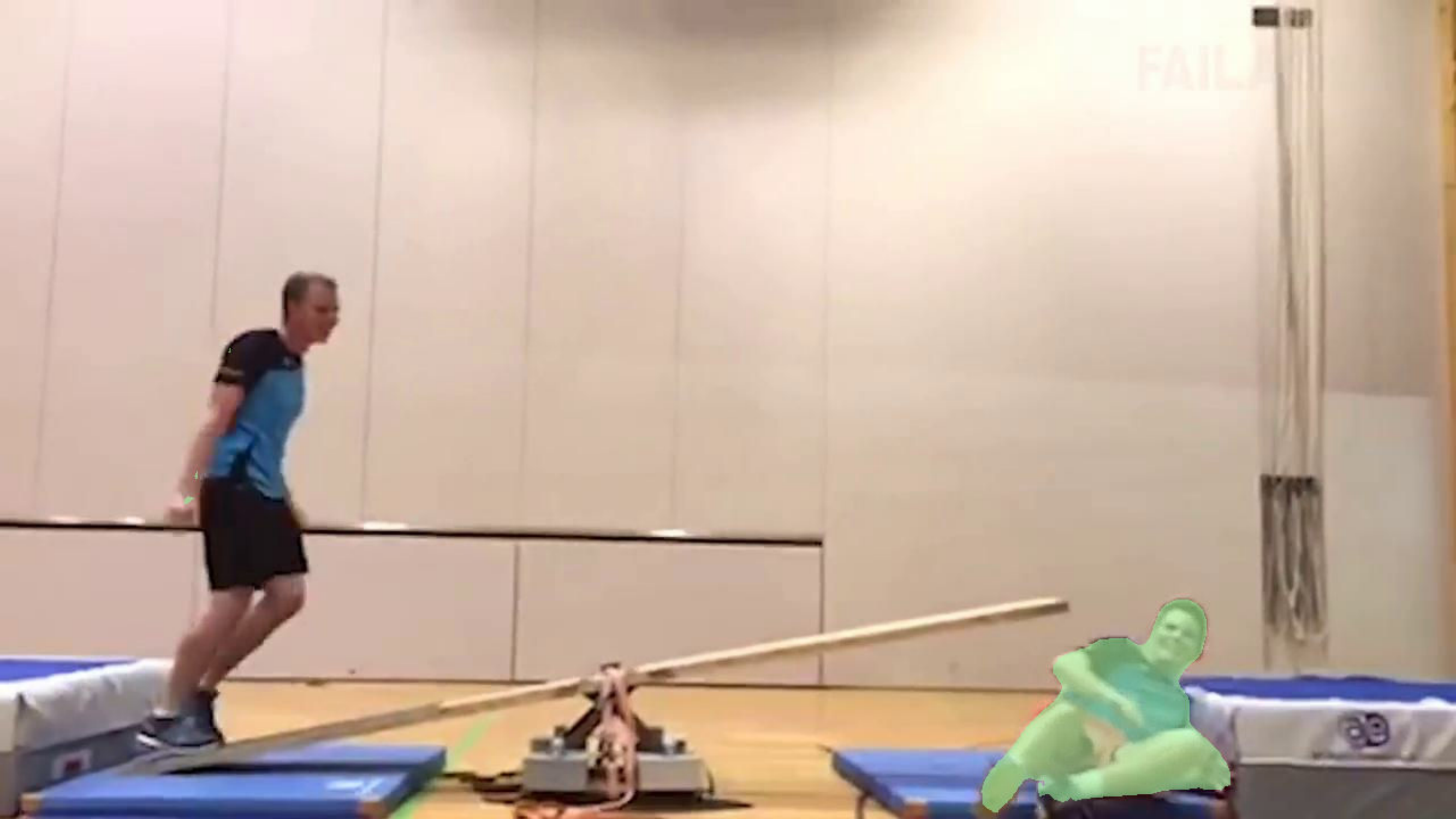}
        \vspace{-16pt}
    \end{subfigure}
    \vspace{5pt}
    
    \begin{subfigure}[b]{0.32\linewidth}
        \includegraphics[width=\mysize\linewidth]{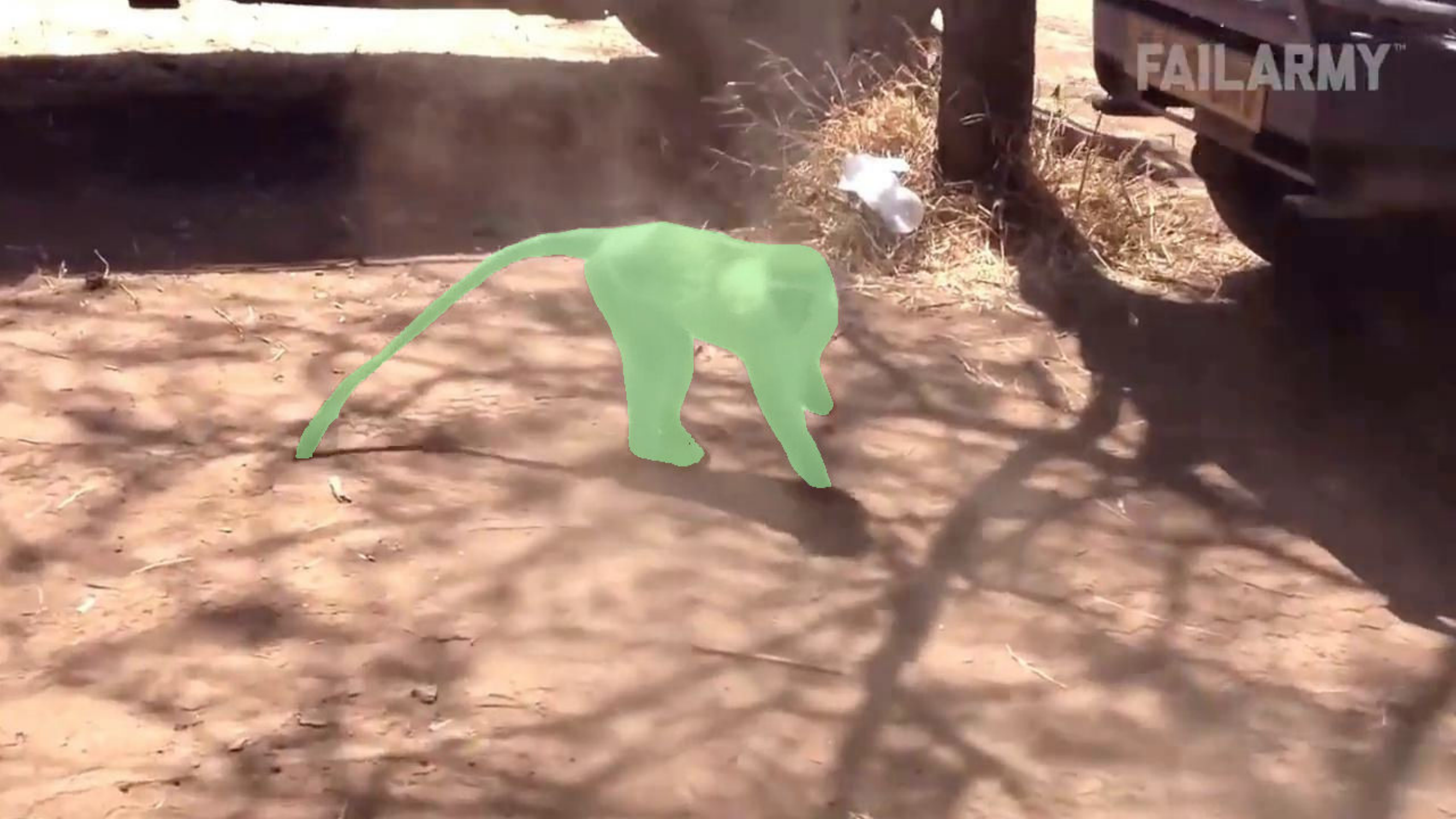}
        \vspace{-16pt}
    \end{subfigure}
    \begin{subfigure}[b]{0.32\linewidth}
        \includegraphics[width=\mysize\linewidth]{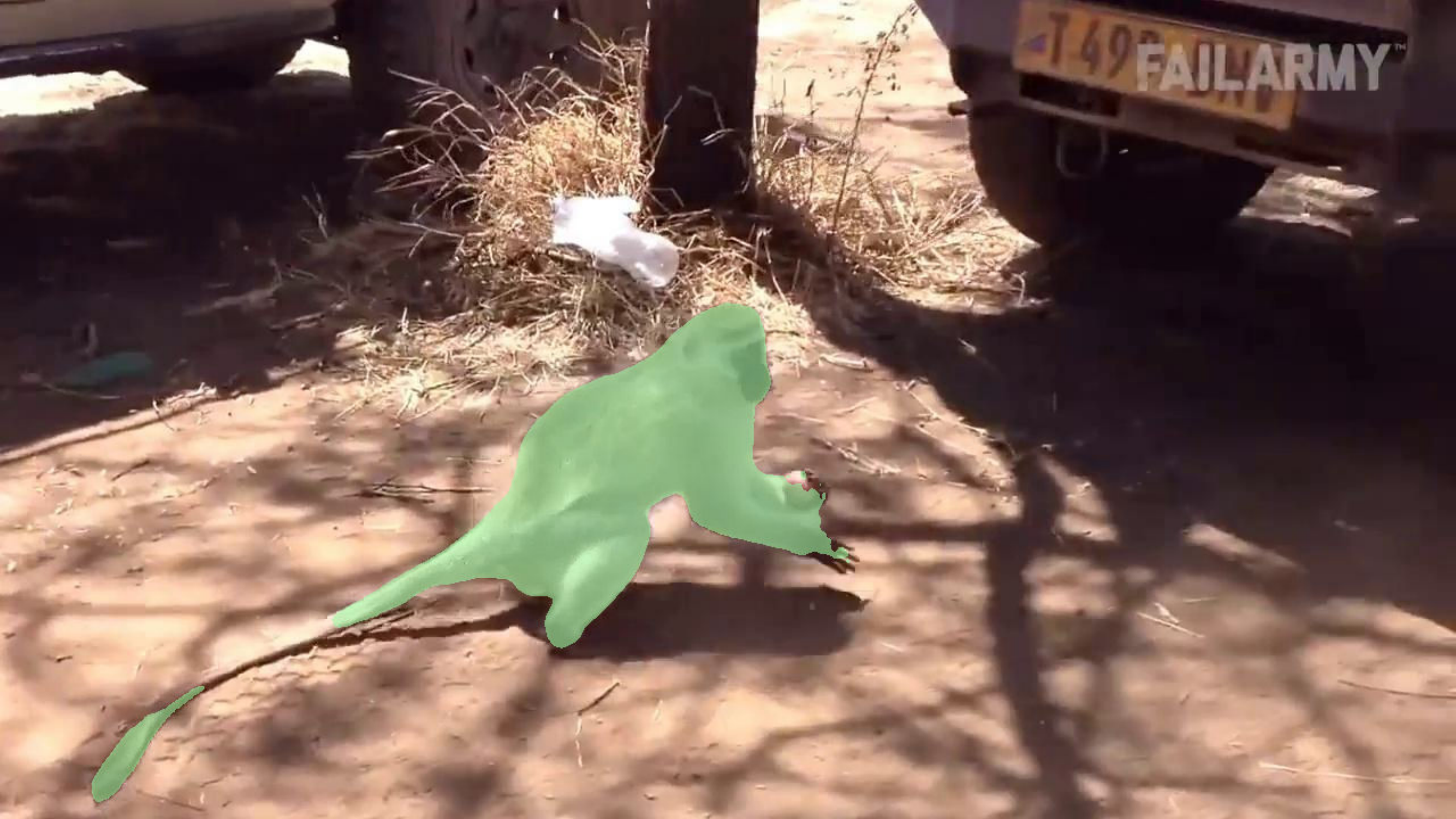}
        \vspace{-16pt}
    \end{subfigure}
    \begin{subfigure}[b]{0.32\linewidth}
     \includegraphics[width=\mysize\linewidth]{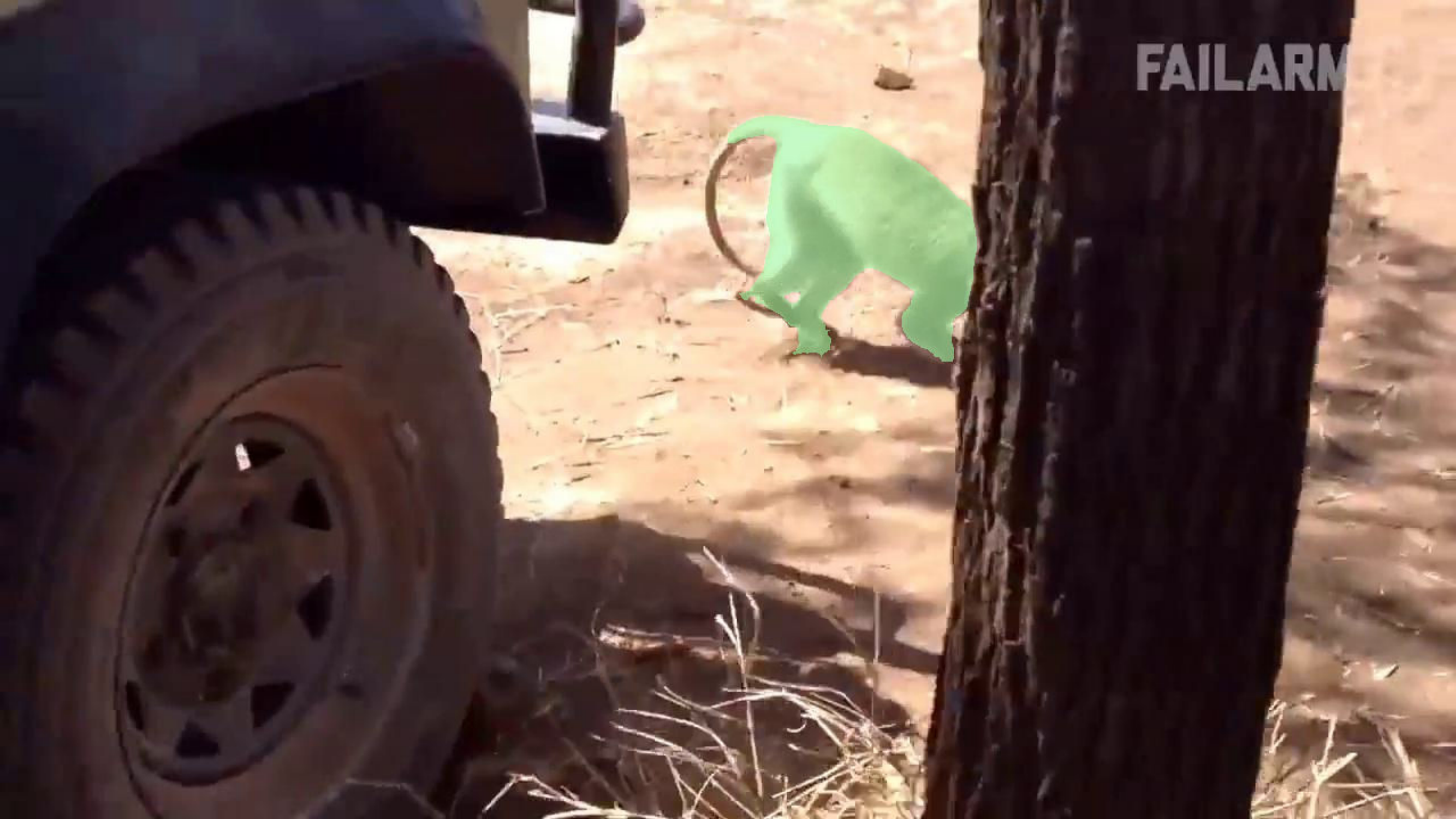}
        \vspace{-16pt}
    \end{subfigure}
    \vspace{5pt}

    \caption{\textbf{Tracking results of STCN~\cite{cheng2021stcn} on \pvoops{}.} The model is trained on \pvoops{} with~\dynamite~\cite{RanaMahadevan23Arxiv} pseudo-masks, then evaluated on 10 points setup.}
    \label{fig:pseudo_results_oops}
\end{figure*}

\begin{figure*}[ht!]
    \centering
    \setlength{\fboxsep}{0.32pt}
    \begin{subfigure}[b]{0.32\linewidth}
        \includegraphics[width=\mysize\linewidth]{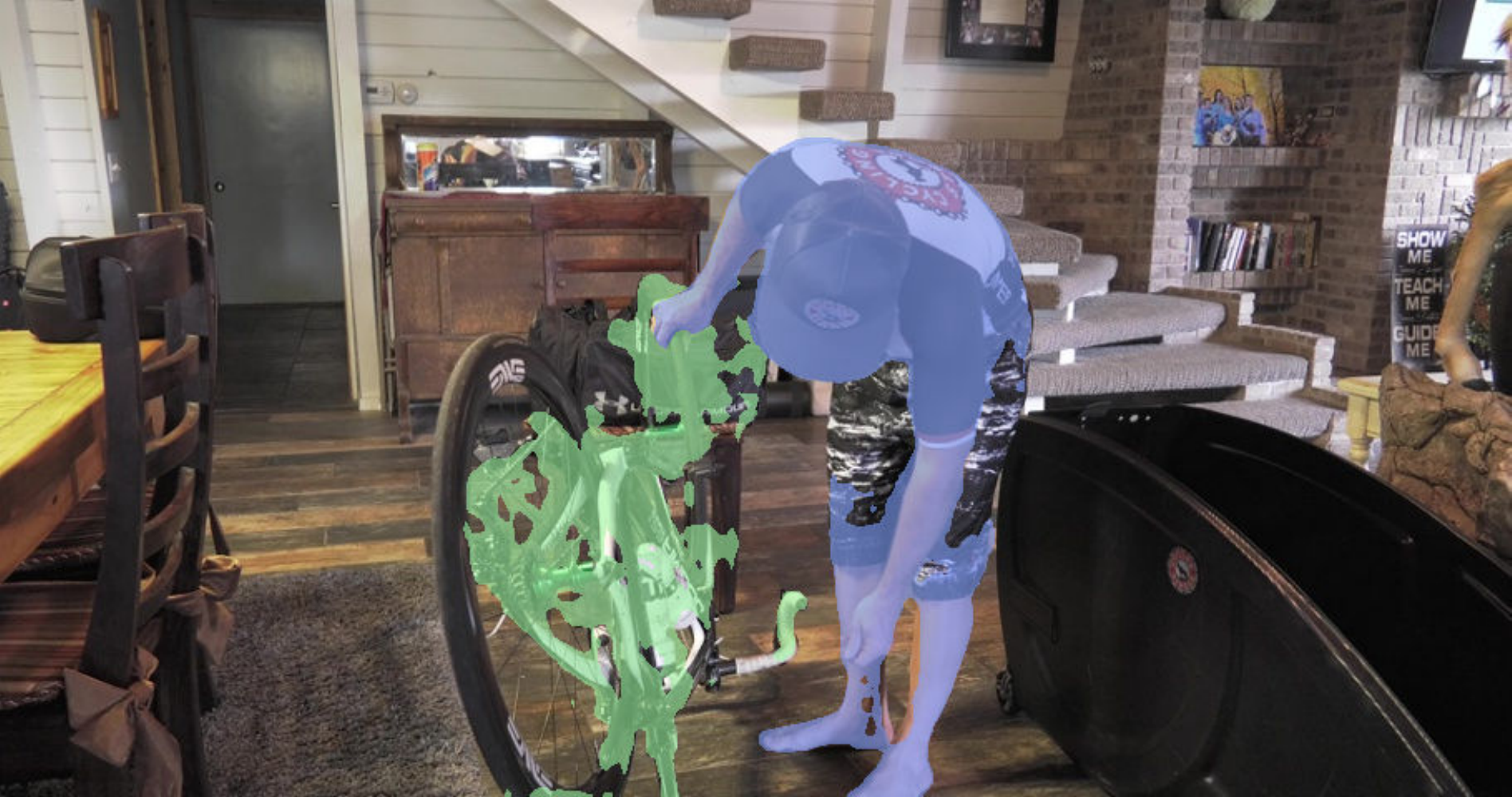}
        \vspace{-16pt}
    \end{subfigure}
    \begin{subfigure}[b]{0.32\linewidth}
     \includegraphics[width=\mysize\linewidth]{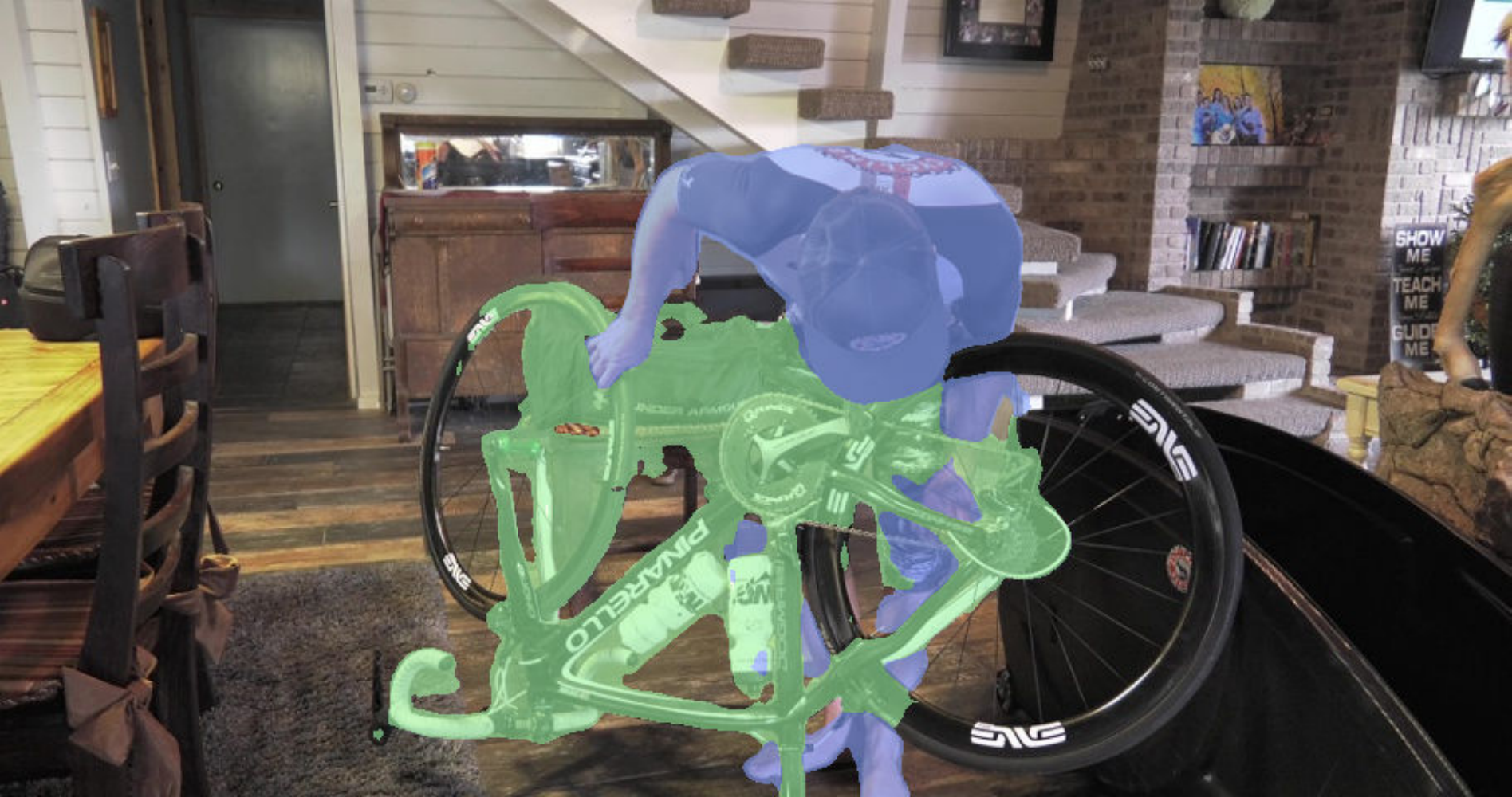}
        \vspace{-16pt}
    \end{subfigure}
    \begin{subfigure}[b]{0.32\linewidth}
     \includegraphics[width=\mysize\linewidth]{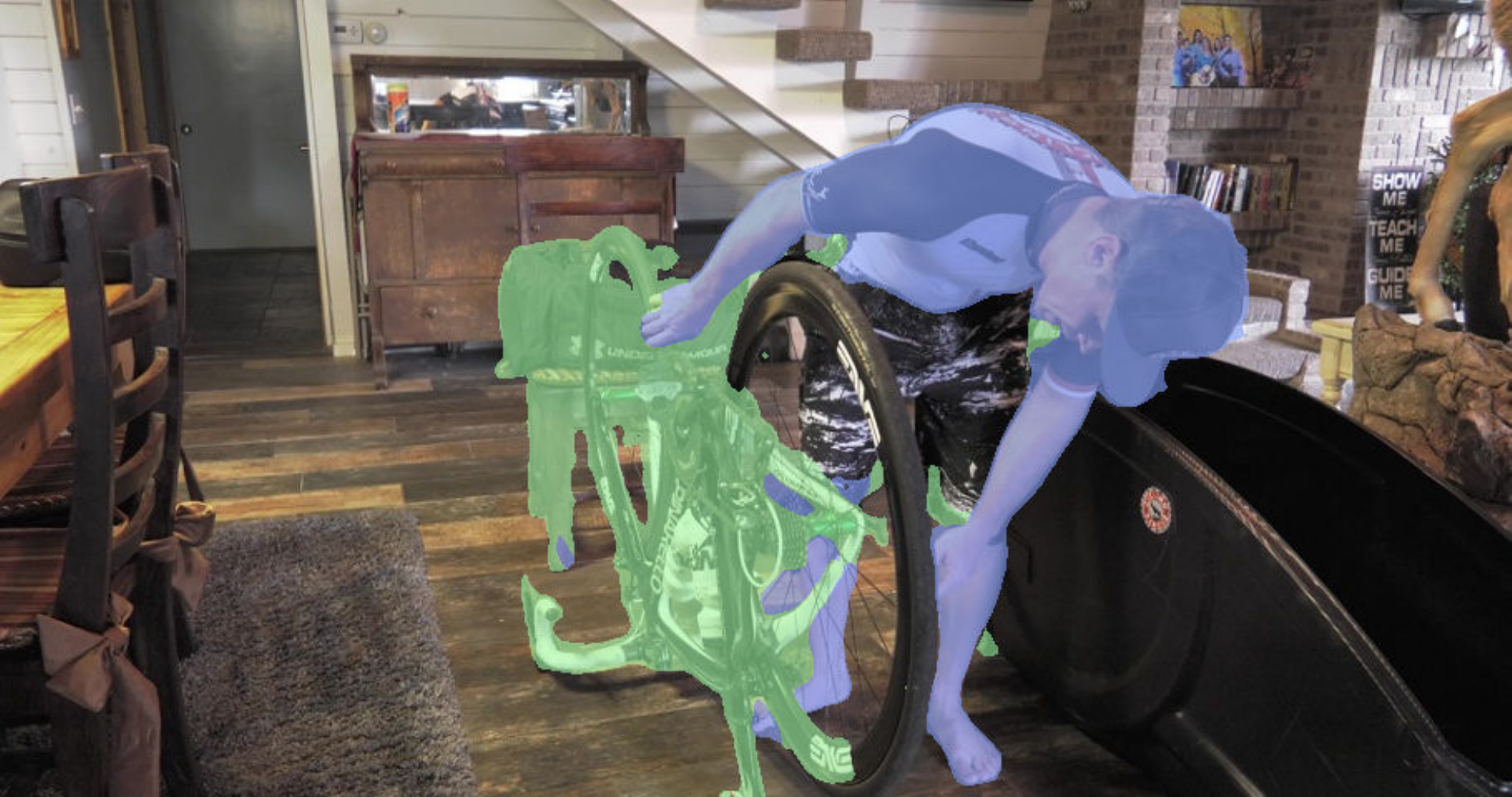}
        \vspace{-16pt}
    \end{subfigure}
    \vspace{5pt}

    \begin{subfigure}[b]{0.32\linewidth}
        \includegraphics[width=\mysize\linewidth]{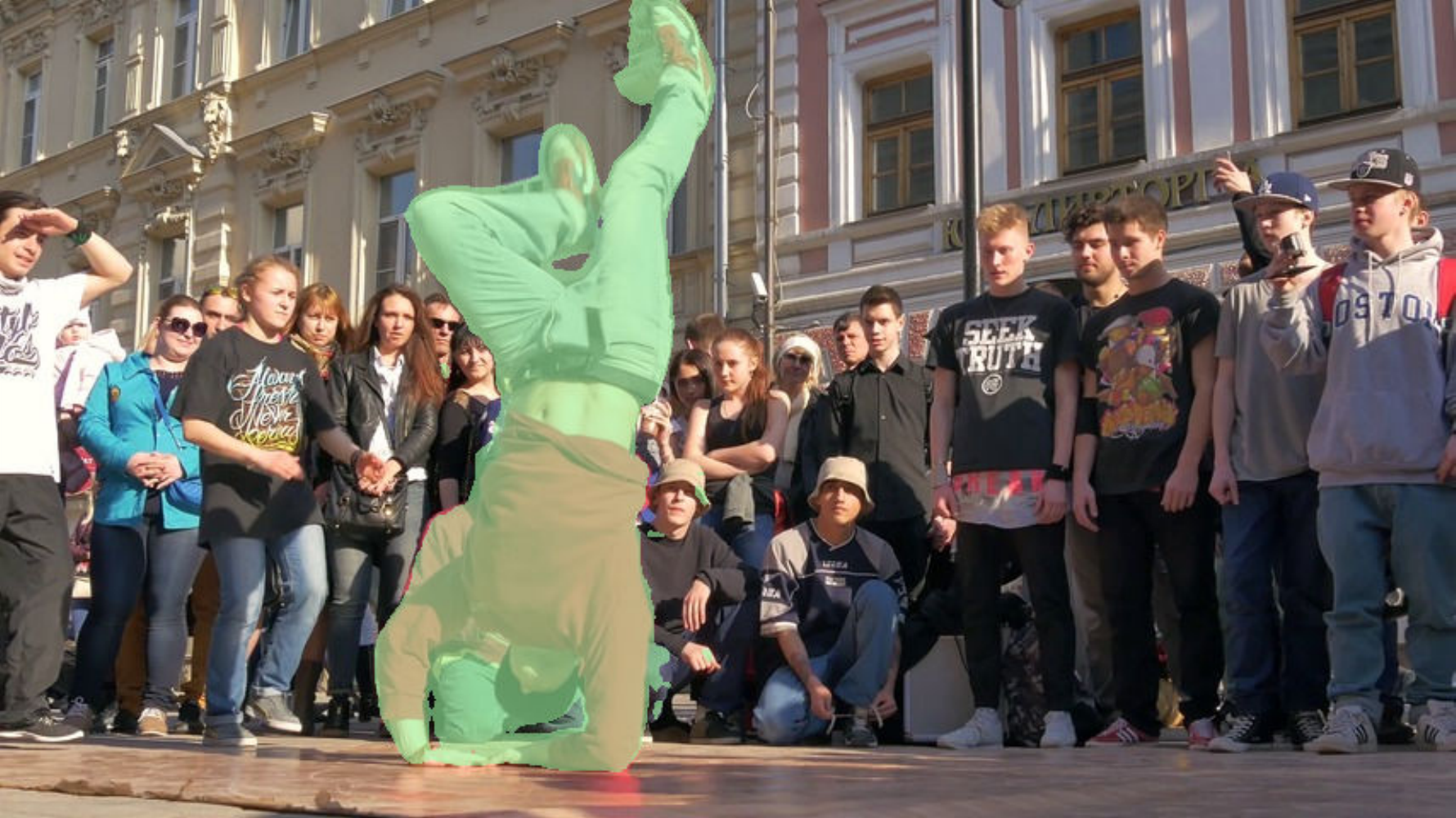}
        \vspace{-16pt}
    \end{subfigure}
    \begin{subfigure}[b]{0.32\linewidth}
     \includegraphics[width=\mysize\linewidth]{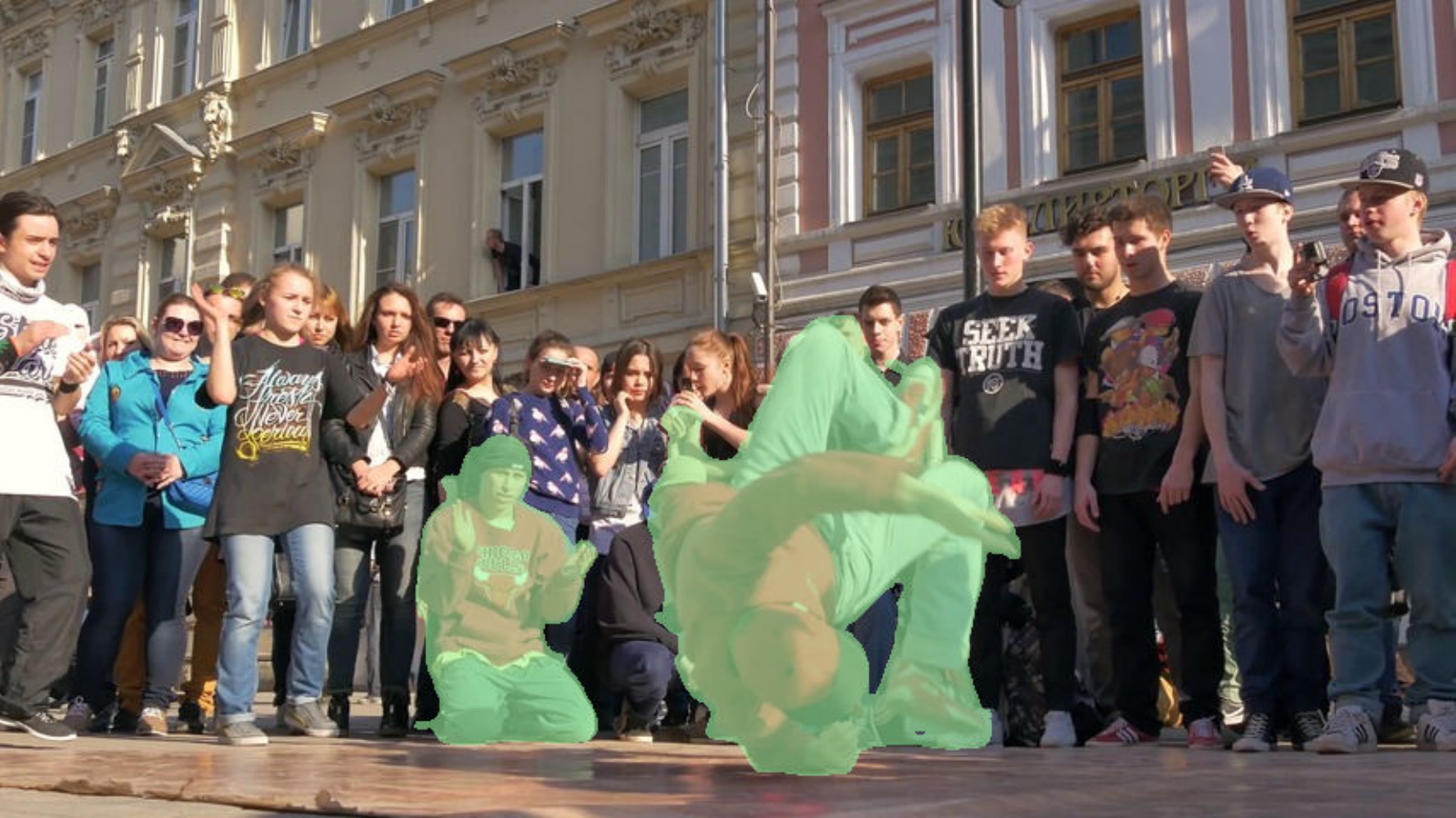}
        \vspace{-16pt}
    \end{subfigure}
    \begin{subfigure}[b]{0.32\linewidth}
     \includegraphics[width=\mysize\linewidth]{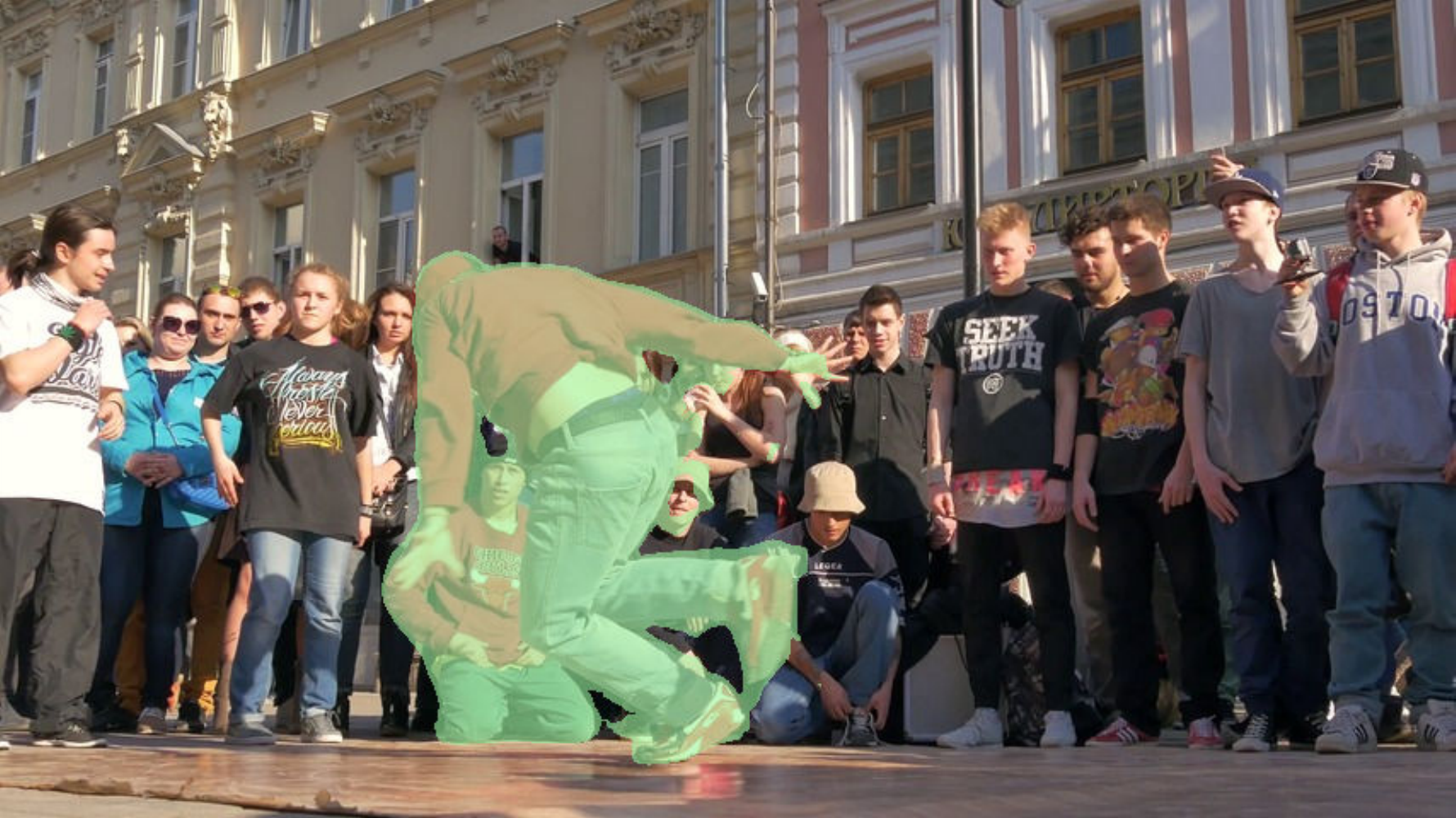}
        \vspace{-16pt}
    \end{subfigure}
    \vspace{5pt}

    \begin{subfigure}[b]{0.32\linewidth}
        \includegraphics[width=\mysize\linewidth]{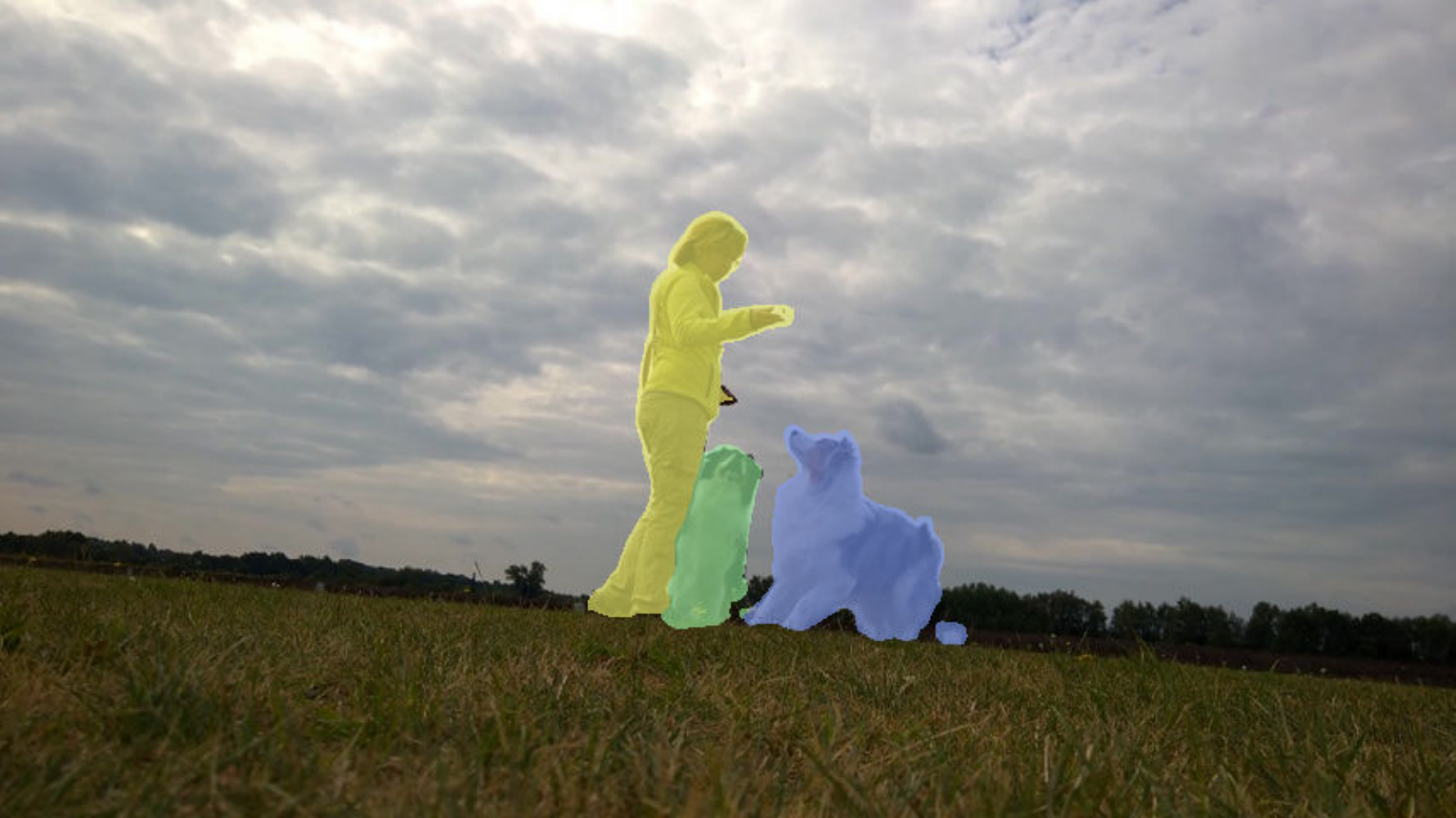}
        \vspace{-16pt}
    \end{subfigure}
    \begin{subfigure}[b]{0.32\linewidth}
        \includegraphics[width=\mysize\linewidth]{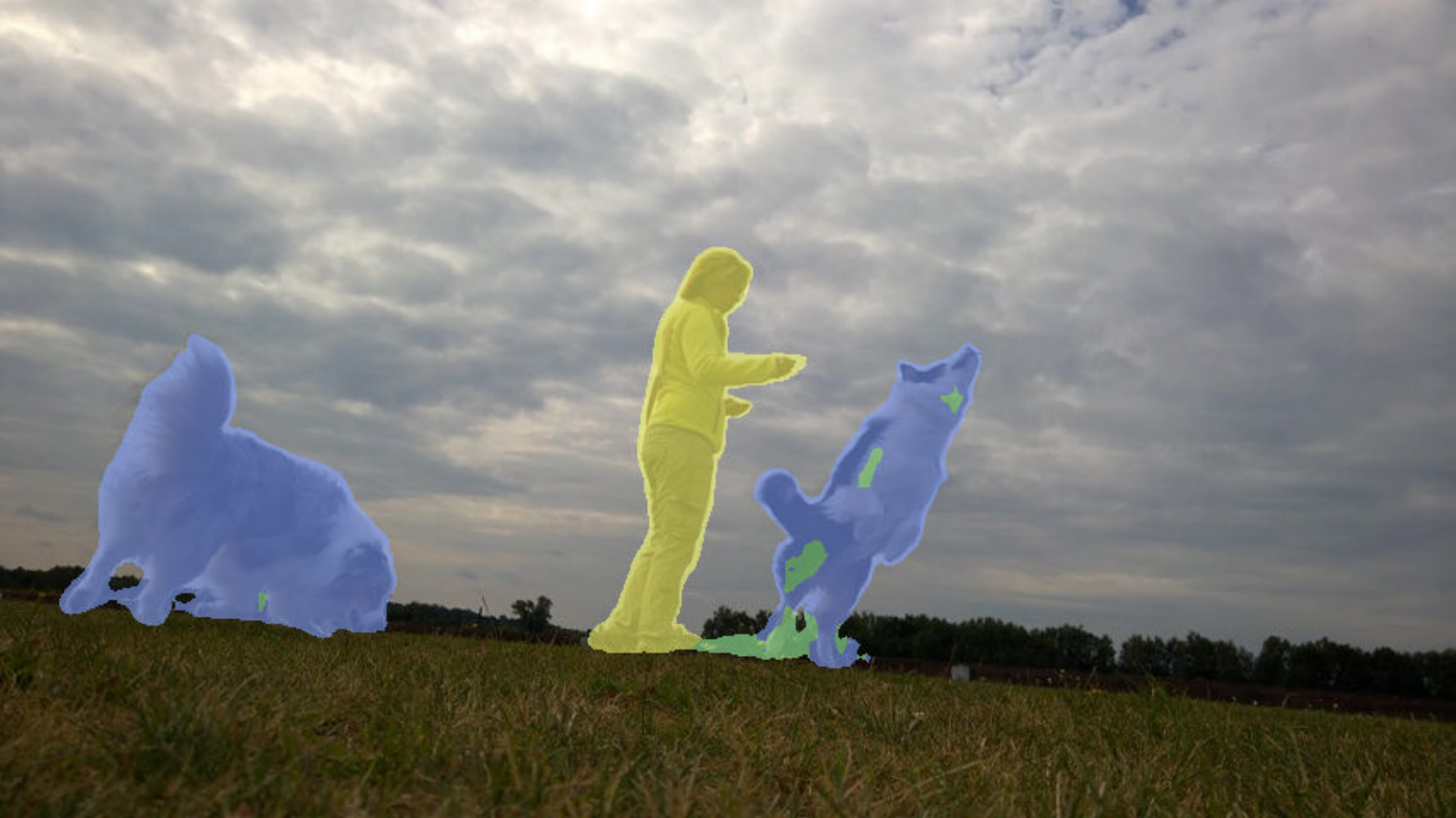}
        \vspace{-16pt}
    \end{subfigure}
    \begin{subfigure}[b]{0.32\linewidth}
     \includegraphics[width=\mysize\linewidth]{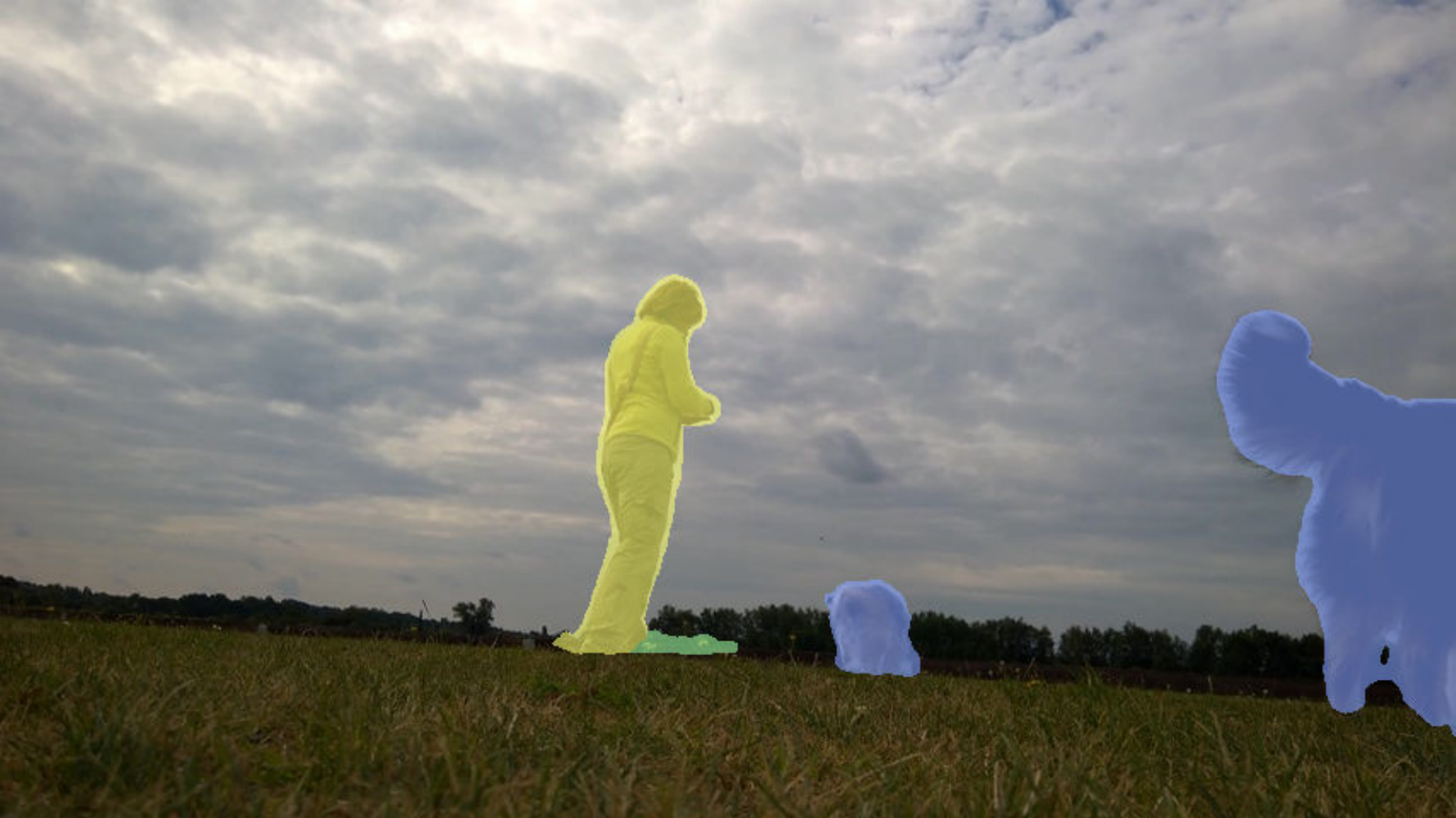}
        \vspace{-16pt}
    \end{subfigure}
    \vspace{5pt}

    \caption{\textbf{Tracking results of \pointstcn{} on \pvdavis{}.} The model is first pre-trained on \pvoops{} and \pvkinetics{} with points, then fine-tuned on \pvdavis{} and \pvytvos{} with points, and finally evaluated on the 10-point setup.}
    \label{fig:point_results_davis}
\end{figure*}

\begin{figure*}[ht!]
    \centering
    \setlength{\fboxsep}{0.32pt}
    \begin{subfigure}[b]{0.32\linewidth}
        \includegraphics[width=\mysize\linewidth]{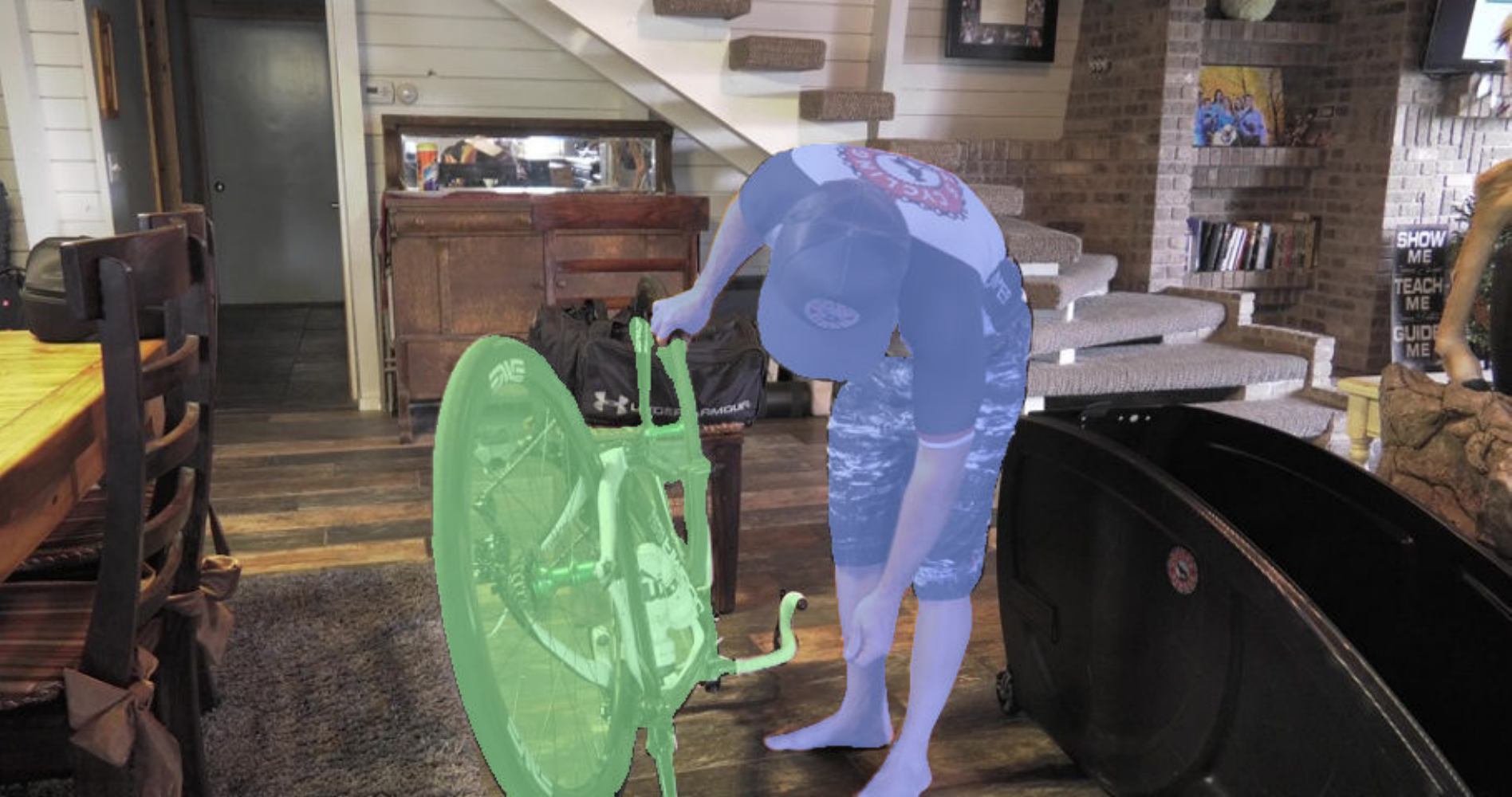}
        \vspace{-16pt}
    \end{subfigure}
    \begin{subfigure}[b]{0.32\linewidth}
     \includegraphics[width=\mysize\linewidth]{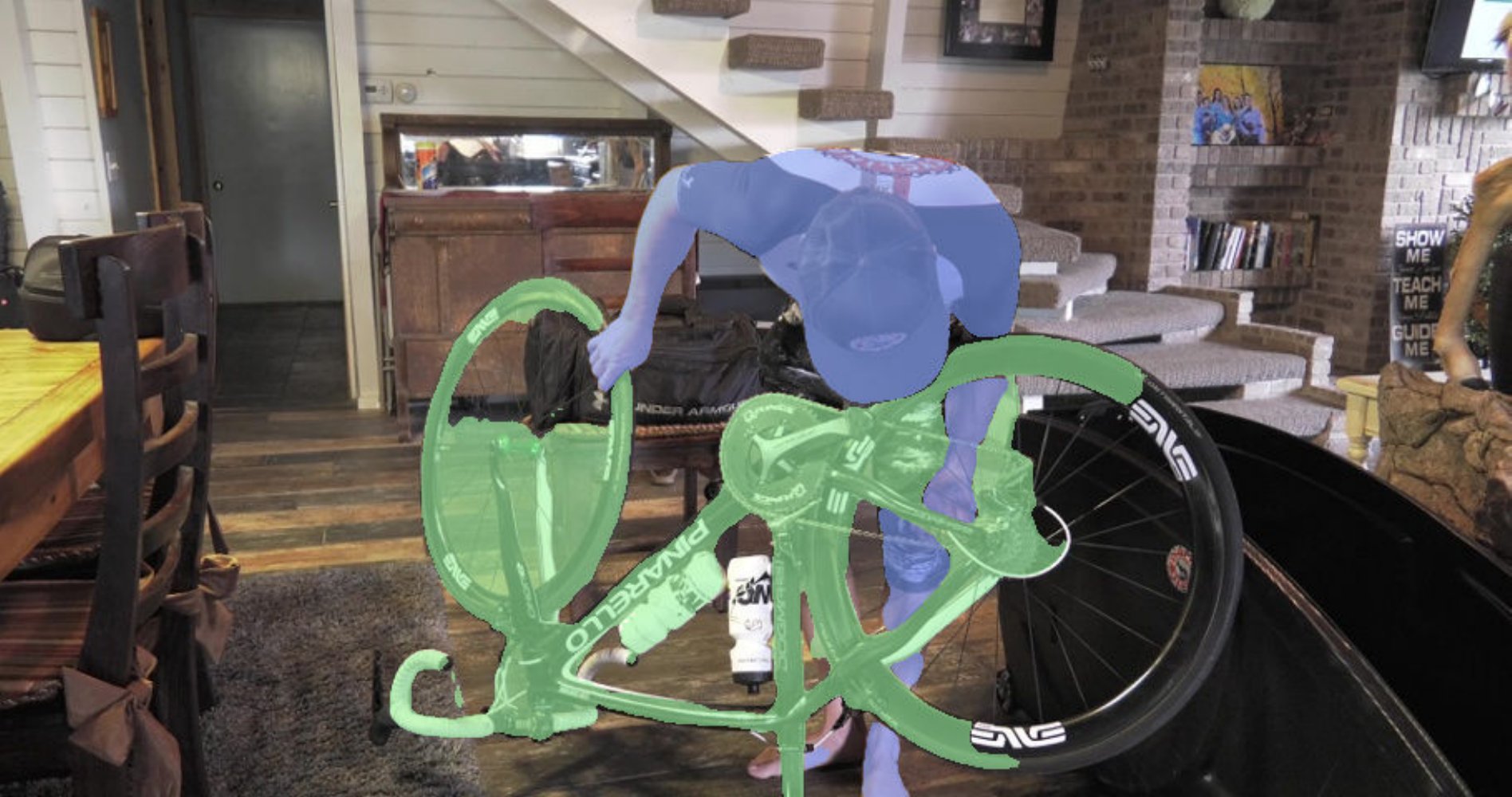}
        \vspace{-16pt}
    \end{subfigure}
    \begin{subfigure}[b]{0.32\linewidth}
     \includegraphics[width=\mysize\linewidth]{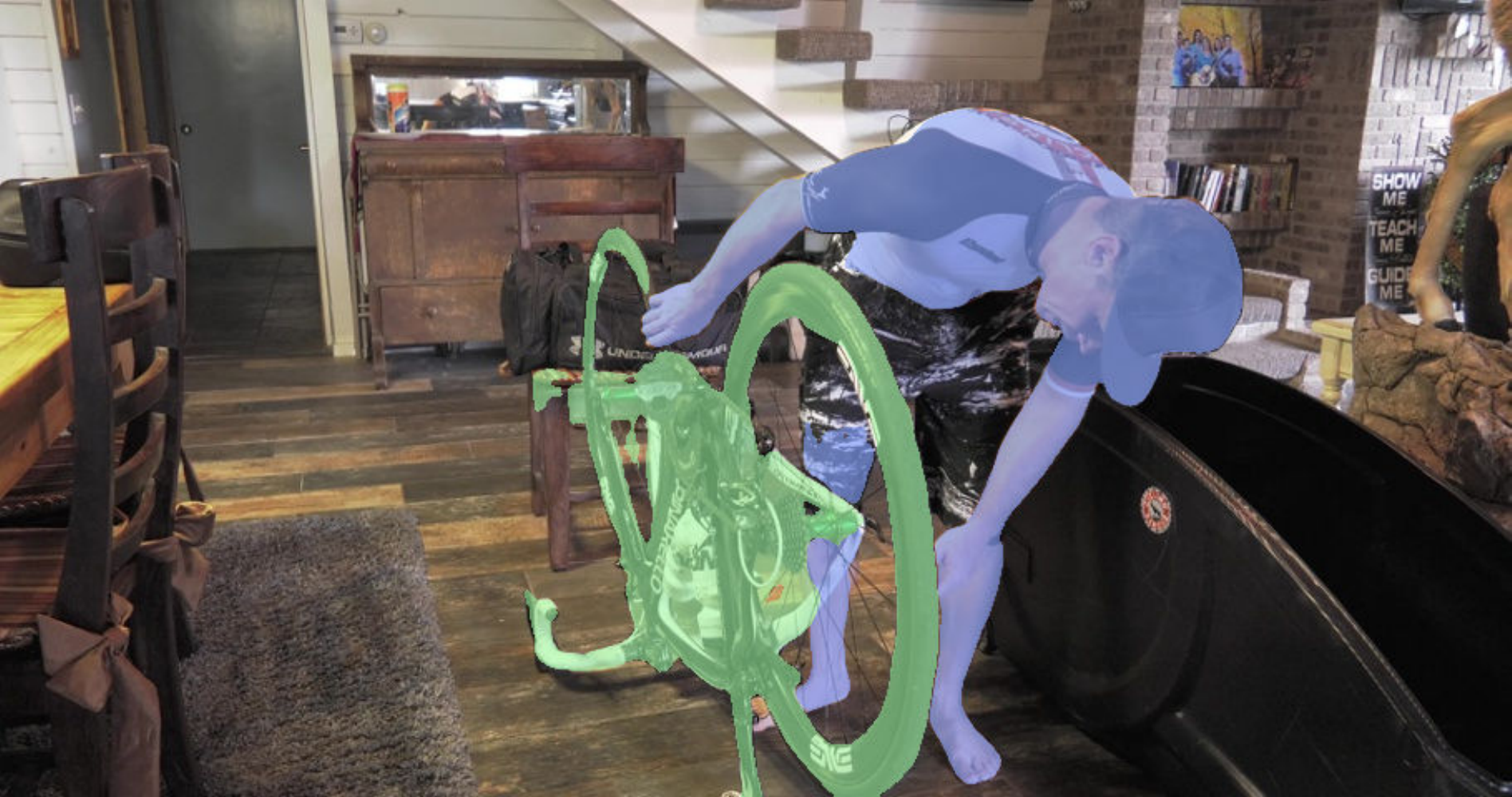}
        \vspace{-16pt}
    \end{subfigure}
    \vspace{5pt}

    \begin{subfigure}[b]{0.32\linewidth}
        \includegraphics[width=\mysize\linewidth]{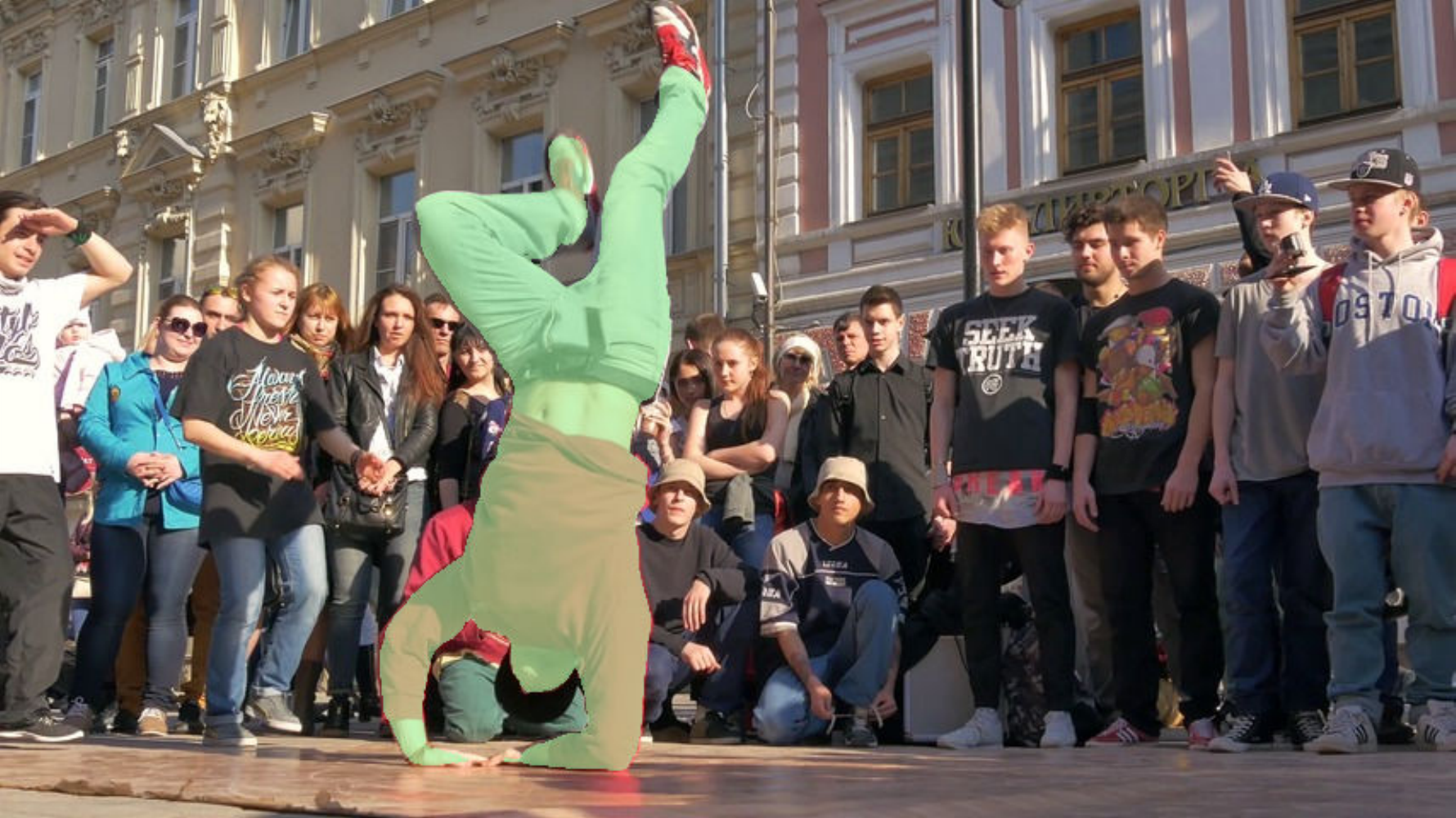}
        \vspace{-16pt}
    \end{subfigure}
    \begin{subfigure}[b]{0.32\linewidth}
     \includegraphics[width=\mysize\linewidth]{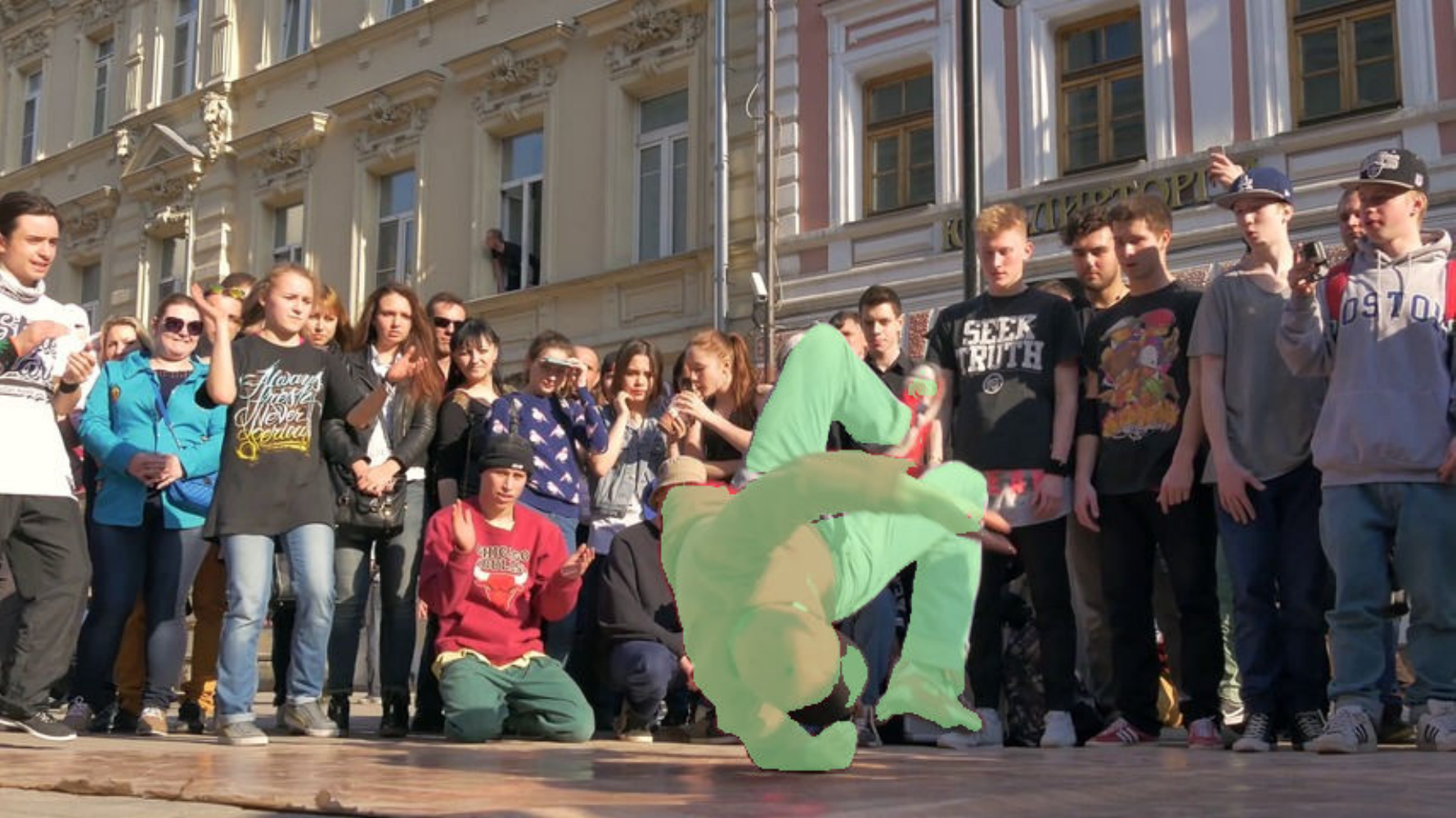}
        \vspace{-16pt}
    \end{subfigure}
    \begin{subfigure}[b]{0.32\linewidth}
     \includegraphics[width=\mysize\linewidth]{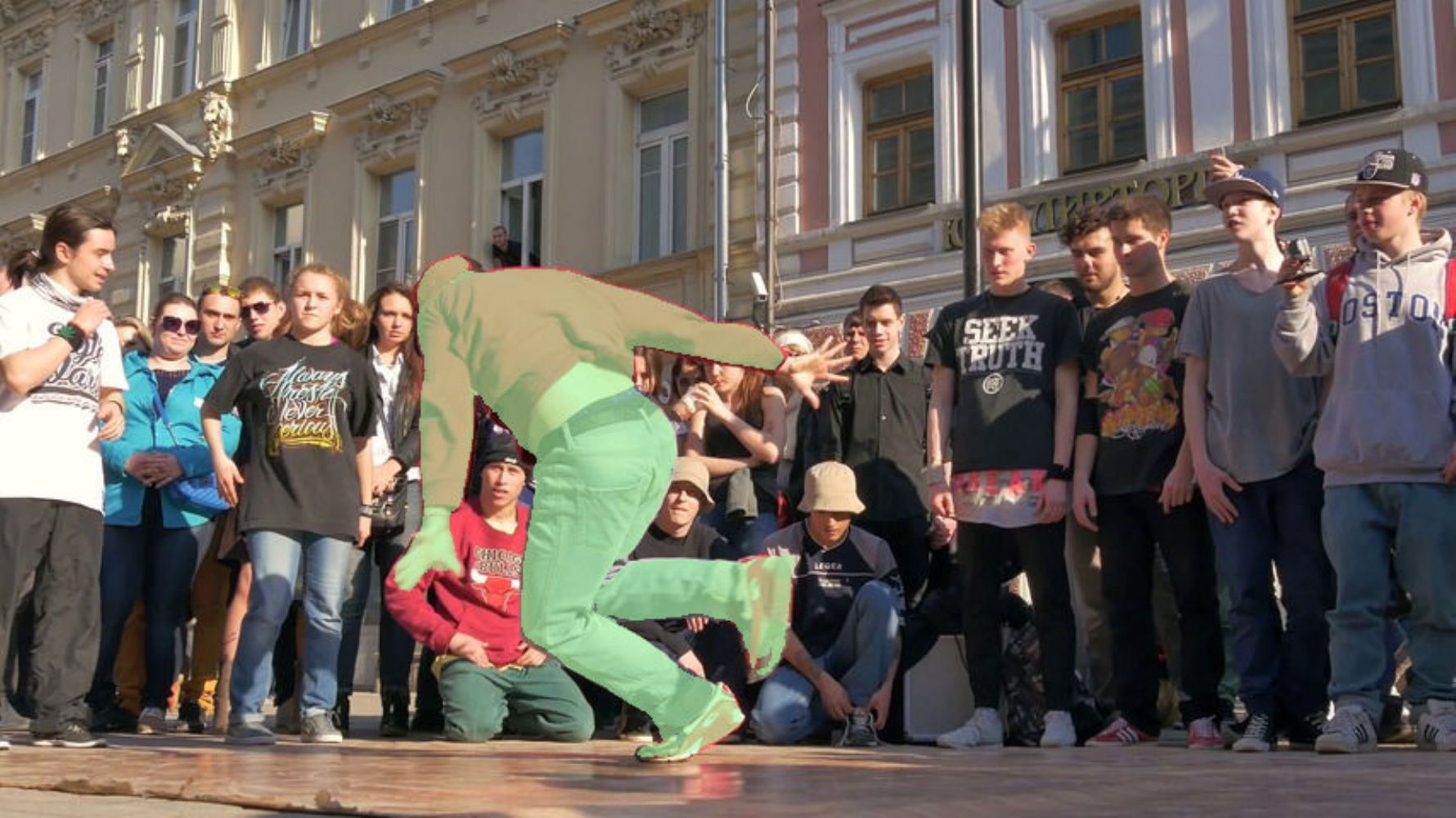}
        \vspace{-16pt}
    \end{subfigure}
    \vspace{5pt}

    \begin{subfigure}[b]{0.32\linewidth}
        \includegraphics[width=\mysize\linewidth]{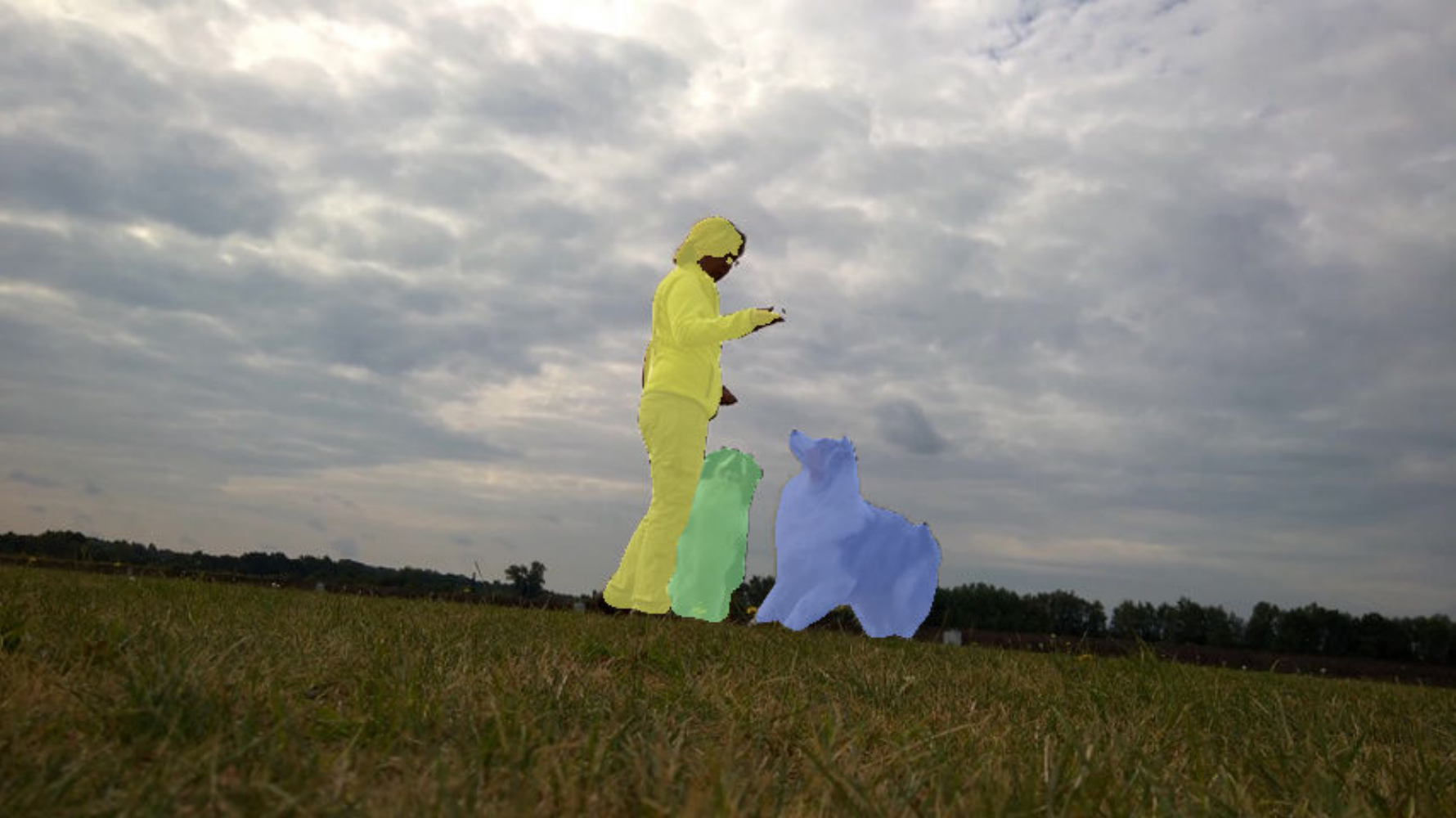}
        \vspace{-16pt}
    \end{subfigure}
    \begin{subfigure}[b]{0.32\linewidth}
        \includegraphics[width=\mysize\linewidth]{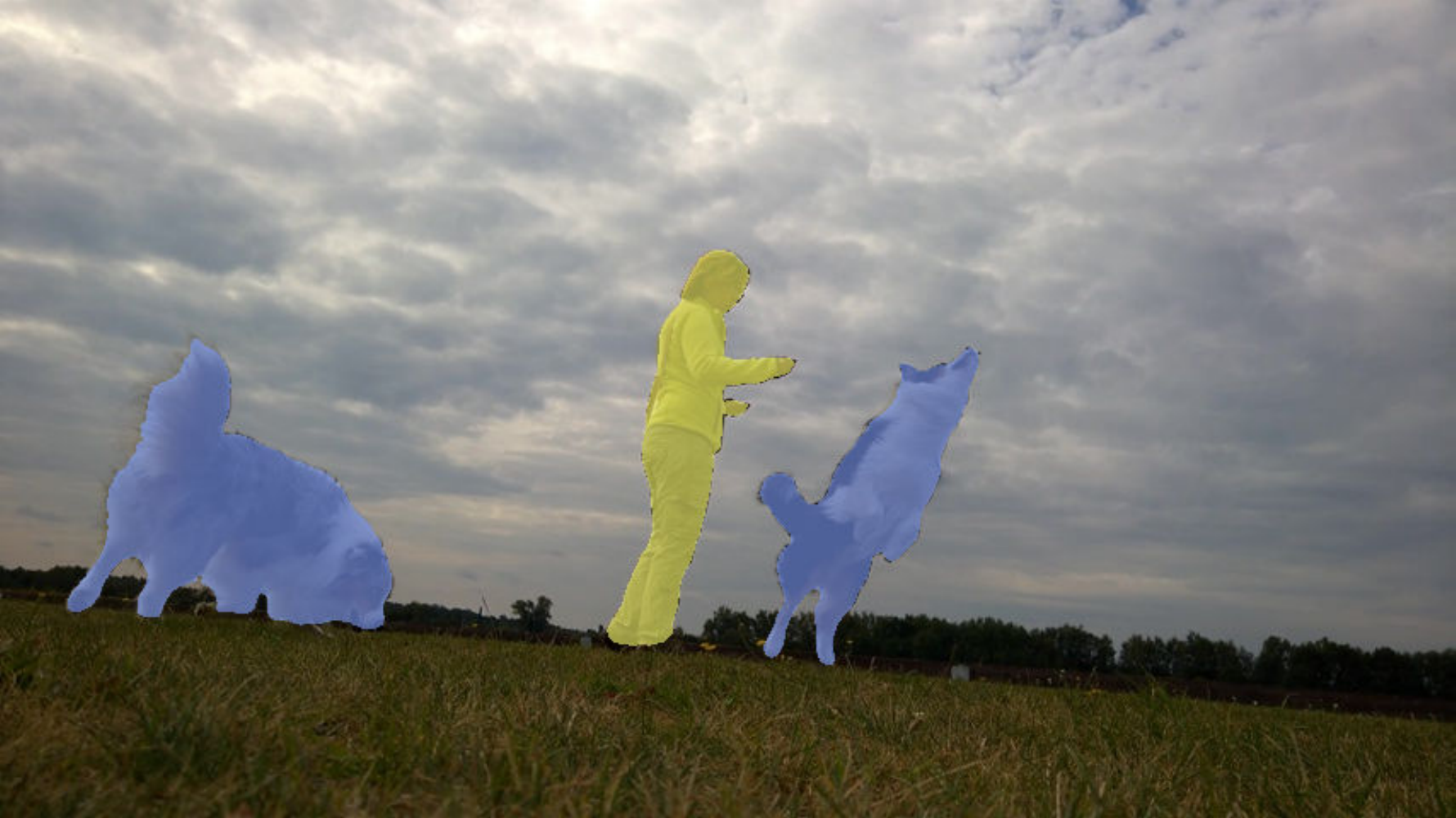}
        \vspace{-16pt}
    \end{subfigure}
    \begin{subfigure}[b]{0.32\linewidth}
     \includegraphics[width=\mysize\linewidth]{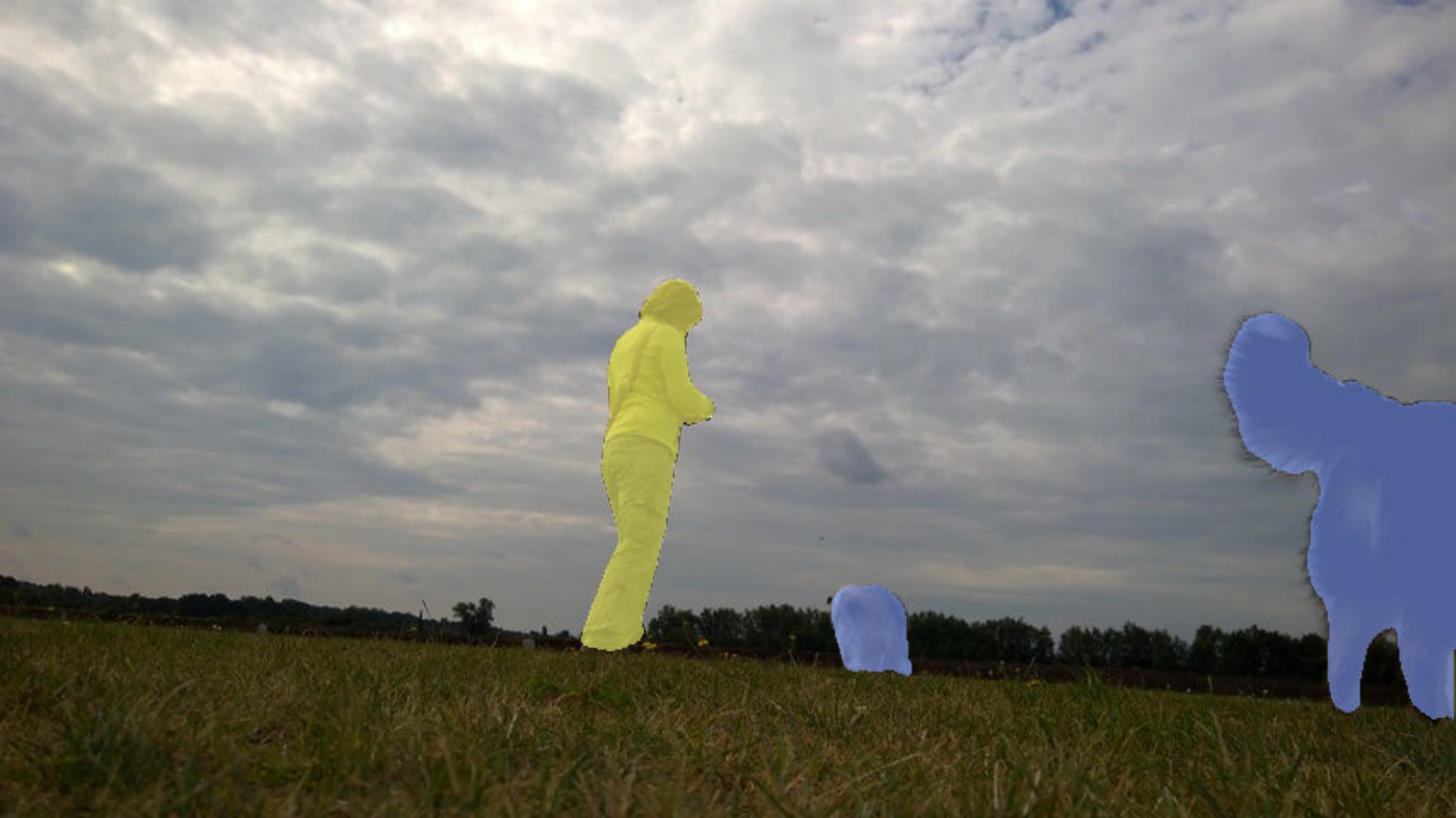}
        \vspace{-16pt}
    \end{subfigure}
    \vspace{5pt}

    \caption{\textbf{Tracking results of STCN~\cite{cheng2021stcn} on \pvdavis{}.} The model is pre-trained on \pvoops{} and \pvkinetics{} with pseudo-masks, then fine-tuned on \pvdavis{} and \pvytvos{} with pseudo-masks, and finally evaluated on 10 points setup. The pseudo-masks are generated from SAM~\cite{kirillov2023sam}.}
    \label{fig:pseudo_results_davis}
\end{figure*}

\begin{figure*}[ht!]
    \centering
    \setlength{\fboxsep}{0.32pt}
    \begin{subfigure}[b]{0.32\linewidth}
        \includegraphics[width=\mysize\linewidth]{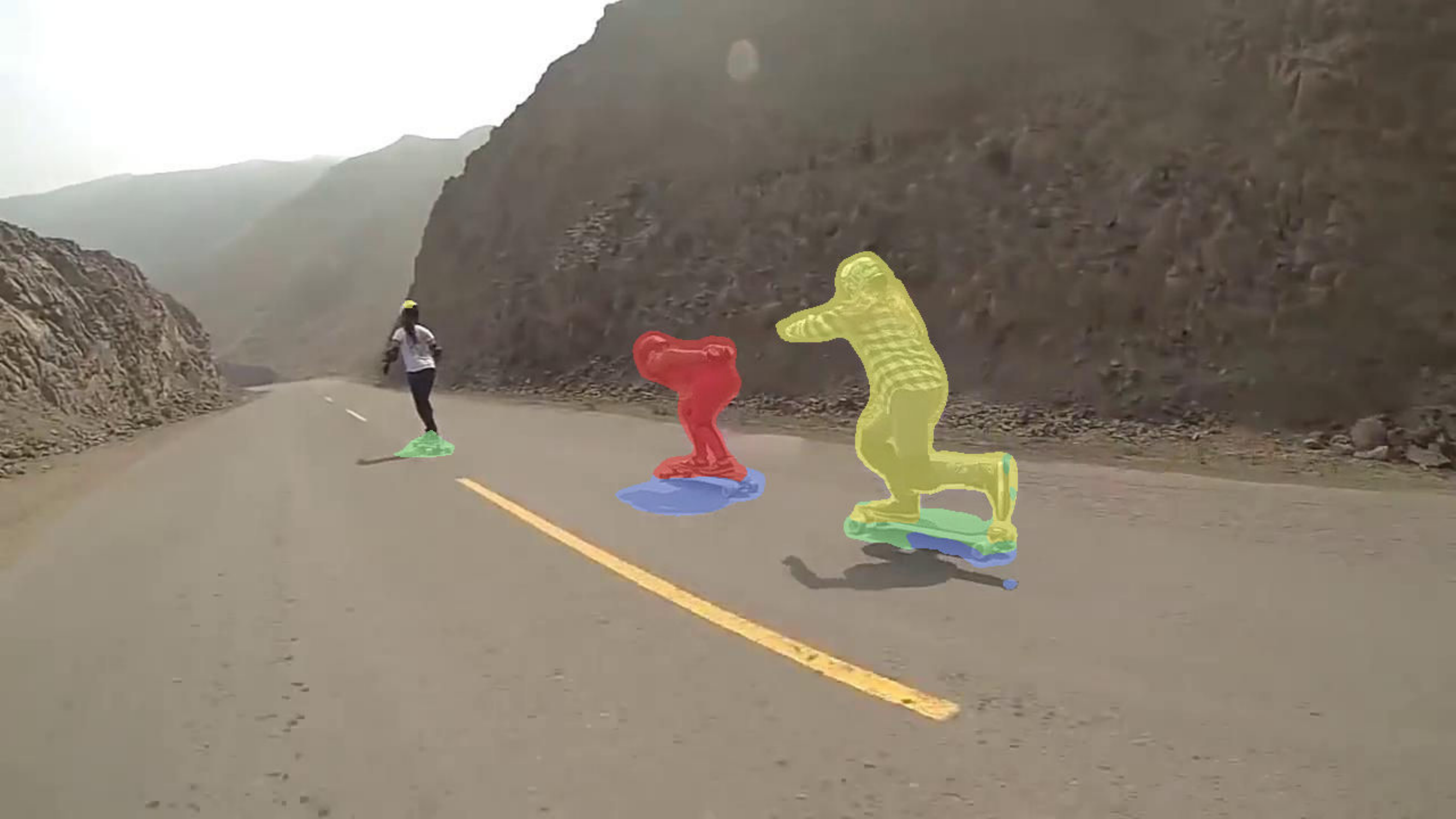}
        \vspace{-16pt}
    \end{subfigure}
    \begin{subfigure}[b]{0.32\linewidth}
     \includegraphics[width=\mysize\linewidth]{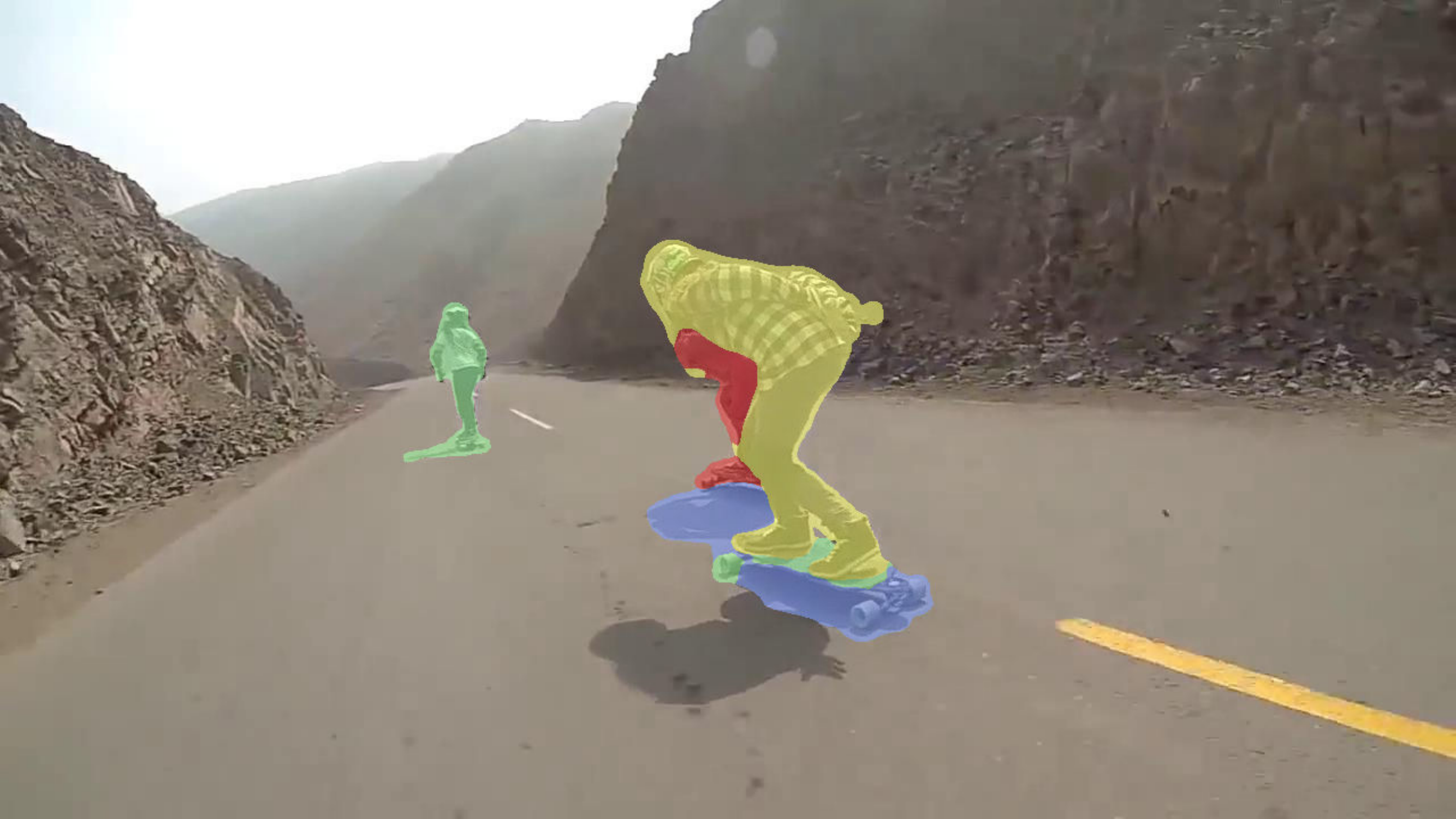}
        \vspace{-16pt}
    \end{subfigure}
    \begin{subfigure}[b]{0.32\linewidth}
     \includegraphics[width=\mysize\linewidth]{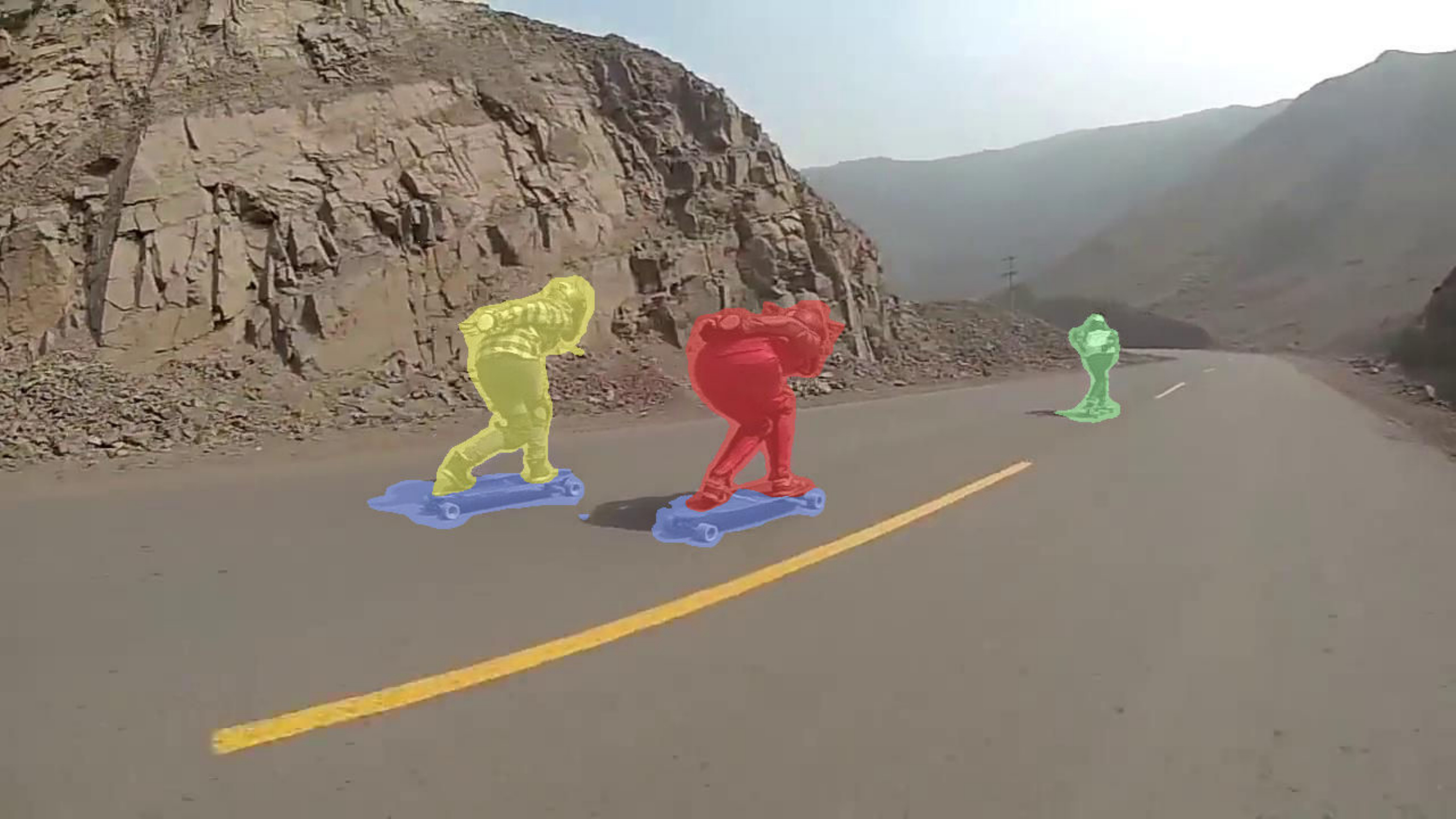}
        \vspace{-16pt}
    \end{subfigure}
    \vspace{5pt}

    \begin{subfigure}[b]{0.32\linewidth}
        \includegraphics[width=\mysize\linewidth]{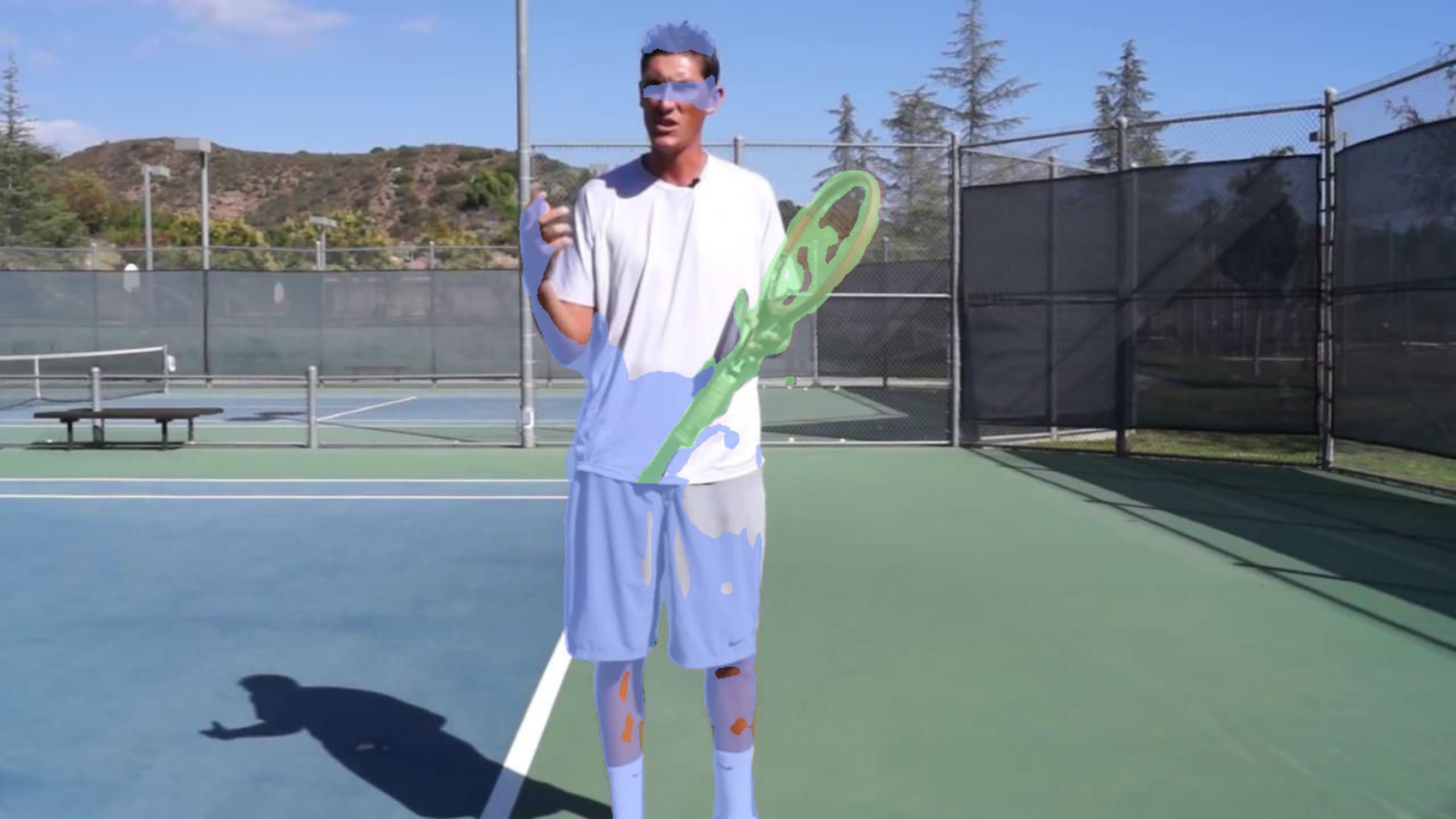}
        \vspace{-16pt}
    \end{subfigure}
    \begin{subfigure}[b]{0.32\linewidth}
     \includegraphics[width=\mysize\linewidth]{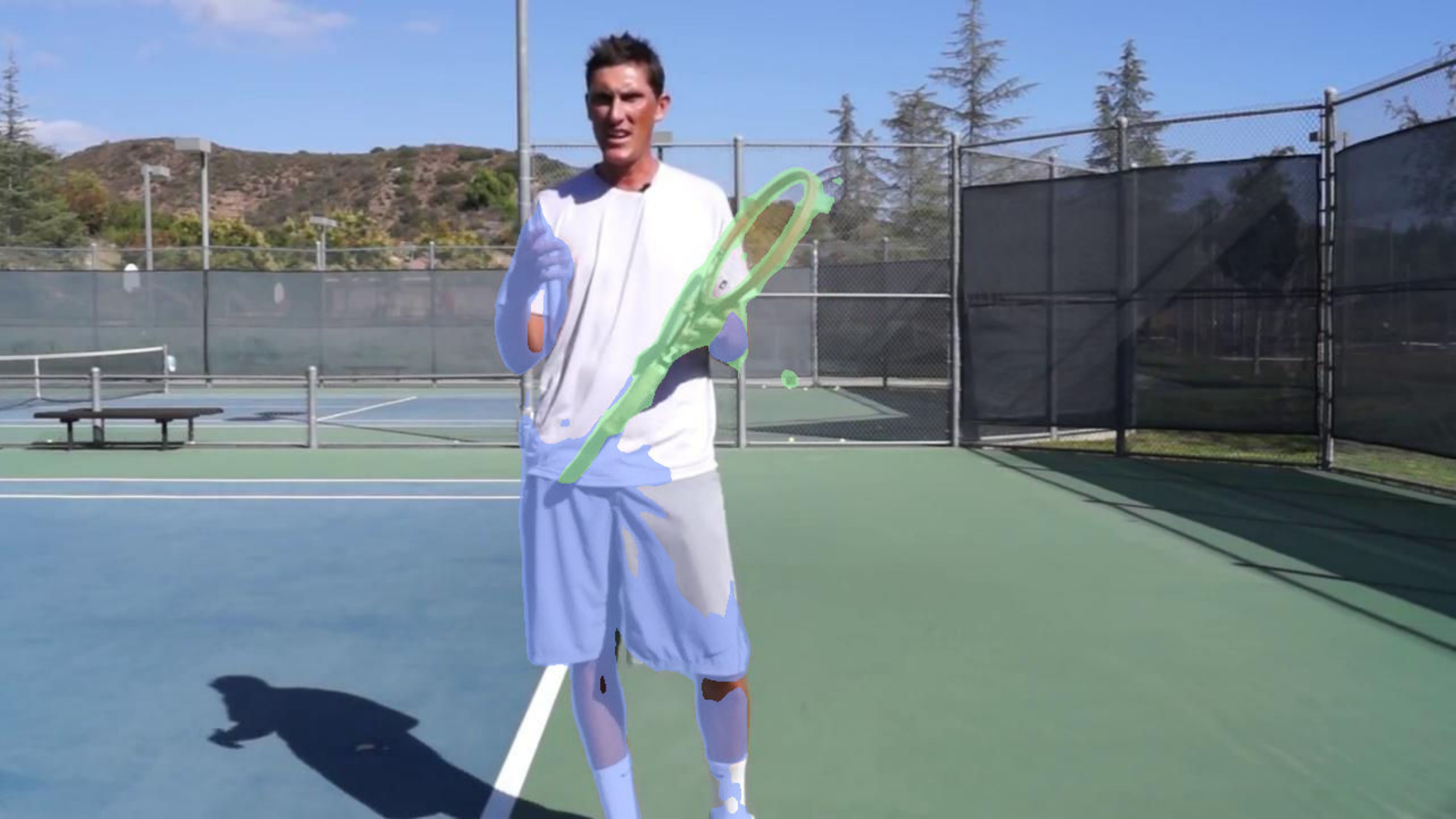}
        \vspace{-16pt}
    \end{subfigure}
    \begin{subfigure}[b]{0.32\linewidth}
     \includegraphics[width=\mysize\linewidth]{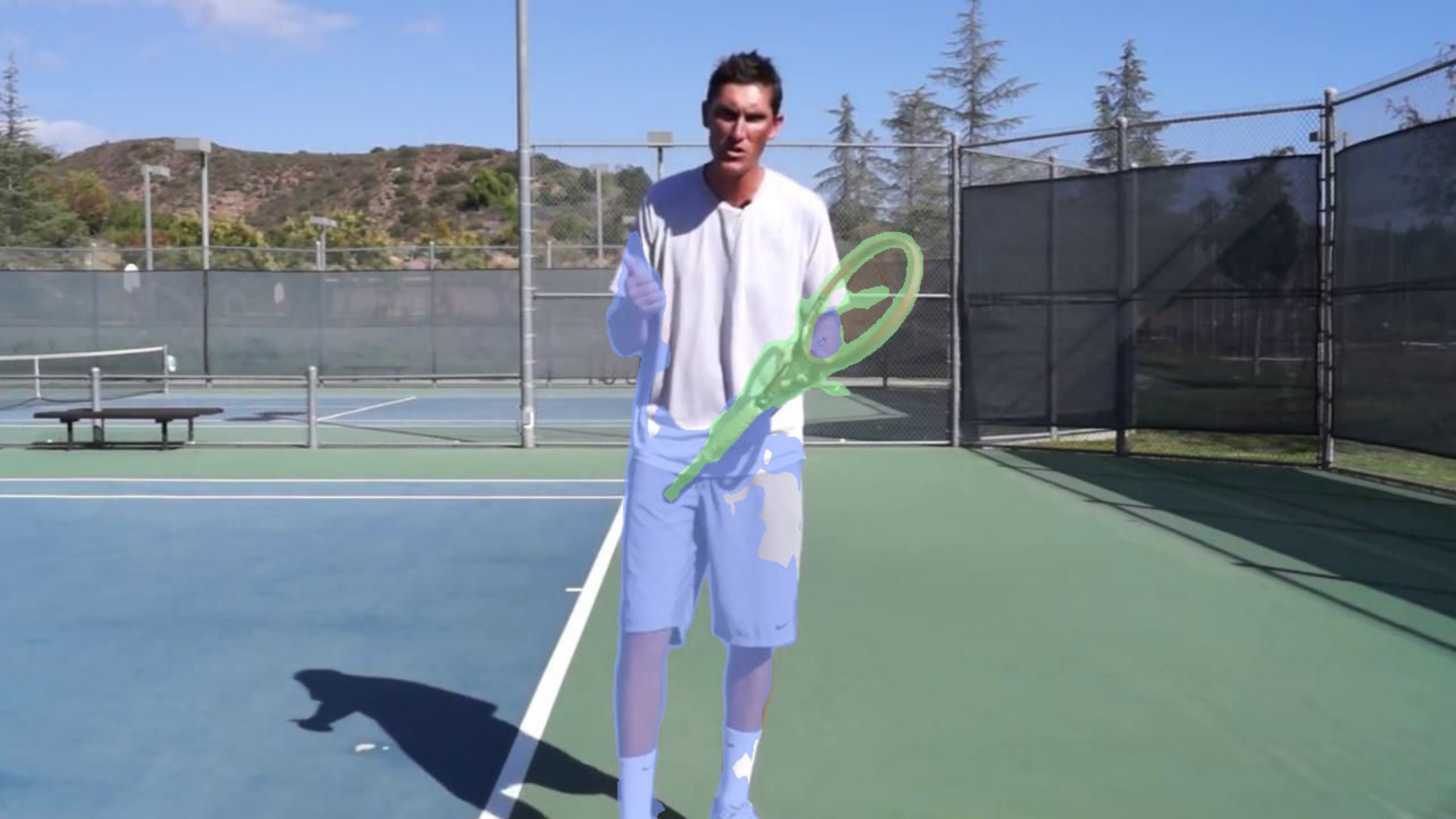}
        \vspace{-16pt}
    \end{subfigure}
    \vspace{5pt}

    \begin{subfigure}[b]{0.32\linewidth}
        \includegraphics[width=\mysize\linewidth]{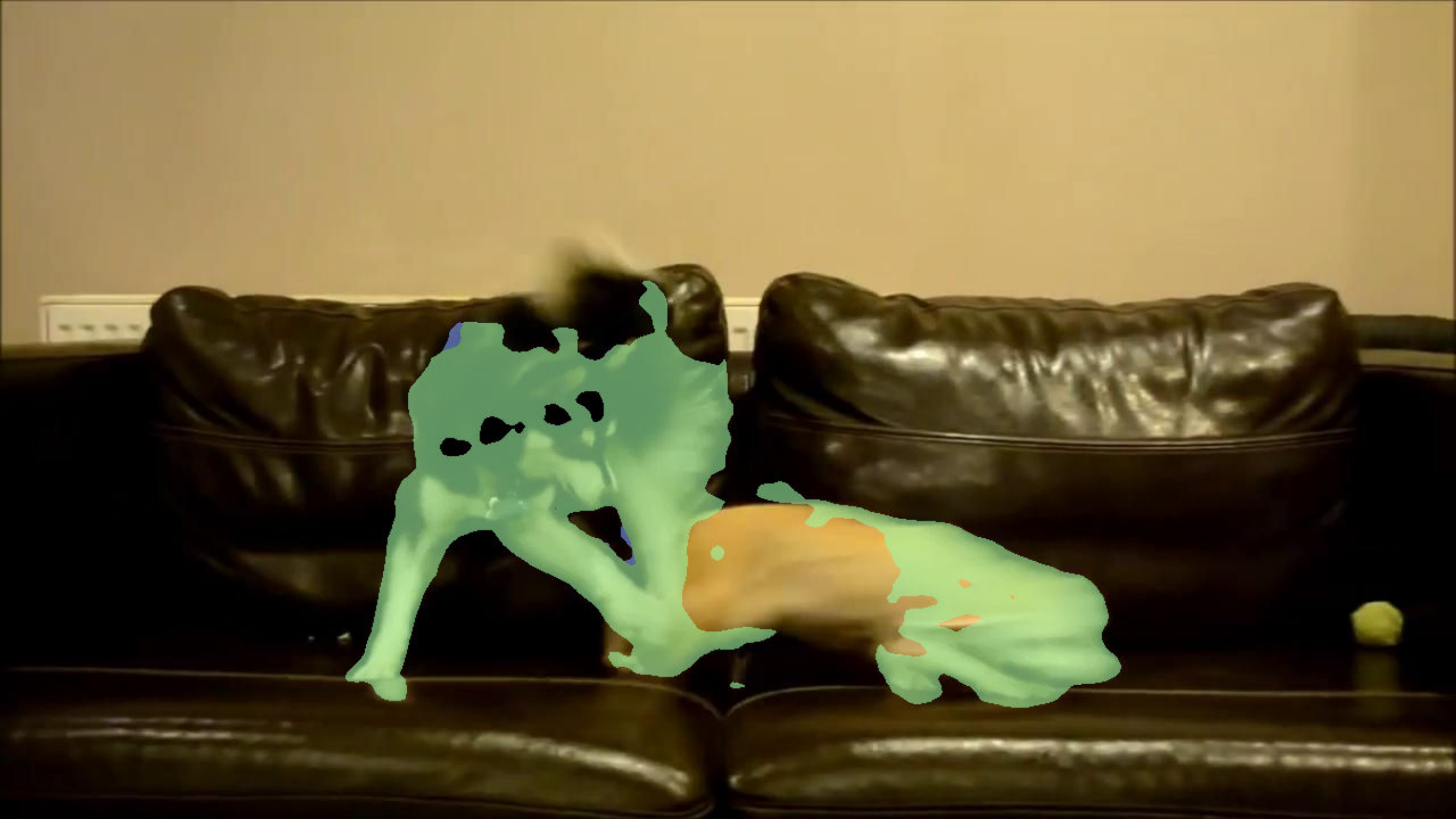}
        \vspace{-16pt}
    \end{subfigure}
    \begin{subfigure}[b]{0.32\linewidth}
        \includegraphics[width=\mysize\linewidth]{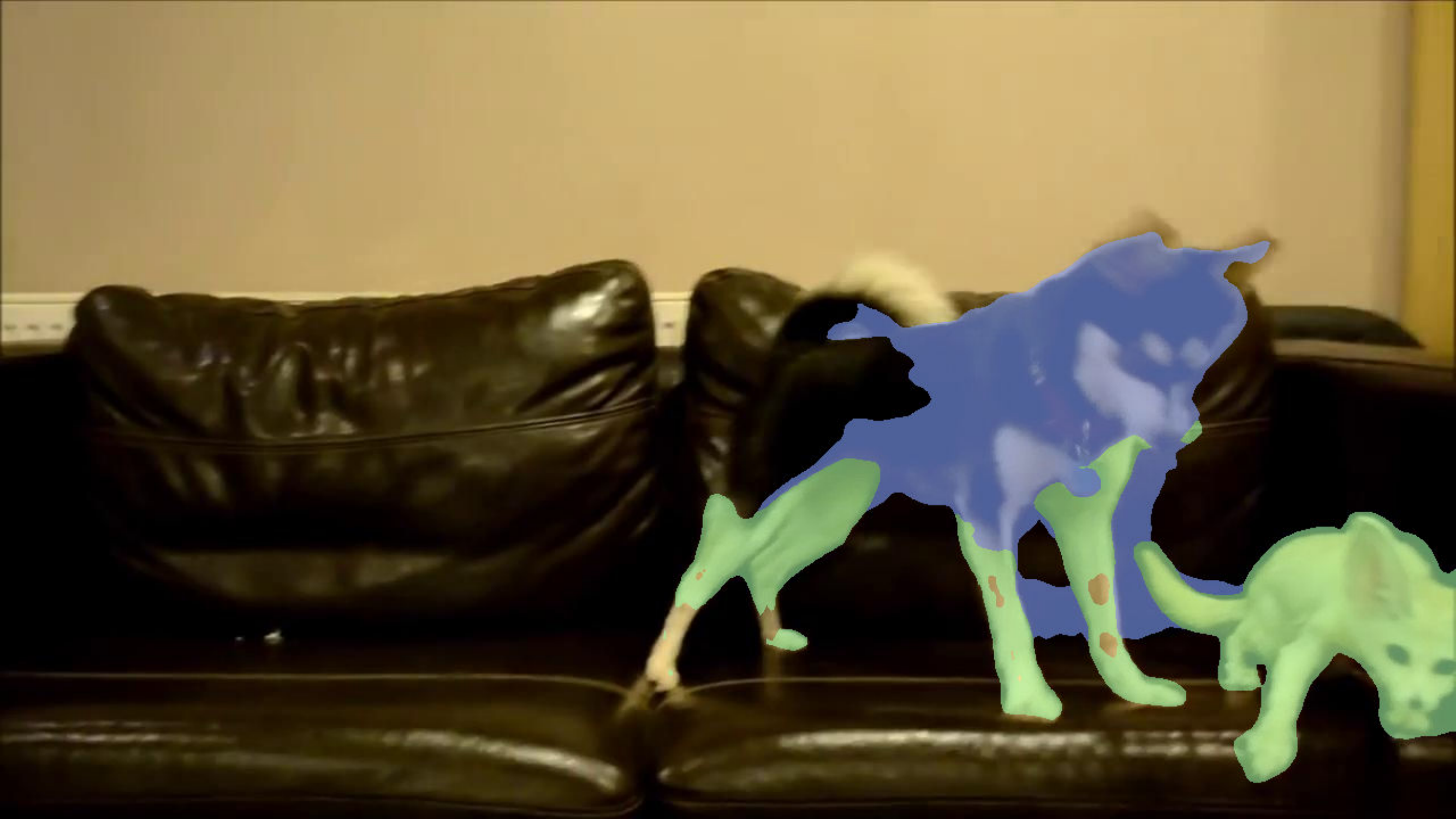}
        \vspace{-16pt}
    \end{subfigure}
    \begin{subfigure}[b]{0.32\linewidth}
     \includegraphics[width=\mysize\linewidth]{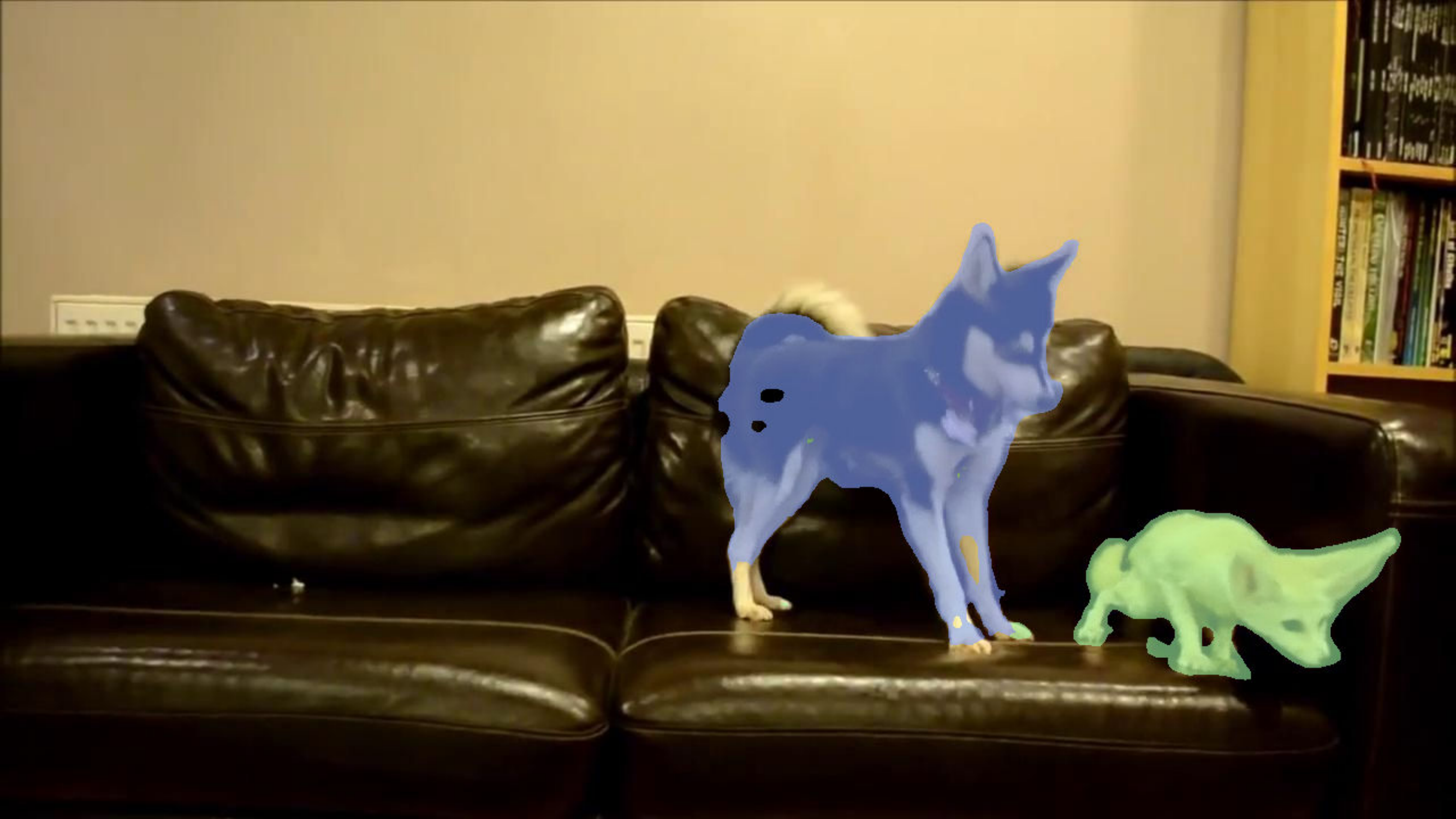}
        \vspace{-16pt}
    \end{subfigure}
    \vspace{5pt}

    \caption{\textbf{Tracking results of \pointstcn{} on \pvytvos{}.} The model is first pre-trained on \pvoops{} and \pvkinetics{} with points, then fine-tuned on \pvdavis{} and \pvytvos{} with points, and finally evaluated on the 10-point setup.}
    \label{fig:point_results_ytvos}
\end{figure*}

\begin{figure*}[ht!]
    \centering
    \setlength{\fboxsep}{0.32pt}
    \begin{subfigure}[b]{0.32\linewidth}
        \includegraphics[width=\mysize\linewidth]{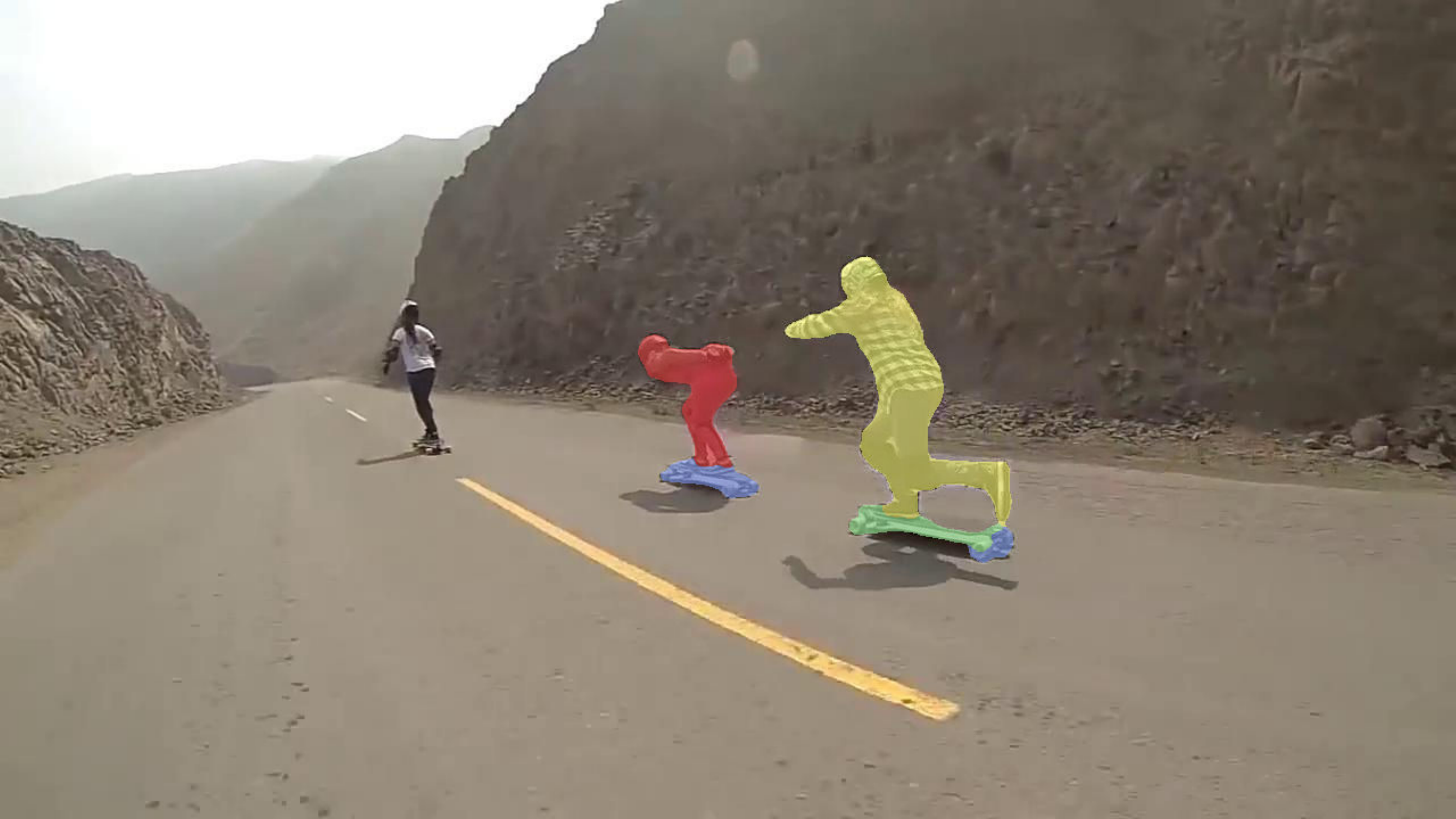}
        \vspace{-16pt}
    \end{subfigure}
    \begin{subfigure}[b]{0.32\linewidth}
     \includegraphics[width=\mysize\linewidth]{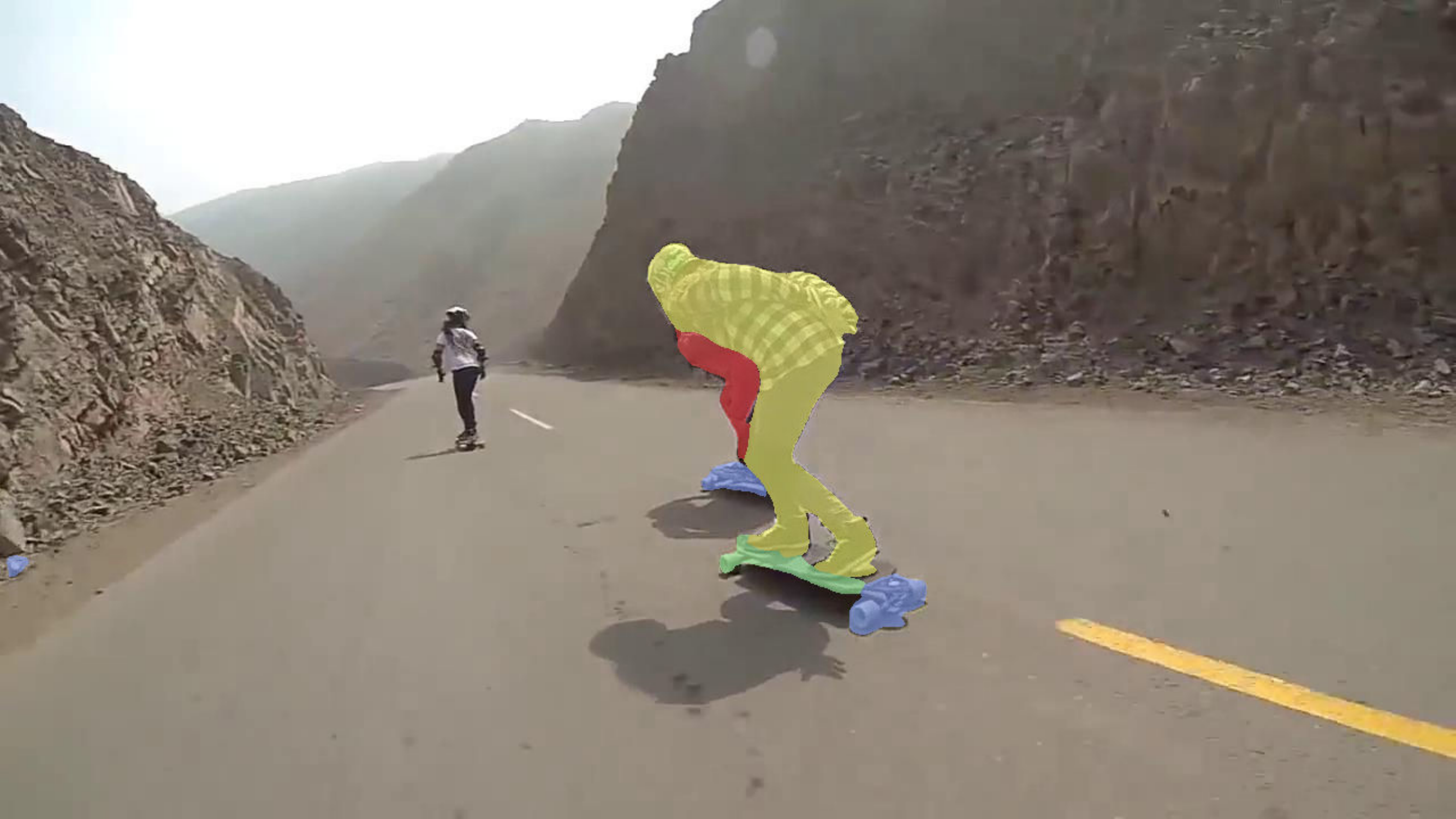}
        \vspace{-16pt}
    \end{subfigure}
    \begin{subfigure}[b]{0.32\linewidth}
     \includegraphics[width=\mysize\linewidth]{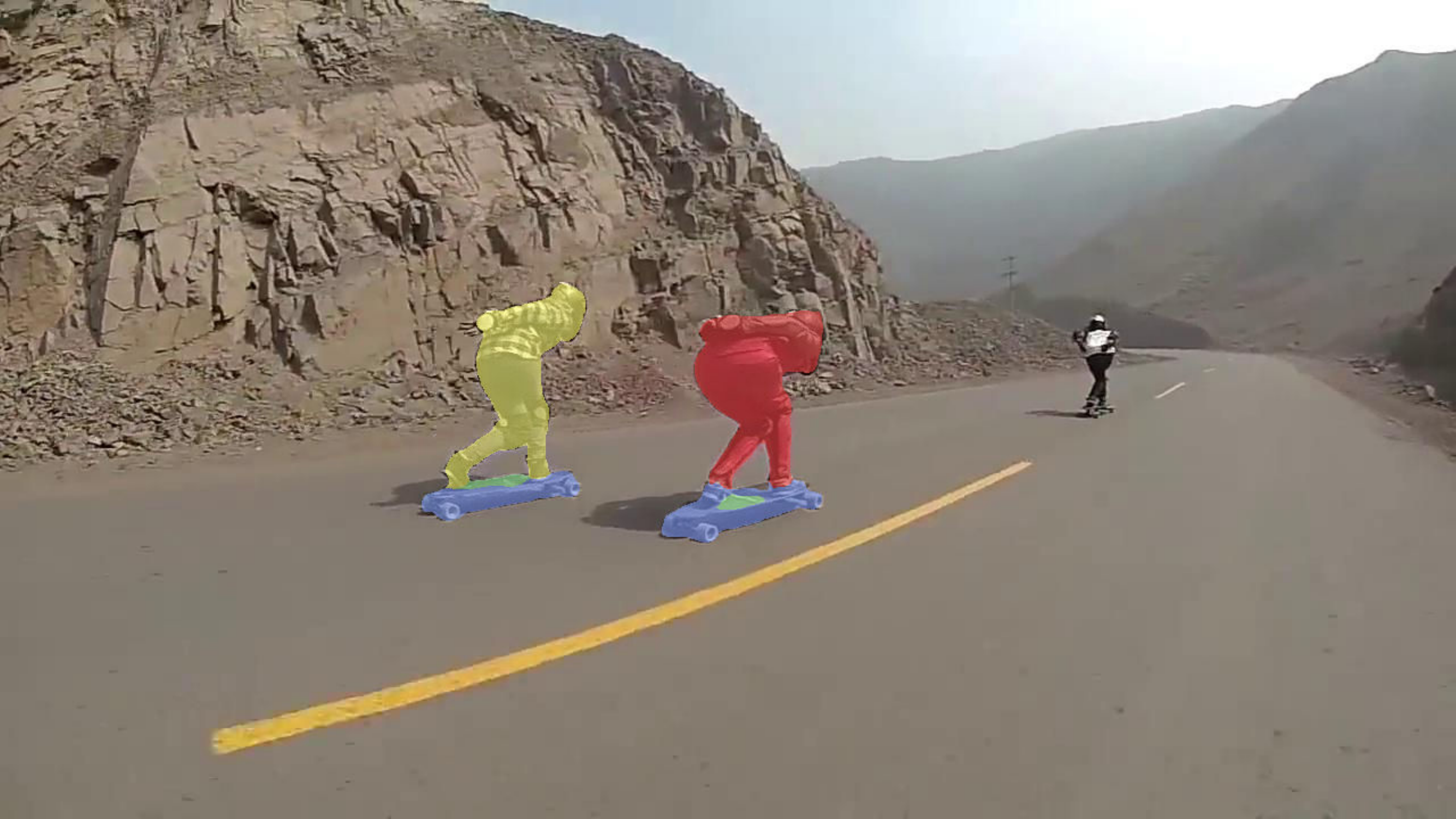}
        \vspace{-16pt}
    \end{subfigure}
    \vspace{5pt}

    \begin{subfigure}[b]{0.32\linewidth}
        \includegraphics[width=\mysize\linewidth]{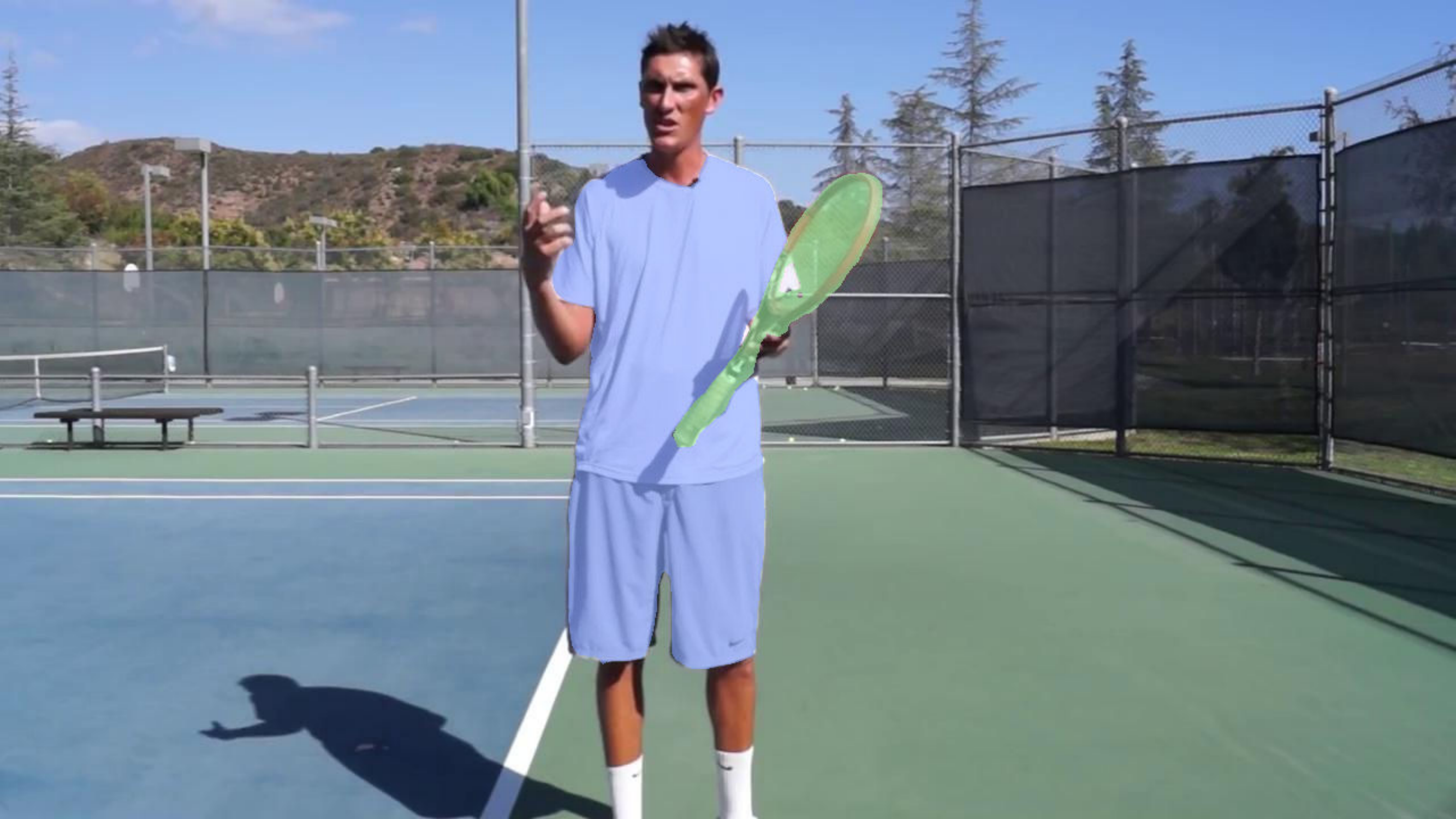}
        \vspace{-16pt}
    \end{subfigure}
    \begin{subfigure}[b]{0.32\linewidth}
     \includegraphics[width=\mysize\linewidth]{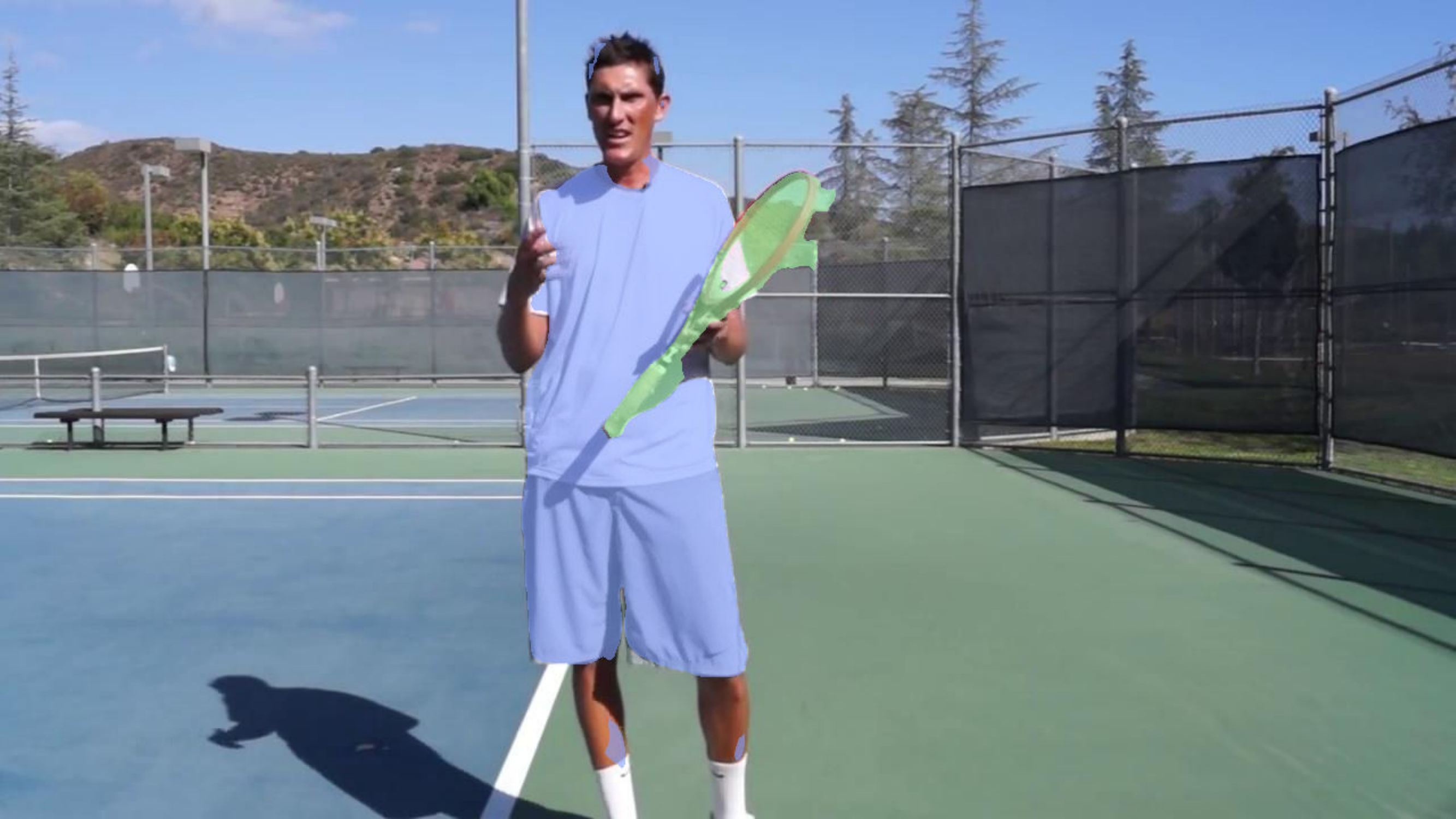}
        \vspace{-16pt}
    \end{subfigure}
    \begin{subfigure}[b]{0.32\linewidth}
     \includegraphics[width=\mysize\linewidth]{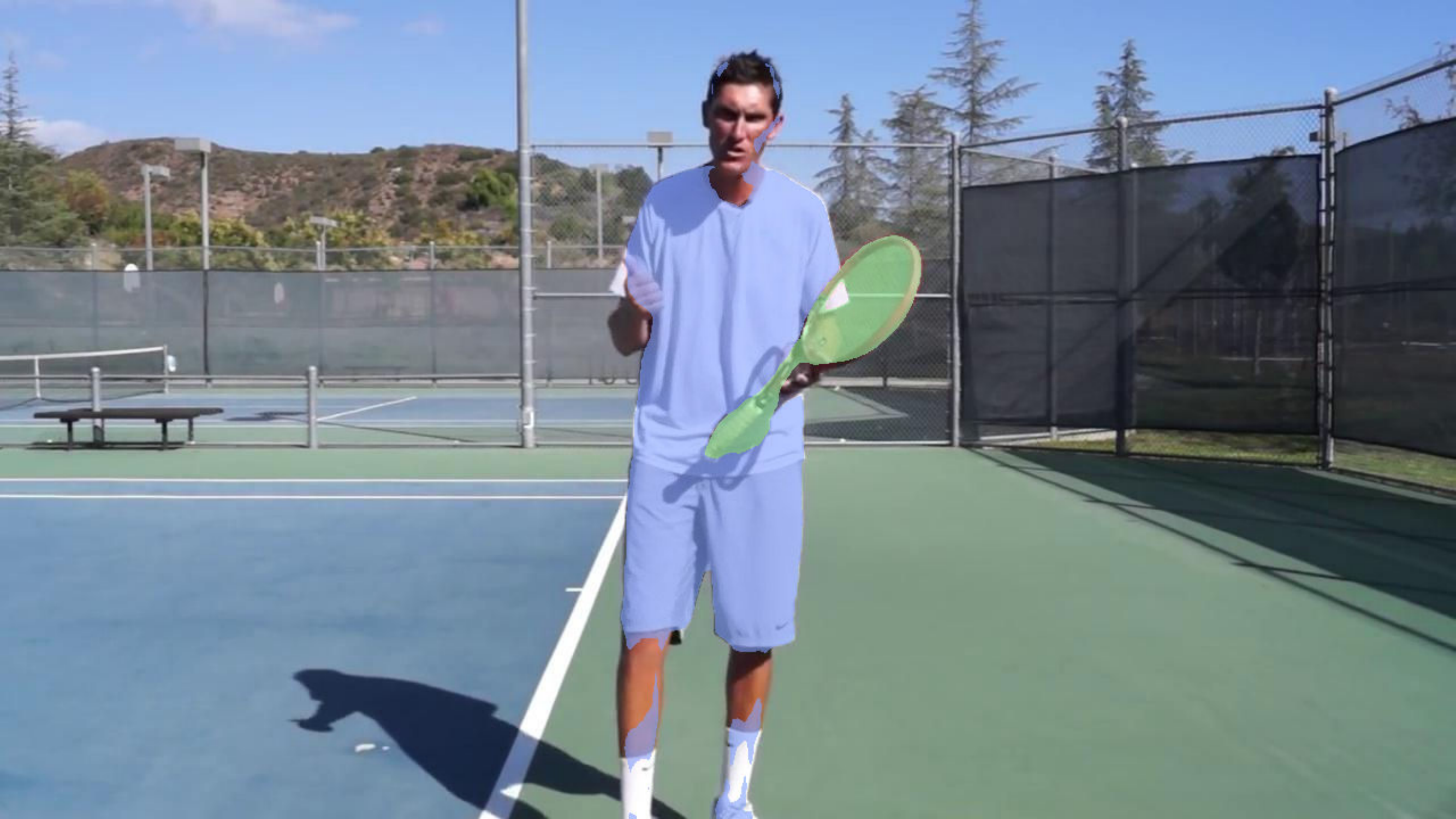}
        \vspace{-16pt}
    \end{subfigure}
    \vspace{5pt}

    \begin{subfigure}[b]{0.32\linewidth}
        \includegraphics[width=\mysize\linewidth]{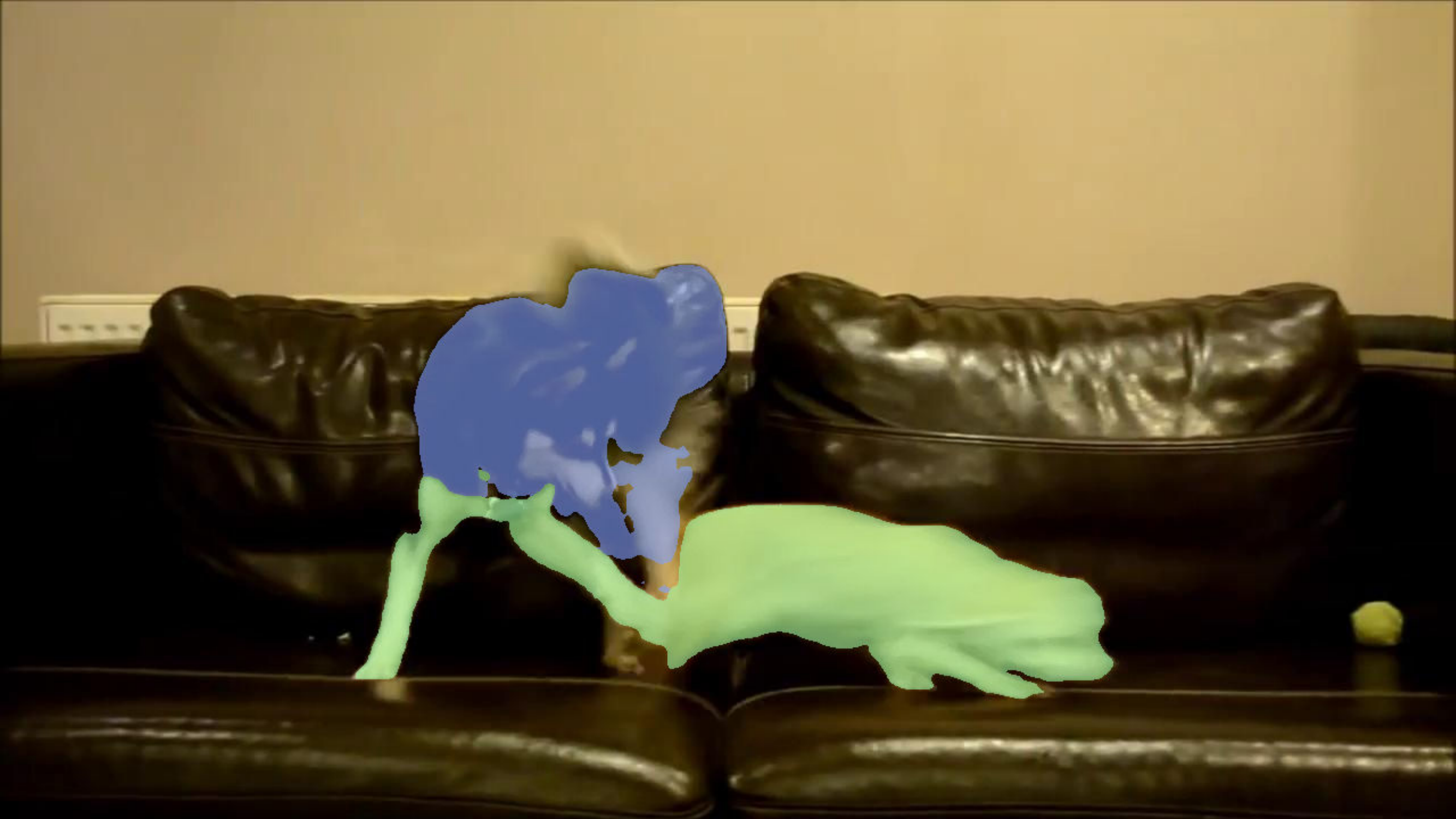}
        \vspace{-16pt}
    \end{subfigure}
    \begin{subfigure}[b]{0.32\linewidth}
        \includegraphics[width=\mysize\linewidth]{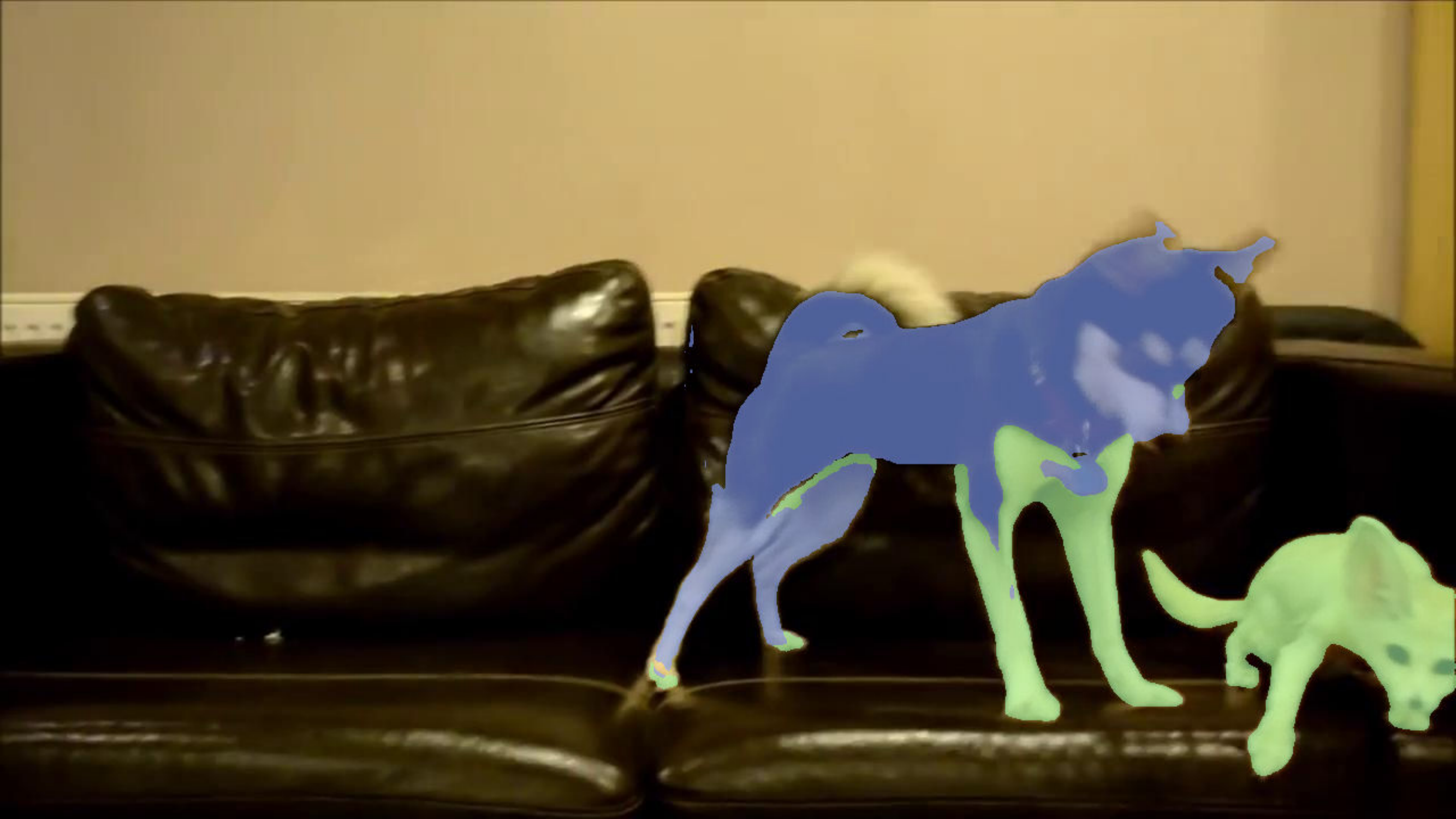}
        \vspace{-16pt}
    \end{subfigure}
    \begin{subfigure}[b]{0.32\linewidth}
     \includegraphics[width=\mysize\linewidth]{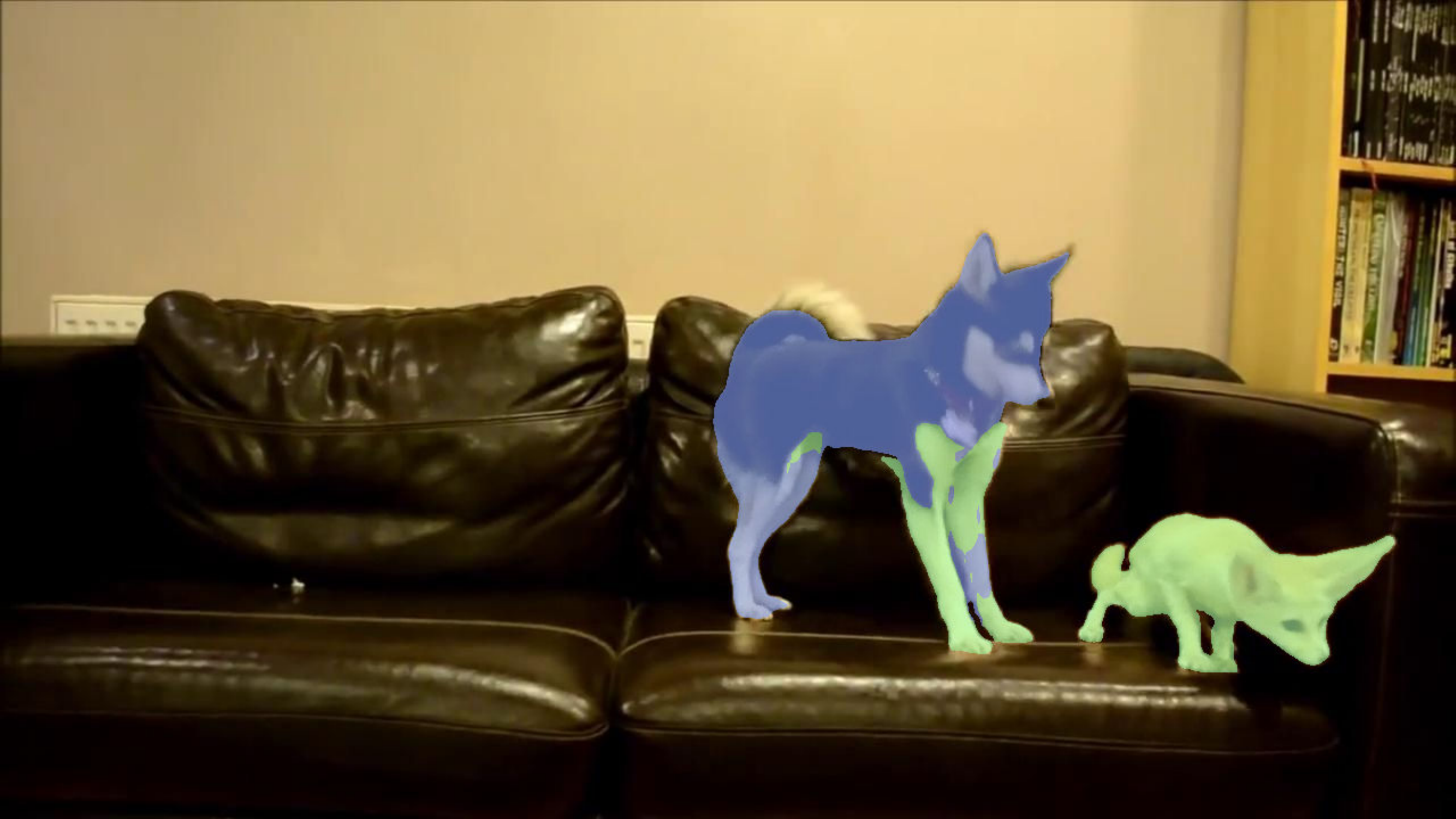}
        \vspace{-16pt}
    \end{subfigure}
    \vspace{5pt}

    \caption{\textbf{Tracking results of STCN~\cite{cheng2021stcn} on \pvytvos{}.} The model is pre-trained on \pvoops{} and \pvkinetics{} with pseudo-masks, then fine-tuned on \pvdavis{} and \pvytvos{} with pseudo-masks, and finally evaluated on 10 points setup. The pseudo-masks are generated from SAM~\cite{kirillov2023sam}.}
    \label{fig:pseudo_results_ytvos}
\end{figure*}

In~\cref{fig:point_results_oops}, ~\cref{fig:point_results_davis}, and~\cref{fig:point_results_ytvos}, we present the tracking results of \pointstcn{} (trained with points) on \pointvos{} Oops (\pvoops{}), \pointvos{} DAVIS (\pvdavis{}) and \pointvos{} YouTube (\pvytvos{}), respectively. Also, in~\cref{fig:pseudo_results_oops}, ~\cref{fig:pseudo_results_davis} and~\cref{fig:pseudo_results_ytvos}, we demonstrate the results of
\stcn{}~\cite{cheng2021stcn} (trained with pseudo-masks) on \pvoops{}, \pvdavis{} and \pvytvos{}, respectively.

\definecolor{figgreen}{RGB}{0, 177, 29}  
\definecolor{figred}{RGB}{12, 34, 238}

\definecolor{figgreen}{RGB}{0, 177, 29}  
\definecolor{figred}{RGB}{12, 34, 238}
\definecolor{figgreen}{RGB}{0, 177, 29}  
\definecolor{figred}{RGB}{12, 34, 238}

\definecolor{figgreen}{RGB}{0, 177, 29}  
\definecolor{figred}{RGB}{12, 34, 238}
\definecolor{figgreen}{RGB}{0, 177, 29}  
\definecolor{figred}{RGB}{12, 34, 238}

\definecolor{figgreen}{RGB}{0, 177, 29}  
\definecolor{figred}{RGB}{12, 34, 238}
\definecolor{figgreen}{RGB}{0, 177, 29}  
\definecolor{figred}{RGB}{12, 34, 238}

\end{document}